\documentclass[10pt]{article} 
\usepackage[accepted]{_format_tmlr/tmlr}

\usepackage{amsmath,amsfonts,bm}

\def\eqref#1{equation~\ref{#1}}

\def\1{\bm{1}}

\DeclareMathAlphabet{\mathsfit}{\encodingdefault}{\sfdefault}{m}{sl}
\SetMathAlphabet{\mathsfit}{bold}{\encodingdefault}{\sfdefault}{bx}{n}

\usepackage{graphicx}
\usepackage{subcaption}
\usepackage{booktabs}
\usepackage{tikz}
\usepackage{acro}
\usepackage{amsthm}
\usepackage{nicefrac}
\usepackage{enumitem}
\usepackage{bbm}
\usepackage{pifont}
\usepackage{placeins}
\usepackage{xcolor}
\usepackage{multirow}
\usepackage{rotating}
\usepackage{makecell}
\usepackage{threeparttable}

\acsetup{
  first-style=long-short,
  list/display=used,  
  make-links=false
}

\usepackage[accsupp]{axessibility}  
\usepackage[hidelinks]{hyperref}
\usepackage[capitalise]{cleveref}
\usepackage{url}

\title{From Keypoints to Predictive Distributions:\\ Post-Hoc Uncertainty for YOLO-Pose Models}

\author{\name Alexej Klushyn\textsuperscript{1} \quad
        Juan Rivero Sesma\textsuperscript{2} \quad
        Florian Seligmann\textsuperscript{3}\\[0.35em]
        Richard Kurle\textsuperscript{4} \quad
        Kinh Tieu\textsuperscript{5} \quad
        Jayant Sen Gupta\textsuperscript{1} \\[1.1em]
        \addr \textsuperscript{1}Airbus Central Research \& Technology \quad
              \textsuperscript{2}Airbus Commercial Aircraft \\[0.35em]
        \addr \textsuperscript{3}Karlsruhe Institute of Technology\quad
              \textsuperscript{4}Mistral AI \quad
              \textsuperscript{5}Acubed
}

\newcommand{\blfootnote}[1]{\begingroup\renewcommand\thefootnote{}\footnote{#1}\addtocounter{footnote}{-1}\endgroup}

\definecolor{mygreen}{RGB}{0,150,90}

\newtheorem{lemma}{Lemma}

\crefname{section}{Sec.}{Secs.}
\crefname{table}{Tab.}{Tabs.}

\def\month{11}  
\def\year{2026} 
\def\openreview{\url{https://openreview.net/forum?id=MHG2uIP3hB}} 

\DeclareAcronym{nll}{
  short = NLL,
  long  = negative log-likelihood
}

\DeclareAcronym{gnll}{
  short = GNLL,
  long  = Gaussian negative log-likelihood
}

\DeclareAcronym{nms}{
  short = NMS,
  long  = non-maximum suppression
}

\DeclareAcronym{uq}{
  short = UQ,
  long  = uncertainty quantification
}

\DeclareAcronym{oks}{
  short = OKS,
  long  = object keypoint similarity
}

\DeclareAcronym{ks}{
  short = KS,
  long  = keypoint similarity
}

\DeclareAcronym{iou}{
  short = IoU,
  long  = intersection over union
}

\DeclareAcronym{iw}{
  short = IW,
  long  = importance weighting
}

\DeclareAcronym{ap}{
  short = AP,
  long  = average precision
}

\DeclareAcronym{ar}{
  short = AR,
  long  = average recall
}

\DeclareAcronym{tp}{
  short = TP,
  long  = true positive
}

\DeclareAcronym{roc}{
  short = ROC,
  long  = receiver operating characteristic
}

\DeclareAcronym{fp}{
  short = FP,
  long  = false positive
}

\DeclareAcronym{akp}{
  short = AKP,
  long  = average keypoint precision
}

\DeclareAcronym{ace}{
  short = ACE,
  long  = average coverage error
}

\DeclareAcronym{ence}{
  short = ENCE,
  long  = expected normalized calibration error
}

\DeclareAcronym{ece}{
  short = ECE,
  long  = expected calibration error
}

\DeclareAcronym{qq}{
  short = Q--Q,
  long  = quantile--quantile
}

\DeclareAcronym{vbl}{
  short = VBL,
  long  = vision-based landing
}

\DeclareAcronym{sf}{
  short = SF,
  long  = sensor fusion
}

\DeclareAcronym{pnp}{
  short = PnP,
  long  = Perspective-\texorpdfstring{$n$}{n}-Point
}

\DeclareAcronym{gbas}{
  short = GBAS,
  long  = ground-based augmentation system
}

\DeclareAcronym{gls}{
  short = GLS,
  long  = GBAS landing system
}

\DeclareAcronym{gnss}{
  short = GNSS,
  long  = global navigation satellite system
}

\DeclareAcronym{ils}{
  short = ILS,
  long  = instrument landing system
}

\DeclareAcronym{irs}{
  short = IRS,
  long  = inertial reference system
}

\DeclareAcronym{ra}{
  short = RA,
  long  = radio altimeter
}

\DeclareAcronym{atc}{
  short = ATC,
  long  = air traffic control
}

\DeclareAcronym{rle}{
  short = RLE,
  long  = residual log-likelihood estimation
}

\DeclareAcronym{ekf}{
  short = EKF,
  long  = extended Kalman filter
}

\begin{document}

\maketitle
\blfootnote{Corresponding author: \texttt{alexej.klushyn@airbus.com}}

\begin{abstract}
YOLO-Pose models provide efficient multi-object keypoint localization, but do not quantify the associated spatial uncertainty.
We introduce a lightweight post-hoc probabilistic extension that augments a trained YOLO-Pose model with calibrated bivariate predictive distributions over keypoint locations, centered at the model's original predictions.
Concretely, we train additional probabilistic heads with an importance-weighted negative log-likelihood to predict an input-dependent $2\times2$ dispersion matrix for each keypoint, followed by Gaussian calibration for broad downstream compatibility or Student-$t$ calibration for distributional fidelity.
Complementing this, we propose an evaluation protocol that combines a suite of distributional calibration diagnostics with \acf{akp}, a keypoint-level extension of the COCO \acs{ap} protocol for assessing reliability rankings.
Experiments on COCO with YOLO-NAS-Pose and YOLO26-Pose show that the learned uncertainty estimates enable keypoint-level reliability ranking, Student-$t$ calibration best captures the empirical residual distribution, and uncertainty-based pruning removes unreliable keypoints.
In vision-based aircraft landing, calibrated predictive distributions over runway keypoints support uncertainty-aware position estimation and downstream sensor fusion.
\end{abstract}
\section{Introduction}
\label{sec:intro}

YOLO-Pose models~\citep{maji2022yolopose} are widely used in real-time applications because they predict multiple object instances alongside their keypoints in a single forward pass, while benefiting from highly optimized implementations and deployment pipelines~\citep{supergradients,ultralytics}.
This makes them attractive whenever low latency and reliable keypoint localization must be achieved under practical deployment constraints.
However, each keypoint is represented by a single predicted location, and the instance-level confidence score does not quantify keypoint-specific spatial uncertainty.
To support localization reliability assessment and informed keypoint selection or weighting in downstream applications, localization uncertainty must be quantified at the level of individual keypoints.

Within the YOLO family, uncertainty quantification has largely focused on object detection, typically by jointly learning bounding-box predictions and their associated uncertainty~\citep[e.g.][]{kraus2019uq-boundingbox,choi2019gaussianyolo}.
The sole alternative at the keypoint level is normalizing-flow-based \ac{rle}~\citep{li2021keypoint-normalizing-flows}, built into YOLO26-Pose~\citep{jocher2026ultralyticsyolo26}.
As we show in \cref{sec:coco}, its uncertainty estimates are uncalibrated and capture input dependence less accurately than our approach.
Moreover, the joint-learning paradigm is not always desirable.
First, many models are already trained, tuned, validated, or deployed, so changing their predictions may require renewed evaluation or certification.
Second, replacing a task-specific loss with a likelihood-based objective is not guaranteed to preserve prediction performance.
If the assumed residual distribution is misspecified, the induced error weighting may degrade prediction accuracy.
In addition, joint mean--variance optimization, for example, can compromise the mean fit unless carefully regularized~\citep{stirn2023faithful,seitzer2022pitfalls}.

\paragraph{Post-Hoc Probabilistic Extension for YOLO-Pose Models.}
Our formulation separates uncertainty estimation from keypoint localization.
Starting from an already trained YOLO-Pose model, we keep its parameters fixed and learn bivariate predictive distributions over keypoint locations, centered at the model's original predictions.
To this end, we propose a lightweight post-hoc probabilistic extension that introduces additional heads into the YOLO architecture to estimate a heteroscedastic positive-definite $2\times2$~\mbox{dispersion} matrix for each keypoint.
This compact distributional representation provides input-dependent uncertainty estimates that can be learned from finite data and used directly in downstream processing.
An importance-weighted negative log-likelihood shifts dispersion learning toward the prediction distribution expected after inference-time post-processing and targets balanced calibration across object scales.

The input data encountered at inference may differ from the data used for dispersion learning.
To compensate for this, we calibrate the learned dispersion matrices on held-out data representative of the intended operating conditions.
Furthermore, keypoint residuals can be heavy-tailed due to occlusion, truncation, ambiguity, annotation noise, and the robust, saturating form of the \acl{oks} loss~\citep{maji2022yolopose} used in YOLO-Pose models.
We therefore consider two complementary calibration schemes: Gaussian calibration for broad downstream compatibility with covariance-based methods, and Student-$t$ calibration for improved distributional fidelity to the empirical residual distribution.
Both calibration schemes are applied to the same frozen YOLO-Pose model and trained dispersion model.
Using calibrated uncertainty estimates as a keypoint-level reliability signal, we further propose uncertainty-based keypoint pruning to remove high-uncertainty keypoints before downstream processing.
Since the probabilistic heads depend only on the YOLO architecture's neck features, the pipeline generalizes across YOLO-Pose variants.
To our knowledge, this is the first method to provide calibrated per-keypoint predictive distributions for YOLO-Pose models.

\begin{figure}[t]
    \centering
    \includegraphics[width=\linewidth]{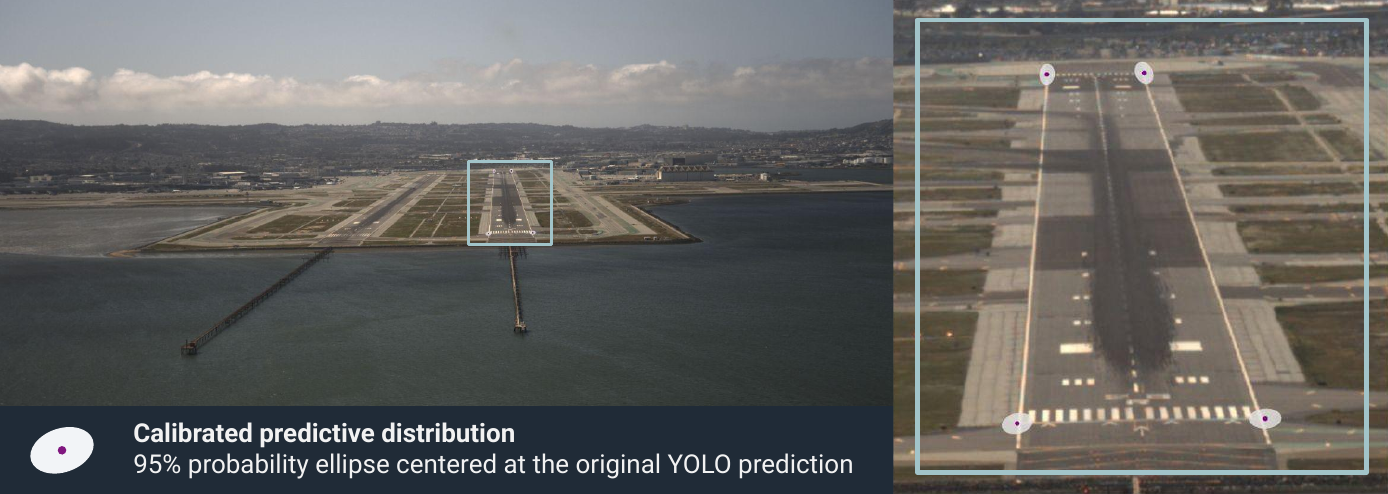}
    \caption{
    Vision-based aircraft landing.
    Given a trained YOLO-Pose model, our post-hoc probabilistic extension provides calibrated predictive distributions over runway keypoint locations, centered at the model's original predictions.
    Ellipses denote the corresponding 95\% probability regions.
    }
    \label{fig:intro}
\end{figure}

\paragraph{Contributions:}
\begin{enumerate}[leftmargin=*, itemsep=0pt, topsep=2pt]
    \item \textbf{Post-hoc dispersion heads for YOLO-Pose.}
    Starting from a frozen YOLO-Pose model, lightweight heads predict a per-keypoint $2\times2$ dispersion matrix via Cholesky parameterization.
    An importance-weighted \acl{nll} aligns dispersion learning with inference-time post-processing and balances calibration across object scales (\cref{subsec:iw-nll}).
    \item \textbf{Gaussian and Student-$t$ calibration.}
    Two calibration schemes operate on the same trained dispersion model: Gaussian temperature scaling for downstream covariance compatibility, and Student-$t$ calibration for distributional fidelity to the heavy-tailed residuals induced by the saturating \acs{oks} loss (\cref{subsec:calibration}).
    \item \textbf{Uncertainty-based keypoint pruning.}
    Calibrated uncertainty serves as a per-keypoint reliability signal to selectively remove unreliable keypoints before downstream processing (\cref{subsec:kp_pruning}).
    \item \textbf{Evaluation protocol for bivariate keypoint distributions.}
    We propose \acf{akp}, a keypoint-level extension of the COCO \acf{ap} protocol, to evaluate whether uncertainty estimates provide reliable rankings for keypoint selection and pruning.
    Distributional calibration is assessed through coverage, variance calibration, and \acl{qq} diagnostics.
    The protocol is applicable to any method producing bivariate keypoint distributions (\cref{sec:uq_eval}).
    \item \textbf{Cross-architecture validation and baseline comparison on COCO.}
    We evaluate on COCO across two architecturally distinct YOLO-Pose families: YOLO-NAS-Pose~\citep{supergradients} (one-to-many label assignment) and the recent YOLO26-Pose~\citep{jocher2026ultralyticsyolo26} (one-to-one label assignment), confirming generalization across YOLO-Pose training paradigms.
    Compared to the built-in \ac{rle} in YOLO26-Pose, our approach captures input-dependent uncertainty substantially better and achieves stronger calibration and reliability ranking (\cref{sec:coco}).
    \item \textbf{Application to vision-based aircraft landing.}
    We further apply the probabilistic extension to vision-based aircraft landing (\cref{fig:intro}), a safety-critical setting where predictive keypoint distributions are used for uncertainty-aware \acl{pnp} estimation and downstream sensor fusion (\cref{sec:app_vbl}).
\end{enumerate}

\section{Related Work}
\label{sec:rel_work}

\paragraph{Uncertainty Quantification Within the YOLO Family.}
Uncertainty quantification for YOLO-based models has mainly focused on object detection by augmenting bounding-box predictions with localization uncertainty.
\citet{kraus2019uq-boundingbox} develop a probabilistic extension of the YOLOv3 object detector~\citep{redmon2018yolov3} that predicts variance for bounding-box position and size and captures epistemic uncertainty with Monte Carlo dropout~\citep{gal2016mc-dropout}.
They show that lower uncertainty correlates with higher overlap between predicted and ground-truth bounding boxes.
However, their uncertainty estimates are not calibrated across heads and prior boxes, which limits the use of global calibration metrics.
Their approach modifies and retrains the detector, whereas our method is post hoc and leaves the base detector's mean predictions unchanged.
\citet{choi2019gaussianyolo} focus on localization uncertainty by predicting variance for all bounding-box parameters and training the detector with a scale-weighted Gaussian negative log-likelihood loss.
They demonstrate that incorporating predicted localization uncertainty into detection confidence can improve the model's \ac{ap}.
\citet{he2019kl-loss} model bounding-box regression probabilistically and use a Kullback--Leibler loss that allows ambiguous boxes to be assigned larger variance and therefore lower effective regression penalty.
Overall, these approaches operate at the bounding-box level and modify the detector by jointly training localization and uncertainty estimates.
Such modifications are undesirable when detectors are already trained, tuned, or deployment-constrained, since changing their predictions or training objective can require renewed validation and may compromise the established operating point.

Beyond object detection, likelihood-based objectives have also been explored for keypoint localization.
\citet{li2021keypoint-normalizing-flows} propose \ac{rle}, which pairs per-keypoint diagonal standard deviations with a normalizing flow shared across all keypoints.
The flow models the two-dimensional residual density and can capture $x$--$y$ correlations.
Since a single flow is shared across all keypoints, this correlation structure is identical for every keypoint.
Per-keypoint heteroscedasticity enters only through the predicted diagonal scale.
As originally proposed, the density serves as an internal training signal that improves localization accuracy but is neither calibrated nor exposed as an uncertainty estimate at inference time.
To the best of our knowledge, the \ac{rle} approach integrated into YOLO26-Pose is the only keypoint-level mechanism in a YOLO-Pose model that yields a predictive distribution over keypoint locations, making it the sole available benchmark for our method (\cref{sec:coco}).
By contrast, we provide calibrated, explicit per-keypoint covariance matrices suitable for downstream uncertainty propagation, weighting, and sensor fusion (\cref{subsec:iw-nll,subsec:calibration}).
Because our method is post hoc, it requires no retraining of the base model and leaves its keypoint predictions unchanged.

\paragraph{Calibration Evaluation for Probabilistic Regression Models.}
While a large body of work focuses on probabilistic classification models and their calibration~\citep{guo2017cimnns}, the literature on probabilistic regression models is comparatively sparse.
For classification models, the \ac{ece} measures the discrepancy between predicted confidence and the observed frequency of correctness~\citep{guo2017cimnns}.
For object detection, \citet{Kuppers2020mccod} propose a detection-specific extension of \ac{ece} that conditions confidence calibration on detection attributes such as location and scale.
However, their metric calibrates scalar confidence with respect to detection correctness rather than a continuous predictive distribution over localization parameters.
Extending calibration to probabilistic regression, \citet{kuleshov2018accurate} propose \emph{confidence calibration}, which assesses whether nominal predictive quantiles agree with empirical coverage levels.
\citet{song2019distribution-calibration} show that quantile-based calibration alone does not necessarily ensure that the full predictive distribution is calibrated and propose distribution calibration as a stronger criterion.
\citet{levi2022ence} propose the \ac{ence} as a regression analogue of the \ac{ece}, comparing predicted and empirical error magnitudes within uncertainty bins.
These works, however, primarily focus on scalar confidence calibration, univariate regression, or generic regression settings, whereas keypoint detection additionally requires evaluating the joint calibration of two-dimensional localization distributions across multiple keypoints and detections.

The lack of widely adopted calibration protocols has also shaped prior evaluations of localization uncertainty.
\citet{kraus2019uq-boundingbox} evaluate their probabilistic YOLOv3 model's bounding-box uncertainty estimates primarily qualitatively, since the uncertainties are not calibrated across different detection heads.
\citet{choi2019gaussianyolo} evaluate the uncertainty estimates of their object detector by plotting the average uncertainty in the bounding-box parameters against the \ac{iou}.
This indicates whether uncertainty estimates reflect localization error, but not whether they are calibrated as predictive distributions.
The authors also modify the confidence of each prediction by a heuristic quantity derived from the model's uncertainty and evaluate the resulting \ac{ap}.
By contrast, in this work, we evaluate full bivariate predictive distributions over keypoint localization errors, rather than only localization accuracy.
Our evaluation protocol combines complementary diagnostics for marginal and joint calibration~(\cref{subsec:calib_metrics}) with \acf{akp}~(\cref{subsec:akp}), a keypoint-level metric that evaluates whether the estimated uncertainty reflects the reliability of keypoint predictions.
Our keypoint pruning method~(\cref{subsec:kp_pruning}) provides a principled alternative to heuristic confidence modifications~\citep{choi2019gaussianyolo} that operates at the keypoint level rather than the instance level, without changing the post-processing pipeline.

\section{Post-Hoc Probabilistic Extension for YOLO-Pose Models}
\label{sec:prob_extension}
\begin{figure}[t]
    \centering
    \includegraphics[width=0.99\linewidth]{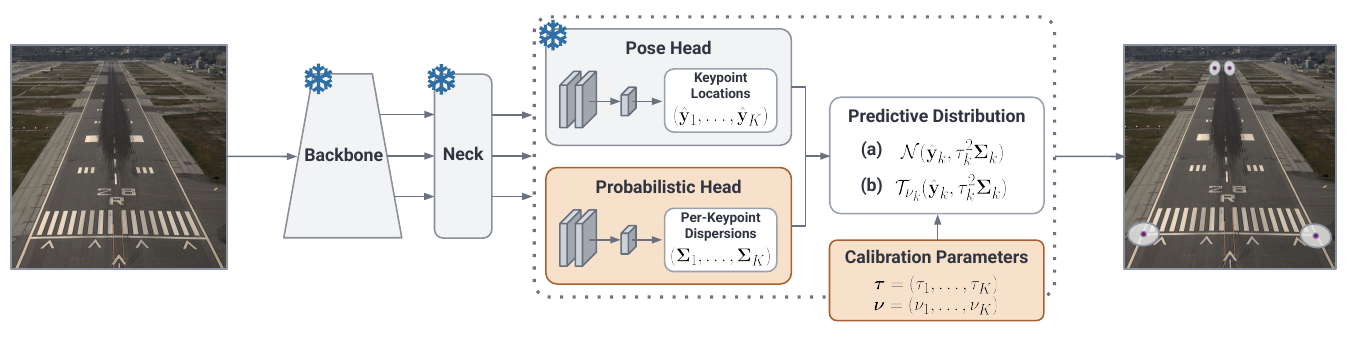}
    \caption{Post-hoc probabilistic extension for YOLO-Pose models. For clarity, only one set of~keypoint predictions is shown, and other task heads are omitted. We freeze (\raisebox{-0.45ex}{{\Large\ding{100}}}) a trained YOLO-Pose model and augment each predicted keypoint $\hat{\mathbf{y}}$, used as mean location, with an input-dependent $2\times2$ dispersion matrix~$\mathbf{\Sigma}$ calibrated by a temperature $\tau$ and, optionally, \mbox{Student-$t$} degrees of freedom~$\nu$. This yields either \textbf{(a)} a Gaussian predictive distribution $\mathcal{N}(\hat{\mathbf{y}}, \tau^2\mathbf{\Sigma})$ for broad downstream compatibility or \textbf{(b)} a heavy-tailed~\mbox{Student-$t$} distribution $\mathcal{T}_\nu(\hat{\mathbf{y}}, \tau^2\mathbf{\Sigma})$ for distributional fidelity to the empirical residuals. In the full~model, such candidate pose predictions are produced at every spatial location on the neck's multi-resolution feature~maps.}
    \label{fig:main}
\end{figure}

We assume access to a trained YOLO-Pose model that follows the standard single-stage YOLO design with a backbone--neck--head architecture and predicts keypoint locations, object bounding boxes, and instance-level confidence scores.
For keypoint regression, the model is trained using the \acf{oks} loss~\citep{maji2022yolopose}.
Representative examples include SuperGradients' \emph{YOLO-NAS-Pose}~\citep{supergradients} and Ultralytics' \emph{YOLO-Pose}~\citep{ultralytics}.
Starting from this baseline, we freeze the original YOLO-Pose model so that its predicted keypoint locations, bounding boxes, and confidence scores remain unchanged, and attach a lightweight post-hoc probabilistic extension, visualized in \cref{fig:main}. It augments each predicted keypoint $\hat{\mathbf{y}}$, used as mean location, with a heteroscedastic uncertainty estimate in the form of an input-dependent dispersion matrix~$\mathbf{\Sigma}$.

\paragraph{Architecture Extension.} We propose additional probabilistic heads (see \cref{fig:main}) that mirror the architecture of the original pose heads and reuse the shared multi-resolution neck features to predict a heteroscedastic $2\times 2$ dispersion matrix for each predicted keypoint.
The predicted dispersion matrices are index-aligned with the keypoint predictions, ensuring a one-to-one correspondence at each resolution and spatial location.
Positive definiteness is enforced via a Cholesky parameterization of the predicted dispersion matrices.
We ablate this design against alternative architectures and simpler isotropic and diagonal parameterizations~in~\cref{sec:coco}.

\paragraph{Training \& Calibration.}
In \cref{subsec:iw-nll}, we introduce an importance-weighted \ac{nll} for training the probabilistic heads; specifically, we justify using a Gaussian likelihood to learn the conditional dispersion structure.
In \cref{subsec:calibration}, we propose Gaussian calibration for broad downstream compatibility with covariance-based methods and Student-$t$ calibration for distributional fidelity to the empirical localization errors.
Both calibration schemes share the same frozen YOLO-Pose model and trained dispersion model.
After calibration, the predicted dispersion matrix~$\mathbf{\Sigma}$ represents a covariance matrix under the Gaussian model and a scale matrix under the Student-$t$ model.

\paragraph{Inference.}
At inference time, the post-processing pipeline closely follows the original YOLO design. Our key refinement at this stage is an additional pruning mechanism for uncertain keypoints, described in \cref{subsec:kp_pruning}.

\subsection{Importance-Weighted NLL for Learning Keypoint Uncertainty}
\label{subsec:iw-nll}
We aim to extend a trained YOLO-Pose model post-hoc with heteroscedastic uncertainty estimates while keeping the YOLO-Pose model frozen and preserving its original predictions.
In deep learning, predictive uncertainty is commonly decomposed into an aleatoric and an epistemic component~\citep{kendall2017uncertainty-cv}. 
In keypoint detection, aleatoric uncertainty primarily arises from irreducible ambiguity in the target annotations, for example due to occlusions, truncation, or subjective labeling. 
Epistemic uncertainty, by contrast, reflects uncertainty about the predictor itself, e.g., due to incomplete coverage of the deployment domain or suboptimal training. 
We do not attempt to disentangle these sources.
Methods like Monte~Carlo dropout~\citep{gal2016mc-dropout} or deep ensembles~\citep{lakshminarayanan2017simple} isolate epistemic uncertainty via model disagreement and target out-of-distribution detection rather than residual-distribution approximation.
Instead, we learn a model of the conditional residual distribution around the keypoint predictions of the frozen YOLO-Pose model, with parameters estimated from training-data residuals and therefore specific to the in-distribution regime defined by the training data.
This modeled distribution captures annotation ambiguity, while also reflecting systematic, input-dependent residual structure arising from the imperfect approximation and optimization of the frozen YOLO-Pose model~\citep[cf.][]{jimenez2026epistemic}.

\subsubsection{A Probabilistic Surrogate for the OKS Loss} 
We learn a parametric approximation to the conditional residual distribution around the frozen YOLO-Pose model's keypoint predictions.
This compact distributional representation can be learned from finite data and yields input-dependent uncertainty estimates that can be used directly in downstream processing.
The observed residuals reflect the data, annotations, and limitations of the learned model, but their distribution is substantially shaped by the keypoint regression objective~\citep[cf.][]{hooker2012learned}.
The objective determines how strongly residuals across keypoints, object scales, and error magnitudes influence training and therefore which errors are corrected or allowed to persist~\citep{huber2009robustbook,barron2019robustloss}. 
In YOLO-Pose models, the \ac{oks} loss~\citep{maji2022yolopose} has become a common default~\citep[cf.][]{supergradients, ultralytics}. 
For a single prediction--ground-truth instance pair indexed by~$i$, the \ac{oks} is given by
\begin{equation}
\label{eq:oks}
\mathrm{OKS}
\;=\; 
\frac{1}{\lvert\mathcal{K}_i\rvert}
\sum_{k\in\mathcal{K}_i}
\exp\!\left(
    -\frac{
        \mathbf{r}_{i,k}^\top\mathbf{r}_{i,k}
    }{
        2\,s_i^2\,\kappa_k^2
    }
\right),
\end{equation}
\begin{equation}
\label{eq:oks_loss}
\mathcal{L}_{\mathrm{OKS}}
\;=\; 
1 - \mathrm{OKS},
\end{equation}
where $\mathbf{r}_{i,k} \triangleq \mathbf{y}_{i,k} - \hat{\mathbf{y}}_{i,k} \in\mathbb{R}^2$ is the residual between the ground-truth and predicted keypoint locations, $s_i^2$~corresponds to the object area, and $\kappa_k$ is a keypoint-specific hyperparameter that accounts for the label noise.
Here, ${\mathcal{K}_i=\big\{k\in\{1, \dots, K\}\mid v_{i,k} > 0\big\}}$ denotes the index set of labeled keypoints, where $K$ is the number of keypoints per object and $v_{i,k}$ is a visibility flag to exclude unlabeled keypoints.

The \ac{oks} in \cref{eq:oks} can be interpreted as a Gaussian kernel in the keypoint residual, i.e.,
\begin{equation}
\exp\!\left(-\frac{\mathbf{r}_{i,k}^\top\mathbf{r}_{i,k}}{2\,s_i^2\,\kappa_k^2}\right)
\:\propto\:
\mathcal{N}\!\left(\mathbf{y}_{i,k};\hat{\mathbf{y}}_{i,k},\,s_i^2\kappa_k^2\mathbf{I}_2\right), 
\end{equation}
and thus induces a scale-normalized squared-distance geometry. This shared quadratic form motivates a Gaussian likelihood as a surrogate for learning the conditional dispersion structure of the residuals.
In \cref{subsec:calibration}, we revisit this modeling choice and derive how to calibrate the resulting predictive distribution in order to address distributional misspecification beyond second moments, particularly tail heaviness.

Although the kernel above is isotropic, the induced keypoint residuals can be anisotropic and exhibit cross-axis correlation (e.g., due to structured annotation noise), as we verify experimentally in \cref{subsec:coco_ablation}, \cref{tab:coco_main_ablation_results}.
We therefore model the uncertainty for each keypoint using an input-dependent full covariance $\mathbf{\Sigma}_{i,k}\in\mathbb{S}^2_{++}$, where $\mathbb{S}^2_{++}$ denotes the set of symmetric positive definite $2\times 2$ matrices, with isotropy as a special case.
Accordingly, we adopt a bivariate \ac{gnll} per keypoint as a principled probabilistic surrogate for the \ac{oks} loss:
\begin{equation}
\label{eq:gnll}
\mathcal{L}_{\mathrm{GNLL}}(\mathbf{\Sigma}_{i,k}; \mathbf{r}_{i,k})
\;=\;
\frac{1}{2}\,\mathbf{r}_{i,k}^{\top}\,\mathbf{\Sigma}_{i,k}^{-1}\,\mathbf{r}_{i,k}
+
\frac{1}{2}\ln\lvert\mathbf{\Sigma}_{i,k}\rvert
+ 
\ln Z_\mathcal{N},
\end{equation}
where we fix the mean to the frozen YOLO keypoint predictions, i.e., ${\boldsymbol{\mu}_{i,k}\triangleq\hat{\mathbf{y}}_{i,k}}$, and $\ln Z_{\mathcal{N}}=\ln(2\pi)$ denotes the corresponding normalization constant.

\subsubsection{Importance Weighting for Alignment With Deployment Conditions}
\label{subsubsec:iw}
The residuals $\mathbf{r}_{i,k}$ required for minimizing the \ac{gnll} in~\cref{eq:gnll} are obtained by matching predictions of the frozen YOLO-Pose model to ground-truth instances.
To this end, we reuse the training-time label assignment of the underlying YOLO framework to align the probabilistic heads with the frozen YOLO pose heads, while preserving model-specific assignment semantics and providing a rich training signal.
This applies to both the standard one-to-many label assignment~\citep[cf.][]{feng2021tood} used in most frameworks, including SuperGradients, and the more recent one-to-one label assignment~\citep{wang2024yolov10realtimeendtoendobject}, as used in Ultralytics' implementation of \emph{YOLO26}~\citep{sapkota2026yolo26keyarchitecturalenhancements}.
Each matched prediction--ground-truth instance pair defines a set $\mathcal{M}_i$ of associated quantities, comprising the YOLO-Pose predictions, the matched ground-truth annotations, and any derived quantities required by the loss.
The training-time label assignment therefore induces an empirical distribution $p_\mathrm{train}(\mathcal{M})$ over such matched-pair sets.

At inference time, label assignment is no longer used; instead, the post-processing pipeline retains predictions by confidence-score thresholding. Models with one-to-many label assignment additionally require \ac{nms}, whereas models with one-to-one assignment, such as YOLO26, are \ac{nms}-free by design.
Post-processing therefore determines which predictions are retained and shapes the residual distribution observed after evaluation-time matching.
We denote the corresponding empirical distribution over evaluation-time matched prediction--ground-truth instance pairs by $p_\mathrm{inference}(\mathcal{M})$.
To obtain reliable uncertainty estimates under the intended deployment conditions, we define the target distribution
\begin{equation}
\label{eq:p_target}
p_\mathrm{target}(\mathcal{M})
\;\propto\;
p_\mathrm{inference}(\mathcal{M})\cdot\pi(\mathcal{M}),
\qquad\text{with}\qquad
\pi(\mathcal{M})\;\geq\;0,
\end{equation}
where $\pi(\mathcal{M})$ encodes optional deployment-relevant importance of a matched-pair set $\mathcal{M}$ in the target distribution.
For example, this importance may depend on object scale, occlusion level, or image quality.
In this work, we choose $\pi(\mathcal{M})$ to target balanced calibration across object scales, a reported issue in modern detection models~\citep{Kuppers2020mccod,gilg2021bsb} and a critical requirement for vision-based aircraft landing, where the image-plane runway scale varies substantially during approach, as discussed in~\cref{sec:app_vbl}.

Our experiments suggest a substantial mismatch, $p_\mathrm{train}(\mathcal{M})\neq p_\mathrm{target}(\mathcal{M})$.
To compensate for this mismatch and minimize the \ac{gnll} under the target distribution, we employ \ac{iw}:
\begin{equation}
\label{eq:importance_weighting}
\mathbb{E}_{p_\mathrm{target}(\mathcal{M})}
\big[
\mathcal{L}_{\mathrm{GNLL}}(\mathbf{\Sigma}; \mathcal{M})
\big]
\;\propto\;
\mathbb{E}_{p_\mathrm{train}(\mathcal{M})}
\big[
w(\mathcal{M})\cdot
\mathcal{L}_{\mathrm{GNLL}}(\mathbf{\Sigma}; \mathcal{M})
\big].
\end{equation}
We define the importance weights up to normalization as
\begin{equation}
\label{eq:importance_weight}
w(\mathcal{M})
\;\propto\;
\frac{p_\mathrm{target}(\mathcal{M})}{p_\mathrm{train}(\mathcal{M})}
\;\propto\;
\frac{p_\mathrm{inference}(\mathcal{M})}{p_\mathrm{train}(\mathcal{M})}
\cdot
\pi(\mathcal{M}),
\qquad\text{with}\qquad
w(\mathcal{M})
\;\triangleq\;
w_{\mathrm{ret}}(\mathcal{M})\cdot w_{\mathrm{scale}}(\mathcal{M}),
\end{equation}
where the retention weight $w_{\mathrm{ret}}(\mathcal{M})$ approximates the density ratio $\tfrac{p_\mathrm{inference}(\mathcal{M})}{p_\mathrm{train}(\mathcal{M})}$, and $w_{\mathrm{scale}}(\mathcal{M})$ implements $\pi(\mathcal{M})$ through reweighting by object scale.
In this sense, our approach combines covariate-shift importance weighting and
group reweighting~\citep{SHIMODAIRA2000227,JMLRsugiyama07a,sagawa2020distributionallyrobustneuralnetworks}.

\paragraph{Retention and Object-Scale Importance Weights.} 
An assigned prediction should contribute more strongly when it is likely to be retained after inference-time post-processing.
To approximate this retention likelihood, we propose a retention weight for each prediction--ground-truth instance pair matched by the training-time label assignment:
\begin{equation}
\label{eq:w_r}
w_{\mathrm{ret}}(\mathcal{M}_i)
\;\triangleq\;
c_{\mathrm{det}, i}^{\alpha},
\end{equation}
where $c_{\mathrm{det}, i}$ is YOLO's instance-level confidence score used to rank and select detections at inference time.
Analogous to the well-established anchor alignment metric of~\citet{feng2021tood}, the hyperparameter $\alpha$ controls the strength of confidence-based reweighting in the training objective.

Consistent with prior work by~\citet{Kuppers2020mccod, gilg2021bsb}, we observe that smaller objects exhibit noticeably worse uncertainty calibration than larger ones, see \cref{subsec:coco_ablation}, \cref{fig:coco_ablation_area_weight}.
We hypothesize that larger instances have a greater impact on optimization, as detailed in Apps.~\ref{app:iw_object_scale_balance_1} and \ref{app:iw_object_scale_balance_2}.
To counteract this, we introduce a second multiplicative weight based on the object area of the assigned ground-truth instance, motivated by a log-uniform weighting profile over object area (App.~\ref{app:iw_object_scale_balance_3}):
\begin{equation}
\label{eq:w_s}
w_{\mathrm{scale}}(\mathcal{M}_i) \;\triangleq\; \frac{1}{s^2_i}.
\end{equation}

Both $w_{\mathrm{ret}}$ and $w_{\mathrm{scale}}$ are heuristic reweighting terms and are not guaranteed to yield an unbiased estimate of $\tfrac{p_\mathrm{target}(\mathcal{M})}{p_\mathrm{train}(\mathcal{M})}$.
Our ablation in \cref{subsec:coco_ablation} (\cref{tab:coco_main_ablation_results}) confirms a substantial mismatch ${p_\mathrm{train}(\mathcal{M})\neq p_\mathrm{target}(\mathcal{M})}$ under both one-to-many and one-to-one label assignment: omitting the importance weights degrades calibration considerably.
Both $w_{\mathrm{ret}}$ and $w_{\mathrm{scale}}$ improve calibration, with $w_{\mathrm{ret}}$ having the greater impact overall, while $w_{\mathrm{scale}}$ reduces the calibration gap between small and large instances.

\paragraph{Importance-Weighted Training Objective.} 
Our post-hoc probabilistic extension, shown in~\cref{fig:main}, introduces probabilistic heads that predict 
${\mathbf{\Sigma}_{i,k} = \mathbf{\Sigma}_{k}(\mathbf{f}_i; \boldsymbol{\theta}_{\Sigma})}$ 
from the neck features~$\mathbf{f}_i$ of the underlying YOLO model.
To learn their parameters $\boldsymbol{\theta}_{\Sigma}$, we approximate
the objective in \cref{eq:importance_weighting} using a mini-batch of matched-pair sets obtained from training-time label assignment on training data.
We denote this mini-batch by the indexed family $(\mathcal{M}_i)_{i\in\mathcal{B}}$, where $\mathcal{B}$ is the corresponding mini-batch index set:
\begin{equation}
\label{eq:iw_gnll}
\mathcal{L}_{\mathrm{IW\text{-}GNLL}}^{\mathcal{B}}
\;=\;
\frac{1}{\sum\limits_{i\in\mathcal{B}} w_i \, \lvert\mathcal{K}_i\rvert}\,
\sum_{i\in\mathcal{B}} w_i
\left[
    \sum_{k\in\mathcal{K}_i} 
    \mathcal{L}_{\mathrm{GNLL}}\big(\mathbf{\Sigma}_{k}(\mathbf{f}_i; \boldsymbol{\theta}_{\Sigma}); \mathbf{r}_{i,k}\big)
\right].
\end{equation}
We minimize $\mathcal{L}_{\mathrm{IW\text{-}GNLL}}^{\mathcal{B}}$ with respect to $\boldsymbol{\theta}_{\Sigma}$ while keeping $\mathbf{f}_i$ and $\mathbf{r}_{i,k}$ fixed, i.e., by freezing the underlying YOLO model.
Here, $w_i\triangleq\tfrac{c_{\mathrm{det}, i}^{\alpha}}{s^2_i}$ denotes the assignment-based importance weight (see \cref{eq:w_r,eq:w_s}) and $\mathcal{K}_i$ the corresponding index set of labeled keypoints.
The normalization by $\sum_{i\in\mathcal{B}} w_i\, \lvert\mathcal{K}_i\rvert$ yields a self-normalized mini-batch estimate, stabilizing the loss scale across batches.

\subsection{Calibrating the Predictive Distribution: Dispersion and Tail Behavior}
\label{subsec:calibration}
While \cref{eq:iw_gnll} provides a principled objective for learning the conditional dispersion structure, the predictions $\mathbf{\Sigma}_{i,k}$ are not necessarily well calibrated on deployment data.
In practice, uncertainty estimates can become overconfident at deployment: residuals observed during training are typically smaller than those encountered under real-world conditions, causing $\mathbf{\Sigma}_{i,k}$ to underestimate inference-time errors, as we show in \cref{subsec:coco_core}, \cref{fig:coco_nano_calibration_diagnostics}.
Moreover, the residual distribution may exhibit distributional misspecification beyond second moments: extreme errors can occur more frequently than implied by Gaussian tails.
As we detail in \cref{subsubsec:heavy_tails_by_design}, this tail behavior is consistent with the robust, outlier-suppressing structure of the \ac{oks}~loss.

We therefore propose two complementary baseline calibration methods, depending on downstream distributional assumptions, as shown in \cref{fig:main}.
When downstream pipelines require a Gaussian likelihood, we apply \emph{Gaussian temperature scaling} to calibrate dispersion while preserving downstream compatibility.
When this constraint is absent, we instead perform \emph{Student-$t$ dispersion--tail scaling} to preserve distributional fidelity to the empirical residuals by jointly calibrating dispersion and tail behavior.
Importantly, both calibration schemes share the same frozen YOLO-Pose model and trained dispersion model; inference-time switching only requires selecting the corresponding stored calibration parameters.

\subsubsection{Gaussian Calibration for Broad Downstream Compatibility}
Gaussian likelihoods are a convenient default for downstream applications due to their broad methodological compatibility.
To obtain reliable uncertainty estimates at deployment, we calibrate the trained heteroscedastic dispersion model on a held-out calibration set that reflects the intended operating conditions.
A systematic comparison of calibration methods is beyond the scope of this work; we employ \emph{Gaussian temperature scaling}~\citep{guo2017cimnns, levi2022ence} as a baseline that preserves the heteroscedastic structure learned in \cref{subsec:iw-nll}:
\begin{equation}
\label{eq:gauss_temp_scaling}
\tilde{\mathbf{\Sigma}}_{i,k}(\tau_k) 
\;\triangleq\;
\tau_k^2\,\mathbf{\Sigma}_{i,k},
\qquad \text{with} \qquad 
\tau_k\;>\;0.
\end{equation}
For calibration, we reuse the training objective in~\cref{eq:iw_gnll} to learn a global vector of keypoint-specific temperatures, ${\boldsymbol{\tau}=(\tau_1,\dots,\tau_K)}$.
Applying the training-time label-assignment procedure to the held-out calibration data yields an indexed family of matched-pair sets $(\mathcal{M}_i)_{i\in\mathcal{C}}$, where $\mathcal{C}$ denotes the corresponding calibration index set.
During calibration, we keep $\mathbf{r}_{i,k}$ and $\mathbf{\Sigma}_{i,k}$ fixed, i.e., freeze both the underlying YOLO model and our probabilistic heads, and optimize only $\boldsymbol{\tau}$ subject to $\boldsymbol{\tau}\in\mathbb{R}^K_{>0}$ by minimizing
\begin{equation}
\label{eq:calib_iw_gnll}
\mathcal{L}_{\mathrm{IW\text{-}GNLL}}^{\mathcal{C}}
\;=\;
\frac{1}{\sum\limits_{i\in\mathcal{C}} w_i \, \lvert\mathcal{K}_i\rvert}\,
\sum_{i\in\mathcal{C}} w_i
\left[
    \sum_{k\in\mathcal{K}_i} 
    \mathcal{L}_{\mathrm{GNLL}}\!\left(\tilde{\mathbf{\Sigma}}_{i,k}(\tau_k); \mathbf{r}_{i,k}\right)
\right].
\end{equation}

Empirically, Gaussian temperature scaling provides a useful yet limited calibration baseline because tail mismatch with the residual distribution remains; see \cref{subsec:coco_core}, \cref{tab:coco_main_results}.
In App.~\ref{app:residual_finetune}, we propose \emph{residual distribution fine-tuning} as a complementary upstream step to improve calibration quality under the Gaussian model.

\subsubsection{OKS Loss: Heavy Tails by Design}
\label{subsubsec:heavy_tails_by_design}
The limitations of the Gaussian model can be understood from the geometry of the \ac{oks} loss and its suppression of large residuals.
In particular, the \ac{oks} loss in \cref{eq:oks_loss} is not a \ac{gnll}: it maximizes a Gaussian kernel similarity rather than minimizing the corresponding negative log-likelihood.

To contrast the two objectives, we adopt the perspective of robust M-estimation and consider the $\psi$-function, ${\psi(\mathbf{r})\triangleq\nabla_{\mathbf{r}}\,\mathcal{L}(\mathbf{r})}$, which quantifies the influence of a single residual $\mathbf{r}$ on the optimization update~\citep{huber2009robustbook,barron2019robustloss}. Differentiating the \ac{oks} loss in \cref{eq:oks_loss} yields
\begin{equation}
\label{eq:grad_oks}
\psi_{\mathrm{OKS}}(\mathbf{r})
\;=\;
\nabla_{\mathbf r}\,\mathcal{L}_{\mathrm{OKS}}
\;\propto\;
\exp\!\left(-\tfrac{\mathbf{r}^\top\mathbf{r}}{2\,s^2\,\kappa^2}\right)\mathbf{r},
\end{equation}
up to constant factors.
In M-estimation terms, $\psi_{\mathrm{OKS}}(\mathbf{r})$ is \emph{redescending}: for $\|\mathbf{r}\|_2\to\infty$ its magnitude is exponentially suppressed, such that $\|\psi_{\mathrm{OKS}}(\mathbf{r})\|_2\to 0$ and gross outliers exert vanishing influence during training.
By contrast, under the \ac{gnll} in \cref{eq:gnll}, the per-residual influence is
\begin{equation}
\label{eq:grad_gnll}
\psi_{\mathrm{GNLL}}(\mathbf{r})
\;=\;
\nabla_{\mathbf{r}}\,\mathcal{L}_{\mathrm{GNLL}}
\;=\;
\mathbf{\Sigma}^{-1}\mathbf{r}.
\end{equation}
In the heteroscedastic setting, predicting a larger $\mathbf{\Sigma}$ can reduce $\|\psi_{\mathrm{GNLL}}(\mathbf{r})\|_2$, but this is counterbalanced by the $\ln|\mathbf{\Sigma}|$ term in \cref{eq:gnll} and thus cannot reproduce the exponential suppression in $\psi_{\mathrm{OKS}}(\mathbf{r})$.
Consequently, $\psi_{\mathrm{GNLL}}(\mathbf{r})$ is \emph{non-redescending}: 
for $\|\mathbf r\|_2\to\infty$, the influence diverges, $\|\psi_{\mathrm{GNLL}}(\mathbf{r})\|_2\to\infty$.

The \ac{oks} loss therefore induces a robust, outlier-suppressing training signal, reducing the incentive to correct large residuals and thereby making heavier-tailed residual behavior more plausible than under a \ac{gnll}.
More fundamentally, $\mathcal{L}_{\mathrm{OKS}}$ cannot even be interpreted as a valid \ac{nll}, as we formalize in \cref{lem:non-normalizable}.
\begin{lemma}[Non-Normalizability]
\label{lem:non-normalizable}
Since $\mathcal{L}_{\mathrm{OKS}}\in[0,1)$ is bounded, it holds that $\exp(-\mathcal{L}_{\mathrm{OKS}})\ge e^{-1}$ for all $\mathbf{r}$.
For any $R>0$, let $B_R=\left\{\mathbf r\in\mathbb R^2\,\middle|\,\|\mathbf r\|_2\le R\right\}$. Then
\[
\int_{\mathbb R^2}\!\exp(-\mathcal{L}_{\mathrm{OKS}})\,d\mathbf r
\;\ge\;
\int_{B_R}\!e^{-1}\,d\mathbf r
\;=\;
e^{-1}\,\pi R^2
\;\xrightarrow{R\to\infty}\;\infty.
\]
Hence a putative density $p(\mathbf r)\propto \exp(-\mathcal{L}_{\mathrm{OKS}})$ is not normalizable; consequently, $\mathcal{L}_{\mathrm{OKS}}$ cannot be interpreted as a negative log-likelihood on $\mathbb{R}^2$ under the standard identification $p(\mathbf r)\propto \exp(-\mathcal{L}(\mathbf r))$.
\end{lemma}
To model \ac{oks}-induced residuals with a well-defined likelihood, an elliptically contoured heavy-tailed distribution is a natural choice. In particular, the Student\nobreakdash-$t$ likelihood provides a normalizable surrogate that more faithfully captures the observed tail behavior, while also admitting a principled covariance interpretation. We justify this modeling choice in App.~\ref{app:student-t_motivation} and empirically validate it in \cref{subsec:coco_core}, \cref{tab:coco_main_results}.

\subsubsection{Student-\texorpdfstring{$t$}{t} Calibration for Distributional Fidelity}
To account for the heavy-tailed residual distribution, we calibrate a global tail and scale adjustment under a Student-$t$ likelihood while retaining our Gaussian-trained heteroscedastic dispersion model.
The Student-$t$ distribution generalizes the Gaussian by introducing a degrees-of-freedom parameter $\nu\in\mathbb{R}_{>0}$, which controls tail heaviness, and a positive definite scale matrix $\mathbf{S}$.
For the bivariate case, the implied covariance is ${\mathrm{Cov}[\mathbf{r}]=\frac{\nu}{\nu-2}\,\mathbf{S}}$ for $\nu>2$, whereas it is not defined for $\nu\leq 2$. 
The corresponding \ac{nll} is
\begin{equation}
\label{eq:tnll}
\mathcal{L}_{t\text{-}\mathrm{NLL}}(\nu, \mathbf{S}; \mathbf{r})
\;=\;
\frac{\nu+2}{2}\ln\!\left(1+\frac{1}{\nu}\,\mathbf{r}^{\top}\mathbf{S}^{-1}\mathbf{r}\right)
+
\frac{1}{2}\ln\lvert\mathbf{S}\rvert
+
\ln Z_{t}(\nu),
\end{equation}
where $\ln Z_t(\nu)=\ln\Gamma\!\left(\frac{\nu}{2}\right)-\ln\Gamma\!\left(\frac{\nu+2}{2}\right)+\ln(\nu\pi)$ is the normalization term, with gamma function $\Gamma(\cdot)$.

The Student-$t$ likelihood admits an equivalent Gaussian scale-mixture representation: 
$\mathbf{r} \mid \lambda \sim \mathcal{N}(\mathbf{0},\,\nicefrac{\mathbf{S}}{\lambda})$ with
$\lambda \sim \operatorname{Gamma}(\nicefrac{\nu}{2},\nicefrac{\nu}{2})$ yields
$\mathbf{r} \sim \mathcal{T}_\nu(\mathbf{0},\mathbf{S})$~\citep{Kotz_Nadarajah_2004}.
Hence, the Student-$t$ distribution remains elliptically contoured: $\mathbf{S}$ sets orientation and anisotropy through the Mahalanobis term $\mathbf{r}^\top \mathbf{S}^{-1}\mathbf{r}$, while $\nu$ controls tail heaviness.
This justifies using our Gaussian-trained heteroscedastic dispersion model as an \emph{uncalibrated} Student-$t$ scale model.
As a baseline calibration method, we introduce \emph{Student-$t$ dispersion--tail scaling}:
\begin{equation}
\label{eq:t_temp_scaling_scale}
\tilde{\mathbf{S}}_{i,k}(\tau_k)
\;\triangleq\;
\tau_k^2\,\mathbf{\Sigma}_{i,k},
\qquad\text{with}\qquad
\tau_k \;>\; 0,
\end{equation}
\begin{equation}
\label{eq:t_temp_scaling_cov}
\tilde{\mathbf{\Sigma}}_{i,k}(\nu_k,\tau_k)
\;\triangleq\;
\frac{\nu_k}{\nu_k-2}\,\tilde{\mathbf{S}}_{i,k}(\tau_k),
\qquad\text{with}\qquad
\nu_k \;>\; 2,
\end{equation}
where $\nu_k>2$ ensures a well-defined calibrated covariance $\tilde{\mathbf{\Sigma}}_{i,k}$; in our experiments, all fitted $\nu_k$ satisfy this bound by a wide margin (see \cref{subsec:coco_core}).
Consequently, our approach is a generalization of Gaussian temperature scaling (cf.\ \cref{eq:gauss_temp_scaling}), which is recovered in the limit $\nu_k\to\infty$.

In order to learn the global degrees of freedom ${\boldsymbol{\nu}=(\nu_1,\dots,\nu_K)}$ and temperature ${\boldsymbol{\tau}=(\tau_1,\dots,\tau_K)}$, we reuse the importance-weighting approach from \cref{eq:iw_gnll} and minimize
\begin{equation}
\label{eq:calib_t_gnll}
\mathcal{L}_{\mathrm{IW}\text{-}t\text{-}\mathrm{NLL}}^{\mathcal{C}}
\;=\;
\frac{1}{\sum\limits_{i\in\mathcal{C}} w_i \, \lvert\mathcal{K}_i\rvert}\,
\sum_{i\in\mathcal{C}} w_i
\left[
    \sum_{k\in\mathcal{K}_i} 
    \mathcal{L}_{t\text{-}\mathrm{NLL}}\!\left(\nu_k,\tilde{\mathbf{S}}_{i,k}(\tau_k); \mathbf{r}_{i,k}\right)
\right]
\end{equation}
subject to $\boldsymbol{\nu}\in\mathbb{R}^K_{>2}$ and $\boldsymbol{\tau}\in\mathbb{R}^K_{>0}$ on a held-out calibration set, while keeping $\mathbf{r}_{i,k}$ and $\mathbf{\Sigma}_{i,k}$ fixed, i.e., by freezing both the underlying YOLO model and our probabilistic heads.

\subsection{Pruning Uncertain Keypoints}
\label{subsec:kp_pruning}
Keypoints are typically used as intermediate measurements in downstream pipelines rather than as an end product, and a single badly localized keypoint can dominate the estimate and cause severe downstream failures.
YOLO-Pose models already provide a rejection mechanism for occluded or truncated keypoints beyond reliable inference~\citep{maji2022yolopose}.
They predict a per-keypoint confidence score trained on visibility flags $v_{i,k}$, which is thresholded at inference time; we refer to this as \emph{visibility-score thresholding}.
Pruning keypoints that pass the visibility-score threshold but are likely to be poorly localized would substantially improve the robustness of downstream pipelines.

Our probabilistic extension provides a principled pruning signal: for each instance and keypoint, we obtain a calibrated heteroscedastic covariance $\tilde{\mathbf{\Sigma}}_{i,k}$ that quantifies input-dependent localization uncertainty. 
Concretely, in the Gaussian case, we set $\tilde{\mathbf{\Sigma}}_{i,k}=\tau_k^2\,\mathbf{\Sigma}_{i,k}$ (\cref{eq:gauss_temp_scaling}), whereas for Student-$t$ with $\nu_k>2$, we convert the scale matrix to a covariance,
$\tilde{\mathbf{\Sigma}}_{i,k}=\frac{\nu_k}{\nu_k-2}\,\tau_k^2\,\mathbf{\Sigma}_{i,k}$ (\cref{eq:t_temp_scaling_cov}).
We then prune keypoints with large $\tilde{\mathbf{\Sigma}}_{i,k}$, capturing anisotropy and keypoint-specific ambiguity while exposing a simple reliability--coverage trade-off.

\paragraph{Localization Confidence Score.}
We convert $\tilde{\mathbf{\Sigma}}_{i,k}$ into a scalar localization confidence score by combining standard covariance scalarizations with an \ac{oks}-motivated normalization:
\begin{equation}
\label{eq:kp_uq_scalar_conf}
c_{i,k}
\;\triangleq\;
-\frac{\phi\bigl(\tilde{\mathbf{\Sigma}}_{i,k}\bigr)}{s_i^{2}\,\kappa_k^{2}},
\qquad
\phi(\mathbf{\Sigma})
\in
\Bigl\{
\operatorname{tr}(\mathbf{\Sigma}),
\ \sqrt{\det(\mathbf{\Sigma})},
\ \lambda_{\max}(\mathbf{\Sigma})
\Bigr\}.
\end{equation}
By default, we use the total variance $\phi(\mathbf{\Sigma})=\operatorname{tr}(\mathbf{\Sigma})$, i.e., the sum of eigenvalues.
In \cref{subsec:coco_ablation}, \cref{tab:coco_scalarization_ablation}, we ablate two alternatives: the ellipse-area scale $\sqrt{\det(\mathbf{\Sigma})}$ and the worst-direction variance $\lambda_{\max}(\mathbf{\Sigma})$.
We emphasize that the best scalarization can be dataset- and application-dependent.

\paragraph{Pruning-Threshold Selection.}
We prune keypoints by thresholding $c_{i,k}$ (\cref{eq:kp_uq_scalar_conf}).
To select a pruning threshold, we apply evaluation-time matching (see~\cref{subsec:akp}) on a held-out dataset, such as the calibration set, yielding an indexed family of matched-pair sets $(\mathcal{M}_i)_{i\in\mathcal{H}}$, where $\mathcal{H}$ denotes the corresponding index set.
We then cast pruning as a classification problem by defining the binary label
\begin{equation}
\label{eq:kp_prune_binary_label}
z_{i,k} 
\;\triangleq\;
\mathbbm{1}\!\left[\mathrm{KS}_{i,k}\ge \gamma_{\scriptscriptstyle \mathrm{KS}}\right],
\qquad\text{with}\qquad
\mathrm{KS}_{i,k}
\;=\;
\exp\!\left(-\tfrac{\mathbf{r}_{i,k}^\top\mathbf{r}_{i,k}}{2\,s_i^2\,\kappa_k^2}\right),
\end{equation}
for labeled keypoints $k\in\mathcal{K}_i$, where $\gamma_{\scriptscriptstyle \mathrm{KS}}\in(0,1)$ denotes a prespecified \ac{ks} threshold.
We then select the pruning threshold $\gamma_c$, applied to localization confidence scores $c_{i,k}$, as
\begin{equation}
\label{eq:kp_prune_threshold}
\gamma_c
\;=\; 
\arg\max_{\gamma}\, J\left(\gamma;\left\{\bigl(c_{i,k}, z_{i,k}\bigr)\,\middle|\, i\in\mathcal{H},\; k\in\mathcal{K}_i\right\}\right),
\end{equation}
where $J(\gamma)=\mathrm{TPR}(\gamma)-\mathrm{FPR}(\gamma)$ is Youden's $J$ statistic, a standard threshold-selection criterion based on the \ac{roc} curve. 
In \cref{subsec:coco_ablation}, \cref{tab:coco_threshold_selection_ablation}, we ablate two alternative operating-point criteria: the minimum distance to the ideal \ac{roc} point $(0,1)$ and the maximum F1 score.

\paragraph{Inference-Time Pruning.}
The pruning threshold $\gamma_c$ is fixed after selection and used at inference time. After post-processing, we retain for each instance $i$,
\begin{equation}
\label{eq:kp_prune_rule}
\mathcal{K}_i^{\mathrm{keep}}
\;=\;
\left\{k\in\mathcal{K}_i^\mathrm{vis}\,\middle|\, c_{i,k}\geq \gamma_c\right\},
\end{equation}
where $\mathcal{K}_i^{\mathrm{vis}}$ denotes the set of keypoints after visibility-score thresholding.

\section{Evaluation Protocol for Keypoint Uncertainty Estimates}
\label{sec:uq_eval}
Keypoint detectors are typically evaluated as point estimators using OKS-based \acf{ap} and \acf{ar}, as standardized in the COCO evaluation framework~\citep{lin2014mscoco}, which prioritizes keypoint localization accuracy over uncertainty quantification. 
Consequently, there is no widely accepted protocol for assessing keypoint uncertainty, especially for models that parameterize heteroscedastic, anisotropic predictive distributions.

This gap matters in safety-critical pipelines, where uncertainty informs downstream decisions such as gating, tracking, and sensor fusion. Overconfident estimates risk propagating errors, whereas underconfident estimates can lead to unnecessary rejection or overly conservative behavior. Neither failure mode is revealed by \ac{ap}/\ac{ar} alone. 
To this end, we introduce a new COCO-style metric that evaluates keypoint-level reliability ranking (\cref{subsec:akp}), and propose a set of calibration diagnostics tailored to heteroscedastic keypoint regression under Gaussian and the more general Student-$t$ likelihoods (\cref{subsec:calib_metrics}).

\subsection{Average Keypoint Precision: A COCO-Style Evaluation Metric}
\label{subsec:akp}
The COCO evaluation framework provides no metric for assessing the quality of per-keypoint confidence scores. The reason is structural: \ac{ap}/\ac{ar} evaluate keypoint-detection performance at the instance level based on \ac{oks} (\cref{eq:oks}), which aggregates per-keypoint similarity over all labeled keypoints of a detected instance. Consequently, the COCO evaluation protocol has no notion of keypoint-level abstention. In particular, \ac{ap}/\ac{ar} cannot distinguish between a detector that always predicts all keypoints and one that correctly ranks poorly detected keypoints for pruning.

This is undesirable for downstream pipelines where fewer but more reliable keypoints are preferable to a complete but noisy pose estimate. To address this limitation, we introduce \acf{akp}, a COCO-style evaluation metric defined at keypoint level. \ac{akp} quantifies how well a confidence score ranks keypoints by correctness.

\paragraph{Evaluation-Time Matching.}
First, we follow the COCO protocol and apply greedy one-to-one matching between post-processed detections and ground-truth instances on the evaluation data.
This yields an indexed family of matched-pair sets $(\mathcal{M}_i)_{i\in\mathcal{E}}$, where $\mathcal{E}$ denotes the corresponding evaluation index set.
In contrast to \ac{ap}/\ac{ar}, we evaluate \ac{akp} only on true-positive detections by requiring the match \ac{oks} to exceed the lowest COCO threshold, i.e., $\mathrm{OKS}\geq 0.5$. This decouples keypoint-level reliability from instance-level mislocalization and thus complements \ac{ap}/\ac{ar}.

\paragraph{Keypoint-Level Binary Correctness.}
We then define binary correctness for labeled keypoints $k\in\mathcal{K}_i$ by reusing the \acl{ks} in \cref{eq:kp_prune_binary_label}:
\begin{equation}
z_{i,k}(\gamma_{\scriptscriptstyle \mathrm{KS}}) 
\;\triangleq\; 
\mathbbm{1}\!\left[\mathrm{KS}_{i,k}\ge \gamma_{\scriptscriptstyle \mathrm{KS}}\right].
\end{equation}

\paragraph{Average Keypoint Precision.}
Analogous to COCO's \ac{ap}, our proposed metric is based on the 101-point interpolated average precision~\citep{lin2014mscoco}, denoted by $\mathrm{AP}_{101}$.
We average $\mathrm{AP}_{101}$ over the standard COCO threshold set, applied here to keypoint similarity.
For each keypoint-similarity threshold $\gamma_{\scriptscriptstyle \mathrm{KS}}$, $\mathrm{AP}_{101}$ requires a confidence score $c_{i,k}$ for ranking.
In general, $c_{i,k}$ may be any keypoint-level confidence score.
In our setting, we use the keypoint-level localization confidence from \cref{eq:kp_uq_scalar_conf},
which is derived from the calibrated covariances~$\tilde{\mathbf{\Sigma}}_{i,k}$. 
Accordingly, we define
\begin{equation}
\label{eq:akp}
\mathrm{AKP}
\;=\;
\frac{1}{\lvert \mathcal{G} \rvert}
\sum_{\gamma_{\scriptscriptstyle \mathrm{KS}} \in \mathcal{G}}
\mathrm{AP}_{101}\!\left(
\left\{
\bigl(z_{i,k}(\gamma_{\scriptscriptstyle \mathrm{KS}}),\, c_{i,k}\bigr)
\;\middle|\;
i\in\mathcal{E}\,,\, k\in\mathcal{K}_i
\right\}
\right),
\end{equation}
where $\mathcal{G}=\{0.50,0.55,\ldots,0.95\}$ is the standard COCO threshold set.

\subsection{Calibration Diagnostics for Heteroscedastic Keypoint Regression}
\label{subsec:calib_metrics}
Calibrated uncertainty estimates are essential when keypoints are used in downstream tasks.
A comprehensive evaluation of uncertainty calibration requires multiple complementary diagnostics, particularly in the heteroscedastic regression setting induced by our probabilistic extension in \cref{sec:prob_extension}, where predictive uncertainty is input-dependent and may vary substantially across object instances as well as their scales and orientations.
Although \ac{nll} is a proper scoring rule and the standard training objective for probabilistic models, as a single scalar it conflates accuracy and uncertainty and can therefore obscure whether a model is over- or underconfident, especially under likelihood mismatch, where variance calibration and tail coverage may diverge.
We therefore introduce a suite of \emph{coverage}, \emph{\acf{qq}}, and \emph{variance-calibration} diagnostics, comprising both plots and quantitative metrics.

\paragraph{Evaluation-Time Matching.}
We use the same evaluation-time matching procedure as in \cref{subsec:akp} to obtain $\mathcal{E}$, thereby restricting calibration analysis to true-positive instances.
This isolates the calibration of keypoint-localization uncertainty from instance-level detection failures, for which no meaningful localization residual is defined.
Moreover, $\mathcal{E}$ may be restricted to subsets of interest, e.g., specific object-scale regimes, while $\mathcal{K}_i$ may be restricted to a single keypoint category to enable per-keypoint calibration analysis.

\paragraph{Coverage Diagnostics.}
Coverage diagnostics assess whether nominal confidence levels match empirical frequencies \citep{gneiting2007probabilistic,kuleshov2018accurate}.
A model is perfectly calibrated if the empirical coverage $\hat{C}(\alpha)$ at nominal coverage level $\alpha$ satisfies $\hat{C}(\alpha)=\alpha$ for all $\alpha\in(0,1)$.
For our bivariate Gaussian and Student-$t$ predictive distributions, we compute the empirical joint coverage as
\begin{equation}
\label{eq:gauss_coverage}
\hat{C}_\mathcal{N}(\alpha)
\;=\;
\frac{1}{\sum_{i\in\mathcal{E}}\, \lvert\mathcal{K}_i\rvert}
\sum_{i\in\mathcal{E},\, k\in\mathcal{K}_i}\!
\mathbbm{1}\!\left[
\mathbf{r}_{i,k}^\top \tilde{\mathbf{\Sigma}}_{i,k}^{-1}\mathbf{r}_{i,k} \leq \chi^2_{2,\alpha}
\right],
\end{equation}
\begin{equation}
\label{eq:t_coverage}
\hat{C}_\mathcal{T}(\alpha)
\;=\;
\frac{1}{\sum_{i\in\mathcal{E}}\, \lvert\mathcal{K}_i\rvert}
\sum_{i\in\mathcal{E},\, k\in\mathcal{K}_i}\!
\mathbbm{1}\!\left[
\mathbf{r}_{i,k}^\top \tilde{\mathbf{S}}_{i,k}^{-1}\mathbf{r}_{i,k} \leq 2\,F^{-1}_{2,\nu_k}(\alpha)
\right],
\end{equation}
where $\tilde{\mathbf{S}}_{i,k}$ is the calibrated scale matrix (\cref{eq:t_temp_scaling_scale}), and $\chi^2_{2,\alpha}$ and $F^{-1}_{2,\nu_k}(\alpha)$ denote the corresponding $\alpha$-quantiles of the $\chi^2$ and $F$ distributions with the appropriate degrees of freedom.
The corresponding definitions of marginal empirical coverage for each standardized coordinate are provided in App.~\ref{app:coverage}.
We visualize joint and marginal calibration using coverage plots over a grid of nominal coverage levels, e.g., $\mathcal{A}=\{0.01, 0.02, \dots, 0.99\}$, and summarize deviations using the \ac{ace},
\begin{equation}
\label{eq:ace}
\mathrm{ACE}
\;=\;
\frac{1}{\lvert \mathcal{A} \rvert}\sum_{\alpha\in\mathcal{A}}
\left|
\hat{C}(\alpha)-\alpha
\right|,
\end{equation}
where lower values indicate better calibration.
The \ac{ace} corresponds to a discrete approximation of the area between the observed coverage curve and the diagonal of perfect calibration.
Further details on coverage diagnostics are provided in App.~\ref{app:coverage}.

\paragraph{Q--Q Goodness-of-Fit Analysis.}
Complementary \ac{qq} analysis assesses whether the full distribution of normalized residual statistics agrees with the corresponding reference law, revealing shape and tail mismatches beyond confidence-region probabilities alone \citep{wilk1968probability}.
Concretely, the sorted Mahalanobis distances from \cref{eq:gauss_coverage,eq:t_coverage} are plotted against the quantiles of the reference $\chi^2_2$ or $F_{2,\nu_k}$ distribution, where systematic deviations from the identity line indicate distributional misfit.
We perform \ac{qq} analysis jointly and marginally; full details are provided in App.~\ref{app:qq}.

\paragraph{Variance-Calibration Diagnostics.}
Coverage and \ac{qq} diagnostics can indicate perfect calibration even when the predicted uncertainty is uninformative in the heteroscedastic sense \citep{levi2022ence}.
Variance-calibration diagnostics therefore complement coverage analysis by assessing whether the calibrated covariance predictions $\tilde{\mathbf{\Sigma}}_{i,k}$, obtained either from the Gaussian model (\cref{eq:gauss_temp_scaling}) or the Student-$t$ model (\cref{eq:t_temp_scaling_cov}), match empirical squared residuals across the range of predicted uncertainty levels.
For our multivariate setting, the global moment-matching condition is
\begin{equation}
\label{eq:moment_matching}
\mathbb{E}\!\left[\mathbf{r}\mathbf{r}^\top\right]
\;\approx\;
\mathbb{E}\!\left[\tilde{\mathbf{\Sigma}}\right].
\end{equation}

For a stratified evaluation across uncertainty levels, we sort predictions by their uncertainty magnitude and compare predicted variances and empirical squared residuals within bins.
To this end, we use the total variance~$\operatorname{tr}(\tilde{\mathbf{\Sigma}}_{i,k})$ as a scalar uncertainty score for joint variance calibration, analogous to the marginal definitions of \citet{levi2022ence}, which are detailed in App.~\ref{app:variance_calibration}.
This choice is natural because the trace is a rotation-invariant aggregate of the marginal variances and thus the multivariate counterpart of the scalar variance used by \citet{levi2022ence}.
Accordingly, we sort all evaluated keypoints by $\operatorname{tr}(\tilde{\mathbf{\Sigma}}_{i,k})$ and use~$(n)$ for the corresponding sorted index such that $\operatorname{tr}\!\left(\tilde{\mathbf{\Sigma}}_{(1)}\right)\leq\cdots\leq\operatorname{tr}\!\left(\tilde{\mathbf{\Sigma}}_{(N)}\right)$, where $N=\sum_{i\in\mathcal{E}} |\mathcal{K}_i|$.
We then partition the sorted samples into $B$ equally sized bins $\{\mathcal{B}_b\}_{b=1}^B$ and define the mean empirical squared residuals and mean predicted total variance in bin $\mathcal{B}_b$ as
\begin{equation}
\label{eq:joint_var_bins}
\hat{e}_b
\;=\;
\frac{1}{|\mathcal{B}_b|}
\sum_{n\in\mathcal{B}_b}
\frac{1}{2}\|\mathbf{r}_{(n)}\|_2^2,
\qquad
\hat{v}_b
\;=\;
\frac{1}{|\mathcal{B}_b|}
\sum_{n\in\mathcal{B}_b}
\frac{1}{2}\operatorname{tr}\!\left(\tilde{\mathbf{\Sigma}}_{(n)}\right),
\end{equation}
where the factor $\tfrac{1}{2}$ puts the joint quantity on the same scale as the marginal variance-calibration diagnostics.
We visualize variance calibration by plotting $\hat{e}_b$ against $\hat{v}_b$, where perfect calibration is reflected by points lying close to the diagonal $\hat{e}_b=\hat{v}_b$ for all $b\in\{1,2,\dots,B\}$.
For a scalar summary, we follow \citet{levi2022ence} and compute the \acf{ence} on the corresponding root-mean quantities,
\begin{equation}
\label{eq:joint_ence}
\mathrm{ENCE}
\;=\;
\frac{1}{B}\sum_{b=1}^B
\frac{|\sqrt{\hat{e}_b}-\sqrt{\hat{v}_b}|}{\sqrt{\hat{v}_b}}.
\end{equation}
Lower values indicate better agreement between predicted and empirical variance across uncertainty levels.
Further details on joint and marginal variance-calibration diagnostics are provided in App.~\ref{app:variance_calibration}.

\FloatBarrier
\section{Benchmark Experiments on COCO}
\label{sec:coco}

We evaluate our probabilistic extension on COCO~\citep{lin2014mscoco} to assess its behavior in a standardized and widely used multi-person pose-estimation setting. 
COCO provides diverse object scales, poses, occlusions, and image conditions, making it well suited for evaluating keypoint uncertainty across a broad distribution of localization errors.
We split the original validation set into calibration and evaluation subsets at a 20/80 ratio; all reported metrics are computed on the evaluation subset.

As the underlying YOLO-Pose models, we use SuperGradients' \emph{YOLO-NAS-Pose}~\citep{supergradients} and Ultralytics' \emph{YOLO26-Pose}~\citep{jocher2026ultralyticsyolo26}.
YOLO-NAS-Pose follows the standard YOLO-Pose training scheme: one-to-many label assignment via the TOOD alignment metric~\citep{feng2021tood} and multi-scale neck features, both shared across popular variants~\citep{maji2022yolopose,ultralytics}.
It provides a competitive keypoint detection baseline and, through its Apache-2.0 license, a practical foundation for releasing open-source code.
YOLO26-Pose is a more recent architecture that uses one-to-one label assignment and natively integrates normalizing-flow-based \ac{rle} for uncertainty estimation~\citep{li2021keypoint-normalizing-flows}.
To the best of our knowledge, this built-in \ac{rle} is the only existing keypoint uncertainty-quantification baseline in a YOLO-Pose model, making it a direct benchmark for our post-hoc approach.
We evaluate Nano and Large variants of both models.
In all cases, the YOLO-Pose models are kept frozen and only the probabilistic heads are trained, leaving the original keypoint predictions unchanged.
Implementation details including training hyperparameters, calibration setup, and hardware are provided in App.~\ref{app:coco_impl}.

We first report the core benchmark results in terms of uncertainty calibration diagnostics and \acl{akp} in \cref{subsec:coco_core}. 
We then evaluate whether uncertain keypoints can be identified and pruned reliably, and how this impacts calibration, in \cref{subsec:coco_kpp}.
Finally, \cref{subsec:coco_ablation} presents ablation studies on the importance-weighting scheme, covariance parameterization, probabilistic-head architecture, and uncertainty scalarization to support the main design choices of our approach.

\subsection{Core Results}
\label{subsec:coco_core}

\begin{figure}[tb]
    \centering

    \begin{minipage}[t]{0.3765\linewidth}
        \centering
        \includegraphics[
            width=\linewidth,
            trim={77px 314px 85px 0px},
            clip
        ]{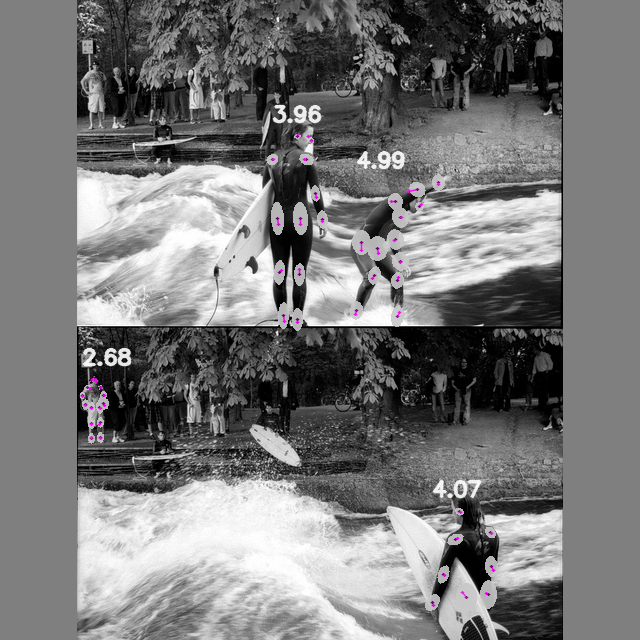}
    \end{minipage}
    \hfill
    \begin{minipage}[t]{0.2279\linewidth}
        \centering
        \includegraphics[
            width=\linewidth,
            trim={35px 0px 37px 0px},
            clip
        ]{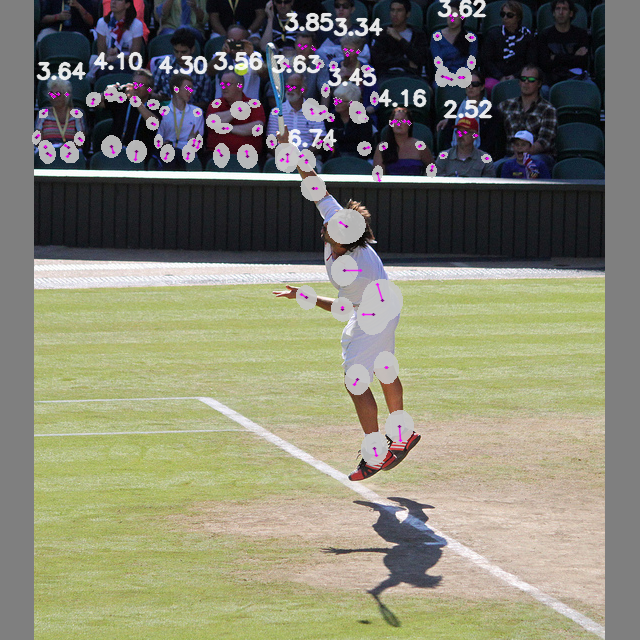}
    \end{minipage}
    \hfill
    \begin{minipage}[t]{0.365\linewidth}
        \centering
        \includegraphics[
            width=\linewidth,
            trim={0px 95px 0px 95px},
            clip
        ]{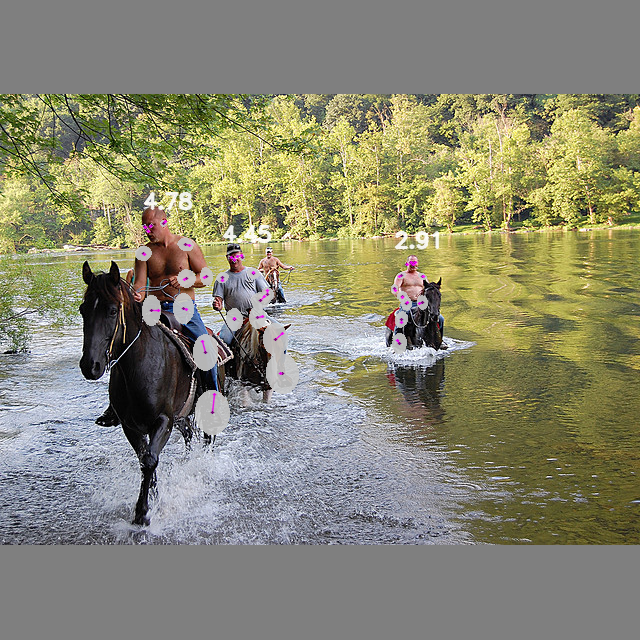}
    \end{minipage}

    \caption{
        Example keypoint predictions produced by our probabilistic extension.
        The gray ellipses enclose 95\% of the probability mass of the predicted keypoint distributions, while the purple lines visualize the residuals between predicted and ground-truth keypoints. The displayed values represent the associated \acp{nll}. 
        Aggregate calibration metrics are summarized in \cref{tab:coco_main_results}, corresponding calibration diagnostics are shown in \cref{fig:coco_nano_calibration_diagnostics}, and per-keypoint calibration metrics are reported in \cref{fig:coco_per_kp_calibration}.
    }
    \label{fig:coco_example_preds}
\end{figure}

\begin{table}[tb]
\caption{
Core uncertainty metrics on COCO for YOLO-NAS-Pose and YOLO26-Pose.
We compare the uncalibrated dispersion model against Gaussian and Student-$t$ calibration.
For YOLO26-Pose, we additionally compare against its built-in RLE~\citep{li2021keypoint-normalizing-flows} and a variant calibrated with temperature scaling.
Joint calibration plots are shown in \cref{fig:coco_nano_calibration_diagnostics} for YOLO-NAS-Pose Nano; the full joint and marginal diagnostics for all model variants and calibration methods are provided in App.~\ref{app:coco}, \crefrange{fig:coco_nas_n_base_diag}{fig:coco_y26_l_rle_cal_diag}.
For reference, we additionally report $\mathrm{AKP}_\mathrm{vis}$ and $\mathrm{AKP}_\mathrm{rand}$ in the table footnote, i.e., the \acp{akp} for visibility-score-based ranking and random ranking.
}
\label{tab:coco_main_results}
\centering
\begin{threeparttable}
\begin{tabular}{clccccc}
\toprule
Model & Calibration method
& ACE $\downarrow$
& ENCE $\downarrow$
& NLL $\downarrow$
& AKP\tnote{\dag}~~$\uparrow$ \\
\midrule
\multirow{3}{*}{\makecell{YOLO-NAS-Pose\\[2pt]Nano}}
& Uncalibrated dispersion  & $0.0374$ & $0.1061$ & $5.67$ & $\mathbf{0.935}$ \\
& Gaussian calibration           & $0.0582$ & $0.0570$ & $5.62$ & $\mathbf{0.935}$ \\
& Student-$t$ calibration        & $\mathbf{0.0016}$ & $\mathbf{0.0434}$ & $\mathbf{5.50}$ & $\mathbf{0.935}$ \\
\midrule
\multirow{3}{*}{\makecell{YOLO-NAS-Pose\\[2pt]Large}}
& Uncalibrated dispersion  & $0.0252$ & $0.1538$ & $5.47$ & $\mathbf{0.948}$ \\
& Gaussian calibration           & $0.0602$ & $0.0813$ & $5.42$ & $\mathbf{0.948}$ \\
& Student-$t$ calibration        & $\mathbf{0.0020}$ & $\mathbf{0.0770}$ & $\mathbf{5.27}$ & $\mathbf{0.948}$ \\
\midrule
\multirow{5}{*}{\makecell{YOLO26-Pose\\[2pt]Nano}}
& Uncalibrated dispersion  & $0.0473$ & $0.0533$ & $5.83$ & $\mathbf{0.927}$ \\
& Gaussian calibration           & $0.0554$ & $0.0426$ & $5.82$ & $\mathbf{0.927}$ \\
& Student-$t$ calibration        & $\mathbf{0.0055}$ & $\mathbf{0.0302}$ & $\mathbf{5.70}$ & $\mathbf{0.927}$ \\
\addlinespace[0.3em]
& Native RLE (built-in)        & $0.0421$ & $0.3944$ & $5.75$ & $0.909$ \\
& Calibrated RLE        & $0.0143$ & $0.3546$ & $5.73$ & $0.910$ \\
\midrule
\multirow{5}{*}{\makecell{YOLO26-Pose\\[2pt]Large}}
& Uncalibrated dispersion  & $0.0537$ & $0.0711$ & $5.38$ & $\mathbf{0.950}$ \\
& Gaussian calibration           & $0.0658$ & $\mathbf{0.0608}$ & $5.37$ & $\mathbf{0.950}$ \\
& Student-$t$ calibration        & $\mathbf{0.0037}$ & $0.0740$ & $\mathbf{5.19}$ & $\mathbf{0.950}$ \\
\addlinespace[0.3em]
& Native RLE (built-in)        & $0.0451$ & $0.3933$ & $5.23$ & $0.941$ \\
& Calibrated RLE        & $0.0163$ & $0.3160$ & $5.20$ & $0.941$ \\
\bottomrule
\end{tabular}
\vspace{-7pt}
\begin{tablenotes}\footnotesize
\item[] \leavevmode
\begin{tabular}[t]{@{}l@{\;\;}l@{\;\;}l@{\;\;}l@{}}
$^\dag$\,$\mathrm{AKP}_\mathrm{vis}$\,|\,$\mathrm{AKP}_\mathrm{rand}$ benchmarks: & \quad YOLO-NAS-Pose: & Nano\:\:$0.913$\,|\,$0.791$, & Large\:\:$0.931$\,|\,$0.831$ \\
& \quad YOLO26-Pose: & Nano\:\:$0.899$\,|\,$0.769$, & Large\:\:$0.932$\,|\,$0.846$
\end{tabular}
\end{tablenotes}
\end{threeparttable}
\end{table}

Qualitative examples of the predicted keypoint distributions are shown in \cref{fig:coco_example_preds}.
The uncertainty ellipses visualize the spatial uncertainty estimated by the probabilistic heads and reflect the local ambiguity of the corresponding keypoints.
The core quantitative results are summarized in \cref{tab:coco_main_results} for YOLO-NAS-Pose and YOLO26-Pose, each in Nano and Large variants.
We compare the uncalibrated dispersion model, Gaussian calibration, and Student-$t$ calibration. 
For YOLO26-Pose, we additionally compare against its built-in \ac{rle}-based uncertainty estimates and a temperature-scaled variant to provide a fair comparison under matched calibration conditions. 
The keypoint mean predictions remain unchanged by design.
Thus, differences in \ac{nll}, \ac{ace}, \ac{ence}, and \ac{akp} are attributable to the uncertainty estimates rather than changes in keypoint localization accuracy.

Gaussian calibration improves the predicted covariances primarily in terms of second-moment calibration, reducing \ac{ence} for both Nano and Large.
At the same time, \ac{ace} is larger than for the uncalibrated dispersion model, highlighting that coverage calibration and variance calibration capture complementary aspects of predictive uncertainty under distributional misspecification.
Student-$t$ calibration better aligns with both diagnostics by combining calibrated covariance scale with heavier tails.
It substantially reduces \ac{ace} and improves \ac{nll} across all model variants, and also improves \ac{ence} in most cases; an exception is YOLO26-Pose Large, where Student-$t$ \ac{ence} slightly exceeds the uncalibrated value.
Overall, Student-$t$ calibration yields the best probabilistic performance in terms of \ac{ace} and \ac{nll}, indicating that the empirical residual distribution is better captured by a heavy-tailed Student-$t$ distribution than by a Gaussian approximation.
Compared to YOLO26-Pose's built-in \ac{rle}-based uncertainty estimates, our approach achieves substantially lower \ac{ence} values across both model sizes, even after temperature scaling the RLE outputs. 
This indicates that our per-keypoint full-covariance heads learn a stronger heteroscedastic dispersion model than \ac{rle}, whose shared flow imposes a uniform correlation structure across all keypoints (\cref{sec:rel_work}).
For \ac{akp}, the RLE confidence score is obtained by applying the same default trace scalarization (\cref{eq:kp_uq_scalar_conf}) to the covariance derived from RLE's predicted distribution.
The \ac{akp} values remain essentially unchanged across calibration methods for a fixed model size.
This is expected because \ac{akp} evaluates the uncertainty-based ranking of keypoints, whereas global calibration primarily rescales or reshapes the predictive distributions without changing the underlying keypoint predictions.
The visibility-score baselines, AKP$_\mathrm{vis}$, which rank keypoints by the detector's built-in visibility confidence, fall between our uncertainty-based scores and the random baselines AKP$_\mathrm{rand}$, indicating that the learned covariances capture localization reliability beyond what visibility prediction alone provides.

\begin{figure}[tb]
    \centering

    \begin{subfigure}[t]{\textwidth}
        \centering
        \includegraphics[width=0.32\linewidth]{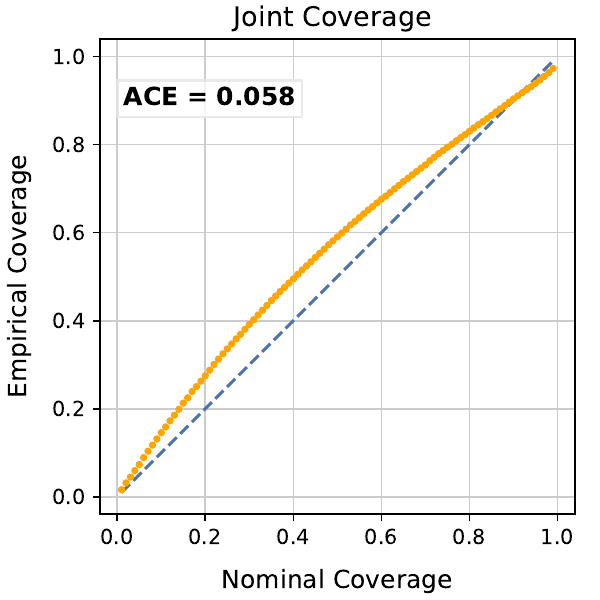}
        \hfill
        \includegraphics[width=0.32\linewidth]{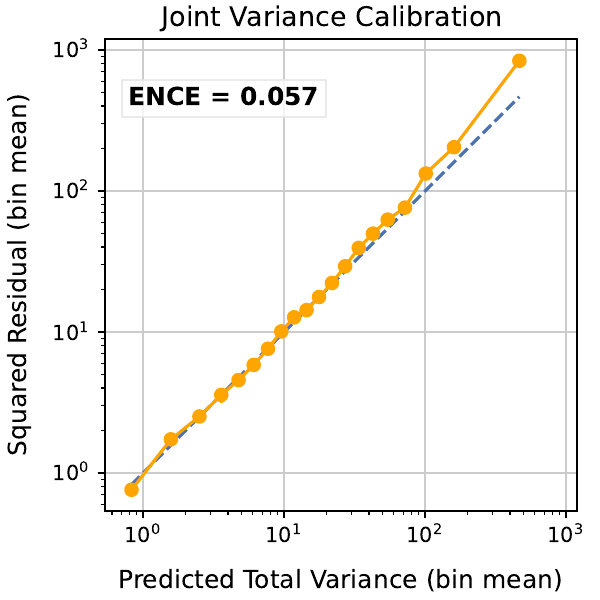}
        \hfill
        \includegraphics[width=0.32\linewidth]{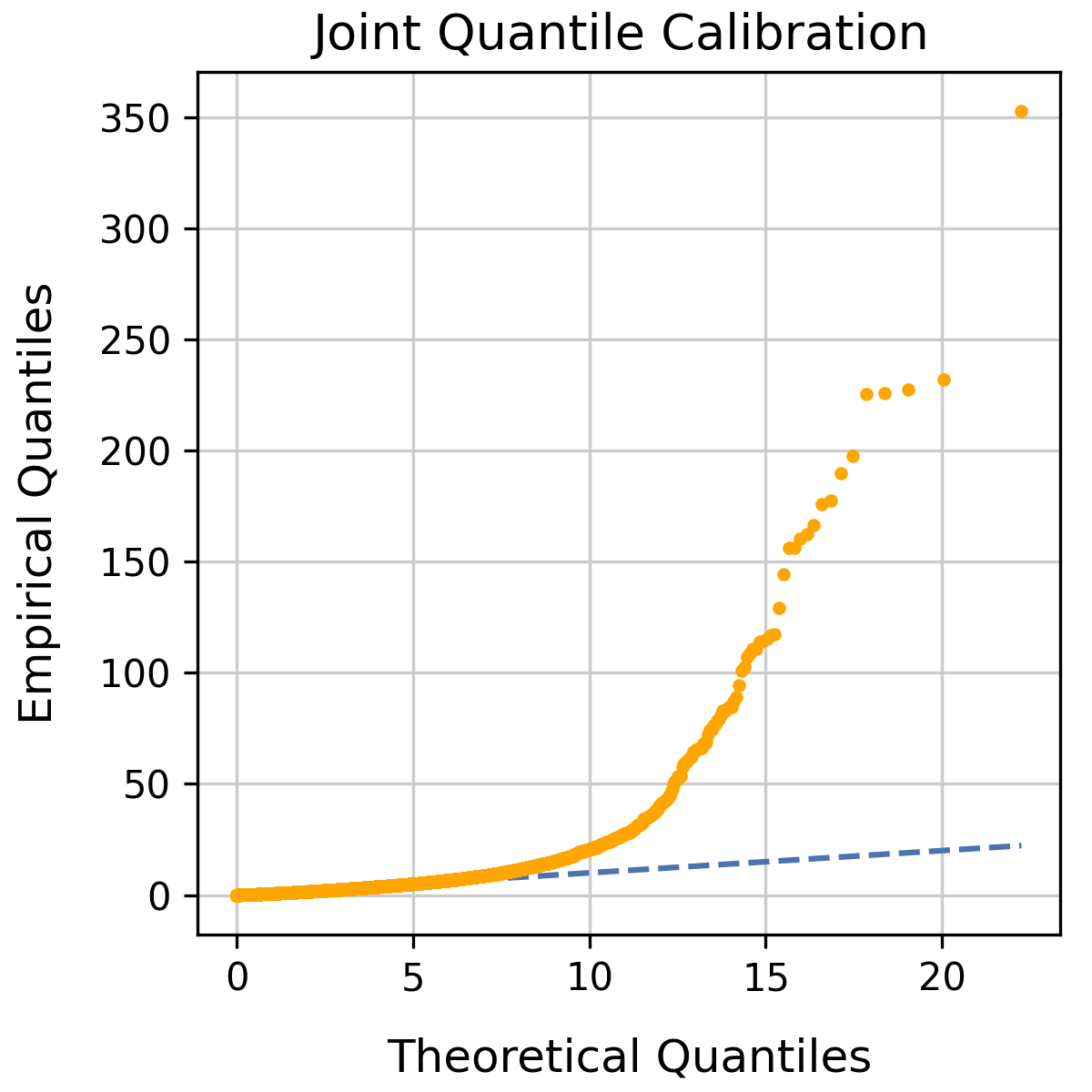}
        \caption{Gaussian calibration}
        \label{fig:coco_nano_g_calib}
    \end{subfigure}

    \vspace{0.5em}

    \begin{subfigure}[t]{\textwidth}
        \centering
        \includegraphics[width=0.32\linewidth]{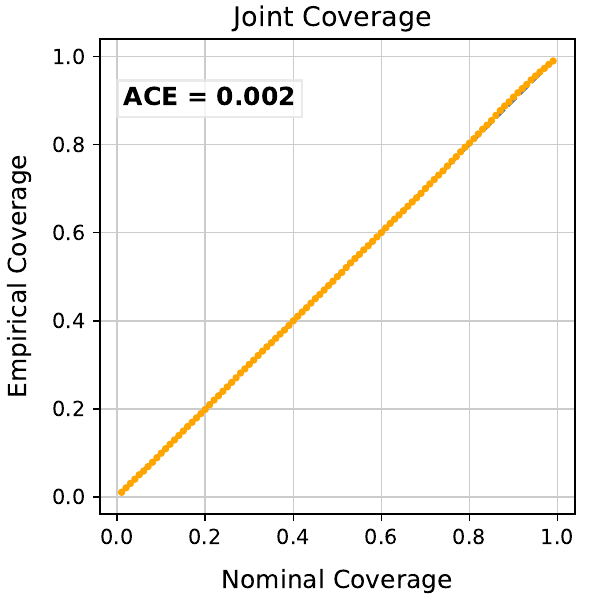}
        \hfill
        \includegraphics[width=0.32\linewidth]{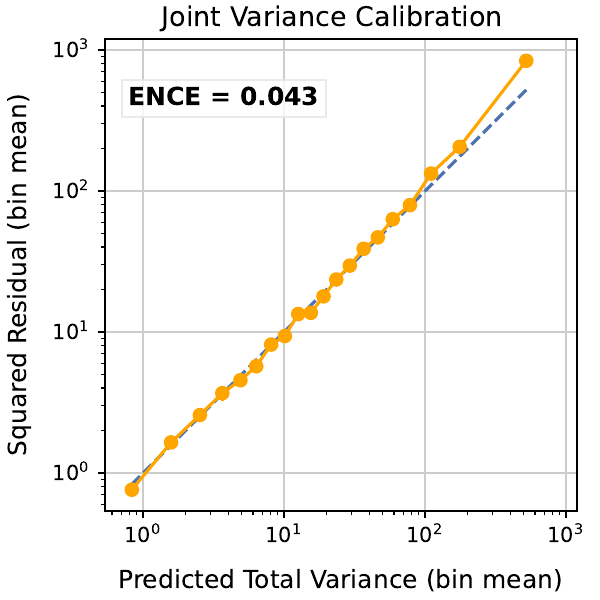}
        \hfill
        \includegraphics[width=0.32\linewidth]{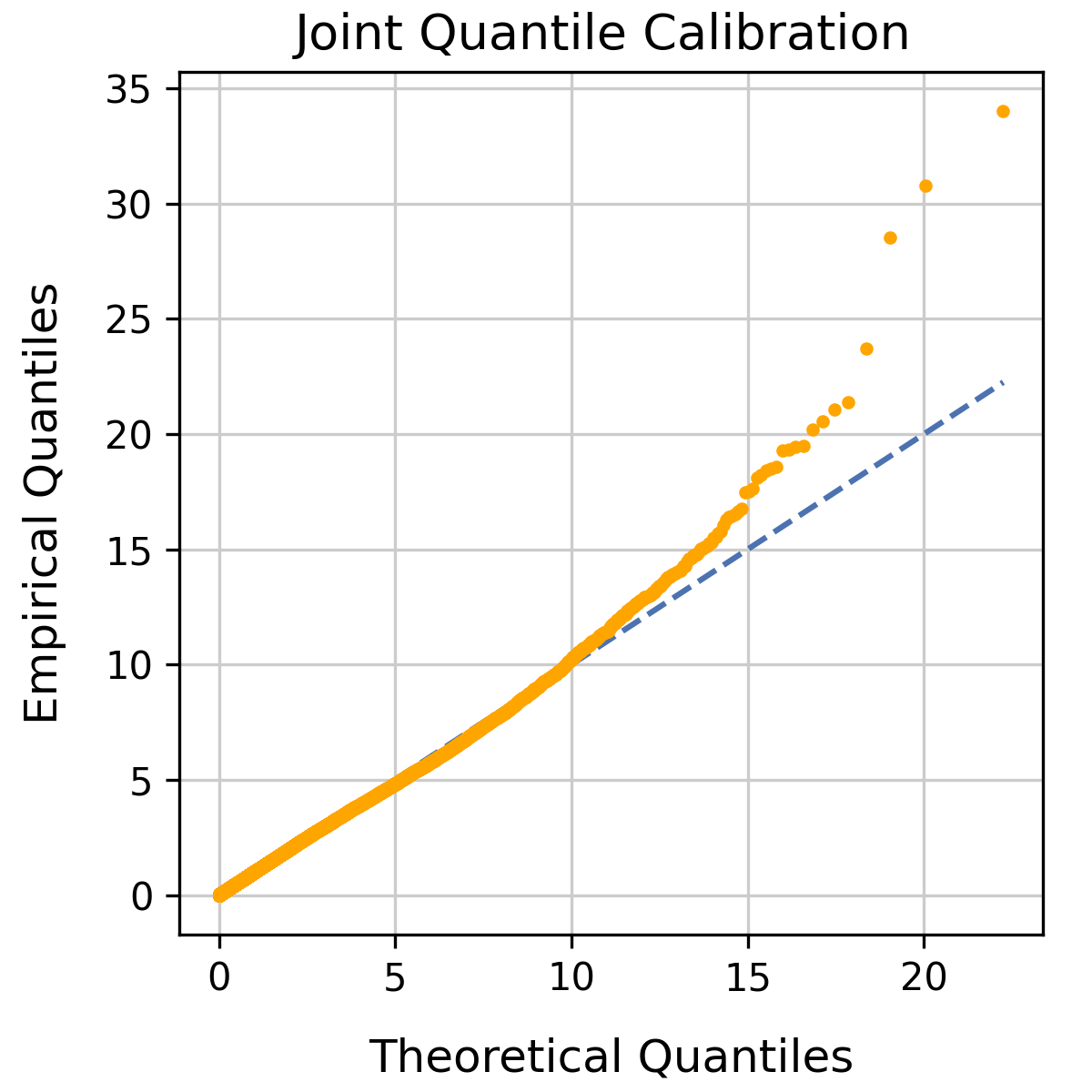}
        \caption{Student-$t$ calibration}
        \label{fig:coco_nano_t_calib}
    \end{subfigure}

    \caption{
        Calibration diagnostics on COCO for our probabilistic extension of YOLO-NAS-Pose Nano.
        The \ac{qq} plots show that the Student-$t$ distribution captures the heavy-tailed residual structure more accurately than the Gaussian.
        The corresponding marginal diagnostics, as well as the full calibration diagnostics for all model variants, can be found in App.~\ref{app:coco}, \crefrange{fig:coco_nas_n_base_diag}{fig:coco_y26_l_rle_cal_diag}.
    }
    \label{fig:coco_nano_calibration_diagnostics}
\end{figure}

\begin{figure}[tb]
    \centering
    \includegraphics[width=0.98\linewidth]{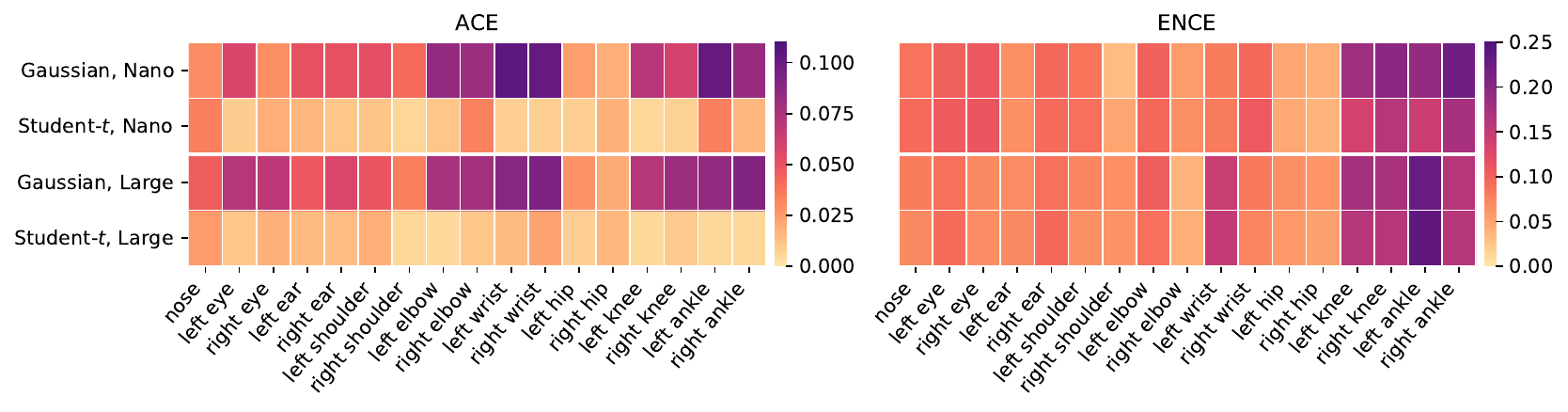}
    \caption{Per-keypoint calibration metrics on COCO for YOLO-NAS-Pose.
    \Ac{ace} and \ac{ence} are reported for each keypoint class under Gaussian and Student-$t$ calibration.
    Lower values indicate better calibration.
    Student-$t$ calibration improves calibration for most keypoint classes, while calibration quality remains keypoint-dependent, with particularly poor variance calibration for knees and ankles.}
    \label{fig:coco_per_kp_calibration}
\end{figure}

\Cref{fig:coco_nano_calibration_diagnostics} visualizes this behavior for YOLO-NAS-Pose Nano.
Gaussian temperature scaling corrects the covariance-scale mismatch of the uncalibrated dispersion model, as reflected by the improved variance-calibration curve.
However, because the empirical residuals deviate from the assumed Gaussian shape, improving the covariance scale does not necessarily improve joint coverage.
The \ac{qq} plots reveal this directly: the Gaussian fit deviates in the tails, whereas the Student-$t$ distribution tracks the empirical quantiles closely.
Student-$t$ calibration mitigates this mismatch by combining calibrated covariance scale with heavier tails, yielding near-nominal coverage and improved variance calibration.
The corresponding full calibration diagnostics for all four model variants can be found in App.~\ref{app:coco}.

Finally, the per-keypoint calibration results in \cref{fig:coco_per_kp_calibration} show that calibration quality varies substantially across keypoint classes, with extremities such as knees and ankles remaining more challenging than more stable body keypoints.
We therefore evaluate in \cref{subsec:coco_kpp} whether predictive uncertainty can be used to identify and remove unreliable keypoint predictions, thereby improving calibration.


\subsection{Pruning Uncertain Keypoints}
\label{subsec:coco_kpp}

\begin{figure}[b]
    \centering

    \begin{subfigure}[b]{0.48\textwidth}
        \centering
        \includegraphics[
            width=0.44\linewidth,
            trim={0px 0px 0px 125px},
            clip        
        ]{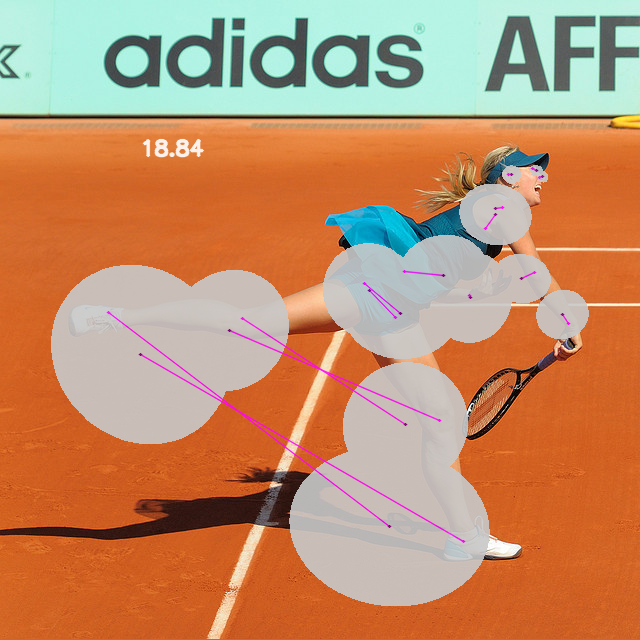}
        \hfill
        \includegraphics[
            width=0.53\linewidth,
            trim={0px 106px 0px 107px},
            clip        
        ]{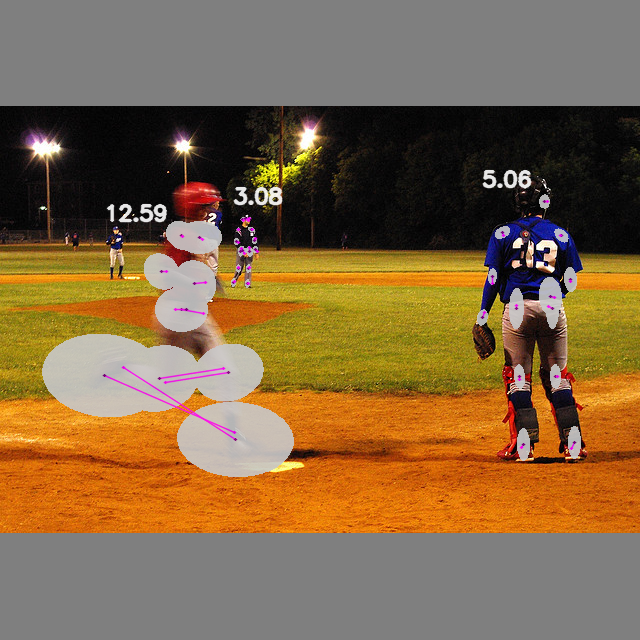}
        \caption{Before keypoint pruning}
        \label{fig:coco_before_pruning}
    \end{subfigure}
    \hfill
    \begin{subfigure}[b]{0.48\textwidth}
        \centering
        \includegraphics[
            width=0.44\linewidth,
            trim={0px 0px 0px 125px},
            clip    
        ]{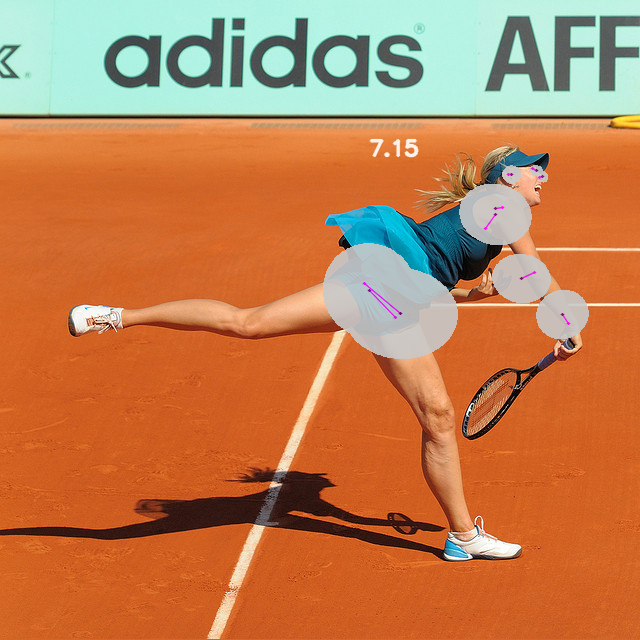}
        \hfill
        \includegraphics[
            width=0.53\linewidth,
            trim={0px 106px 0px 107px},
            clip    
        ]{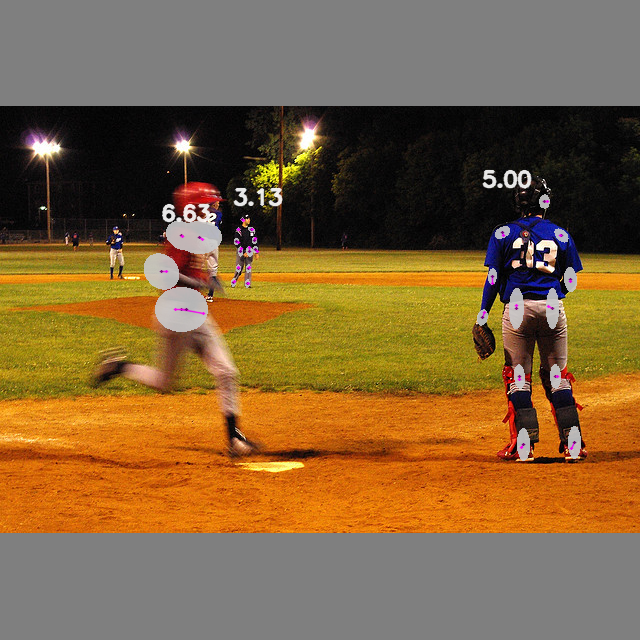}
        \caption{After keypoint pruning}
        \label{fig:coco_after_pruning}
    \end{subfigure}

    \caption{
    Example keypoint predictions before and after pruning uncertain keypoints. 
    The gray ellipses enclose 95\% of the probability mass of the predicted keypoint distributions, while the purple lines visualize the distance between prediction and ground truth. The displayed values correspond to the associated \acp{nll}.
    We observe that left and right legs are frequently confused, leading to misplaced ankle and knee predictions. 
    Uncertainty-based keypoint pruning effectively mitigates these ambiguities, leading to improved calibration for ankles and knees, cf.\ \cref{fig:coco_per_kp_calibration_after_pruning}.
    }
    \label{fig:coco_keypoint_pruning}
\end{figure}

\begin{table}[t]
\centering
\caption{
Calibration improvements from pruning uncertain keypoints on COCO for YOLO-NAS-Pose.
The reported metric values are obtained after applying uncertainty-based keypoint pruning introduced in \cref{subsec:kp_pruning}.
Changes are computed relative to the corresponding models without keypoint pruning reported in \cref{tab:coco_main_results}; negative changes indicate improvement.
The corresponding calibration plots are shown in App.~\ref{app:coco}, \cref{fig:coco_kpp_n_gauss_diag,fig:coco_kpp_n_t_diag,fig:coco_kpp_l_gauss_diag,fig:coco_kpp_l_t_diag}, while changes in the per-keypoint calibration metrics are reported in \cref{fig:coco_per_kp_calibration_after_pruning}.
}
\label{tab:coco_kp_pruning}
\setlength{\tabcolsep}{5.5pt}
\begin{tabular}{ll c cc cc cc}
\toprule
\multirow{2}{*}{Model size}
& \multirow{2}{*}{Calibration method}
& \multirow{2}{*}{Retained}
& \multicolumn{2}{c}{ACE $\downarrow$}
& \multicolumn{2}{c}{ENCE $\downarrow$}
& \multicolumn{2}{c}{NLL $\downarrow$} \\
\cmidrule(lr){4-5}
\cmidrule(lr){6-7}
\cmidrule(lr){8-9}
& & & ACE & $\Delta$ACE
& ENCE & $\Delta$ENCE
& NLL & $\Delta$NLL \\
\midrule
\multirow{2}{*}{Nano}
& Gaussian calibration    & \multirow{2}{*}{83\%}
& 0.0539 & $\mathbf{-0.0043}$
& 0.0467 & $\mathbf{-0.0103}$
& 5.35   & $\mathbf{-0.27}$ \\

& Student-$t$ calibration &
& 0.0049 & $+0.0033$
& 0.0488 & $+0.0054$
& 5.25   & $\mathbf{-0.25}$ \\

\midrule
\multirow{2}{*}{Large}
& Gaussian calibration    & \multirow{2}{*}{84\%}
& 0.0604 & $+0.0002$
& 0.0608 & $\mathbf{-0.0205}$
& 5.20   & $\mathbf{-0.22}$ \\

& Student-$t$ calibration &
& 0.0040 & $+0.0020$
& 0.0560 & $\mathbf{-0.0210}$
& 5.08   & $\mathbf{-0.19}$ \\
\bottomrule
\end{tabular}
\end{table}

We next evaluate whether the predicted uncertainties can be used as a post-hoc reliability criterion for selective keypoint prediction.
Following \cref{subsec:kp_pruning}, we prune keypoints whose predictive uncertainty exceeds the selected threshold and compute the metrics on the retained predictions.
This removes individual unreliable keypoints while leaving the retained predictions unchanged.
The selected threshold retains approximately 83--84\% of keypoints across model sizes and calibration methods (\cref{tab:coco_kp_pruning}).
Qualitative examples in \cref{fig:coco_keypoint_pruning} show that pruning primarily suppresses predictions with large spatial uncertainty relative to the object scale.
Such cases often arise from occlusions, overlapping persons, or left--right ambiguities, particularly for lower-body keypoints where confused limbs can lead to misplaced knees and ankles.

The quantitative results in \cref{tab:coco_kp_pruning} show that pruning consistently improves \ac{nll} across all calibrated models.
Because \ac{nll} depends on both the predicted mean and covariance, part of this improvement stems from removing keypoints with large localization errors, e.g., due to left--right limb confusion, whose residuals dominate the likelihood regardless of uncertainty quality.
Calibration metrics, by contrast, assess only whether the predicted uncertainty matches the observed error magnitude.
Therefore, they can remain unchanged or even worsen when pruned keypoints carry uncertainty roughly proportional to their (large) errors.
The effect on calibration is accordingly more metric-dependent: pruning generally improves variance calibration, whereas changes in \ac{ace} are smaller and less uniform.
For Student-$t$ calibration, pruning can slightly degrade \ac{ace}, as the heavy-tailed likelihood already accommodates large residuals and selective removal perturbs the calibrated tail behavior of the retained distribution.
This suggests that uncertainty-based pruning primarily removes predictions with poorly matched residual magnitude and covariance scale, rather than the full distributional shape, while coverage calibration is less directly affected by selective removal.
The corresponding full calibration diagnostics are provided in App.~\ref{app:coco}, \cref{fig:coco_kpp_n_gauss_diag,fig:coco_kpp_n_t_diag,fig:coco_kpp_l_gauss_diag,fig:coco_kpp_l_t_diag}.

\begin{figure}[tb]
    \centering
    \includegraphics[width=0.98\linewidth]{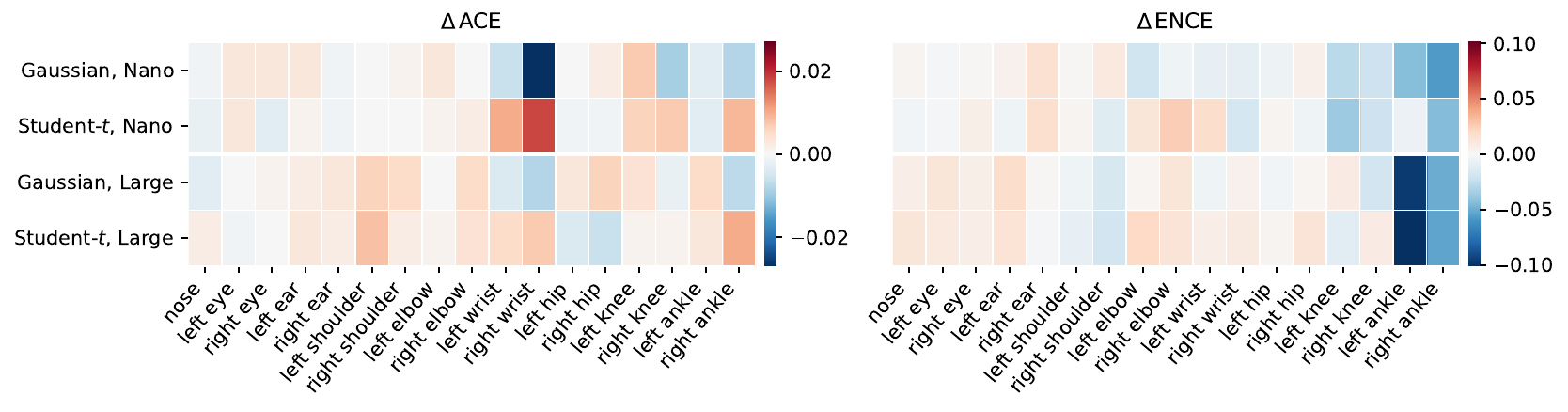}
    \caption{Per-keypoint impact of keypoint pruning on COCO calibration metrics for YOLO-NAS-Pose Nano and Large, corresponding to the unpruned results in \cref{fig:coco_per_kp_calibration}. $\Delta$ACE and $\Delta$ENCE are reported for each keypoint under Gaussian and Student-$t$ calibration, where negative values indicate improved calibration after pruning. Pruning particularly improves calibration for keypoints prone to left--right limb confusion, such as knees and ankles (cf.\ \cref{fig:coco_keypoint_pruning}).}
    \label{fig:coco_per_kp_calibration_after_pruning}
\end{figure}

The per-keypoint analysis in \cref{fig:coco_per_kp_calibration_after_pruning} confirms that the largest improvements occur for knees and ankles, where pose ambiguity is most pronounced, consistent with the qualitative examples in \cref{fig:coco_keypoint_pruning}.
Overall, uncertainty-based pruning provides a simple mechanism for suppressing unreliable keypoint predictions, particularly in cases of limb confusion, and improving the calibration of the retained pose estimates.


\FloatBarrier
\subsection{Ablation Studies}
\label{subsec:coco_ablation}

We ablate the main components of our probabilistic extension on COCO.
Unless stated otherwise, all variants are evaluated after Gaussian calibration, such that differences reflect the underlying training objective, dispersion parameterization, and probabilistic-head design rather than differences in post-hoc calibration.

\begin{table}[!b]
\centering
\caption{
Ablation study on COCO for YOLO-NAS-Pose (one-to-many label assignment) and YOLO26-Pose (one-to-one label assignment).
All configurations are evaluated after Gaussian calibration.
Our default configuration uses full heteroscedastic dispersion matrices, the importance-weighted \ac{gnll} objectives for training and calibration defined in \cref{eq:iw_gnll,eq:calib_iw_gnll}, and the proposed probabilistic head architecture introduced in \cref{fig:main}.
}

\label{tab:coco_main_ablation_results}
\begin{tabular}{clcccc}
\toprule
Model & Ablation & ACE $\downarrow$ & ENCE $\downarrow$ & NLL $\downarrow$ & AKP $\uparrow$ \\
\midrule
\multirow{6}{*}{\makecell{YOLO-NAS-Pose\\[1pt]Nano}}
& \textbf{Default configuration}        & $\mathbf{0.0582}$ & $0.0570$ & $\mathbf{5.62}$ & $0.935$ \\
\addlinespace[0.3em]
& No object-scale IW ($w_\mathrm{scale}=1$)            & $0.0596$ & $0.0579$ & $5.64$ & $\mathbf{0.936}$ \\
& No IW ($w_\mathrm{ret}=1, w_\mathrm{scale}=1$)     & $0.0689$ & $0.0637$ & $5.64$ & $0.935$ \\
\addlinespace[0.3em]
& Diagonal dispersion parameterization   & $0.0617$ & $0.0629$ & $5.67$ & $0.934$ \\
& Isotropic dispersion parameterization  & $0.0652$ & $0.0603$ & $5.71$ & $0.935$ \\
\addlinespace[0.3em]
& Last-layer probabilistic heads        & $0.1017$ & $\mathbf{0.0536}$ & $5.85$ & $0.916$ \\
\midrule
\multirow{6}{*}{\makecell{YOLO-NAS-Pose\\[1pt]Large}}
& \textbf{Default configuration}        & $\mathbf{0.0602}$ & $0.0813$ & $\mathbf{5.42}$ & $\mathbf{0.948}$ \\
\addlinespace[0.3em]
& No object-scale IW ($w_\mathrm{scale}=1$)            & $0.0699$ & $0.0861$ & $5.43$ & $\mathbf{0.948}$ \\
& No IW ($w_\mathrm{ret}=1, w_\mathrm{scale}=1$)     & $0.0796$ & $0.0868$ & $5.43$ & $0.947$ \\
\addlinespace[0.3em]
& Diagonal dispersion parameterization   & $0.0658$ & $0.0778$ & $5.44$ & $\mathbf{0.948}$ \\
& Isotropic dispersion parameterization  & $0.0678$ & $\mathbf{0.0762}$ & $5.48$ & $\mathbf{0.948}$ \\
\addlinespace[0.3em]
& Last-layer probabilistic heads        & $0.0770$ & $0.0811$ & $5.56$ & $0.933$ \\
\midrule
\multirow{2}{*}{\makecell{YOLO26-Pose\\[1pt]Nano}}
& \textbf{Default configuration}        & $\mathbf{0.0554}$ & $0.0426$ & $\mathbf{5.82}$ & $\mathbf{0.927}$ \\
\addlinespace[0.3em]
& No IW ($w_\mathrm{ret}=1, w_\mathrm{scale}=1$)     & $0.0651$ & $\mathbf{0.0406}$ & $5.83$ & $\mathbf{0.927}$ \\
\midrule
\multirow{2}{*}{\makecell{YOLO26-Pose\\[1pt]Large}}
& \textbf{Default configuration}        & $\mathbf{0.0658}$ & $\mathbf{0.0608}$ & $\mathbf{5.37}$ & $\mathbf{0.950}$ \\
\addlinespace[0.3em]
& No IW ($w_\mathrm{ret}=1, w_\mathrm{scale}=1$)     & $0.0677$ & $0.0864$ & $5.39$ & $0.949$ \\
\bottomrule
\end{tabular}
\end{table}

\begin{figure}[!b]
    \centering
    \includegraphics[width=0.75\linewidth]{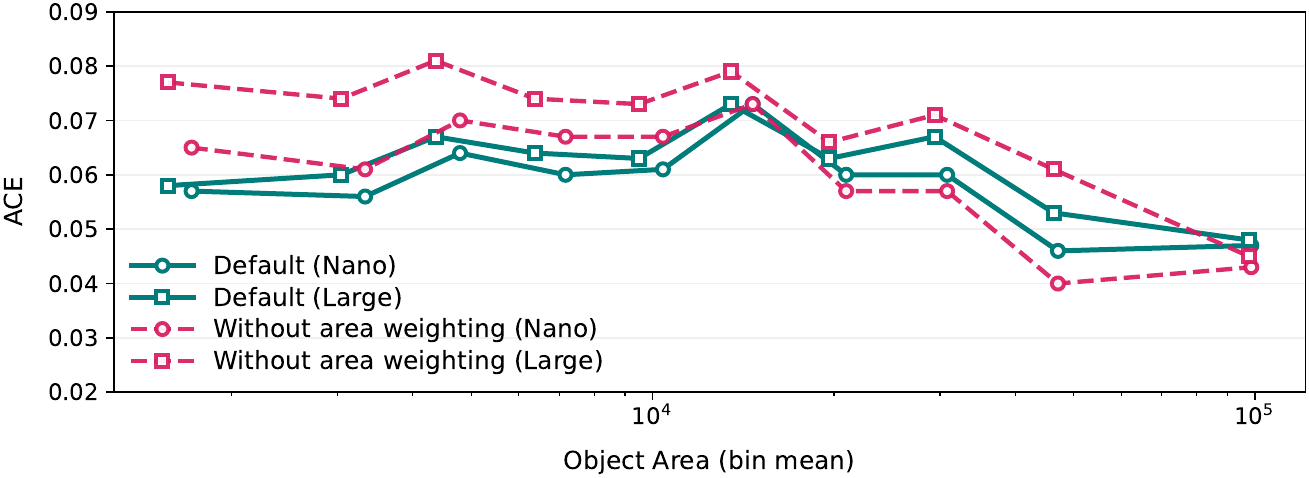}
    \caption{
    Per-scale calibration on COCO for YOLO-NAS-Pose.
    We ablate object-area importance weighting ($w_\mathrm{scale}$) and show that it balances calibration across object scales, substantially improving calibration for smaller objects without degrading calibration for larger objects.
    }
    \label{fig:coco_ablation_area_weight}
\end{figure}

\Cref{tab:coco_main_ablation_results} summarizes the main ablation results for YOLO-NAS-Pose and YOLO26-Pose.
The default configuration provides the best overall trade-off across \ac{ace}, \ac{ence}, \ac{nll}, and \ac{akp}.
For YOLO-NAS-Pose, removing object-scale importance weights (see \cref{eq:w_s}) degrades calibration, while omitting importance weighting altogether, i.e., both object-scale and retention weights (see \cref{eq:w_r}), leads to a further degradation.
For YOLO26-Pose, removing all importance weighting likewise increases \ac{ace} and \ac{nll} for both model sizes, confirming that the benefit generalizes across label-assignment paradigms.
The analysis in \cref{fig:coco_ablation_area_weight} further shows that object-scale importance weights lead to more balanced calibration across object scales and substantially improve calibration for smaller objects.

\begin{table}[!t]
\centering
\caption{
Covariance scalarization methods for deriving keypoint uncertainty scores used by \ac{akp}, evaluated on COCO with YOLO-NAS-Pose.
All variants use the default configuration from \cref{tab:coco_main_ablation_results} and differ only in the scalarization method.
}
\label{tab:coco_scalarization_ablation}
\begin{tabular}{lcccc}
\toprule
& \multicolumn{4}{c}{AKP $\uparrow$} \\
\cmidrule(lr){2-5}
Scalarization
& \multicolumn{2}{c}{Nano}
& \multicolumn{2}{c}{Large} \\
\cmidrule(lr){2-3}
\cmidrule(lr){4-5}
& Gaussian & Student-$t$ & Gaussian & Student-$t$ \\
\midrule
Trace $\operatorname{tr}(\mathbf{\Sigma})$ (\textbf{default})
& $\mathbf{0.935}$ & $\mathbf{0.935}$ & $\mathbf{0.948}$ & $\mathbf{0.948}$ \\
Determinant $\sqrt{\det(\mathbf{\Sigma})}$
& $\mathbf{0.935}$ & $0.934$ & $\mathbf{0.948}$ & $\mathbf{0.948}$ \\
Max-eigenvalue $\lambda_{\max}(\mathbf{\Sigma})$
& $0.933$ & $0.933$ & $0.946$ & $0.946$ \\
\bottomrule
\end{tabular}
\end{table}

\begin{table}[!t]
\centering
\caption{
Threshold-selection ablation for uncertainty-based keypoint pruning on COCO for YOLO-NAS-Pose.
All variants use the default configuration from \cref{tab:coco_main_ablation_results} and differ only in the threshold-selection criterion: Youden's~$J$, minimum distance to the ideal ROC point, and maximum F1.
}
\label{tab:coco_threshold_selection_ablation}
\setlength{\tabcolsep}{5pt}
\begin{tabular}{llcccccc}
\toprule
\multirow{2}{*}{Model size}
& \multirow{2}{*}{Threshold selection}
& \multicolumn{2}{c}{ACE $\downarrow$}
& \multicolumn{2}{c}{ENCE $\downarrow$}
& \multicolumn{2}{c}{NLL $\downarrow$}
\\
\cmidrule(lr){3-4}
\cmidrule(lr){5-6}
\cmidrule(lr){7-8}
&
& Gaussian & Student-$t$
& Gaussian & Student-$t$
& Gaussian & Student-$t$
\\
\midrule
\multirow{3}{*}{Nano}
& Youden's $J$ (\textbf{default})
& \textbf{0.0539} & 0.0049
& 0.0467 & 0.0488
& \textbf{5.35} & \textbf{5.25} \\
& Min. distance to $(0,1)$
& 0.0540 & 0.0049
& \textbf{0.0465} & 0.0488
& 5.36 & \textbf{5.25} \\
& Max. F1
& 0.0580 & \textbf{0.0016}
& 0.0584 & \textbf{0.0446}
& 5.65 & 5.50 \\
\midrule
\multirow{3}{*}{Large}
& Youden's $J$ (\textbf{default})
& 0.0604 & 0.0040
& \textbf{0.0608} & \textbf{0.0560}
& \textbf{5.20} & \textbf{5.08} \\
& Min. distance to $(0,1)$
& 0.0604 & 0.0040
& \textbf{0.0608} & \textbf{0.0560}
& \textbf{5.20} & \textbf{5.08 }\\
& Max. F1
& \textbf{0.0602} & \textbf{0.0021}
& 0.0815 & 0.0771
& 5.42 & 5.26 \\
\bottomrule
\end{tabular}
\end{table}

The dispersion and architecture ablations, reported for YOLO-NAS-Pose, support the proposed design choices.
Restricting the predicted covariance to diagonal or isotropic form generally degrades probabilistic performance, particularly in terms of \ac{nll}, showing that anisotropic uncertainty is beneficial for keypoint localization.
For the isotropic parameterization, the calibration diagnostics in App.~\ref{app:coco}, \cref{fig:coco_isotropic_variance_calib_plots}, further show that acceptable joint variance calibration can mask poor marginal calibration along the individual coordinate axes, highlighting the importance of evaluating both joint and marginal diagnostics.
Replacing the proposed probabilistic heads with last-layer heads, which take the penultimate pose-head features as input and use only a single nonlinear prediction layer, leads to a clear degradation.
This suggests that uncertainty estimation benefits from the additional feature processing capacity of the proposed probabilistic heads.
The probabilistic heads add 4--16\% to GFLOPs and 3--16\% to total parameters depending on architecture and model size, with the lowest overhead on the larger models where the base cost dominates.
App.~\ref{app:coco}, \cref{tab:coco_latency} reports the full latency breakdown.

\Cref{tab:coco_scalarization_ablation} compares covariance scalarizations for deriving the keypoint uncertainty scores used by \ac{akp}.
Trace, determinant, and maximum-eigenvalue scalarization yield similar results.
We use the trace as the default since it directly reflects the total predicted marginal variance.
Complementing this, \cref{tab:coco_threshold_selection_ablation} compares threshold-selection strategies for uncertainty-based keypoint pruning.
Youden's~$J$ statistic and the minimum-distance-to-ideal-ROC-point criterion select non-trivial pruning thresholds and yield similar results, indicating that the pruning behavior is not driven by a fragile operating-point choice.
By contrast, maximizing F1 removes only a few keypoints, resulting in metrics close to the unpruned baseline and making it less suitable for inducing meaningful selective prediction.
We therefore use Youden's~$J$ as the default balanced criterion for separating reliable from unreliable keypoint predictions.

\FloatBarrier
\section{Vision-Based Aircraft Landing}
\label{sec:app_vbl}

\begin{figure}[t]
    \centering
    \includegraphics[width=1.0\linewidth]{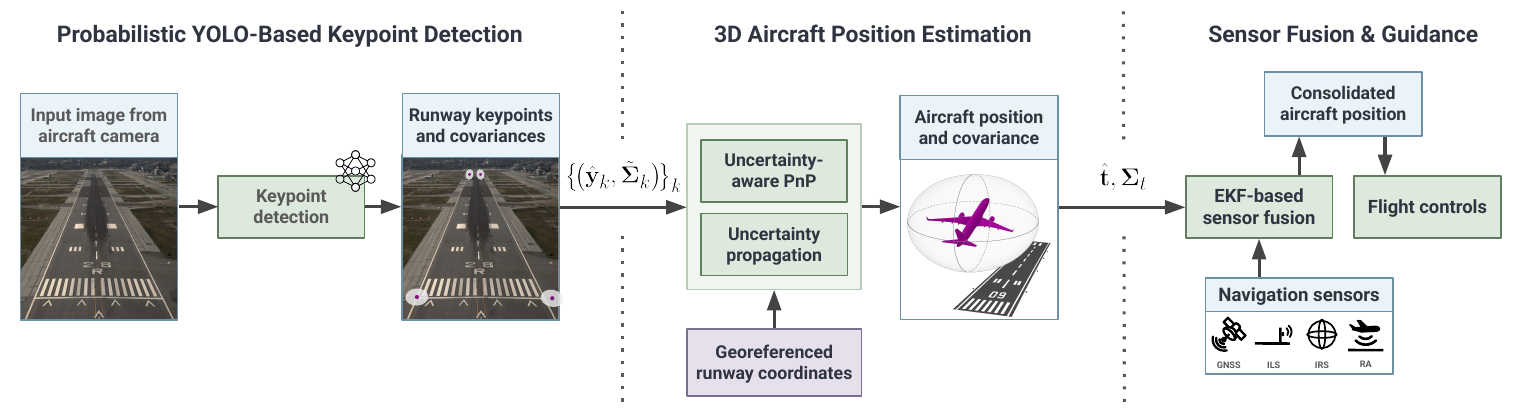}
    \caption{Vision-based aircraft landing. A YOLO-Pose model, augmented with our post-hoc probabilistic extension, predicts runway keypoints together with calibrated 2D covariances. These predictions are used in a covariance-weighted \acs{pnp} step to estimate the 3D position of the aircraft, and the corresponding 3D covariance is obtained by propagating the 2D covariances through the \acs{pnp} mapping. The resulting vision-based measurement is then integrated with other navigation sensors in a downstream \acs{ekf}-based sensor-fusion module to support autonomous landing.}
    \label{fig:vbl_architecture}
\end{figure}

\Ac{vbl} refers to vision-based runway landing guidance or navigation during approach and landing, where visual information from an onboard camera is used to support runway-relative guidance or position estimation~\citep{balduzzi2021runway_landing_guidance}. 
As detailed in App.~\ref{app:vbl_motivation}, \ac{vbl} provides a complementary landing-navigation modality that is independent of the ground-based infrastructure required by the \ac{ils} and does not rely directly on the \ac{gnss}.
In our system, the aircraft's 3D position relative to the runway is estimated from detected image-plane keypoints of runway landmarks with known georeferenced coordinates, such as the four runway corners.
The resulting vision-based position estimate can then be used as an independent measurement source within an \acf{ekf}-based sensor-fusion module, as described by \citet{roumeliotis2000ekf_fusion}.

Since this measurement is fused in an \ac{ekf}-based sensor-fusion module, \acl{uq} is a functional requirement rather than a diagnostic add-on: each vision update must be associated with a well-calibrated covariance estimate compatible with the Gaussian measurement model.
Overconfident covariances can cause the filter to overweight unreliable vision updates, whereas underconfident covariances reduce the contribution of the vision sensor to the fused state estimate.
This is consequential in civil aviation due to certification requirements for deep learning-based systems \citep{easa_ai_concept_paper_issue2_2024,easa_mleap_d4_2024,easa_npa2025_07,faa_ai_safety_assurance_roadmap_2024}.

\begin{figure}[p]
    \centering
    \begin{subfigure}[t]{1.0\linewidth}
        \centering
        \includegraphics[width=1.0\linewidth]{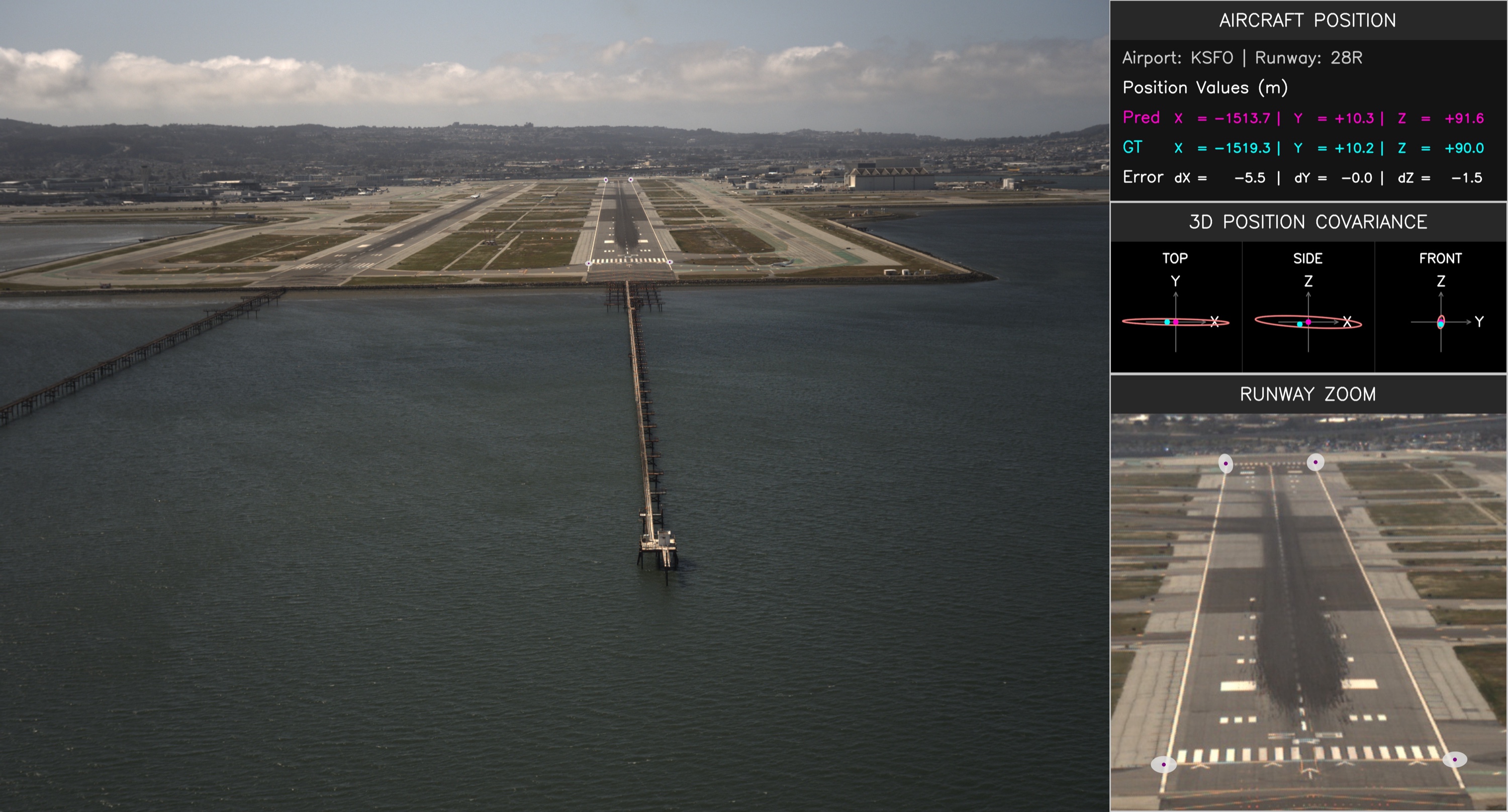}
        \caption{San Francisco International Airport (KSFO), see App.~\ref{app:vbl_seqs_and_detailed_calib_diag}, \cref{fig:vbl_KSFO_seq} for additional KSFO examples.}
    \end{subfigure}

    \vspace{0.3cm}
    
    \begin{subfigure}[t]{1.0\linewidth}
        \centering
        \includegraphics[width=1.0\linewidth]{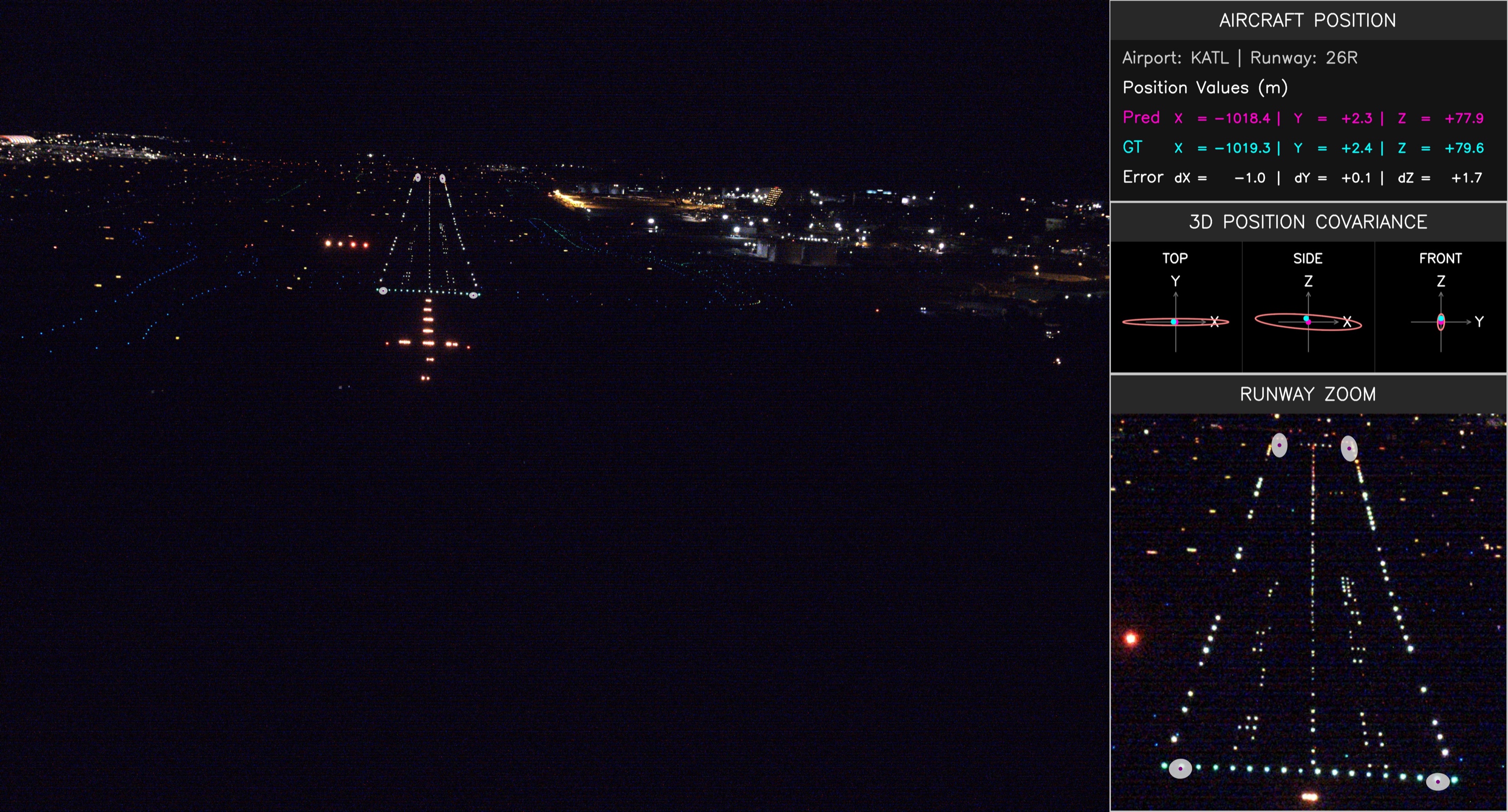}
        \caption{Hartsfield--Jackson Atlanta International Airport (KATL), see App.~\ref{app:vbl_seqs_and_detailed_calib_diag}, \cref{fig:vbl_KATL_seq} for additional KATL examples.}
    \end{subfigure}
    \caption{
    Vision-based aircraft landing examples.
    Detected runway keypoints define the mean locations of the calibrated predictive distributions, visualized as 95\% probability ellipses.
    The 2D--3D matches between image keypoints and georeferenced runway corners are used in a covariance-weighted \ac{pnp} formulation to estimate the aircraft position relative to the runway.
    The propagated 3D position uncertainty is represented by a 95\% probability ellipsoid, shown through top, side, and front projections.
    Positions are reported in the runway coordinate system, where $x$, $y$, and $z$ denote longitudinal, lateral, and vertical directions, respectively.
    \textit{GT} denotes the differential-GPS reference position, \textit{Pred} the \ac{vbl} prediction, and \textit{Diff} their difference.
    }
    \label{fig:vbl_main_landing_vis}
\end{figure}

\begin{figure}[t]
    \centering

    \begin{minipage}[t]{0.32\linewidth}
        \centering
        \begin{subfigure}[t]{0.98\linewidth}
            \includegraphics[width=\linewidth]{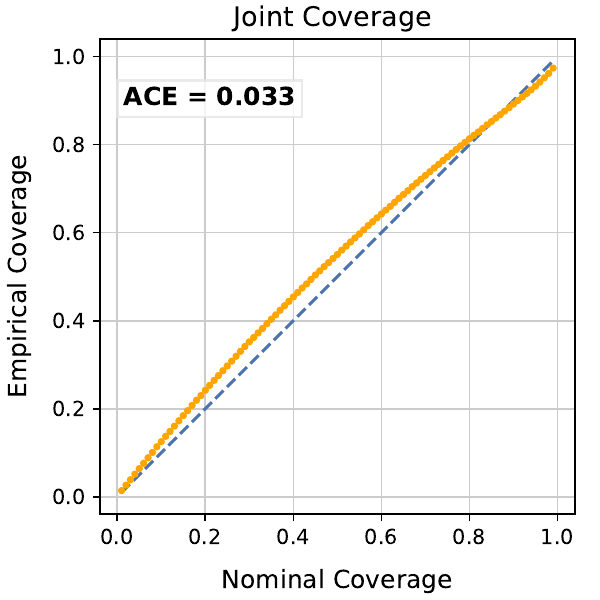}
    
            \vspace{0.7em}
    
            \includegraphics[width=0.98\linewidth]{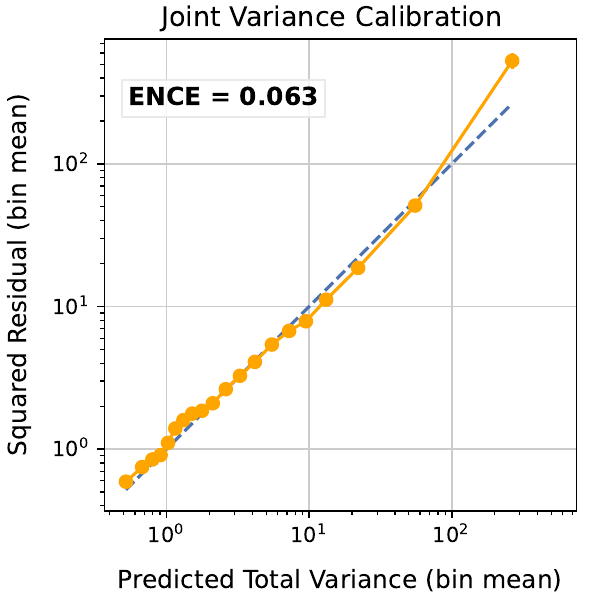}
            \caption{Gaussian calibration}
            \label{fig:vbl_calib_gm}
        \end{subfigure}
    \end{minipage}
    \hfill
    \begin{minipage}[t]{0.64\linewidth}
        \centering

        \begin{subfigure}[t]{\linewidth}
            \centering
            \includegraphics[width=0.49\linewidth]{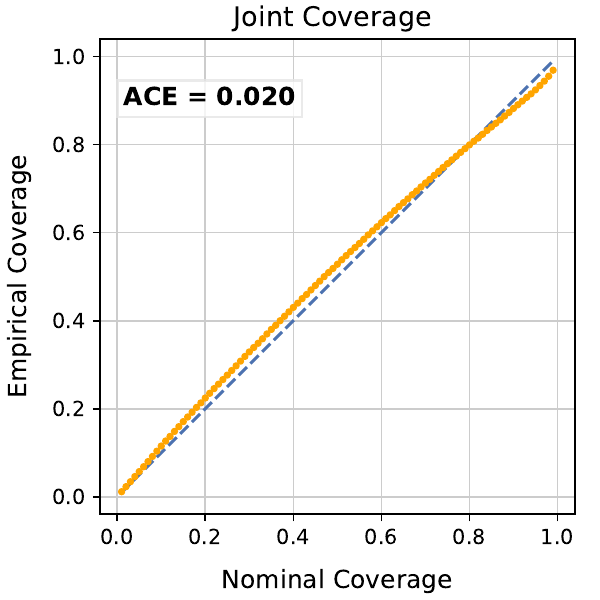}
            \hfill
            \includegraphics[width=0.49\linewidth]{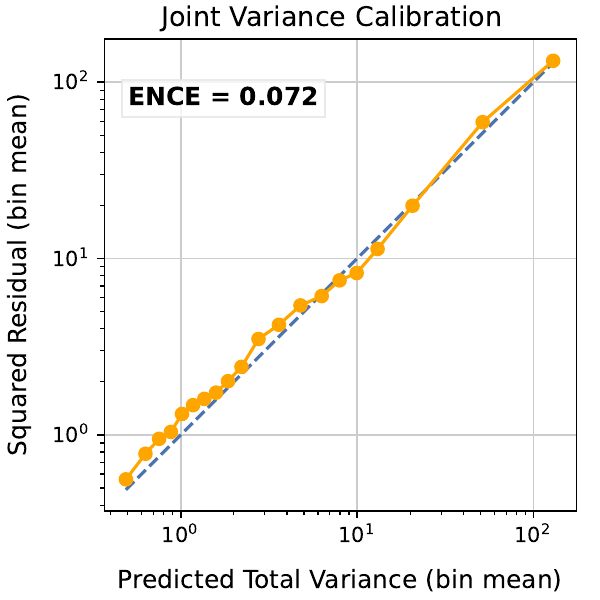}
            \caption{Gaussian calibration + keypoint pruning}
            \label{fig:vbl_calib_gm_kpp}
        \end{subfigure}

        \vspace{1.5em}

        \begin{subfigure}[t]{\linewidth}
            \centering
            \includegraphics[width=0.965\linewidth]{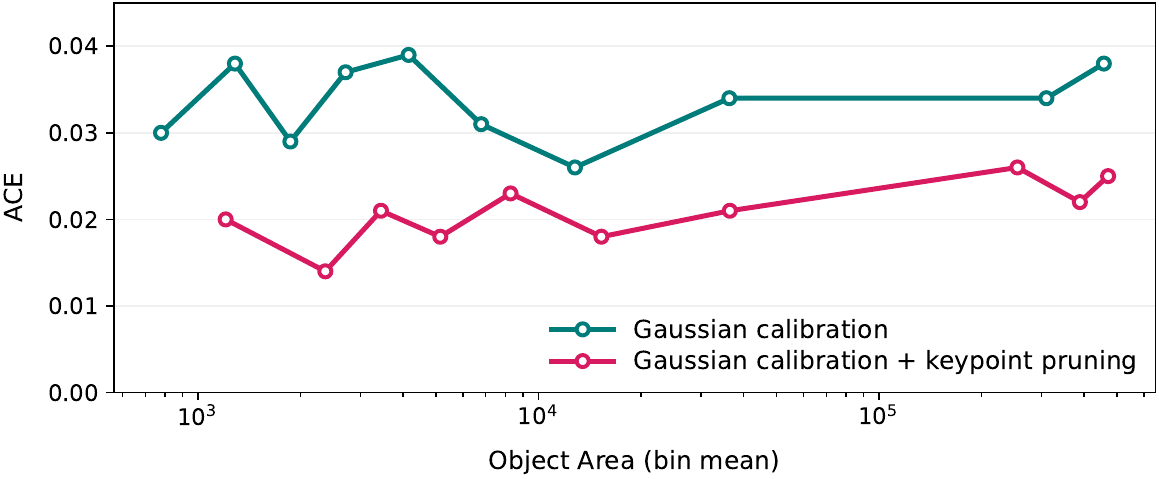}
            \caption{Balanced calibration across object scales}
            \label{fig:vbl_calib_ace_object_area}
        \end{subfigure}

    \end{minipage}
    \caption{
    Calibration diagnostics for vision-based aircraft landing, comparing the Gaussian-calibrated model before and after pruning uncertain keypoints.
    The scale-wise \ac{ace} plot shows that our importance-weighted \ac{gnll} yields balanced calibration across object areas, a key requirement for \ac{vbl}.
    Compared with COCO (cf.~\cref{fig:coco_nano_calibration_diagnostics}), the Gaussian model is substantially better calibrated on the \ac{vbl} data, reflecting the more constrained landing geometry, limited pose and appearance variation, and fewer visibility or occlusion ambiguities of runway keypoints.
    Additional marginal calibration diagnostics and example predictions before and after keypoint pruning are provided in App.~\ref{app:vbl_seqs_and_detailed_calib_diag}, \cref{fig:vbl_calib_diag_with_marg,fig:vbl_qq_plots,fig:vbl_keypoint_pruning}.
    }
    \label{fig:vbl_calibration_diagnostics}
\end{figure}

\Ac{vbl} also imposes strict real-time constraints, making the single-stage YOLO architecture attractive for multi-runway keypoint localization in a single forward pass.
In this setting, uncertainty estimates must not compromise the latency and prediction interface of the underlying YOLO-Pose model.
Our post-hoc probabilistic extension is designed for this constraint: it keeps the YOLO-Pose model frozen, preserves its original keypoint predictions, and adds only lightweight probabilistic heads for calibrated covariance prediction, as shown in \cref{fig:main}.

\paragraph{System Details.}
An overview of the \ac{vbl} system is shown in \cref{fig:vbl_architecture}.
In the following, we consider a single detected runway and omit the instance index $i$ for clarity. 
In this application, we restrict attention to the Gaussian model, since the downstream navigation pipeline is formulated around Gaussian uncertainty representations.
The \ac{vbl} model predicts, for each retained keypoint $k\in\mathcal{K}^{\mathrm{keep}}$ (see \cref{eq:kp_prune_rule}), a 2D keypoint location $\hat{\mathbf{y}}_{k}\in\mathbb{R}^{2}$ together with a calibrated covariance $\tilde{\mathbf{\Sigma}}_{k}\in\mathbb{S}^2_{++}$, where $\mathbb{S}^2_{++}$ denotes the set of symmetric positive definite $2\times2$ matrices.
We estimate the aircraft's 3D position $\hat{\mathbf{t}}\in\mathbb{R}^{3}$ by solving a covariance-weighted \acf{pnp} problem~\citep{FerrazBMN2014} from $\hat{\mathbf{y}}_{k}$, $\tilde{\mathbf{\Sigma}}_{k}$, and the georeferenced runway keypoint coordinates; see App.~\ref{app:pnp}.
We then propagate the image-plane uncertainty through the \ac{pnp} mapping using a first-order linearization~\citep{valentin2024probabilisticparameterestimatorscalibration}, yielding a 3D position covariance $\mathbf{\Sigma}_{\mathbf{t}}\in\mathbb{S}^3_{++}$; see App.~\ref{app:uncertainty_propagation}.
The resulting pair $(\hat{\mathbf{t}}, \mathbf{\Sigma}_{\mathbf{t}})$ is subsequently fused with other navigation sensors; see App.~\ref{app:sensor_fusion}.
\Ac{vbl} therefore provides a realistic application scenario for our central claim: keypoint uncertainty is valuable not only for image-plane diagnostics, but also because it directly affects the reliability of the downstream pipeline.

\paragraph{Experimental Setup.}
We instantiate our approach on SuperGradients' \emph{YOLO-NAS-Pose}, where we adopt the Nano variant to meet the stringent latency requirements of \ac{vbl} in real-time operation.
The YOLO-NAS-Pose Nano was trained on a large-scale internal aircraft-landing dataset, described in App.~\ref{app:vbl_dataset}.
We use a standard train/validation/test split and reserve a separate calibration set for uncertainty calibration.
The post-hoc probabilistic extension is trained without using data augmentations.
This design choice ensures that the learned residual distribution closely reflects the distribution encountered at inference time. Residuals are obtained via SuperGradients' label assignment procedure, and training is performed using the proposed importance-weighted \ac{gnll} objective in \cref{eq:iw_gnll}.
In line with the application setting described above, we restrict attention to the Gaussian model, since the downstream pipeline is designed around Gaussian uncertainty representations. At inference time, we use SuperGradients' default post-processing and optionally apply keypoint pruning as described in \cref{subsec:kp_pruning}.

\paragraph{Results.}
We use \ac{vbl} to assess whether improved keypoint uncertainty estimation transfers to a realistic downstream pipeline.
\cref{fig:vbl_calibration_diagnostics} shows that Gaussian calibration provides a strong fit on the \ac{vbl} data, with joint coverage close to the diagonal and good agreement between predicted and empirical variance.
Compared with the COCO results in \cref{fig:coco_nano_calibration_diagnostics}, the Gaussian model is substantially better calibrated, which is plausible because runway keypoints follow a more constrained projective geometry with limited pose, appearance, visibility, and occlusion variation.
Pruning uncertain keypoints further improves coverage calibration by excluding predictions that tend to exhibit large residuals, reducing the risk of poorly localized keypoints affecting the position estimate.
The scale-wise \ac{ace} results further show that the proposed importance-weighted \ac{gnll} yields balanced calibration across object areas, a key requirement for \ac{vbl} given the wide range of runway scales encountered during approach.
Further examples for San Francisco International Airport (KSFO), Los Angeles International Airport (KLAX), and Hartsfield--Jackson Atlanta International Airport (KATL) are provided in App.~\ref{app:vbl_seqs_and_detailed_calib_diag}, \cref{fig:vbl_KSFO_seq,fig:vbl_KLAX_seq,fig:vbl_KATL_seq}.

Propagating the calibrated image-plane covariances through \ac{pnp} yields informative 3D position covariances, but we observed that their calibration is less accurate than the image-plane calibration.
We attribute this gap to two sources.
First, the first-order uncertainty propagation (\cref{eq:vbl_uq_prop}) assumes that the projection model is locally linear around the estimated position.
At long range, where the runway subtends few pixels, the \ac{pnp} Jacobian $\mathbf{J}$ is ill-conditioned and small image-plane errors are amplified nonlinearly; the linear approximation then becomes inaccurate.
Second, the propagation treats camera intrinsics and extrinsics as exact.
In practice, residual calibration errors contribute a noise term to the 3D covariance that is independent of keypoint uncertainty and therefore cannot be corrected by improving the 2D calibration alone.
Closing the 3D calibration gap would require calibrating the full 2D-to-3D pipeline end to end, accounting for both the learned image-plane uncertainty and the propagation model.
This is outside the scope of the present work, which focuses on calibrated YOLO-Pose keypoint uncertainties.

Finally, \cref{fig:vbl_ood_KCLT_seq} in App.~\ref{app:vbl_seqs_and_detailed_calib_diag} shows vision-based landing in foggy conditions at Charlotte Douglas International Airport (KCLT), representing a distribution shift relative to the training data.
The \ac{vbl} model fails in several frames, and the predicted covariances do not consistently capture this failure.
This illustrates that calibrated keypoint covariances learned in the in-distribution regime do not account for epistemic uncertainty associated with out-of-distribution inputs.
Robust handling of such cases requires additional mechanisms for detecting or accounting for distribution shift, such as epistemic-uncertainty modeling that exposes shift-induced model uncertainty and supports out-of-distribution detection.
This remains an important direction for future work, especially in the context of safety-critical applications.

\FloatBarrier

\section{Conclusion}
\label{sec:conclusion}

We presented a lightweight post-hoc probabilistic extension for YOLO-Pose models that augments a trained YOLO-Pose model with calibrated predictive distributions over keypoint localization errors while preserving its original predictions.
This design is motivated by the task-specific optimization of modern YOLO models through architectures, assignment strategies, losses, and post-processing pipelines.
Rather than replacing these components with a new training objective, which may compromise prediction quality or require revalidating already trained or certification-constrained models, our approach learns heteroscedastic dispersion around the fixed keypoint predictions and calibrates the resulting predictive distribution in a subsequent step.
Based on this predictive distribution, we further introduced uncertainty-based keypoint pruning as a post-hoc mechanism for removing high-uncertainty keypoint predictions.

Complementing the probabilistic extension, we proposed an uncertainty evaluation protocol for assessing both keypoint reliability ranking and distributional calibration.
For reliability ranking, we introduced \ac{akp}, a keypoint-level extension of the COCO \ac{ap} protocol that evaluates whether confidence scores rank individual keypoints according to localization correctness.
For distributional calibration, we evaluated the predicted bivariate keypoint distributions using joint and marginal coverage, Q--Q, and variance-calibration diagnostics.
Together, the protocol assesses whether uncertainty estimates are calibrated and useful for selecting, pruning, or weighting keypoints in downstream tasks.

Our benchmark experiments on COCO show that the separation between point prediction, dispersion learning, and calibration is both practically useful and empirically important.
The same trained dispersion model supports Gaussian and Student-$t$ calibration schemes through stored calibration parameter sets: Gaussian predictions provide covariance compatibility for downstream pipelines, while Student-$t$ predictions better preserve the heavy-tailed residual structure observed for \acs{oks}-trained keypoint detectors.
Gaussian calibration improves variance calibration and \ac{nll}, but remains limited by likelihood mismatch; Student-$t$ calibration better captures heavy-tailed residuals and provides the strongest overall calibration performance.
Uncertainty-based keypoint pruning removes high-uncertainty, empirically unreliable keypoints while leaving retained predictions unchanged, improving calibration of the retained predictions.
The ablation studies confirm the importance of the proposed weighting scheme: retention weighting has the larger effect on overall calibration, while object-scale weighting yields more balanced calibration across object scales.

Finally, the experiments on vision-based aircraft landing demonstrate the downstream relevance of calibrated keypoint uncertainty estimates.
The Gaussian-calibrated model provides well-calibrated image-plane covariances that can be propagated through uncertainty-aware \ac{pnp} and used as measurement uncertainty in downstream \ac{ekf}-based sensor fusion.
At the same time, the foggy landing sequence highlights a fundamental limitation of the present approach: uncertainty estimates learned and calibrated from in-distribution residuals do not, by themselves, capture epistemic uncertainty under distribution shift.
Future work should prioritize epistemic uncertainty estimation and out-of-distribution detection while also exploring more expressive calibration methods.

\paragraph{Broader Impact.}
This limitation carries practical consequences for safety-critical deployment.
A downstream filter that ingests the predicted covariances as measurement noise, such as the \ac{ekf} in \cref{sec:app_vbl}, would overweight unreliable vision updates under distribution shift.
Treating them as a complete safety signal risks false confidence in automated decision making.
Safe integration into certification-constrained pipelines requires complementary mechanisms, whether epistemic uncertainty estimation, input-distribution monitoring, or navigation-level integrity checks, to flag out-of-distribution conditions before they propagate into downstream decision making.

\section*{Acknowledgments}
We thank Thierry Sammour Sawaya, J\'{e}r\'{e}mie Chheang, Pierre Berjon, Don Walker, Matt Sorgenfrei, Leander Fritsch, Tim Brockmeyer, and Luca Mossina for helpful discussions and feedback.

\bibliography{main}
\bibliographystyle{_format_tmlr/tmlr}

\clearpage
\appendix
\section{Target Weighting for Balanced Calibration Across Object Scales}
\label{app:iw_object_scale_balance}

This appendix provides intuition for why, even without an explicit scale imbalance in the dataset, larger instances can contribute a stronger optimization signal when learning heteroscedastic covariances with the \ac{gnll} surrogate in \cref{eq:gnll}. We focus on two effects: (i) the \ac{gnll} gradient scales with residuals, which is coupled to the \ac{oks} geometry, and (ii) larger instances are better resolved at the feature-map stride, yielding more informative features and better-aligned gradients.

\subsection{\ac{gnll} Gradients Scale with Residuals}
\label{app:iw_object_scale_balance_1}
For a single residual $\mathbf{r}\in\mathbb{R}^2$ and covariance $\mathbf{\Sigma}\in\mathbb{R}^{2\times 2}$, the per-keypoint loss is
\begin{equation}
\mathcal{L}_{\mathrm{GNLL}}(\mathbf{\Sigma};\mathbf{r})
\;=\;
\frac{1}{2}\mathbf{r}^{\top}\mathbf{\Sigma}^{-1}\mathbf{r}
+
\frac{1}{2}\ln\lvert\mathbf{\Sigma}\rvert
+ \ln Z_{\mathcal{N}}.
\end{equation}
Using standard matrix derivatives, its gradient w.r.t.\ $\mathbf{\Sigma}$ is
\begin{equation}
\label{eq:gnll_grad_sigma}
\nabla_\mathbf{\Sigma}\, \mathcal{L}_{\mathrm{GNLL}}
\;=\;
\frac{1}{2}\Bigl(
\mathbf{\Sigma}^{-1}
-
\mathbf{\Sigma}^{-1}\mathbf{r}\mathbf{r}^{\top}\mathbf{\Sigma}^{-1}
\Bigr),
\end{equation}
so the dominant term in the gradient magnitude is driven by
$\mathbf{\Sigma}^{-1}\mathbf{r}\mathbf{r}^{\top}\mathbf{\Sigma}^{-1}$, i.e., by the (whitened) residual. Intuitively, large residuals induce larger updates of the covariance parameters.

In YOLO-Pose models, the \ac{oks} geometry normalizes squared residuals by the instance area proxy $s^2$ (cf.\ \cref{eq:oks}). Consequently, a comparable \ac{oks} penalty permits larger absolute residual magnitudes for larger instances. This couples instance scale to the typical residual observed during training, and thus to the typical magnitude of the \ac{gnll} gradient in \cref{eq:gnll_grad_sigma}. As a result, larger instances can disproportionately shape the learned covariance model unless explicitly reweighted.

\subsection{Better Resolution at the Feature-Map Stride Yields Better-Aligned Gradients}
\label{app:iw_object_scale_balance_2}
Let the probabilistic heads predict $\mathbf{\Sigma}_{i,k}$ from neck features $\mathbf{f}_i$ at a certain spatial stride, e.g., 8/16/32 pixels per feature cell in a feature pyramid. Larger instances occupy more pixels and span more feature cells at a given stride, yielding higher signal-to-noise and more informative cues (e.g., keypoint visibility, limb/edge structure, local texture) for uncertainty prediction. By contrast, small instances can become under-resolved, making $\mathbf{f}_i$ less predictive of the true residual dispersion. This increases gradient variance and reduces gradient alignment across small-instance samples, further weakening their effective contribution during optimization compared to well-resolved large instances.

\subsection{Inverse-Area Weighting Yields a Log-Uniform Area Profile}
\label{app:iw_object_scale_balance_3}
To counteract the dominance of large instances, we introduce the inverse-area factor
\begin{equation}
w_{\mathrm{scale}} \;\triangleq\; \frac{1}{s^2}.
\end{equation}
This choice corresponds to targeting a log-uniform weighting profile over area.
To see this, let $a>0$ denote the (proxy) object area, e.g.\ $a\triangleq s^2$, and define the log-area variable $u\triangleq \log a$.
If $a$ has density $f_A(a)$, then by the change-of-variables formula the density of $u$ is
\begin{equation}
\label{eq:log_area_change_of_vars}
f_U(u)
\;=\;
f_A(e^u)\left|\frac{d}{du}e^u\right|
\;=\;
f_A(e^u)\,e^u.
\end{equation}
Now choose a (bounded) weighting profile $f_A(a)\propto 1/a$ on $a\in[a_{\min},a_{\max}]$.
Substituting into \cref{eq:log_area_change_of_vars} yields
\begin{equation}
\label{eq:log_uniform}
f_U(u)\;\propto\; \frac{1}{e^u}e^u\;=\;\mathrm{const},
\end{equation}
so $u=\log a$ is uniform.
Intuitively, this assigns equal mass to equal-width intervals in log-area, i.e., equal mass per multiplicative scale, helping prevent large-scale regimes from dominating the weighted \ac{gnll} objective.
Note that $f_A(a)\propto 1/a$ describes the weighting profile, not the effective training distribution.
If the areas of the matched training instances have density $p_\mathrm{train}(a)$, the weighted objective induces the effective log-area density $f_U(u)\propto p_\mathrm{train}(e^u)\,e^{-u}\,e^{u}=p_\mathrm{train}(e^u)$, which is exactly uniform only if $p_\mathrm{train}(a)$ is uniform in area.
In practice, $w_{\mathrm{scale}}$ therefore shifts weight toward smaller instances; we evaluate its effect empirically in \cref{fig:coco_ablation_area_weight}.
\section{Gaussian Residual Distribution Fine-Tuning}
\label{app:residual_finetune}

If we aim to retain the Gaussian model and are willing to trade a small drop in \ac{ap}/\ac{ar} for substantially improved uncertainty calibration, we can fine-tune the YOLO-Pose model, prior to training the probabilistic heads, such that the resulting keypoint residuals are closer to Gaussian.

The gradients in \cref{eq:grad_oks,eq:grad_gnll} highlight a key mismatch between the \ac{oks} loss and the \ac{gnll}. Unlike the \ac{gnll}, which is a proper probabilistic scoring rule and corresponds to maximum likelihood under an explicit residual model, the \ac{oks} loss aggregates residuals via averages of exponentiated Gaussian-kernel terms over visible keypoints and matched objects in a training batch~$\mathcal{B}$, i.e.,
${\mathcal{L}_{\mathrm{OKS}}^{\mathcal{B}} =\nicefrac{1}{\lvert\mathcal{B}\rvert}\sum_{i\in\mathcal{B}}\mathcal{L}_{\mathrm{OKS}}^{(i)}}$.
Consequently, $\mathcal{L}_{\mathrm{OKS}}^{\mathcal{B}}$ is not a maximum-likelihood objective and does not define a proper probabilistic scoring function for the residual distribution.

At the same time, directly replacing \ac{oks} with a Gaussian NLL introduces two practical pitfalls: (i)~it largely removes the saturation/outlier-robustness induced by the exponential kernel, such that large residuals (e.g., due to occlusions or label noise) receive disproportionately large updates, which typically degrades \ac{ap}/\ac{ar}; and (ii)~the \ac{gnll} is unbounded in contrast to $\mathrm{OKS}\in [0,1]$, which can destabilize YOLO-based multi-task training, where several loss terms are combined with fixed, individual weights.

To enable residual-distribution fine-tuning without substantially altering established detection metrics, such as \acs{ap}/\acs{ar}, we introduce a probabilistic log-likelihood-style \ac{oks} objective that is bounded in $[0,1]$ and preserves the \ac{oks} loss's outlier-suppression mechanism while steering the inlier residual regime towards a Gaussian-like shape:
\begin{equation} 
\mathcal{L}_{\mathrm{pOKS}}^{\mathcal{B}} \;=\; 1-\frac{1}{|\mathcal{K}^\ast|} \sum_{k\in\mathcal{K}^\ast} \exp\!\left( -\frac{1}{|\mathcal{B}_k|} \sum_{i\in\mathcal{B}_k} \frac{\mathbf{r}_{i,k}^\top \mathbf{r}_{i,k}}{2\,s_i^2\,\kappa_k^2} \right), 
\end{equation}
where $\mathcal{K}^\ast=\{\, k\in\{1,\dots,K\}\mid |\mathcal{B}_k|>0 \,\}$ excludes the corner case of empty $\mathcal{B}_k$, and $\mathcal{B}_k=\{\,i\in\mathcal B \mid v_{i,k}>0\,\}$. Notably, for each keypoint $k$, maximizing ${\exp\!\big(-\frac{1}{|\mathcal{B}_k|}\sum_{i\in\mathcal{B}_k}\,\boldsymbol{\cdot}\:\,\big)}$ is equivalent (by monotonicity of $\exp$) to minimizing the corresponding average squared residual, i.e., it induces the same optimum as a constant-variance \ac{gnll} per keypoint.
Crucially, we keep the sum over $k$ outside the exponential: pushing the aggregation over $k$ inside would collapse the objective to a monotone transform of a global constant-variance GNLL and empirically leads to noticeably worse \ac{ap}/\ac{ar} due to reduced robustness.
In practice, $\mathcal{L}_{\mathrm{pOKS}}^{\mathcal{B}}$ attenuates updates whenever a batch contains large errors, limiting the influence of outliers, while still providing a useful learning signal on inliers.

In App.~\ref{app:coco}, we evaluate Gaussian residual-distribution fine-tuning on COCO; see \cref{fig:coco_poks_nano_gauss_diag,fig:coco_poks_nano_gauss_kpp_diag}. Residual fine-tuning substantially improves the calibration of the Gaussian predictive model compared to the original approach presented in \cref{sec:prob_extension}. This indicates that steering the residual distribution towards a more Gaussian shape during training allows the predicted dispersion estimates to better capture the dominant residual structure observed at inference time. As shown in \cref{fig:coco_poks_nano_gauss_diag}, calibration is particularly improved in the central region of the distribution, where empirical and nominal coverage closely agree. However, noticeable deviations remain in the extreme quantiles (cf.~\cref{fig:coco_poks_nano_gauss_diag}), which is expected and suggests that the residual errors retain heavy-tailed characteristics that cannot be fully captured by a Gaussian predictive distribution. Furthermore, \ac{ap} and \ac{ar} decrease slightly, e.g., for the Nano variant from $0.593$ to $0.585$ and from $0.667$ to $0.659$, respectively. This suggests that directly optimizing a probabilistic residual objective may partially trade off mean-prediction accuracy for improved uncertainty calibration. Consequently, our work focuses primarily on post-hoc uncertainty modeling, which preserves the original YOLO-Pose predictions while augmenting them with calibrated uncertainty estimates.
Overall, Gaussian residual fine-tuning improves calibration of the bulk of the error distribution, whereas the treatment of rare, large localization errors remains limited.

\section{Student-\texorpdfstring{$t$}{t} Calibration: A Heavy-Tailed Surrogate for Modeling \ac{oks}-Induced Residuals}
\label{app:student-t_motivation}
This appendix provides an analytical comparison between the \ac{oks} loss, the Gaussian \ac{nll}, and the Student-$t$ \ac{nll}.
We focus on how these objectives downweight large residuals.
This motivates Student-$t$ as a heavy-tailed surrogate likelihood that captures the qualitative outlier downweighting absent under a Gaussian model.

\subsection{Radial Losses and Downweighting}
\label{app:downweighting}

Consider an elliptically contoured residual model, where a loss depends on the residual only through a non-negative radius
\begin{equation}
\label{eq:radius_def}
u \;\triangleq\; \sqrt{\mathbf r^\top \mathbf A\,\mathbf r},
\end{equation}
for some positive definite matrix $\mathbf A\succ 0$.
We have two relevant special cases: ${u=\tfrac{\|\mathbf r\|_2}{(s\,\kappa)}}$ (as in \ac{oks}) and the Mahalanobis radius $u=\sqrt{\mathbf r^\top \mathbf\Sigma^{-1}\mathbf r}$.
Let the per-residual objective be $\rho(u)$ and its derivative be $\psi(u)\triangleq\tfrac{\mathrm{d}\rho(u)}{\mathrm{d}u}$.
By the chain rule, we obtain
\begin{equation}
\label{eq:grad_radial}
\nabla_{\mathbf r}\,\rho(u)
\;=\;
\psi(u)\,\nabla_{\mathbf r}\,u
\;=\;
\psi(u)\,\frac{\mathbf A\,\mathbf r}{u}.
\end{equation}
Thus, the gradient is always aligned with $\mathbf A\,\mathbf r$, while its magnitude is controlled by the scalar factor, which we refer to as the downweighting function
\begin{equation}
\label{eq:downweighting_def}
w(u)\;\triangleq\;\frac{\psi(u)}{u},
\qquad\text{such that}\qquad
\nabla_{\mathbf r}\,\rho(u)\;=\;w(u)\,\mathbf A\,\mathbf r.
\end{equation}
In robust M-estimation terms, $\psi(u)$ is the (radial) $\psi$-function, and $w(u)=\tfrac{\psi(u)}{u}$ is the corresponding weight applied to the residual direction.
A decreasing $w(u)$ suppresses the influence of large residuals.

\subsection{Gaussian, \ac{oks}, and Student-\texorpdfstring{$t$}{t} Objectives}
\label{app:downweighting_examples}
\paragraph{Gaussian \ac{nll}.} Up to an additive constant, the Gaussian \ac{nll} corresponds to $\rho_\mathrm{GNLL}(u)=\tfrac12 u^2$, yielding $\psi_\mathrm{GNLL}(u)= u$ and therefore 
\begin{equation} 
\label{eq:w_gauss} 
w_\mathrm{GNLL}(u)\;=\;1. 
\end{equation} 
Hence, Gaussian objectives do not downweight outliers: the gradient magnitude grows linearly with $u$.

\paragraph{\ac{oks} Loss.}
For a single keypoint, the \ac{oks} loss uses the bounded penalty
\begin{equation}
\label{eq:rho_oks_u}
\rho_{\mathrm{OKS}}(u)
\;=\;
1-\exp\!\left(-\frac{u^2}{2}\right),
\end{equation}
where $u=\tfrac{\|\mathbf r\|_2}{(s\,\kappa)}$.
Differentiating yields $\psi_{\mathrm{OKS}}(u)=\exp\!\left(-\tfrac{u^2}{2}\right)u$ and thus
\begin{equation}
\label{eq:w_oks}
w_{\mathrm{OKS}}(u)
\;=\;
\exp\!\left(-\frac{u^2}{2}\right).
\end{equation}
This exponential downweighting implies a redescending influence: $w_{\mathrm{OKS}}(u)\to 0$ rapidly as $u\to\infty$, so gross outliers contribute vanishing gradient.

\paragraph{Student-$t$ \ac{nll}.}
For a two-dimensional residual, the Student-$t$ \ac{nll} (up to constants) can be written as
\begin{equation}
\label{eq:rho_t_u}
\rho_{t\text{-}\mathrm{NLL}}(u)\;=\;\frac{\nu+2}{2}\,\ln\!\left(1+\frac{u^2}{\nu}\right),
\end{equation}
where $\nu>0$ controls tail heaviness.
Differentiating gives
\begin{equation}
\label{eq:psi_t}
\psi_{t\text{-}\mathrm{NLL}}(u)\;=\;\frac{\nu+2}{\nu+u^2}\,u,
\qquad\text{such that}\qquad
w_{t\text{-}\mathrm{NLL}}(u)\;=\;\frac{\nu+2}{\nu+u^2}.
\end{equation}
Ignoring the constant factor $\tfrac{\nu+2}{\nu}$, the Student-$t$ \ac{nll} downweights outliers as ${w_{t\text{-}\mathrm{NLL}}(u)\propto \tfrac{\nu}{\nu+u^2}}$.
In contrast to \ac{oks}, $w_{t\text{-}\mathrm{NLL}}(u)$ decays only polynomially, ${w_{t\text{-}\mathrm{NLL}}(u)=\mathcal O(u^{-2})}$, while remaining the gradient of a proper likelihood.
Smaller $\nu$ yields stronger downweighting (heavier tails); in the limit $\nu\to\infty$, Student-$t$ recovers Gaussian behavior.

\subsection{Discussion}
\label{app:downweighting_discussion}
\begin{figure}[tbp]
    \centering
    \includegraphics[width=0.72\linewidth]{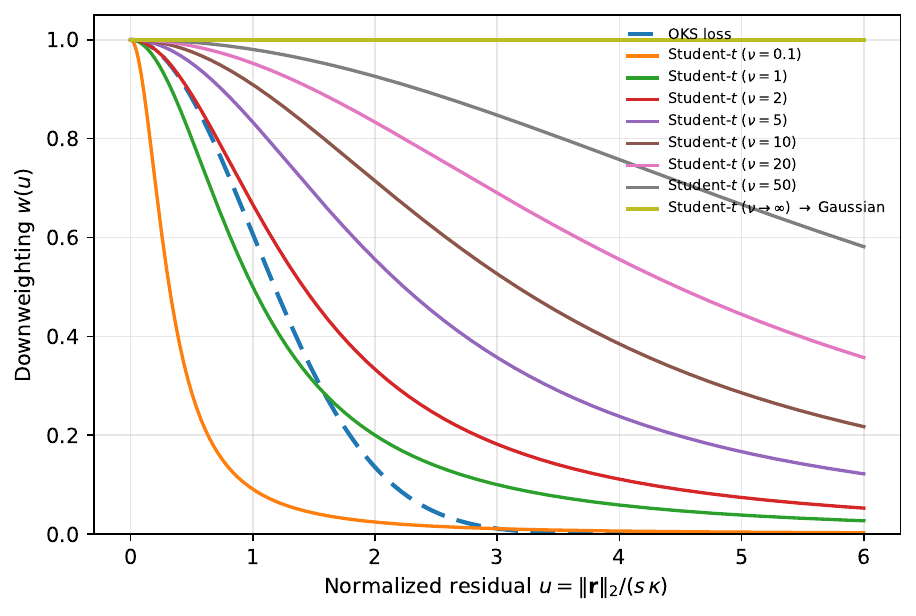}
    \caption{\textbf{Downweighting induced by \ac{oks}, Student-$t$, and Gaussian objectives.}
    We plot the downweighting function $w(u)\triangleq \tfrac{\|\psi(u)\|_2}{u}$ against the normalized residual magnitude $u=\tfrac{\|\mathbf r\|_2}{s\,\kappa}$.
    The \ac{oks} loss yields exponential downweighting, $w_{\mathrm{OKS}}(u)=\exp\!\big(\!-\!\tfrac{u^2}{2}\big)$, and therefore a redescending influence function.
    Student-$t$ induces polynomial downweighting, $w_t(u)\propto \tfrac{\nu}{\nu+u^2}$, with smaller $\nu$ corresponding to heavier tails and stronger outlier suppression; in the limit $\nu\to\infty$, Student-$t$ recovers the Gaussian case.
    By contrast, the Gaussian objective assigns constant weight to all residual magnitudes and therefore does not intrinsically suppress outliers.}
    \label{fig:oks_studentt_downweighting}
\end{figure}
Together, \cref{eq:w_gauss,eq:w_oks,eq:psi_t} clarify the qualitative gap between the Gaussian \ac{nll} and the \ac{oks} objective:
Gaussian training does not suppress outliers, whereas \ac{oks} yields a redescending influence and can leave large residuals less corrected.
Although Student-$t$ does not reproduce the redescending behavior of \ac{oks} exactly, it provides a valid heavy-tailed likelihood with tunable outlier downweighting through $\nu$, and a well-defined covariance for $\nu>2$.
We visualize downweighting for different $\nu$ in \cref{fig:oks_studentt_downweighting}.
This makes Student-$t$ a principled surrogate when a probabilistic model is desired while retaining the qualitative robustness absent under a Gaussian model.

\newpage
\section{Calibration Diagnostics for Heteroscedastic Keypoint Regression}
\label{app:calibration_diagnostics}

\subsection{Coverage Diagnostics and Average Coverage Error}
\label{app:coverage}
Calibrated predictive uncertainties should produce confidence regions whose empirical coverage matches their nominal coverage.
We therefore evaluate calibration both jointly, i.e., for the full bivariate predictive distribution, and marginally, i.e., for each standardized coordinate separately.

\paragraph{Joint Coverage.} 
Let $\mathbf{r}_{i,k}=\mathbf{y}_{i,k}-\hat{\mathbf{y}}_{i,k} \in \mathbb{R}^2$ denote the residual of keypoint $k$ of instance $i$, and let $\tilde{\mathbf{\Sigma}}_{i,k}$ and $\tilde{\mathbf{S}}_{i,k}$ denote the calibrated Gaussian covariance and Student-$t$ scale matrices, respectively.
For the Gaussian model, the squared Mahalanobis distance
\begin{equation}
\label{eq:joint_gauss_perfect_cal}
\mathbf{r}_{i,k}^\top \tilde{\mathbf{\Sigma}}_{i,k}^{-1}\mathbf{r}_{i,k}
\;\sim\;
\chi^2_2
\end{equation}
follows a $\chi^2_2$ distribution under correct calibration.
Thus, for a nominal coverage level $\alpha\in(0,1)$, the empirical joint coverage is (cf.\ \cref{eq:gauss_coverage})
\begin{equation*}
\hat{C}_{\mathcal{N}}(\alpha)
\;=\;
\frac{1}{\sum_{i\in\mathcal{E}}\, \lvert \mathcal{K}_i \rvert}
\sum_{i\in\mathcal{E},\, k\in\mathcal{K}_i}
\mathbbm{1}\!\left[
\mathbf{r}_{i,k}^\top \tilde{\mathbf{\Sigma}}_{i,k}^{-1}\mathbf{r}_{i,k} \leq \chi^2_{2,\alpha}
\right].
\end{equation*}

For the multivariate Student-$t$ model with $\nu_k$ degrees of freedom, the quadratic form satisfies
\begin{equation}
\label{eq:joint_t_perfect_cal}
\frac{1}{2}\,\mathbf{r}_{i,k}^\top \tilde{\mathbf{S}}_{i,k}^{-1}\mathbf{r}_{i,k}
\;\sim\; F_{2,\nu_k},
\end{equation}
so that the corresponding empirical joint coverage is (cf.\ \cref{eq:t_coverage})
\begin{equation*}
\hat{C}_{\mathcal{T}}(\alpha)
\;=\;
\frac{1}{\sum_{i\in\mathcal{E}}\, \lvert \mathcal{K}_i \rvert}
\sum_{i\in\mathcal{E},\, k\in\mathcal{K}_i}
\mathbbm{1}\!\left[
\mathbf{r}_{i,k}^\top \tilde{\mathbf{S}}_{i,k}^{-1}\mathbf{r}_{i,k}
\leq 2\,F^{-1}_{2,\nu_k}(\alpha)
\right].
\end{equation*}

Joint coverage plots are obtained by sweeping $\alpha$ over a grid $\mathcal{A}$ and plotting $\hat C(\alpha)$ against $\alpha$.
Perfect calibration corresponds to the diagonal $\hat C(\alpha)=\alpha$.

\paragraph{Marginal Coverage.}
Joint coverage can hide coordinate-wise miscalibration.
We therefore additionally evaluate each marginal dimension after standardization.

For the Gaussian model, the standardized residual satisfies
\begin{equation}
\label{eq:marg_gauss_perfect_cal}
\frac{r_{i,k,d}}{\sqrt{\tilde{\Sigma}_{i,k,dd}}}
\;\sim\;
\mathcal{N}(0,1)
\end{equation}
under correct calibration.

For the Student-$t$ model, the corresponding standardized residual satisfies
\begin{equation}
\label{eq:marg_t_perfect_cal}
\frac{r_{i,k,d}}{\sqrt{\tilde{S}_{i,k,dd}}}
\;\sim\;
\mathcal{T}_{\nu_k}(0,1)
\end{equation}
under correct calibration.

Thus, for a nominal marginal coverage level $\alpha$, the empirical marginal coverage is
\begin{equation}
\label{eq:marg_gauss_cov}
\hat{C}_{\mathcal{N},d}(\alpha)
\;=\;
\frac{1}{\sum_{i\in\mathcal{E}} \lvert \mathcal{K}_i \rvert}
\sum_{i\in\mathcal{E},\, k\in\mathcal{K}_i}
\mathbbm{1}\!\left[
\left|\tfrac{r_{i,k,d}}{\sqrt{\tilde{\Sigma}_{i,k,dd}}}\right|
\leq
\Phi^{-1}\!\left(\tfrac{1+\alpha}{2}\right)
\right],
\end{equation}
\begin{equation}
\label{eq:marg_t_cov}
\hat{C}_{\mathcal{T},d}(\alpha)
\;=\;
\frac{1}{\sum_{i\in\mathcal{E}} \lvert \mathcal{K}_i \rvert}
\sum_{i\in\mathcal{E},\, k\in\mathcal{K}_i}
\mathbbm{1}\!\left[
\left|\tfrac{r_{i,k,d}}{\sqrt{\tilde{S}_{i,k,dd}}}\right|
\leq
T^{-1}_{\nu_k}\!\left(\tfrac{1+\alpha}{2}\right)
\right],
\end{equation}
where $\Phi^{-1}$ and $T^{-1}_{\nu_k}$ denote the quantile functions of the standard normal and Student-$t$ distributions, respectively.

\paragraph{Average Coverage Error (ACE).}
To complement coverage plots with a scalar summary, we compute the \ac{ace} (cf.\ \cref{eq:ace}), defined as the mean absolute deviation between empirical and nominal coverage over a grid $\mathcal{A}$:
\begin{equation*}
\mathrm{ACE}
\;=\;
\frac{1}{\lvert\mathcal{A}\rvert}\sum_{\alpha\in\mathcal{A}}
\left|
\hat C(\alpha)-\alpha
\right|.
\end{equation*}
This corresponds to a discrete approximation of the area between the observed coverage curve and the diagonal of perfect calibration.
Lower ACE indicates better calibration, with $\mathrm{ACE}=0$ for a perfectly calibrated model.

\paragraph{Practical Protocol.}
In practice, we evaluate coverage on matched instances $\mathcal{E}$ using a fixed grid of nominal coverage levels, typically excluding extreme tails to reduce estimator variance, e.g.\ $\mathcal{A}=\{0.01,0.02,\dots,0.99\}$.
For each $\alpha\in\mathcal{A}$, we compute empirical coverage, plot observed versus nominal coverage, and summarize deviations using ACE.
Joint coverage assesses whether the full predicted covariance captures the correct confidence-region mass, whereas marginal coverage isolates per-dimension calibration.
Together, these diagnostics reveal covariance miscalibration, marginal variance miscalibration, and distributional mismatch that may not be apparent from a scalar \ac{nll} alone.

\subsection{Q--Q Goodness-of-Fit Analysis}
\label{app:qq}
We complement coverage diagnostics with \ac{qq} analysis to assess whether the full distribution of normalized residual statistics agrees with the assumed predictive law.
While coverage evaluates calibration at specific confidence levels, \ac{qq} analysis provides a distributional diagnostic that reveals deviations in shape and tail behavior.

\paragraph{Joint Q--Q Analysis.}
For the Gaussian model, the squared Mahalanobis distance in \cref{eq:joint_gauss_perfect_cal} follows a $\chi^2_2$ distribution under correct calibration.
We therefore collect the empirical statistic
\begin{equation}
u^{\mathcal N}_{i,k}
\;=\;
\mathbf{r}_{i,k}^\top \tilde{\mathbf{\Sigma}}_{i,k}^{-1}\mathbf{r}_{i,k},
\end{equation}
over all $i\in\mathcal E$ and $k\in\mathcal K_i$, sort them as
\begin{equation}
u^{\mathcal N}_{(1)} \le \cdots \le u^{\mathcal N}_{(N)},
\qquad
\text{with}
\qquad
N
\;=\;
\sum_{i\in\mathcal E} |\mathcal K_i|,
\end{equation}
and compare them to the theoretical quantiles
\begin{equation}
q^{\mathcal N}_n
\;=\;
\chi^2_{2,p_n},
\qquad
\text{with}
\qquad
p_n
\;=\;
\frac{n-\nicefrac{1}{2}}{N},
\qquad
n\in\{1,2,\dots,N\}.
\end{equation}
The Gaussian joint \ac{qq} plot is then given by $(q^{\mathcal N}_n,\,u^{\mathcal N}_{(n)})$.

For the Student-$t$ model, \cref{eq:joint_t_perfect_cal} yields the empirical statistic
\begin{equation}
u_{i,k}^{\mathcal T}
\;=\;
\mathbf{r}_{i,k}^\top \tilde{\mathbf{S}}_{i,k}^{-1}\mathbf{r}_{i,k}   
\end{equation}
and the corresponding theoretical quantiles
\begin{equation}
q_{n,k}^{\mathcal T}
\;=\;
2F^{-1}_{2,\nu_k}(p_n).  
\end{equation}
Since the degrees of freedom $\nu_k$ depend on the keypoint category, joint \ac{qq} analysis is performed per keypoint category.

\paragraph{Marginal Q--Q Analysis.}
For the Gaussian and Student-$t$ models, the standardized residuals in \cref{eq:marg_gauss_perfect_cal,eq:marg_t_perfect_cal} follow $\mathcal N(0,1)$ and the standard Student-$t$ distribution $\mathcal T_{\nu_k}(0,1)$ under correct calibration.

For the Gaussian model, we define
\begin{equation}
z^{\mathcal N}_{i,k,d}
\;=\;
\frac{r_{i,k,d}}{\sqrt{\tilde{\Sigma}_{i,k,dd}}},
\end{equation}
sort them as
\begin{equation}
z^{\mathcal N}_{(1),d} \le \cdots \le z^{\mathcal N}_{(N),d},
\end{equation}
and compare them to
\begin{equation}
q^{\mathcal N}_{n,d}
\;=\;
\Phi^{-1}(p_n),
\end{equation}
yielding \ac{qq} points $(q^{\mathcal N}_{n,d},\,z^{\mathcal N}_{(n),d})$.

For the Student-$t$ model, \cref{eq:marg_t_perfect_cal} yields the empirical statistic
\begin{equation}
z_{i,k,d}^{\mathcal T}
\;=\;
\frac{r_{i,k,d}}{\sqrt{\tilde{S}_{i,k,dd}}}    
\end{equation}
and the corresponding theoretical quantiles
\begin{equation}
q_{n,k,d}^{\mathcal T}
\;=\;
T^{-1}_{\nu_k}(p_n).    
\end{equation}
Since the degrees of freedom $\nu_k$ depend on the keypoint category, marginal \ac{qq} analysis is likewise performed per keypoint category.

\paragraph{Interpretation.}
Perfect agreement corresponds to alignment along the diagonal.
Systematic deviations indicate distributional mismatch, such as variance miscalibration or tail misspecification.
\ac{qq} analysis complements coverage diagnostics by revealing distributional deviations that may not be apparent from confidence-region probabilities alone.

\subsection{Variance-Calibration Diagnostics and Expected Normalized Calibration Error}
\label{app:variance_calibration}
Variance-calibration diagnostics assess whether predicted variances match empirical squared residuals across the range of heteroscedastic predictions.
Unlike coverage diagnostics, which evaluate confidence-region frequencies, variance calibration specifically probes whether the model assigns larger uncertainty to samples that indeed incur larger residuals.
We therefore evaluate variance calibration both jointly, i.e., for the full bivariate covariance, and marginally, i.e., for each coordinate separately.

\paragraph{Joint Variance Calibration.}
Let $\mathbf{r}_{i,k}=\mathbf{y}_{i,k}-\hat{\mathbf{y}}_{i,k}\in\mathbb{R}^2$ denote the residual of keypoint $k$ of instance $i$, and let $\tilde{\mathbf{\Sigma}}_{i,k}$ denote the calibrated predictive covariance, obtained either directly from the Gaussian model or from the calibrated Student-$t$ model via its corresponding covariance parameterization.
A necessary global moment-matching condition is
\begin{equation}
\label{eq:joint_var_global_app}
\frac{1}{\sum_{i\in\mathcal{E}} |\mathcal{K}_i|}
\sum_{i\in\mathcal{E},\,k\in\mathcal{K}_i}
\mathbf{r}_{i,k}\mathbf{r}_{i,k}^{\top}
\;\approx\;
\frac{1}{\sum_{i\in\mathcal{E}} |\mathcal{K}_i|}
\sum_{i\in\mathcal{E},\,k\in\mathcal{K}_i}
\tilde{\mathbf{\Sigma}}_{i,k}.
\end{equation}
However, this global condition can mask miscalibration across different uncertainty levels.
We therefore follow the binning strategy of \citet{levi2022ence} and sort predictions according to a scalar uncertainty score.

For the joint setting, we use the total variance $\operatorname{tr}(\tilde{\mathbf{\Sigma}}_{i,k})$, which is a rotation-invariant aggregate of the marginal variances and the natural multivariate counterpart of the scalar predictive variance.
We sort all evaluated keypoints by $\operatorname{tr}(\tilde{\mathbf{\Sigma}}_{i,k})$ and write $(n)$ for the corresponding sorted index such that
\begin{equation}
\label{eq:joint_var_sort_app}
\operatorname{tr}\!\left(\tilde{\mathbf{\Sigma}}_{(1)}\right)
\;\leq\;
\cdots
\;\leq\;
\operatorname{tr}\!\left(\tilde{\mathbf{\Sigma}}_{(N)}\right),
\qquad
\text{with}
\qquad
N
\;=\;
\sum_{i\in\mathcal{E}} |\mathcal{K}_i|.
\end{equation}
We then partition the sorted samples into $B$ equally sized bins $\{\mathcal{B}_b\}_{b=1}^B$ and define the mean empirical squared residual and the mean predicted total variance in bin $\mathcal{B}_b$ as
\begin{equation}
\label{eq:joint_var_bins_app}
\hat{e}_b
\;=\;
\frac{1}{|\mathcal{B}_b|}
\sum_{n\in\mathcal{B}_b}
\frac{1}{2}\|\mathbf{r}_{(n)}\|_2^2,
\qquad
\hat{v}_b
\;=\;
\frac{1}{|\mathcal{B}_b|}
\sum_{n\in\mathcal{B}_b}
\frac{1}{2}\operatorname{tr}\!\left(\tilde{\mathbf{\Sigma}}_{(n)}\right),
\end{equation}
where $\tfrac{1}{2}$ ensures that the joint quantities are on the same scale as the marginal variance-calibration diagnostics.
Joint variance-calibration plots are obtained by plotting $\hat{e}_b$ against $\hat{v}_b$ for $b=1,\dots,B$.
Perfect calibration is reflected by points lying on the diagonal $\hat{e}_b=\hat{v}_b$.

\paragraph{Marginal Variance Calibration.}
Joint variance calibration can hide coordinate-wise miscalibration.
We therefore additionally evaluate each marginal dimension separately.
For a fixed coordinate $d\in\{1,2\}$, we sort all evaluated keypoints by the predicted marginal variance $\tilde{\Sigma}_{i,k,dd}$ and again write $(n)$ for the corresponding sorted index such that
\begin{equation}
\label{eq:marg_var_sort_app}
\tilde{\Sigma}_{(1),dd}
\;\leq\;
\cdots
\;\leq\;
\tilde{\Sigma}_{(N),dd}.
\end{equation}
Using the same bin partition $\{\mathcal{B}_b\}_{b=1}^B$ on the sorted sequence, we define the mean empirical squared residual and mean predicted variance in bin $\mathcal{B}_b$ as
\begin{equation}
\label{eq:marg_var_bins_app}
\hat{e}_{b,d}
\;=\;
\frac{1}{|\mathcal{B}_b|}
\sum_{n\in\mathcal{B}_b}
r_{(n),d}^2,
\qquad
\hat{v}_{b,d}
\;=\;
\frac{1}{|\mathcal{B}_b|}
\sum_{n\in\mathcal{B}_b}
\tilde{\Sigma}_{(n),dd}.
\end{equation}
Marginal variance-calibration plots are obtained by plotting $\hat{e}_{b,d}$ against $\hat{v}_{b,d}$ for each coordinate $d$.
Perfect calibration is again reflected by alignment with the diagonal.

\paragraph{Expected Normalized Calibration Error (ENCE).}
To complement variance-calibration plots with a scalar summary, we follow \citet{levi2022ence} and compute the \ac{ence} on the corresponding root-mean quantities.
For the joint setting, we define
\begin{equation}
\label{eq:joint_ence_app}
\mathrm{ENCE}
\;=\;
\frac{1}{B}\sum_{b=1}^B
\frac{\left|\sqrt{\hat{e}_b}-\sqrt{\hat{v}_b}\right|}{\sqrt{\hat{v}_b}}.
\end{equation}
For the marginal setting, we analogously define
\begin{equation}
\label{eq:marg_ence_app}
\mathrm{ENCE}_{d}
\;=\;
\frac{1}{B}\sum_{b=1}^B
\frac{\left|\sqrt{\hat{e}_{b,d}}-\sqrt{\hat{v}_{b,d}}\right|}{\sqrt{\hat{v}_{b,d}}},
\qquad
d\in\{1,2\}.
\end{equation}
Lower ENCE indicates better agreement between predicted and empirical variance across uncertainty levels, with $\mathrm{ENCE}\approx0$ for a perfectly calibrated model.

\paragraph{Practical Protocol.}
In practice, we evaluate variance calibration on matched instances $\mathcal{E}$.
For the joint diagnostic, predictions are sorted by $\operatorname{tr}(\tilde{\mathbf{\Sigma}}_{i,k})$; for the marginal diagnostics, they are sorted separately by each predicted marginal variance $\tilde{\Sigma}_{i,k,dd}$ for $d\in\{1,2\}$.
The resulting sorted samples are partitioned into $B$ equally sized bins, and we compare predicted and empirical mean variances within each bin using variance-calibration plots and ENCE.
Joint variance calibration assesses whether the predicted aggregate variance tracks the aggregate squared residual, whereas marginal variance calibration isolates per-dimension variance quality.
Together, these diagnostics reveal whether uncertainty estimates are informative in the heteroscedastic sense, complementing coverage and \ac{qq} diagnostics.

\newpage
\section{Benchmark Experiments on COCO}
\label{app:coco}

\subsection{Implementation Details}
\label{app:coco_impl}

All COCO experiments use the official pretrained weights from SuperGradients~\citep{supergradients} for YOLO-NAS-Pose and from Ultralytics~\citep{jocher2026ultralyticsyolo26} for YOLO26-Pose, with input resolution ${640\times640}$.
We split the COCO val2017 set into a calibration subset (20\%) and an evaluation subset (80\%) using a fixed random seed.
All data augmentation is disabled during probabilistic head training and calibration, so that the learned dispersion reflects the clean residual distribution rather than augmentation artifacts.

\paragraph{Probabilistic Head Training.}
The backbone, neck, and original detection/pose heads are frozen; only the probabilistic heads are trainable.
We use AdamW with an initial learning rate of $2\times10^{-3}$, cosine schedule (final ratio $0.05$), and weight decay $10^{-6}$.
Training uses early stopping on NLL (patience 5).
Batch size is 128 for Nano variants and 32 for Large variants.
Mixed precision is disabled to ensure numerical stability of the Cholesky parameterization.

\paragraph{Calibration.}
Temperature scaling and Student-$t$ degree-of-freedom fitting are performed on the calibration subset with all model parameters frozen except the calibration parameters.
We use AdamW with a cosine schedule with early stopping.
The initial learning rate is $10^{-3}$ for Gaussian calibration and $5\times10^{-2}$ (YOLO-NAS-Pose) or $5\times10^{-3}$ (YOLO26-Pose) for Student-$t$ calibration.

\paragraph{Covariance Parameterization.}
Each keypoint's $2\times2$ dispersion matrix is parameterized via three raw outputs converted to a lower-triangular Cholesky factor $\mathbf{L}$, with $\mathbf{\Sigma}=\mathbf{L}\mathbf{L}^\top$.
Positive diagonal entries are enforced via softplus, and cross-correlation is bounded by $0.99\tanh(\cdot)$.

\paragraph{Hardware.}
Training and calibration were performed on NVIDIA A10 and A100 GPUs.
Inference latency measurements are reported in \cref{tab:coco_latency}.


\begin{figure}[p]
    \centering
    \includegraphics[width=0.85\linewidth]{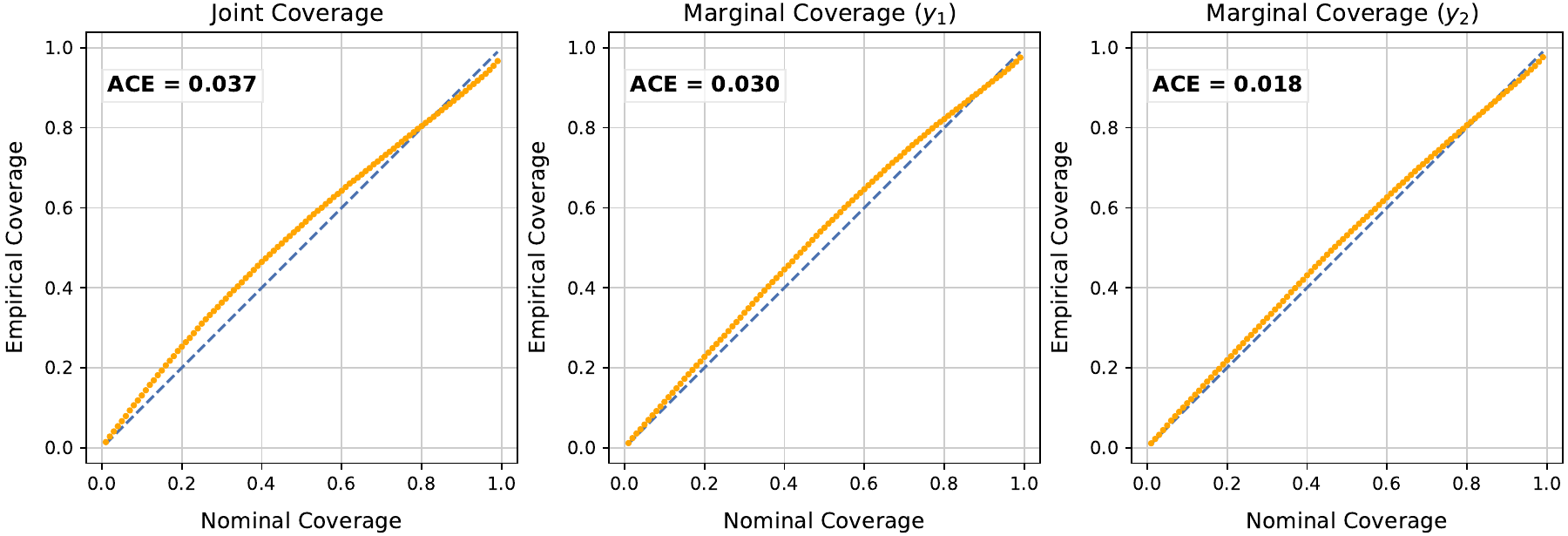}
    \vspace{0.15cm}
    \includegraphics[width=0.85\linewidth]{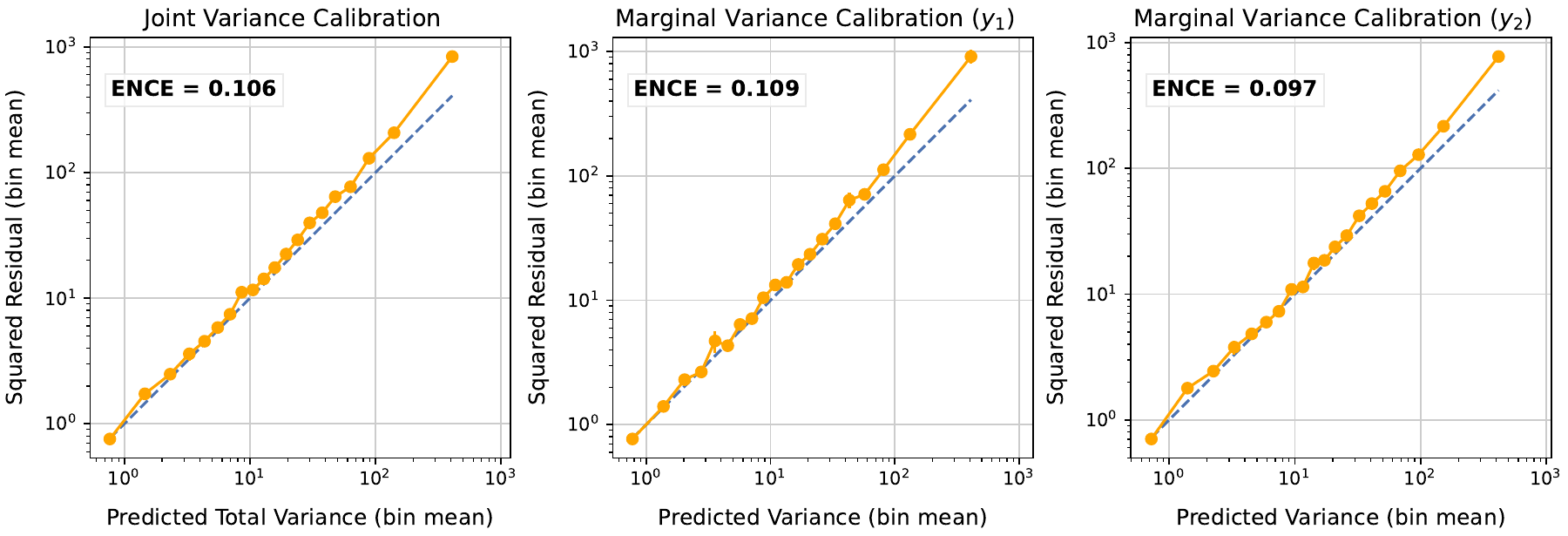}
    \vspace{0.15cm}
    \includegraphics[width=0.85\linewidth]{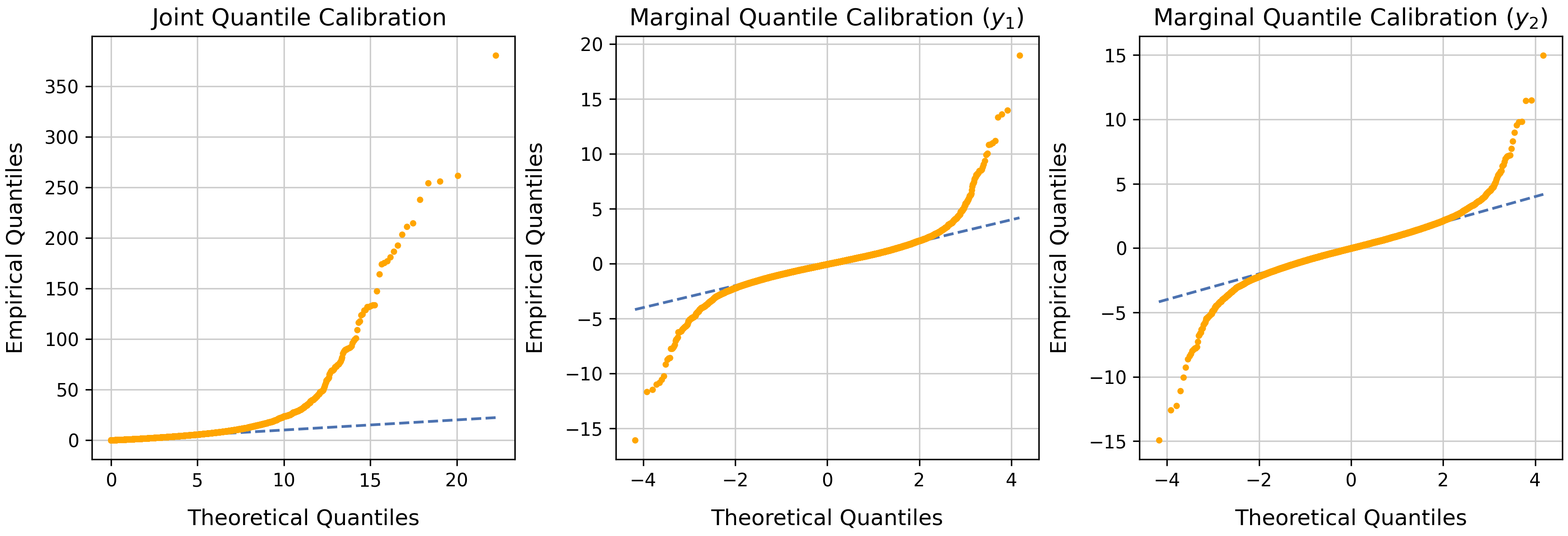}
    \caption{Calibration diagnostics on COCO for YOLO-NAS-Pose Nano: uncalibrated dispersion model.
    Top to bottom: coverage plots, variance calibration plots, and \ac{qq} plots.
    Each row shows the joint (left) and the marginal diagnostics (center and right).
    The corresponding aggregate metrics are reported in \cref{tab:coco_main_results}.}
    \label{fig:coco_nas_n_base_diag}
\end{figure}

\begin{figure}[p]
    \centering
    \includegraphics[width=0.85\linewidth]{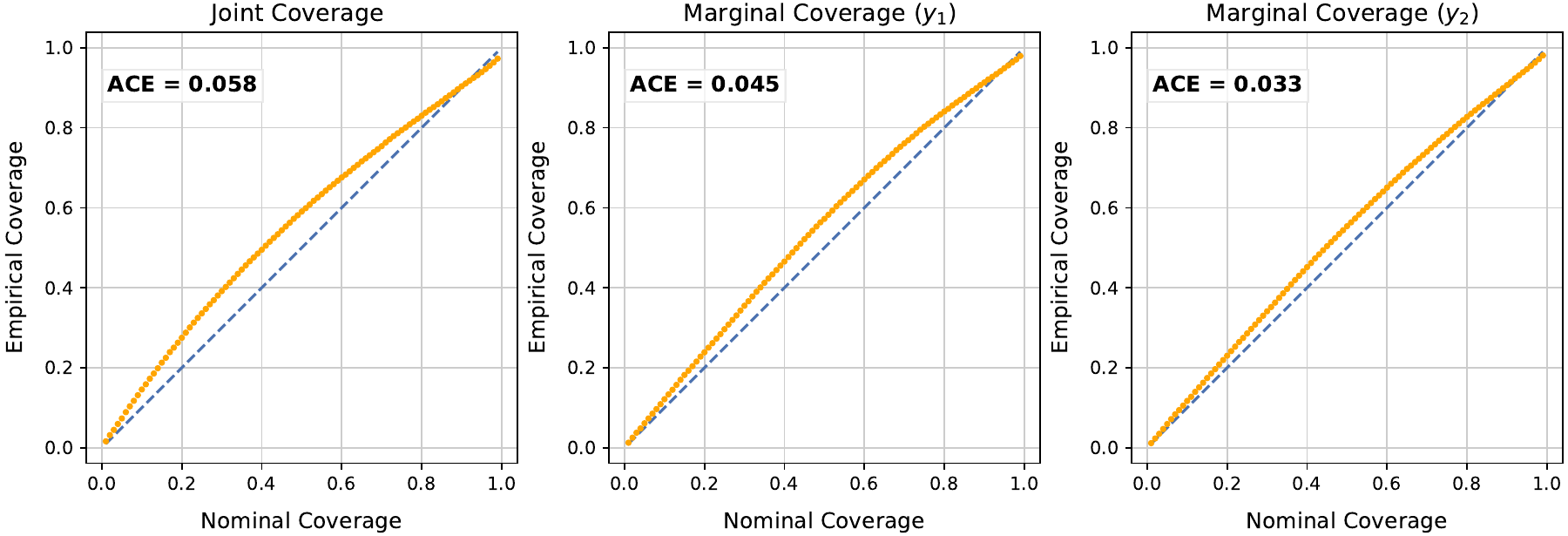}
    \vspace{0.15cm}
    \includegraphics[width=0.85\linewidth]{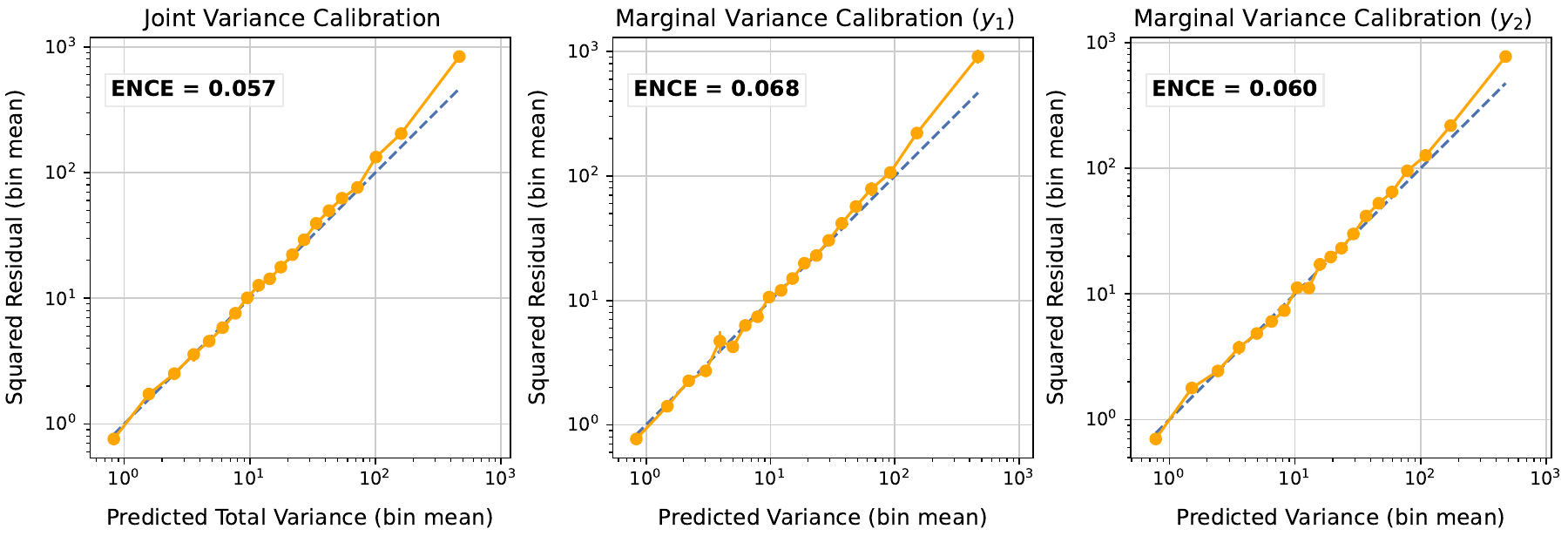}
    \vspace{0.15cm}
    \includegraphics[width=0.85\linewidth]{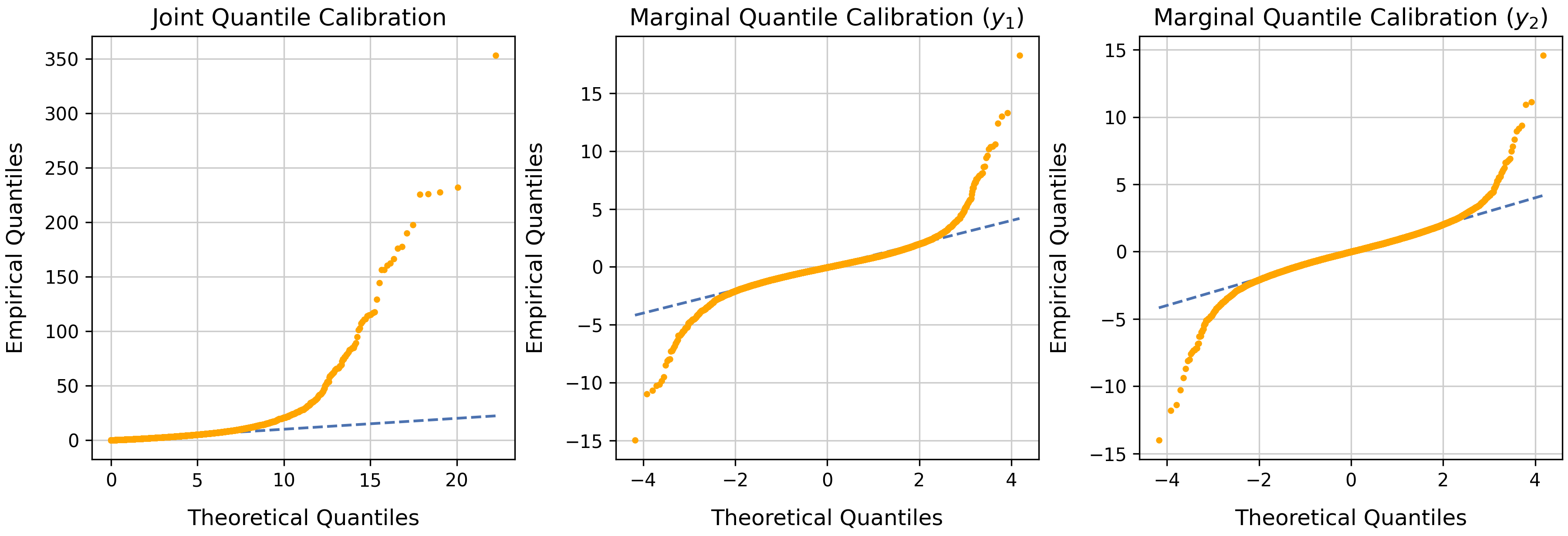}
    \caption{Calibration diagnostics on COCO for YOLO-NAS-Pose Nano: Gaussian calibration. Same layout as \cref{fig:coco_nas_n_base_diag}.}
    \label{fig:coco_nas_n_gauss_diag}
\end{figure}

\begin{figure}[p]
    \centering
    \includegraphics[width=0.85\linewidth]{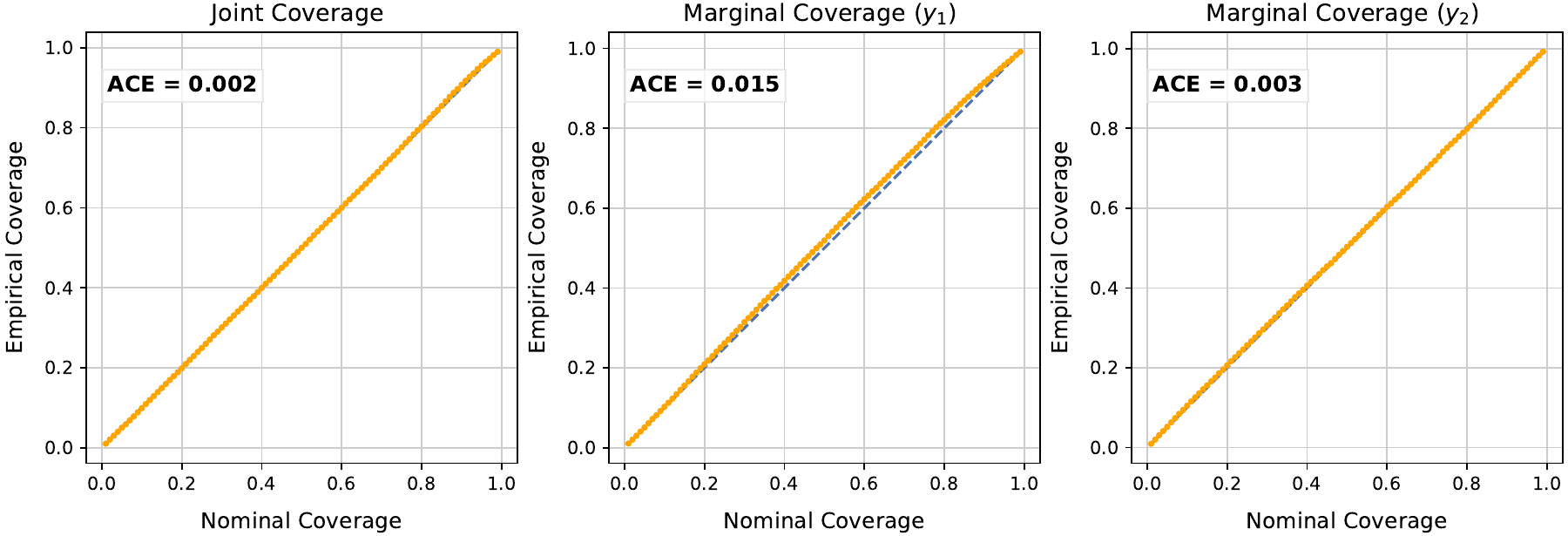}
    \vspace{0.15cm}
    \includegraphics[width=0.85\linewidth]{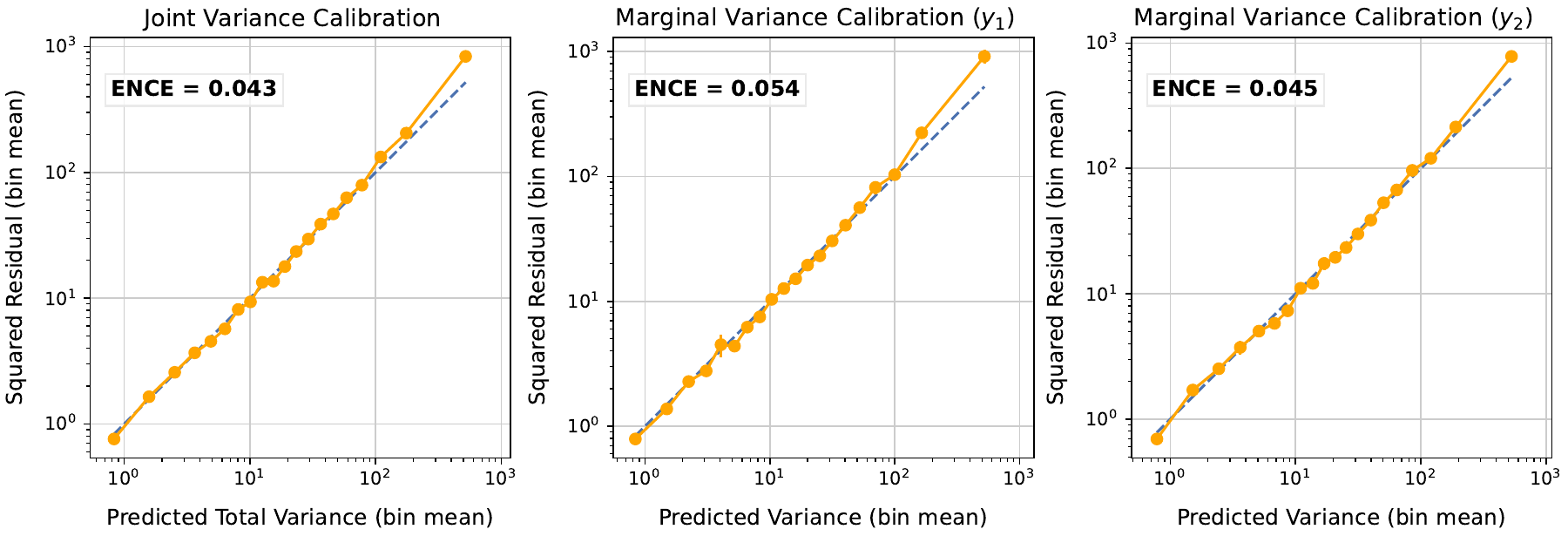}
    \vspace{0.15cm}
    \includegraphics[width=0.85\linewidth]{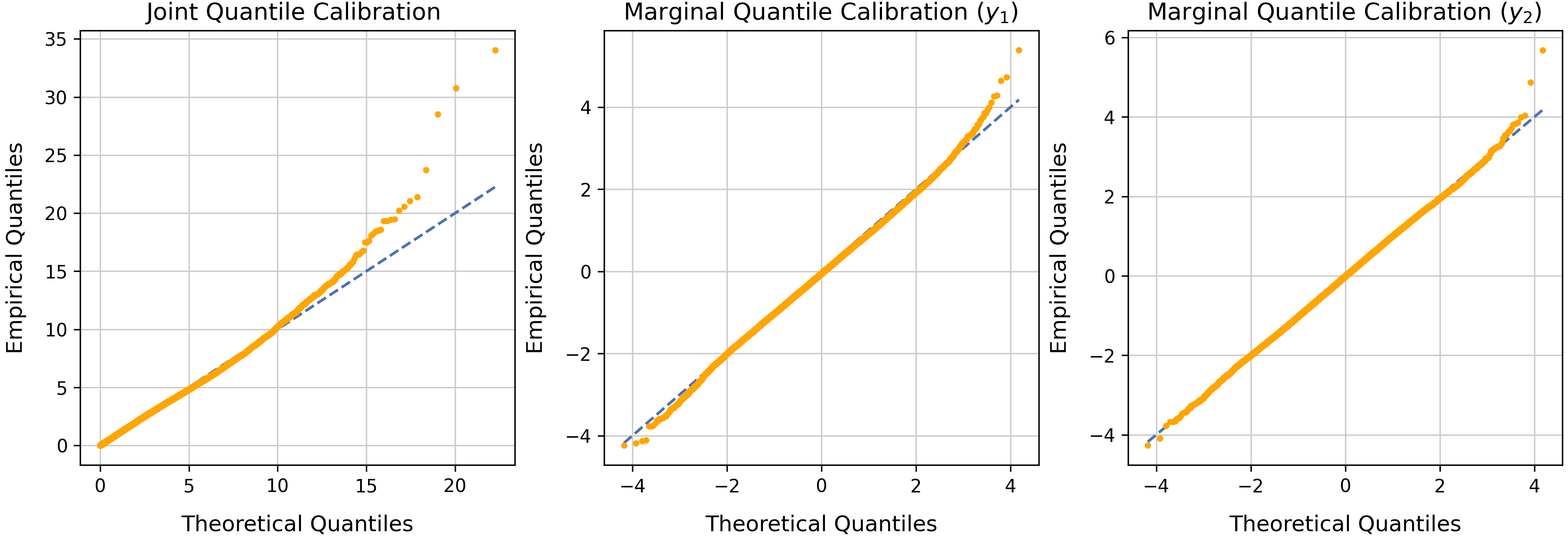}
    \caption{Calibration diagnostics on COCO for YOLO-NAS-Pose Nano: Student-$t$ calibration. Same layout as \cref{fig:coco_nas_n_base_diag}.}
    \label{fig:coco_nas_n_t_diag}
\end{figure}


\begin{figure}[p]
    \centering
    \includegraphics[width=0.85\linewidth]{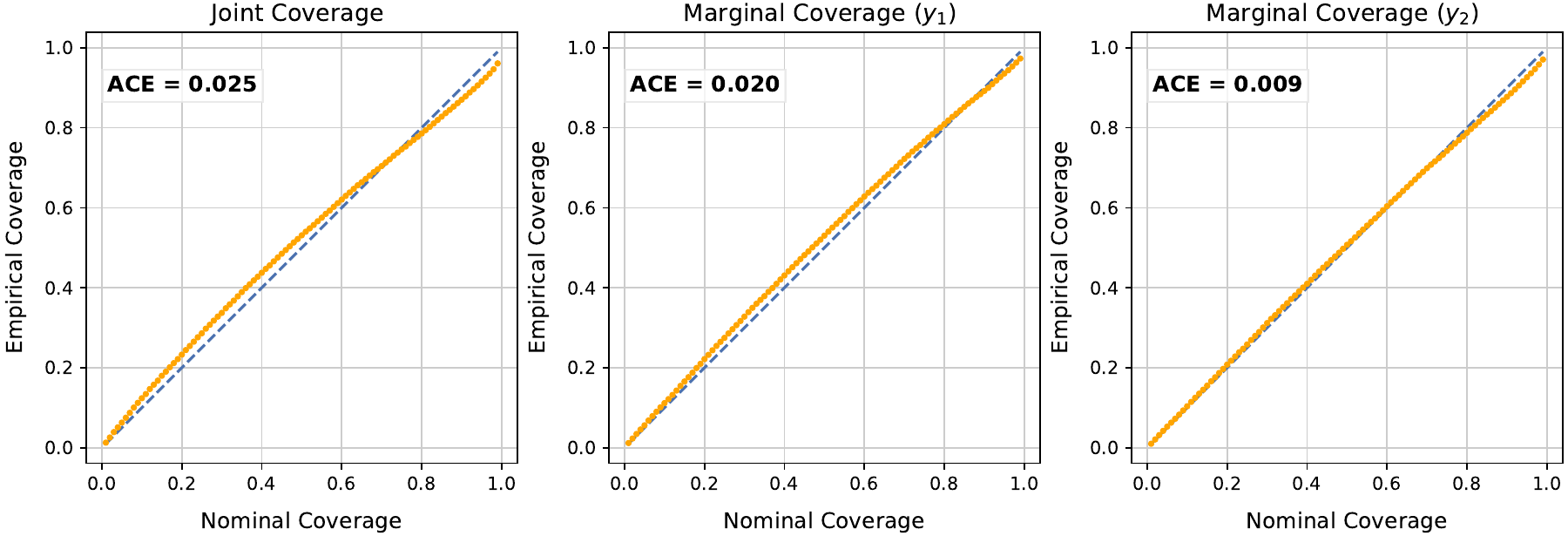}
    \vspace{0.15cm}
    \includegraphics[width=0.85\linewidth]{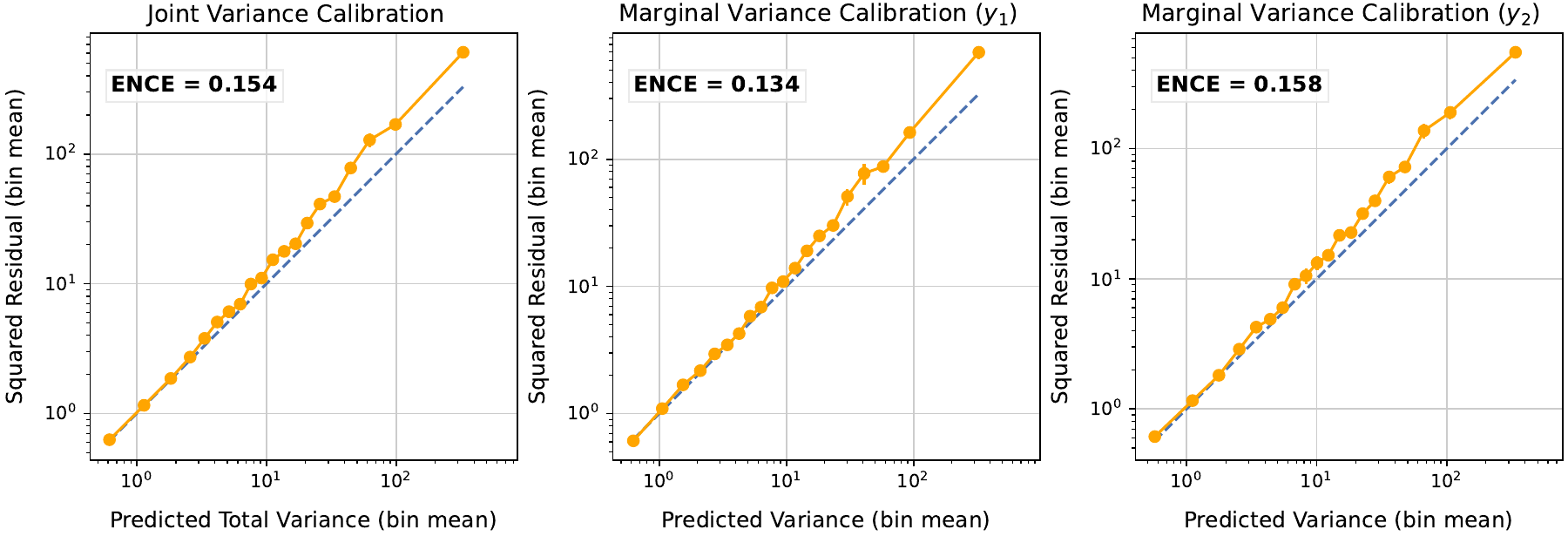}
    \vspace{0.15cm}
    \includegraphics[width=0.85\linewidth]{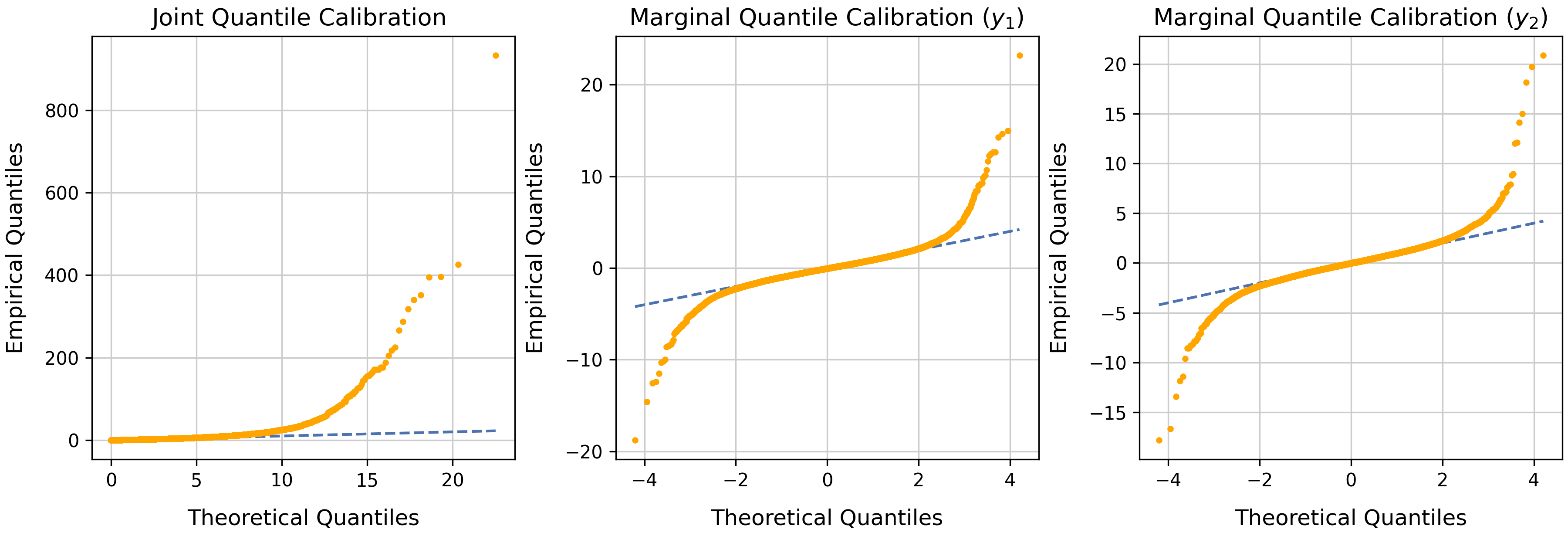}
    \caption{Calibration diagnostics on COCO for YOLO-NAS-Pose Large: uncalibrated dispersion model. Same layout as \cref{fig:coco_nas_n_base_diag}.}
    \label{fig:coco_nas_l_base_diag}
\end{figure}

\begin{figure}[p]
    \centering
    \includegraphics[width=0.85\linewidth]{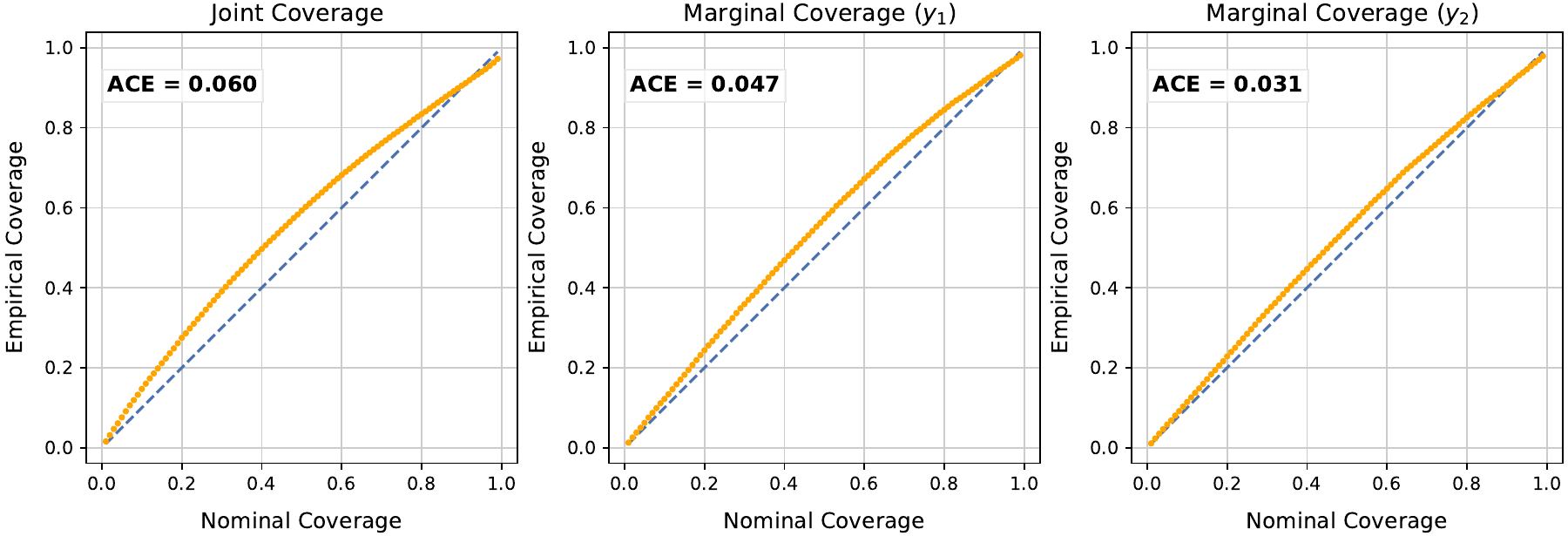}
    \vspace{0.15cm}
    \includegraphics[width=0.85\linewidth]{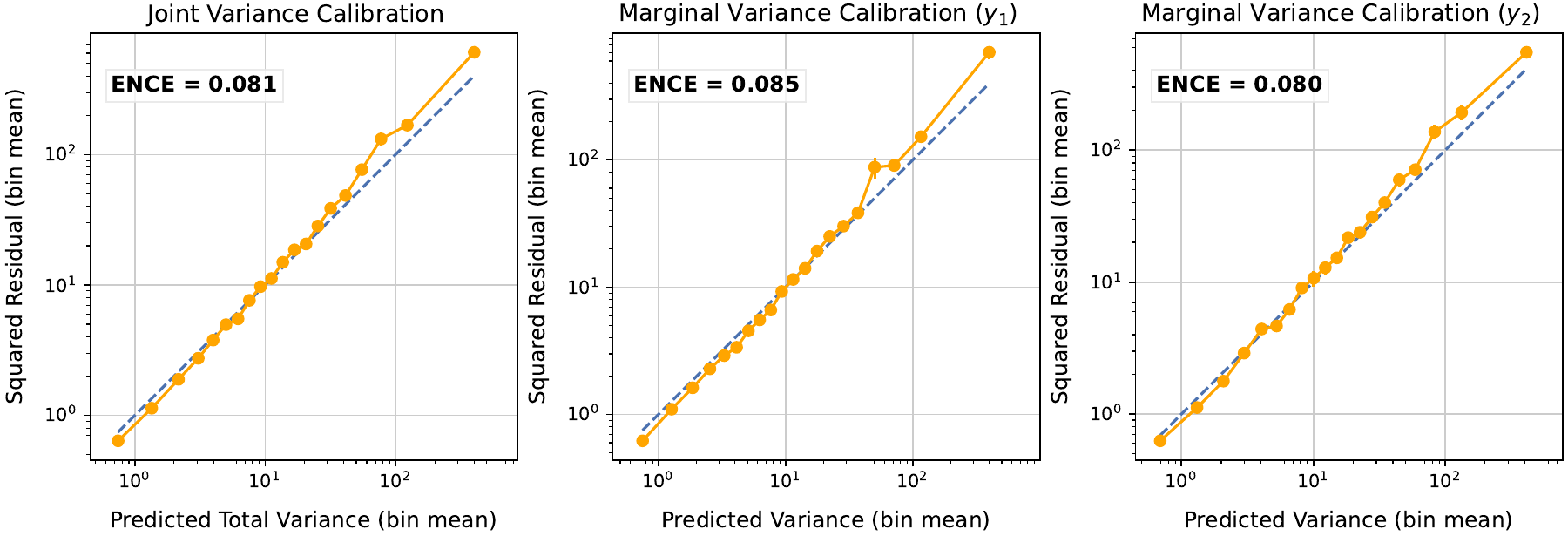}
    \vspace{0.15cm}
    \includegraphics[width=0.85\linewidth]{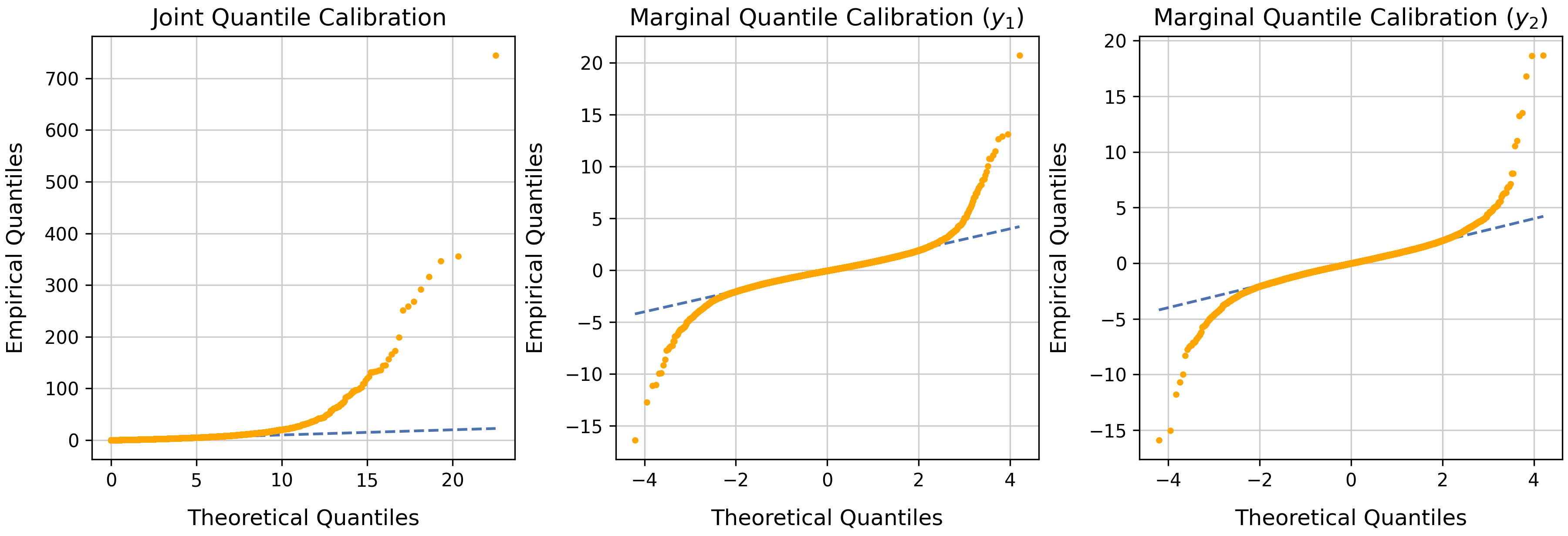}
    \caption{Calibration diagnostics on COCO for YOLO-NAS-Pose Large: Gaussian calibration. Same layout as \cref{fig:coco_nas_n_base_diag}.}
    \label{fig:coco_nas_l_gauss_diag}
\end{figure}

\begin{figure}[p]
    \centering
    \includegraphics[width=0.85\linewidth]{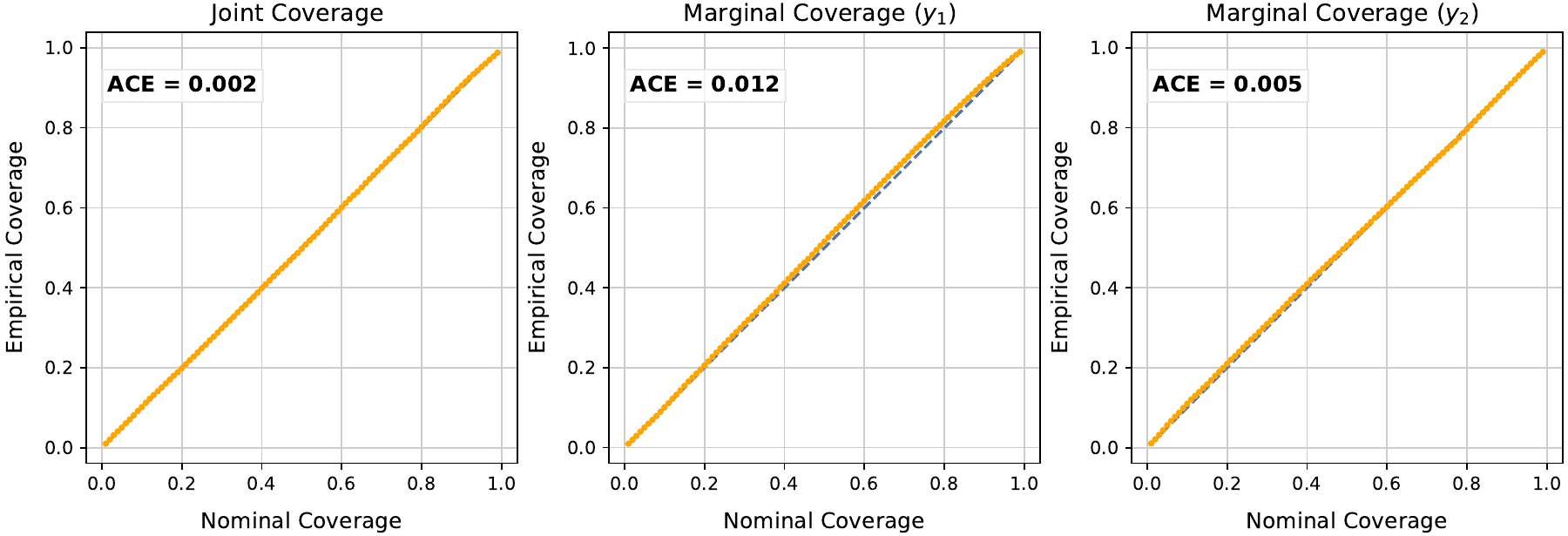}
    \vspace{0.15cm}
    \includegraphics[width=0.85\linewidth]{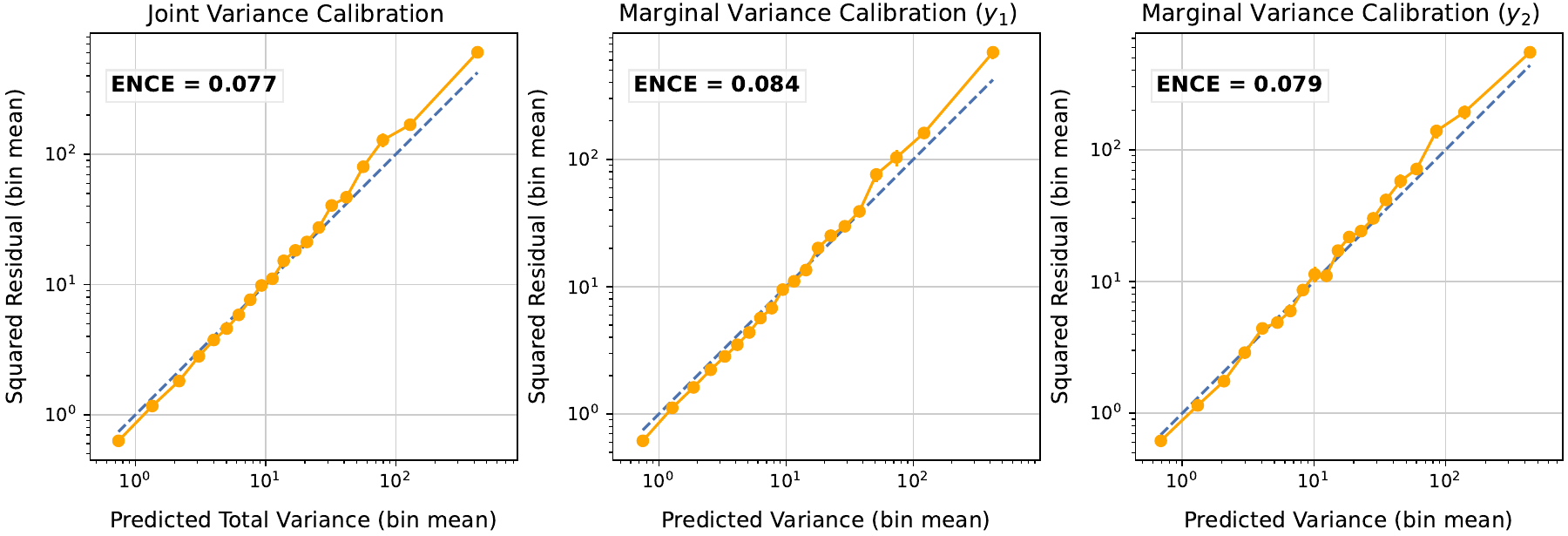}
    \vspace{0.15cm}
    \includegraphics[width=0.85\linewidth]{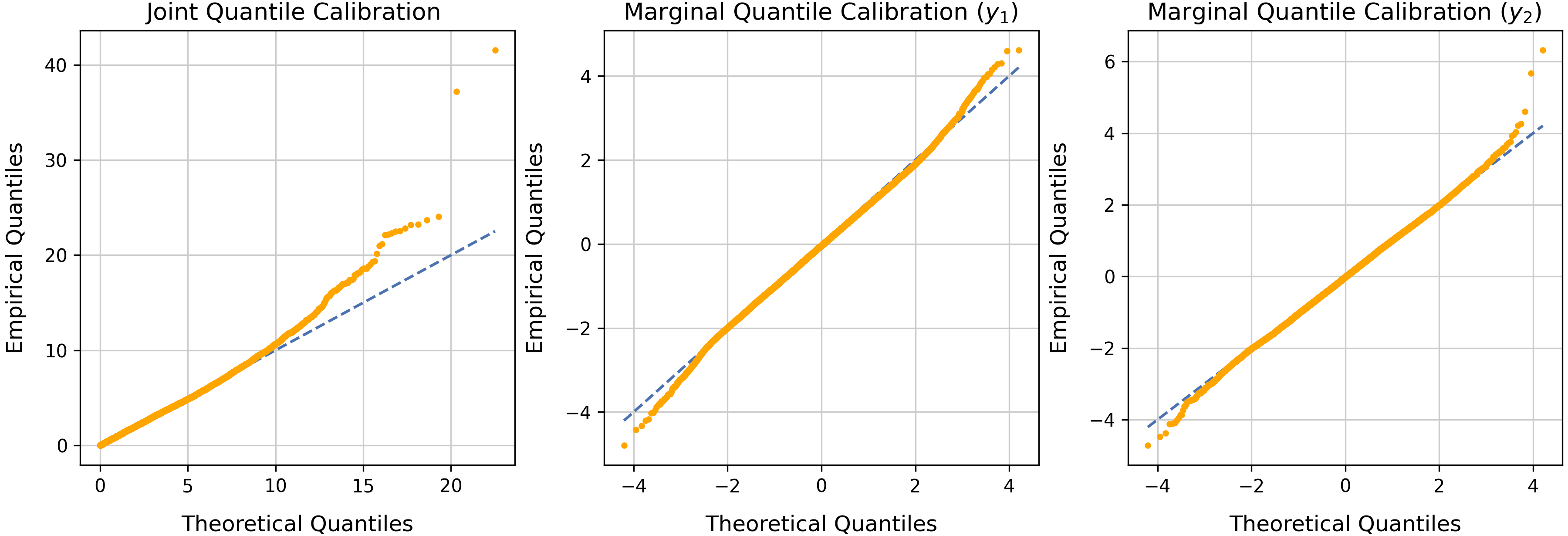}
    \caption{Calibration diagnostics on COCO for YOLO-NAS-Pose Large: Student-$t$ calibration. Same layout as \cref{fig:coco_nas_n_base_diag}.}
    \label{fig:coco_nas_l_t_diag}
\end{figure}


\begin{figure}[p]
    \centering
    \includegraphics[width=0.85\linewidth]{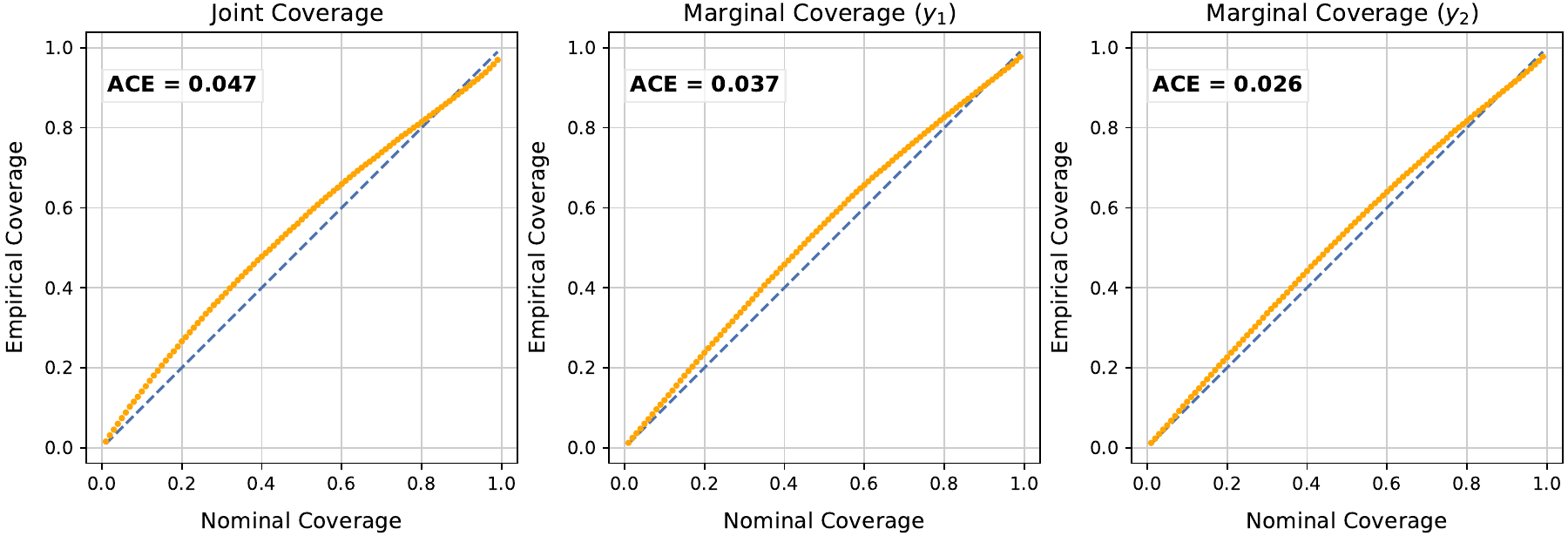}
    \vspace{0.15cm}
    \includegraphics[width=0.85\linewidth]{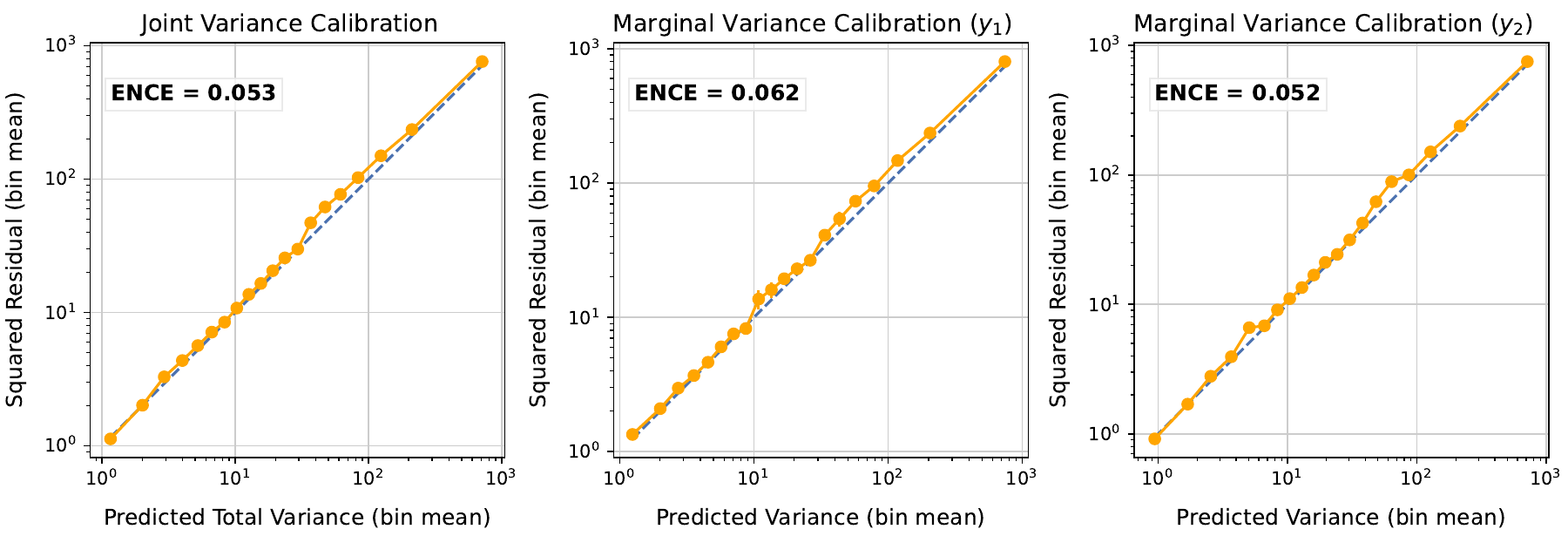}
    \vspace{0.15cm}
    \includegraphics[width=0.85\linewidth]{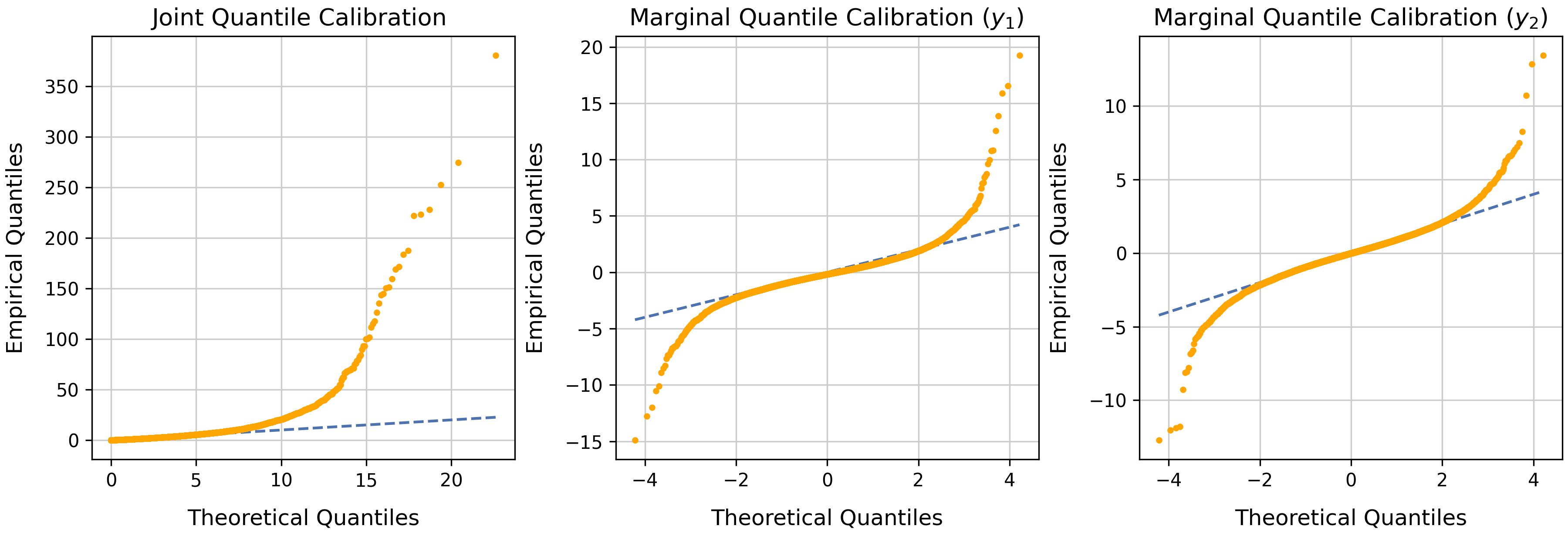}
    \caption{Calibration diagnostics on COCO for YOLO26-Pose Nano: uncalibrated dispersion model. Same layout as \cref{fig:coco_nas_n_base_diag}.}
    \label{fig:coco_y26_n_base_diag}
\end{figure}

\begin{figure}[p]
    \centering
    \includegraphics[width=0.85\linewidth]{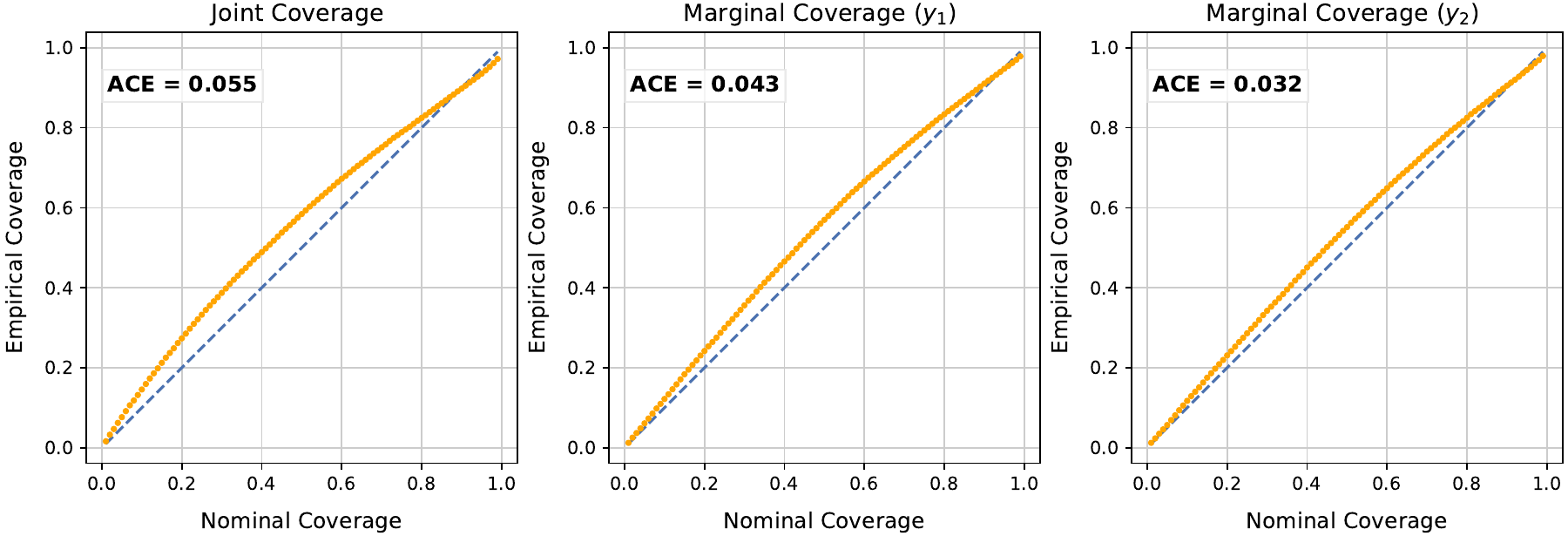}
    \vspace{0.15cm}
    \includegraphics[width=0.85\linewidth]{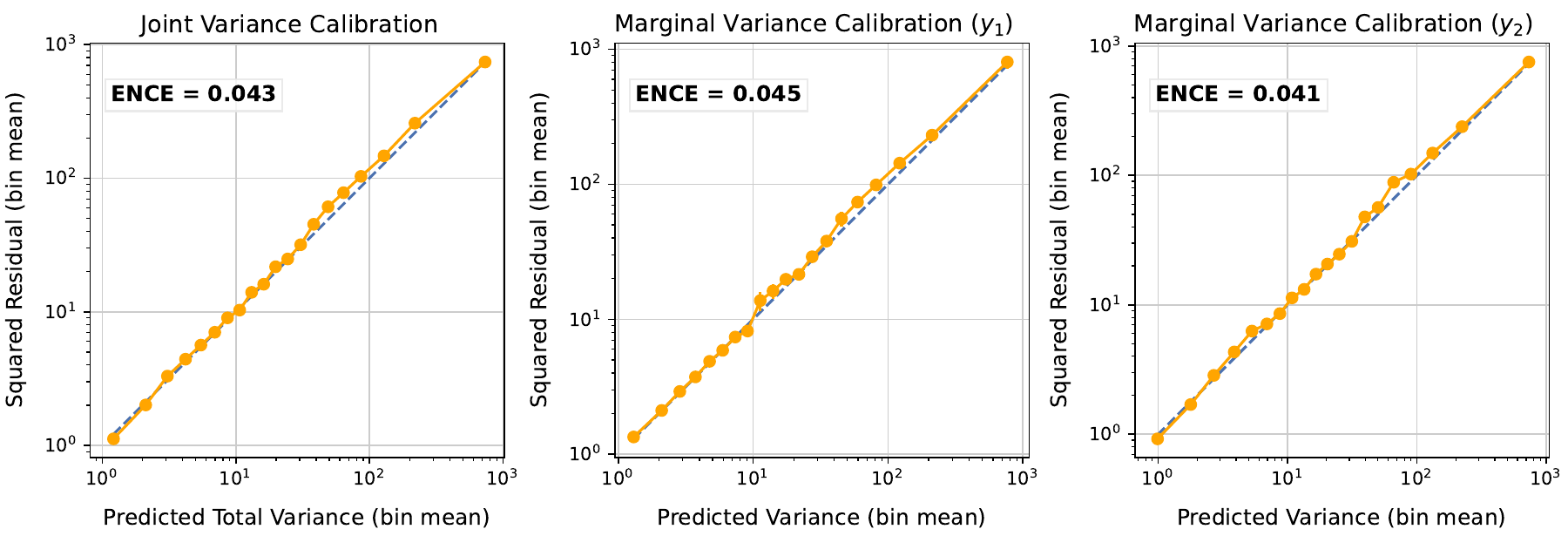}
    \vspace{0.15cm}
    \includegraphics[width=0.85\linewidth]{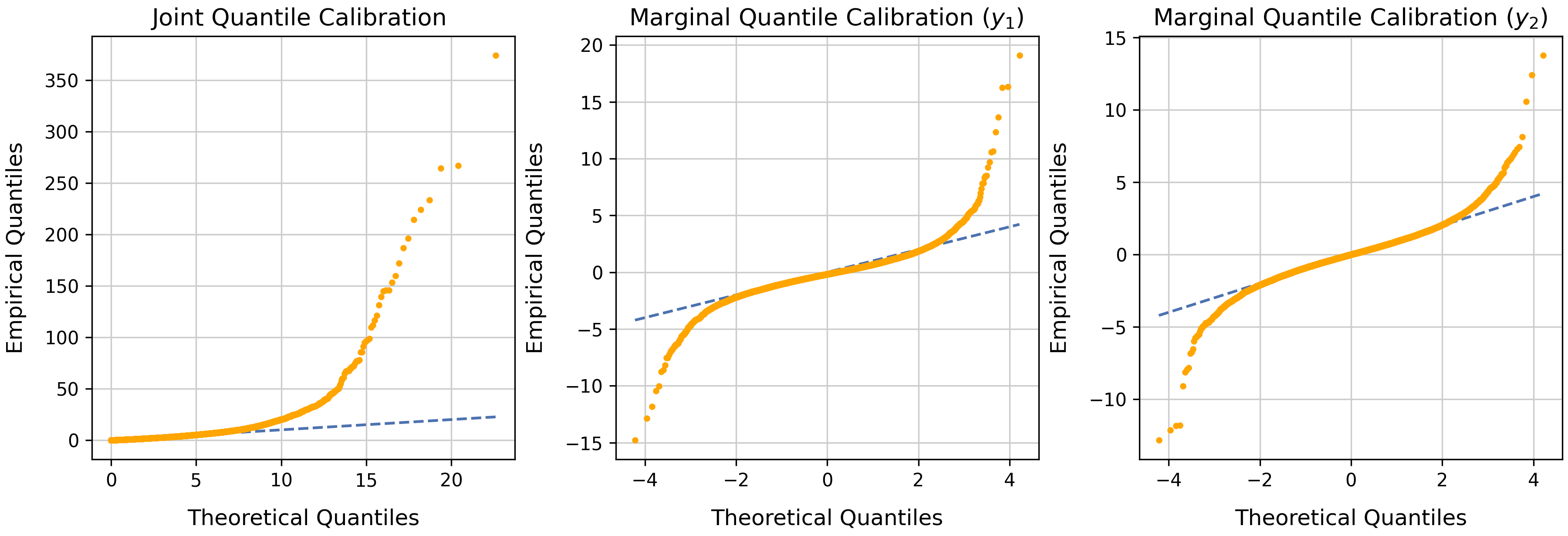}
    \caption{Calibration diagnostics on COCO for YOLO26-Pose Nano: Gaussian calibration. Same layout as \cref{fig:coco_nas_n_base_diag}.}
    \label{fig:coco_y26_n_gauss_diag}
\end{figure}

\begin{figure}[p]
    \centering
    \includegraphics[width=0.85\linewidth]{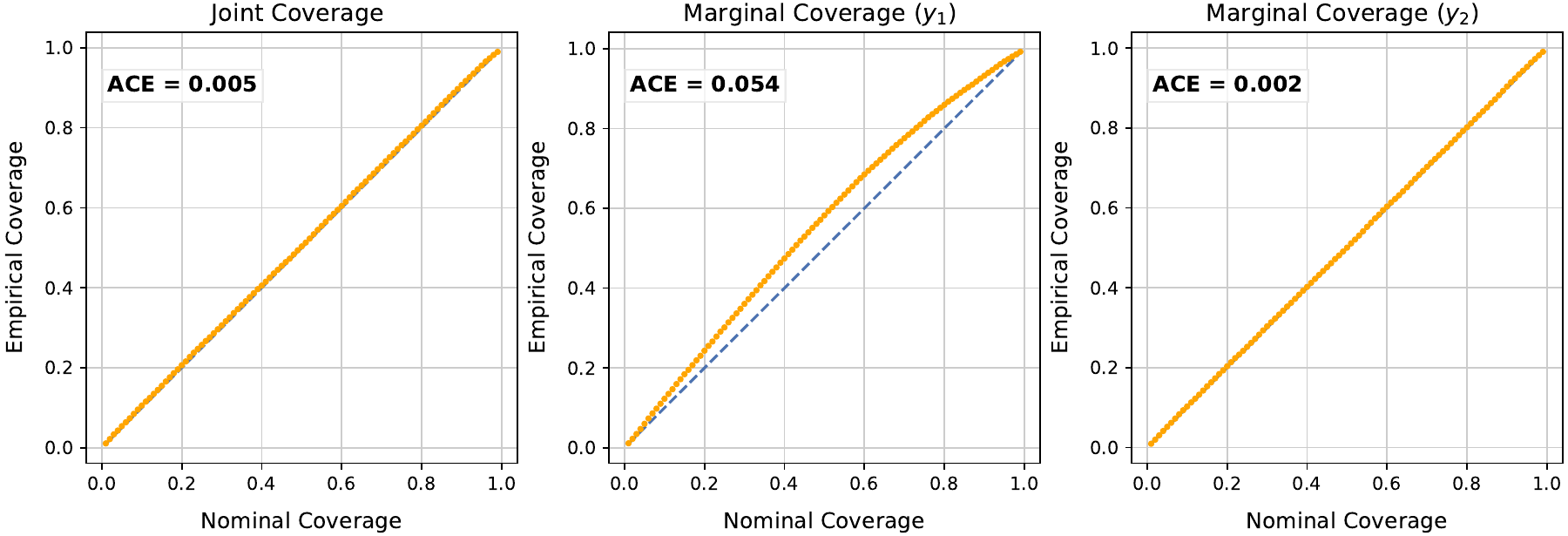}
    \vspace{0.15cm}
    \includegraphics[width=0.85\linewidth]{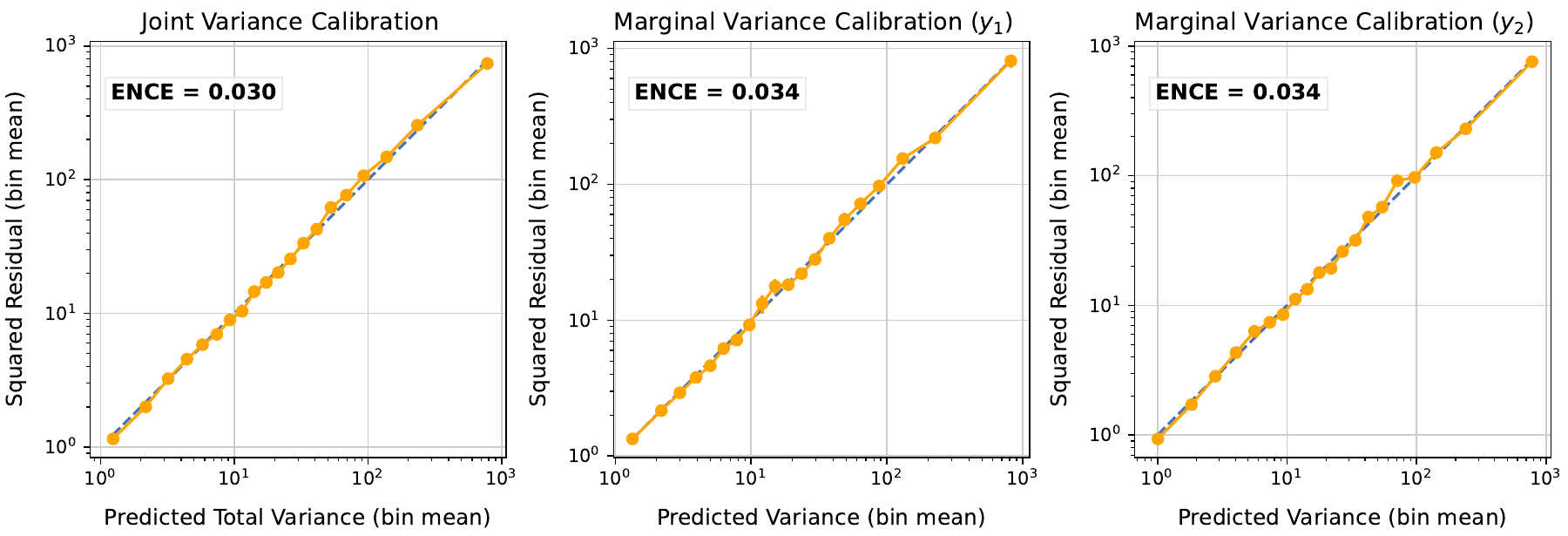}
    \vspace{0.15cm}
    \includegraphics[width=0.85\linewidth]{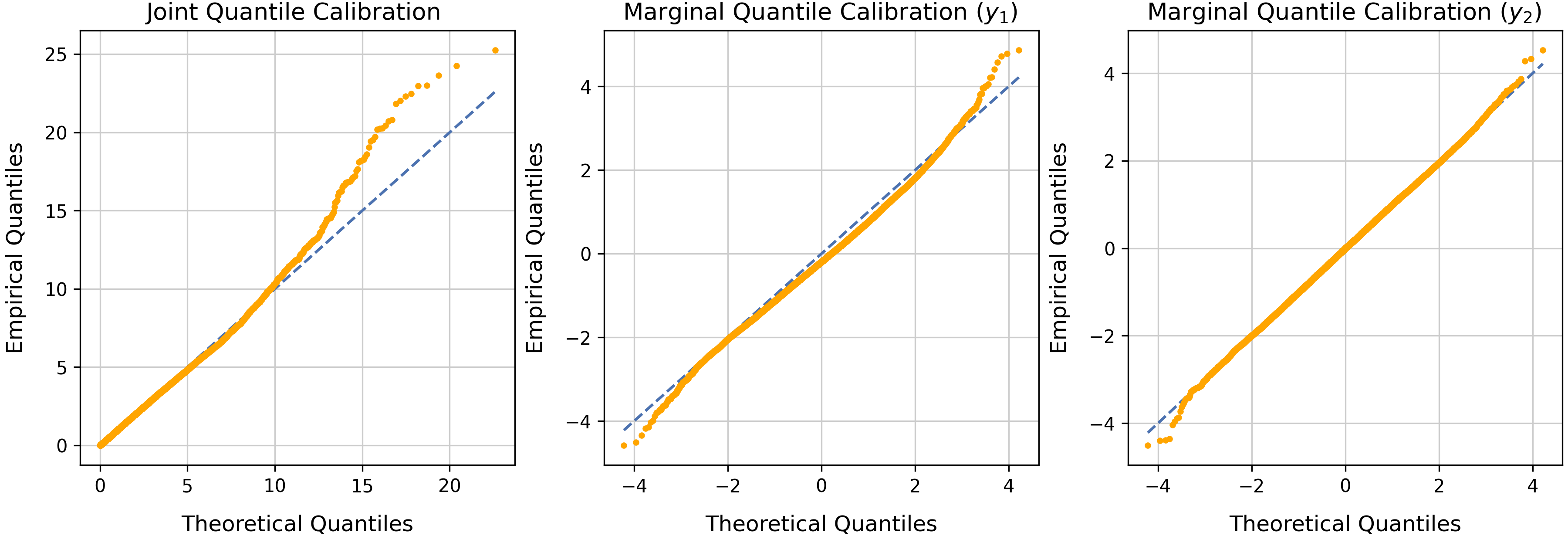}
    \caption{Calibration diagnostics on COCO for YOLO26-Pose Nano: Student-$t$ calibration. Same layout as \cref{fig:coco_nas_n_base_diag}.}
    \label{fig:coco_y26_n_t_diag}
\end{figure}


\begin{figure}[p]
    \centering
    \includegraphics[width=0.85\linewidth]{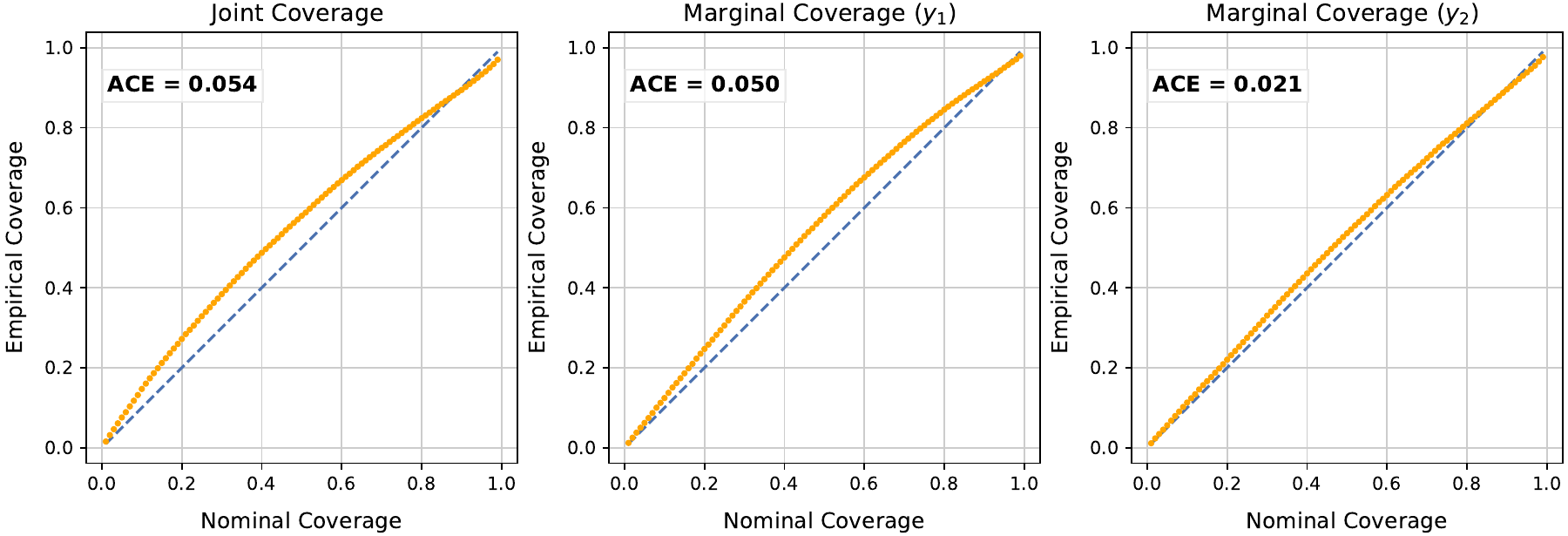}
    \vspace{0.15cm}
    \includegraphics[width=0.85\linewidth]{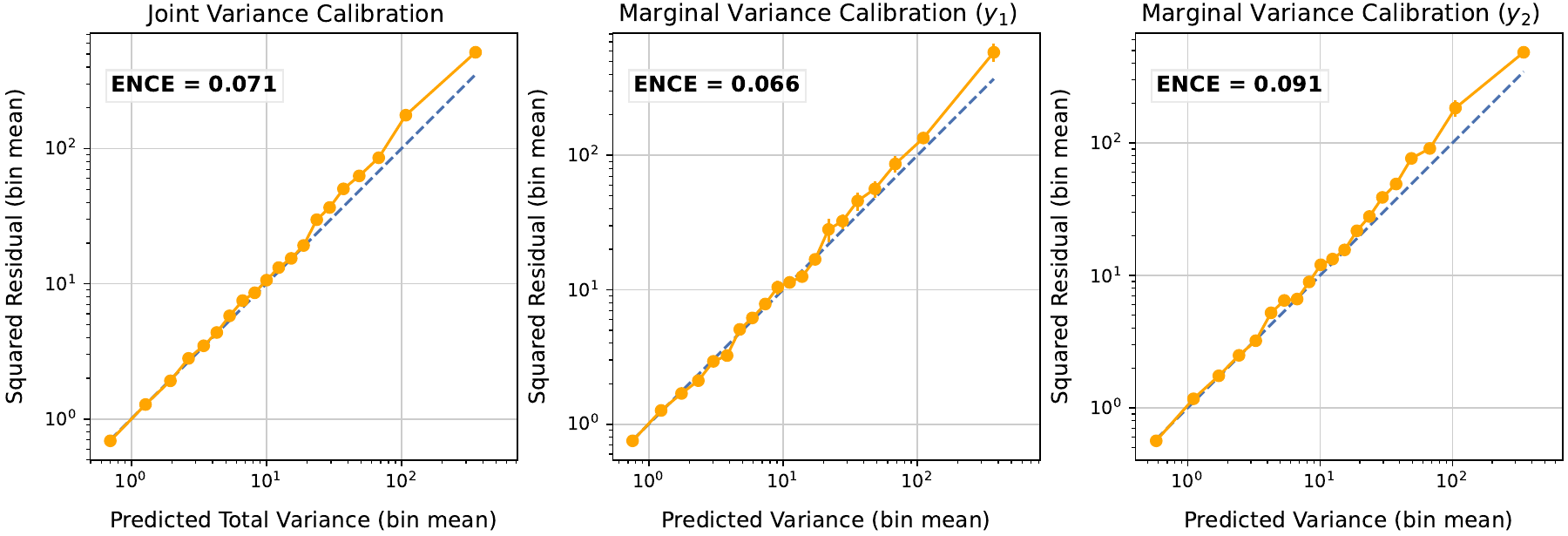}
    \vspace{0.15cm}
    \includegraphics[width=0.85\linewidth]{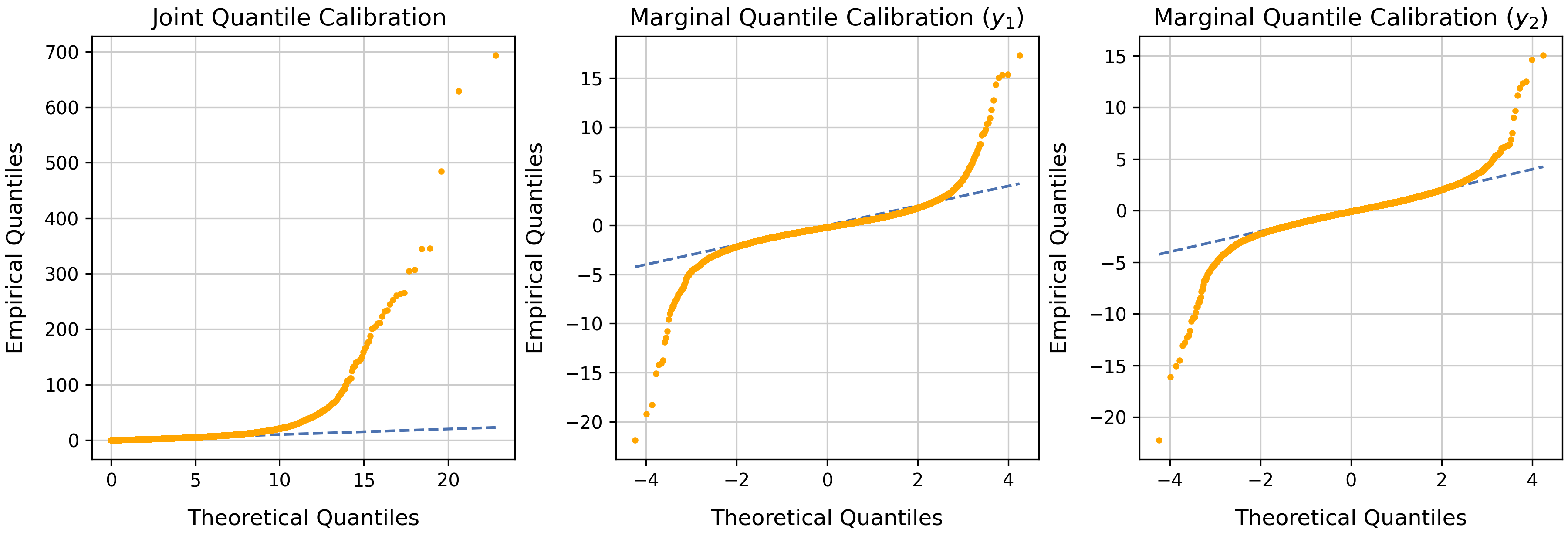}
    \caption{Calibration diagnostics on COCO for YOLO26-Pose Large: uncalibrated dispersion model. Same layout as \cref{fig:coco_nas_n_base_diag}.}
    \label{fig:coco_y26_l_base_diag}
\end{figure}

\begin{figure}[p]
    \centering
    \includegraphics[width=0.85\linewidth]{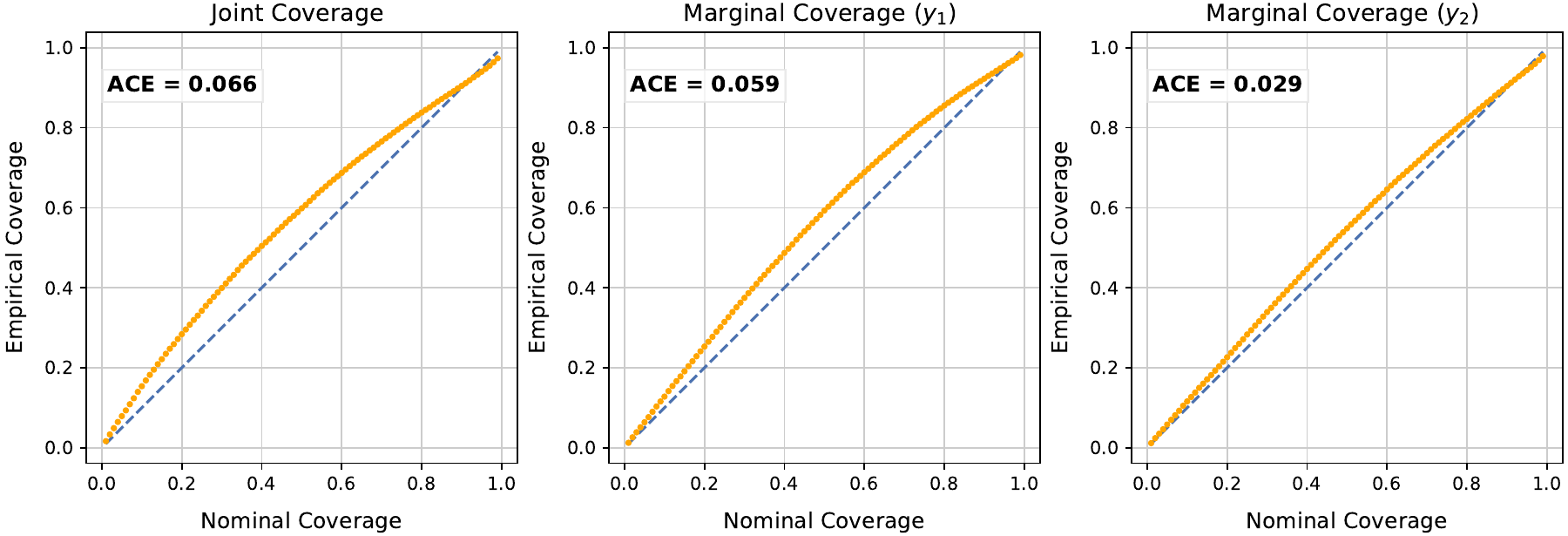}
    \vspace{0.15cm}
    \includegraphics[width=0.85\linewidth]{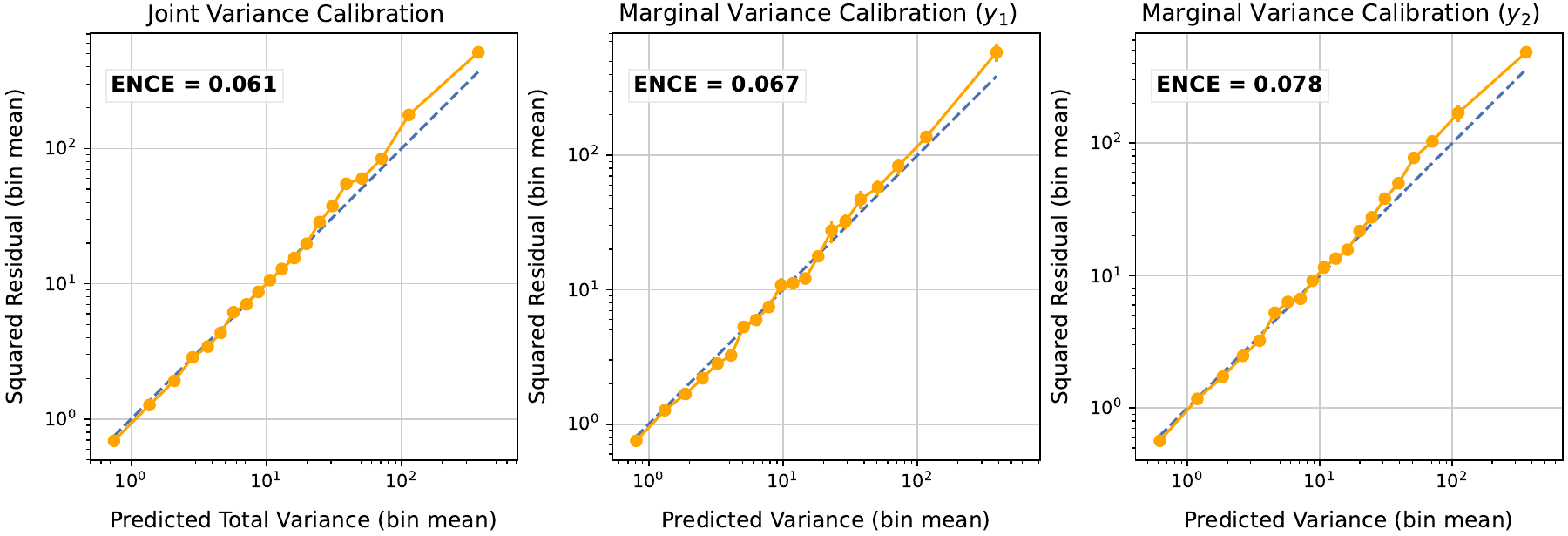}
    \vspace{0.15cm}
    \includegraphics[width=0.85\linewidth]{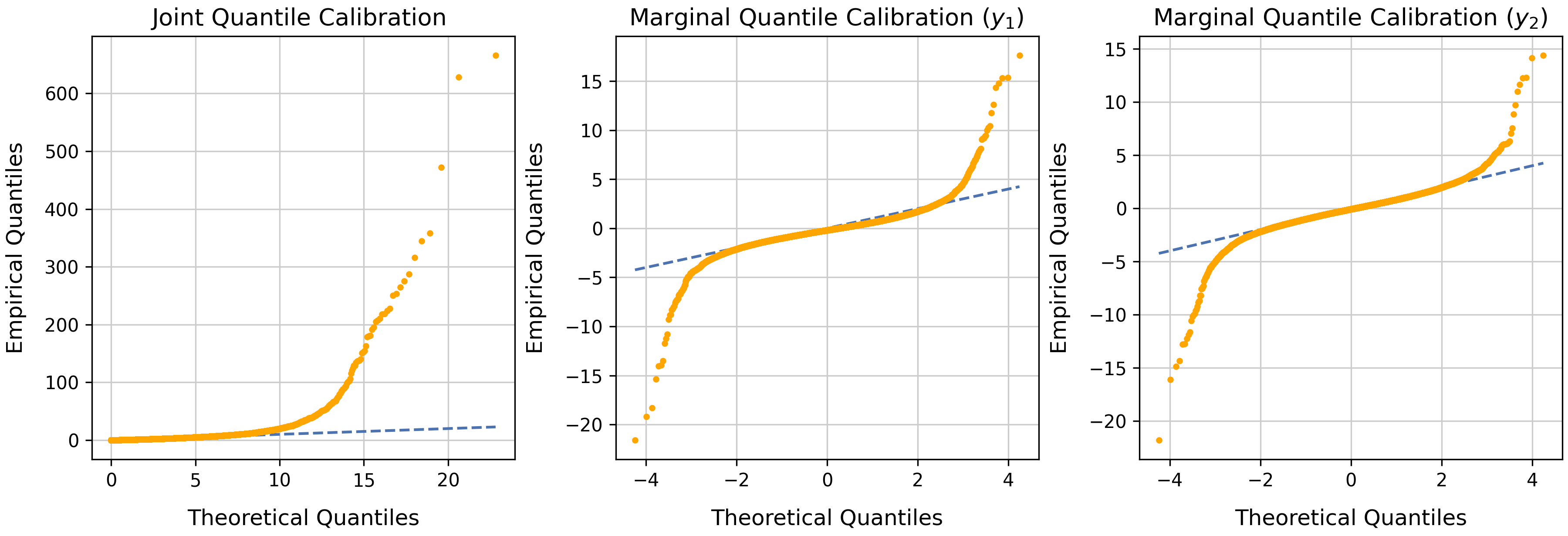}
    \caption{Calibration diagnostics on COCO for YOLO26-Pose Large: Gaussian calibration. Same layout as \cref{fig:coco_nas_n_base_diag}.}
    \label{fig:coco_y26_l_gauss_diag}
\end{figure}

\begin{figure}[p]
    \centering
    \includegraphics[width=0.85\linewidth]{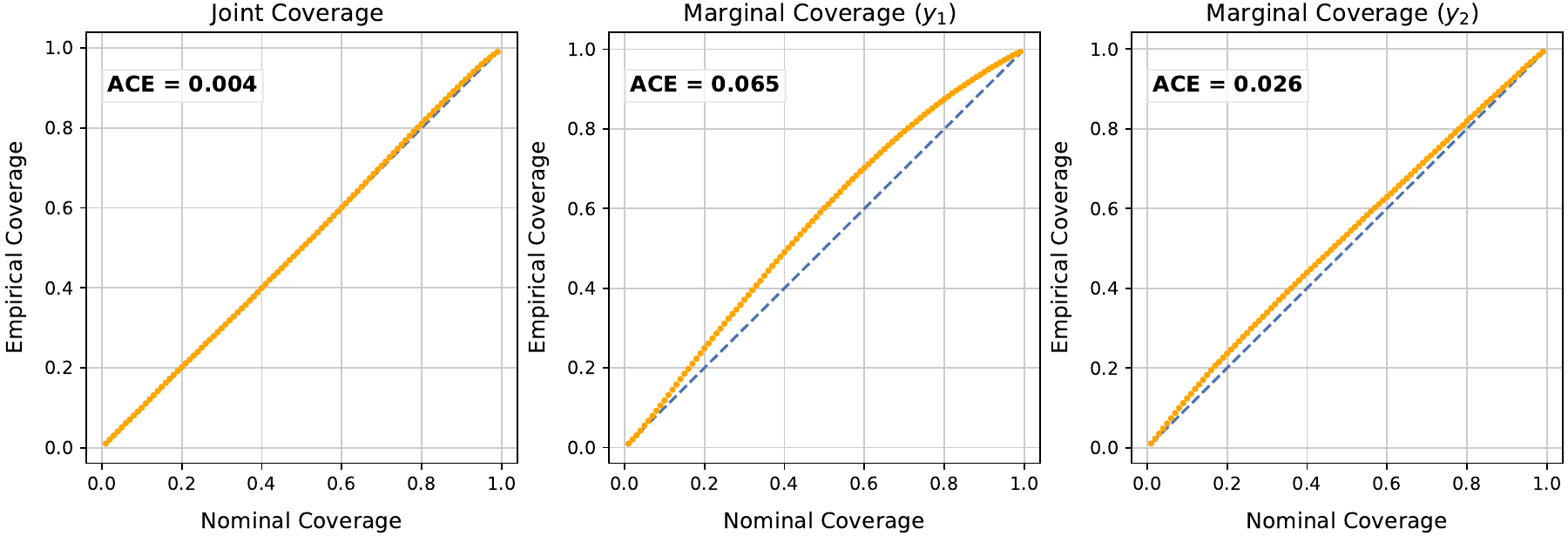}
    \vspace{0.15cm}
    \includegraphics[width=0.85\linewidth]{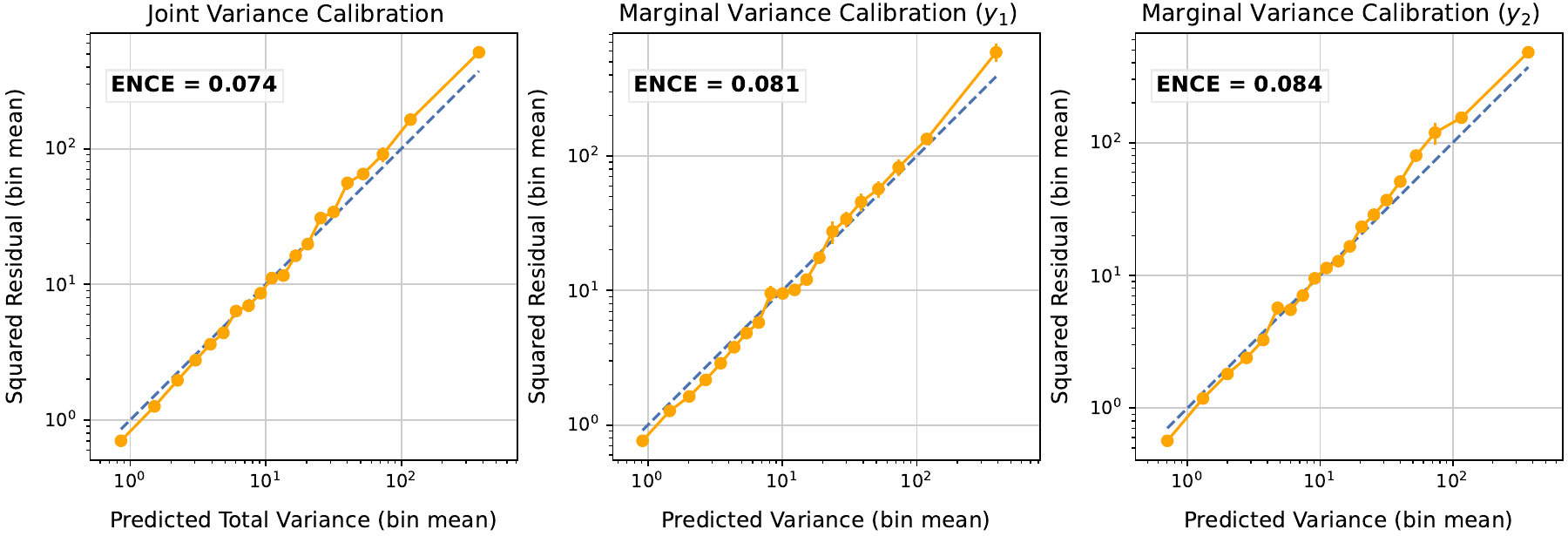}
    \vspace{0.15cm}
    \includegraphics[width=0.85\linewidth]{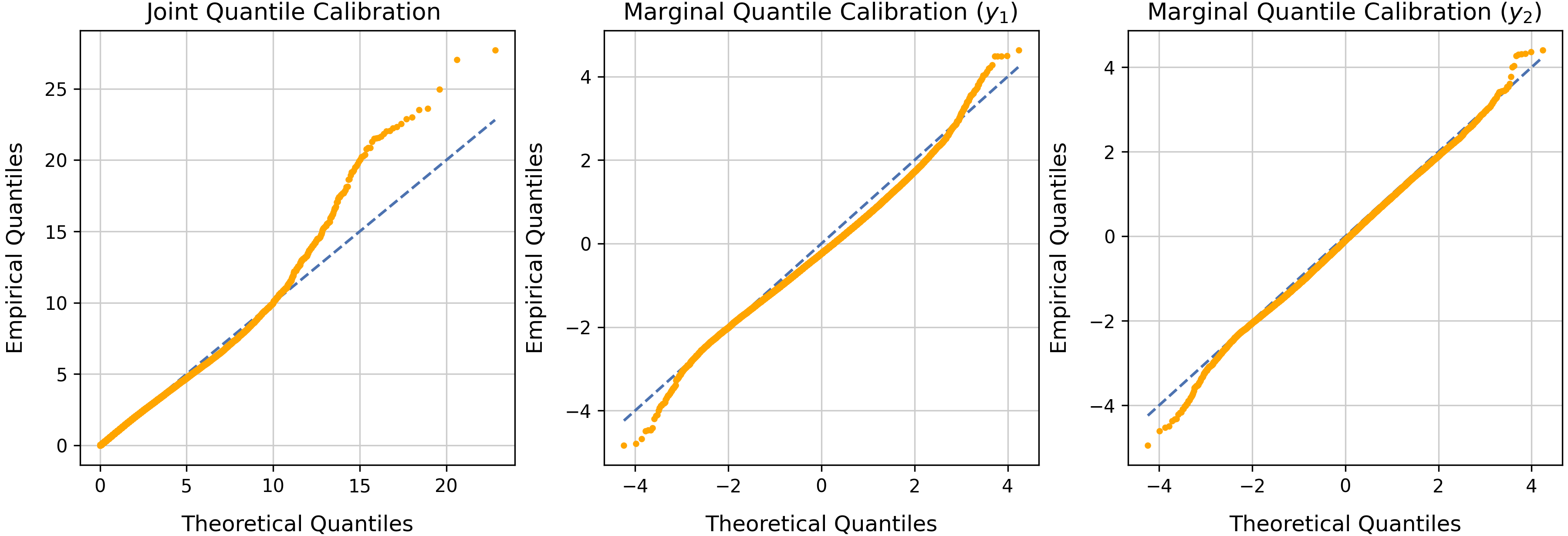}
    \caption{Calibration diagnostics on COCO for YOLO26-Pose Large: Student-$t$ calibration. Same layout as \cref{fig:coco_nas_n_base_diag}.}
    \label{fig:coco_y26_l_t_diag}
\end{figure}


\begin{figure}[p]
    \centering
    \includegraphics[width=0.85\linewidth]{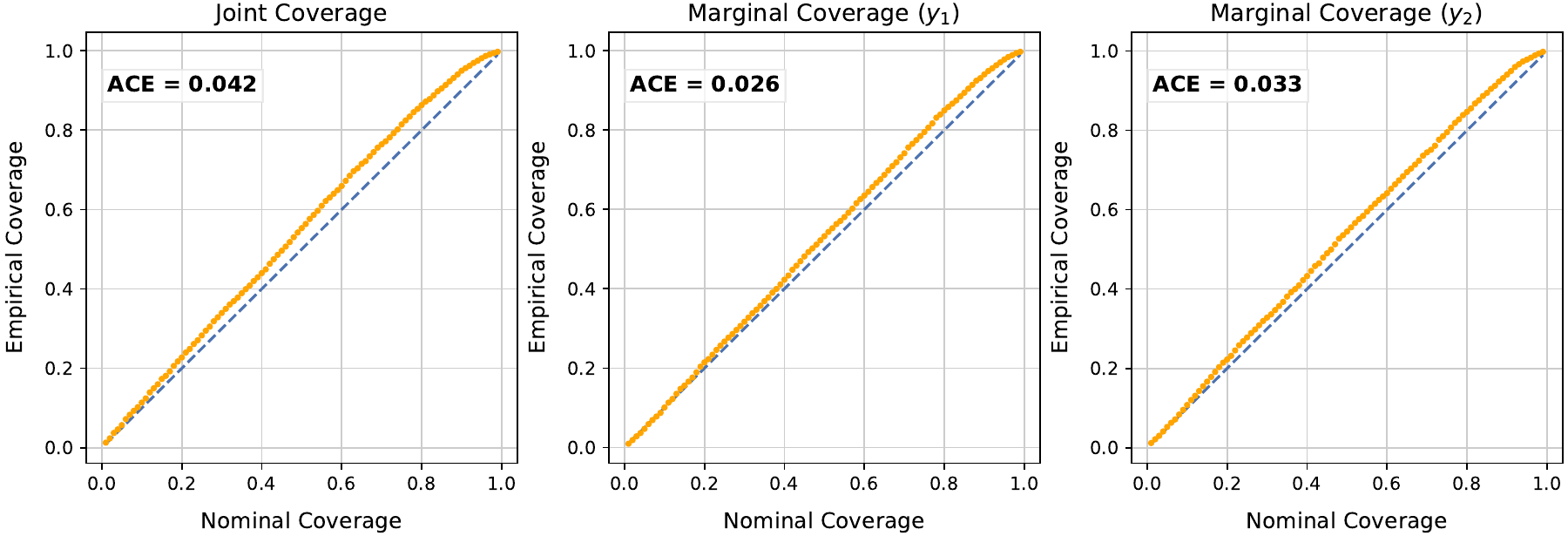}
    \vspace{0.15cm}
    \includegraphics[width=0.85\linewidth]{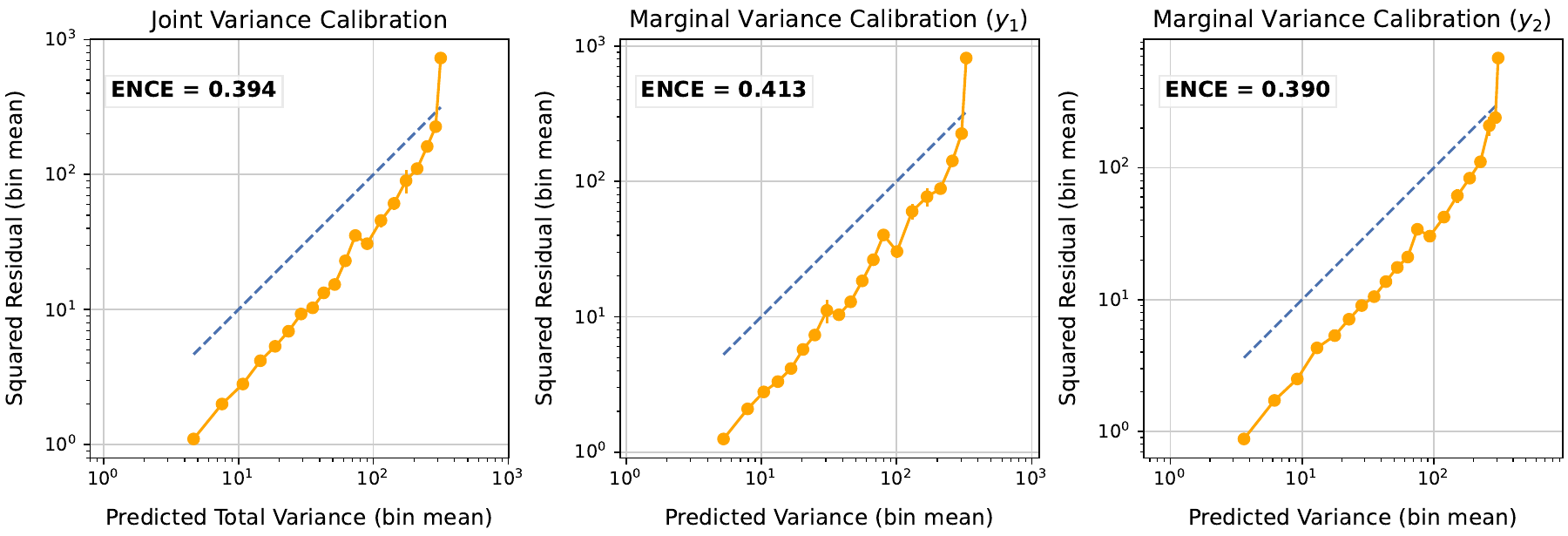}
    \vspace{0.15cm}
    \includegraphics[width=0.85\linewidth]{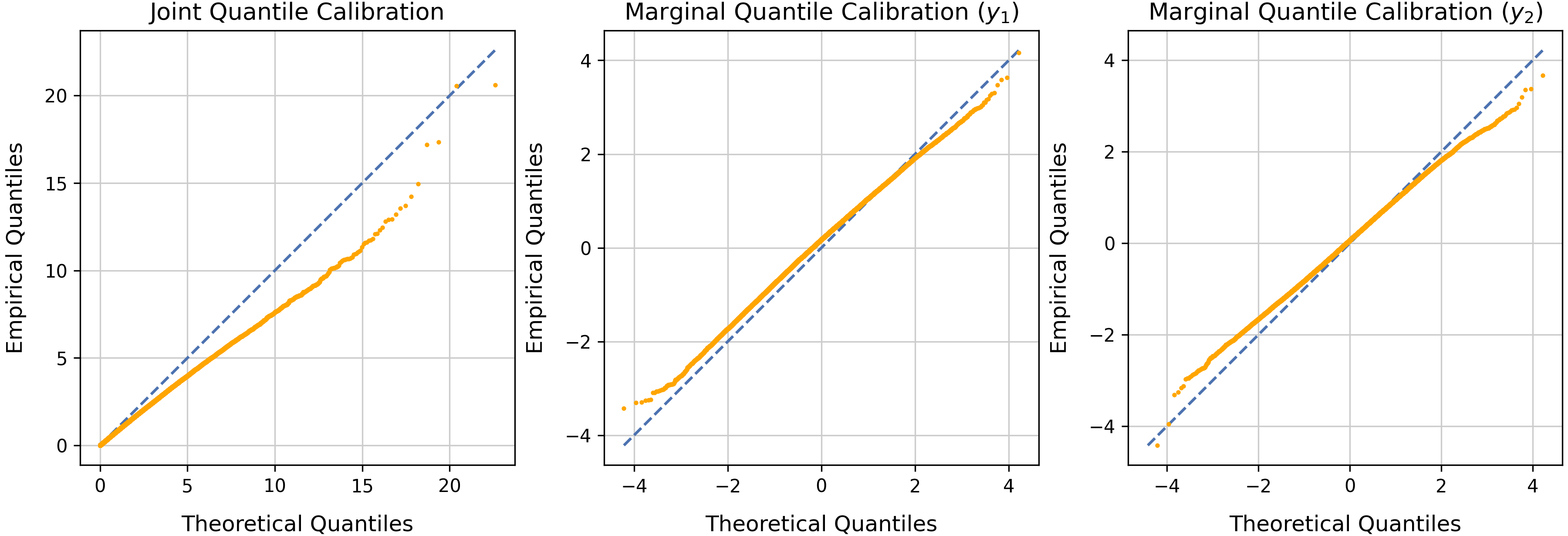}
    \caption{Calibration diagnostics on COCO for YOLO26-Pose Nano: native RLE~\citep{li2021keypoint-normalizing-flows}. Same layout as \cref{fig:coco_nas_n_base_diag}.}
    \label{fig:coco_y26_n_rle_diag}
\end{figure}

\begin{figure}[p]
    \centering
    \includegraphics[width=0.85\linewidth]{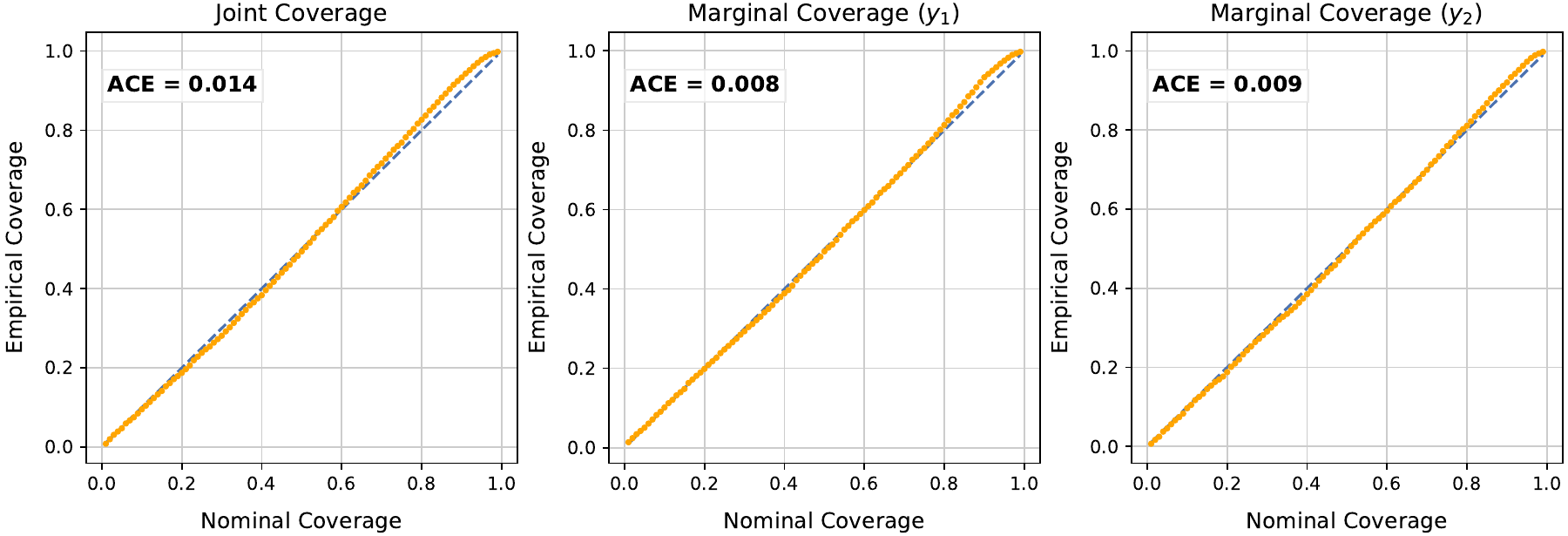}
    \vspace{0.15cm}
    \includegraphics[width=0.85\linewidth]{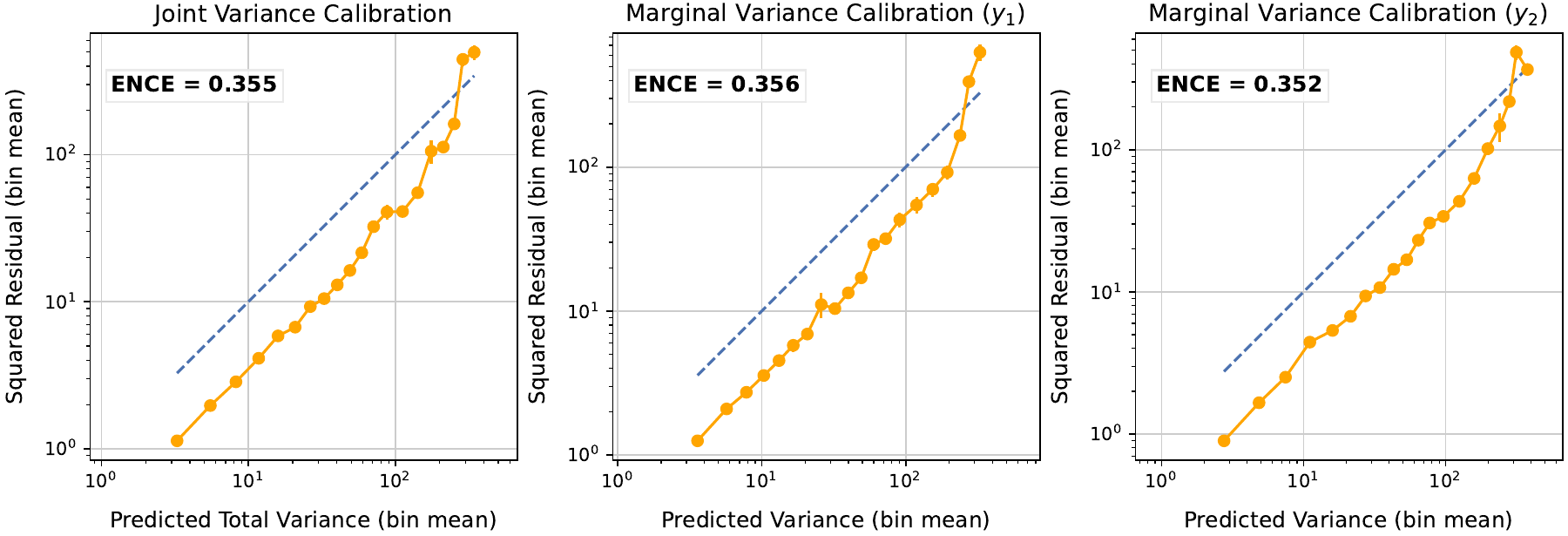}
    \vspace{0.15cm}
    \includegraphics[width=0.85\linewidth]{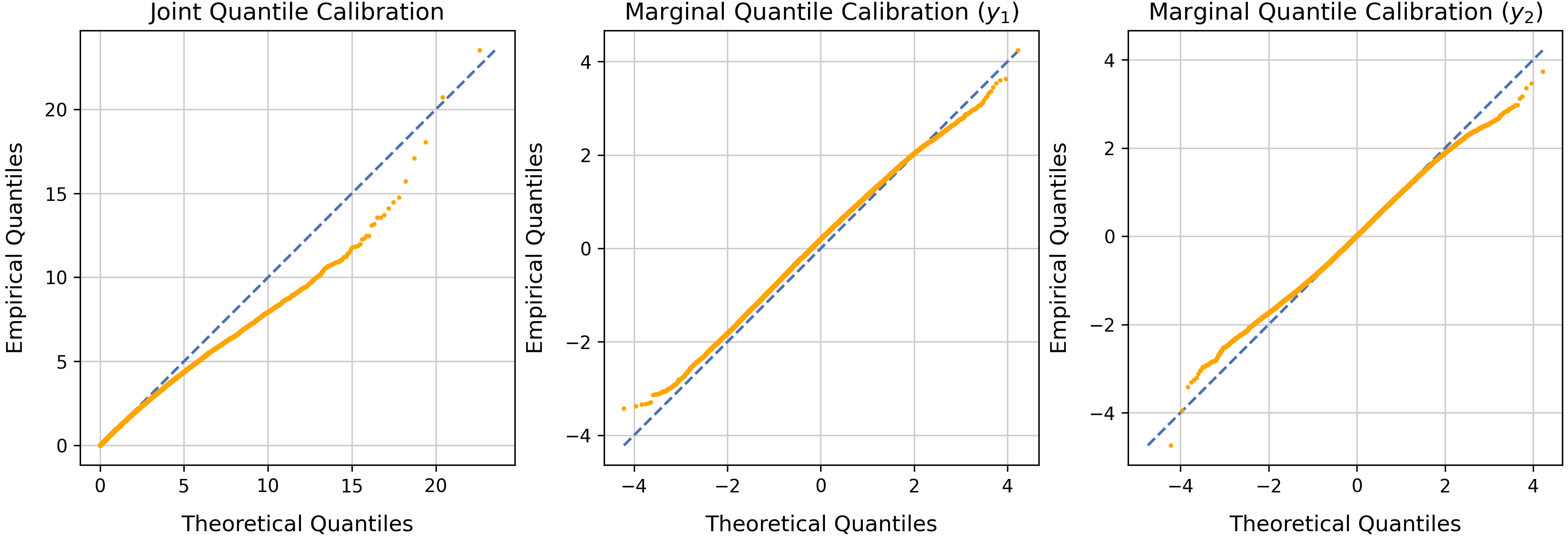}
    \caption{Calibration diagnostics on COCO for YOLO26-Pose Nano: calibrated RLE (temperature-scaled). Same layout as \cref{fig:coco_nas_n_base_diag}.}
    \label{fig:coco_y26_n_rle_cal_diag}
\end{figure}

\begin{figure}[p]
    \centering
    \includegraphics[width=0.85\linewidth]{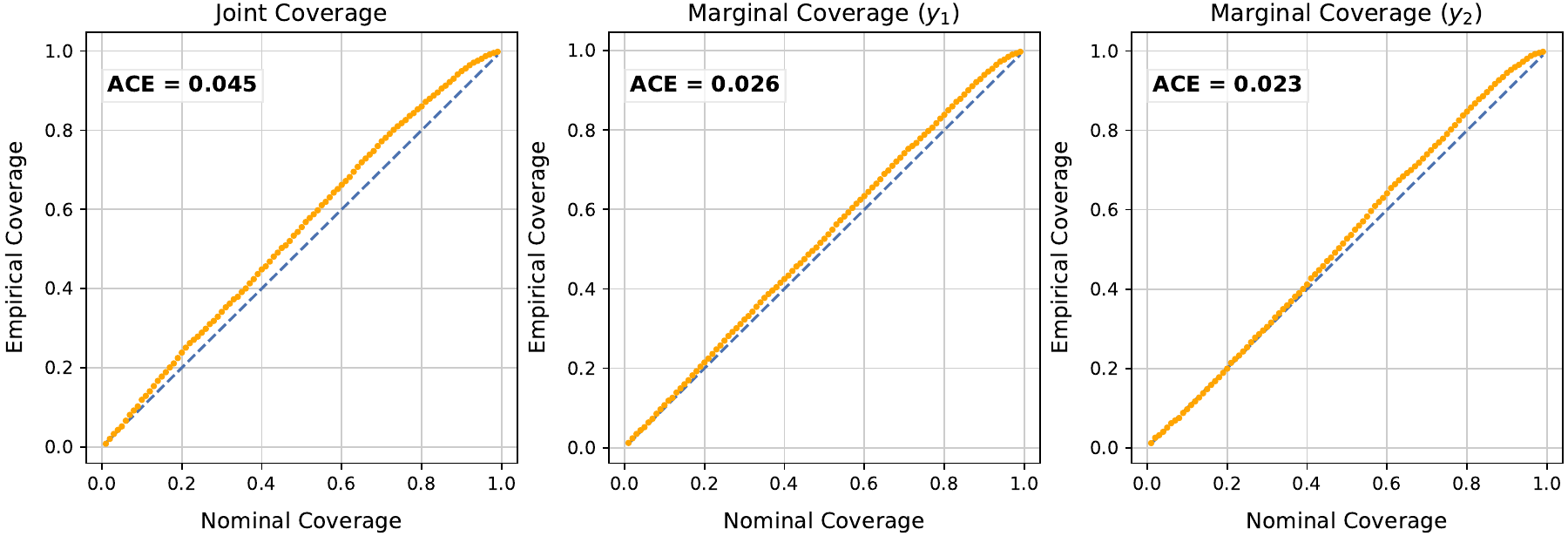}
    \vspace{0.15cm}
    \includegraphics[width=0.85\linewidth]{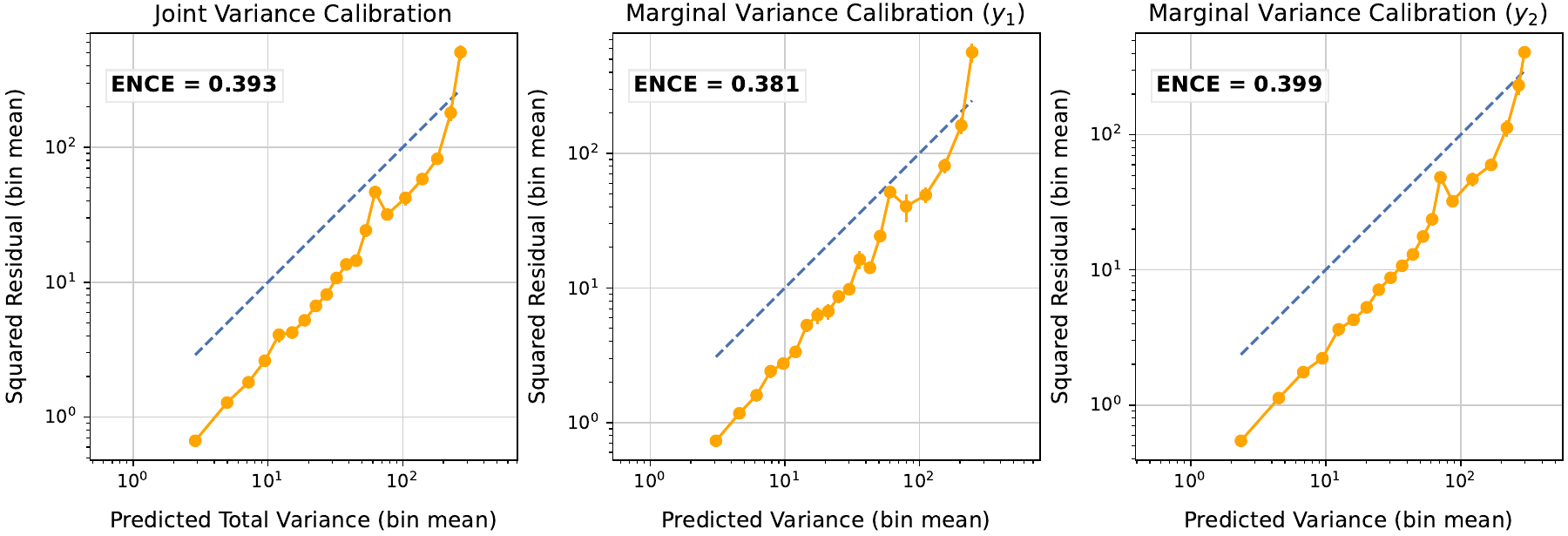}
    \vspace{0.15cm}
    \includegraphics[width=0.85\linewidth]{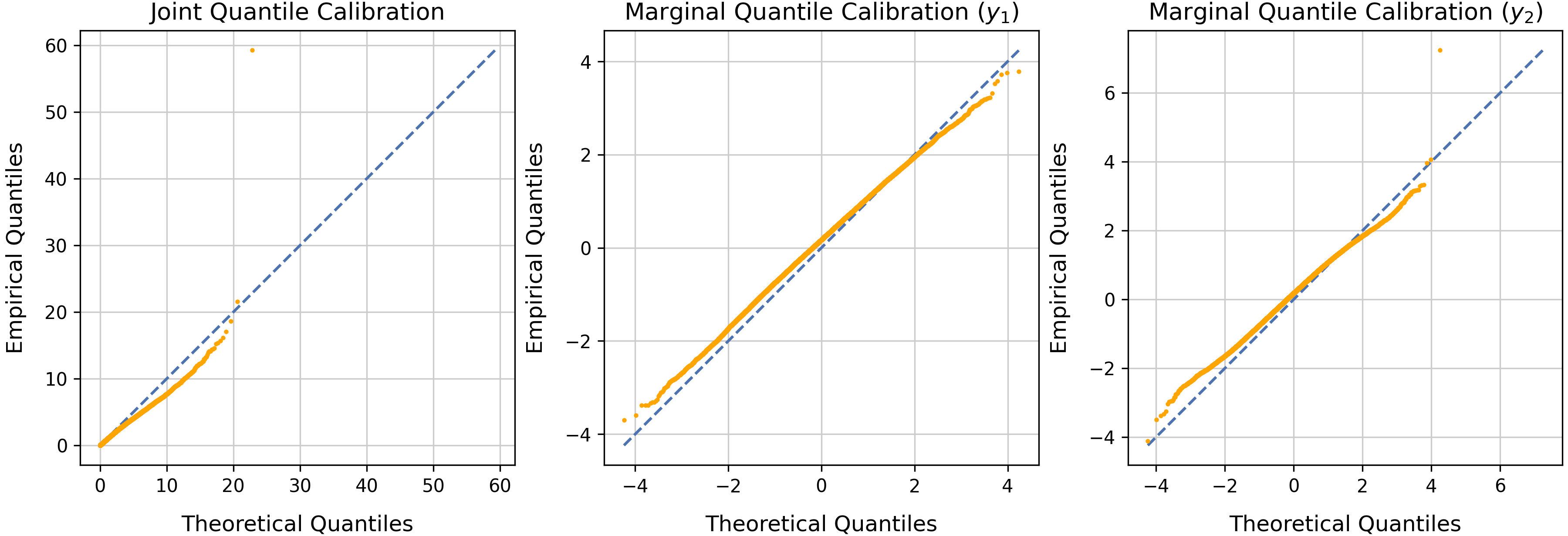}
    \caption{Calibration diagnostics on COCO for YOLO26-Pose Large: native RLE~\citep{li2021keypoint-normalizing-flows}. Same layout as \cref{fig:coco_nas_n_base_diag}.}
    \label{fig:coco_y26_l_rle_diag}
\end{figure}

\begin{figure}[p]
    \centering
    \includegraphics[width=0.85\linewidth]{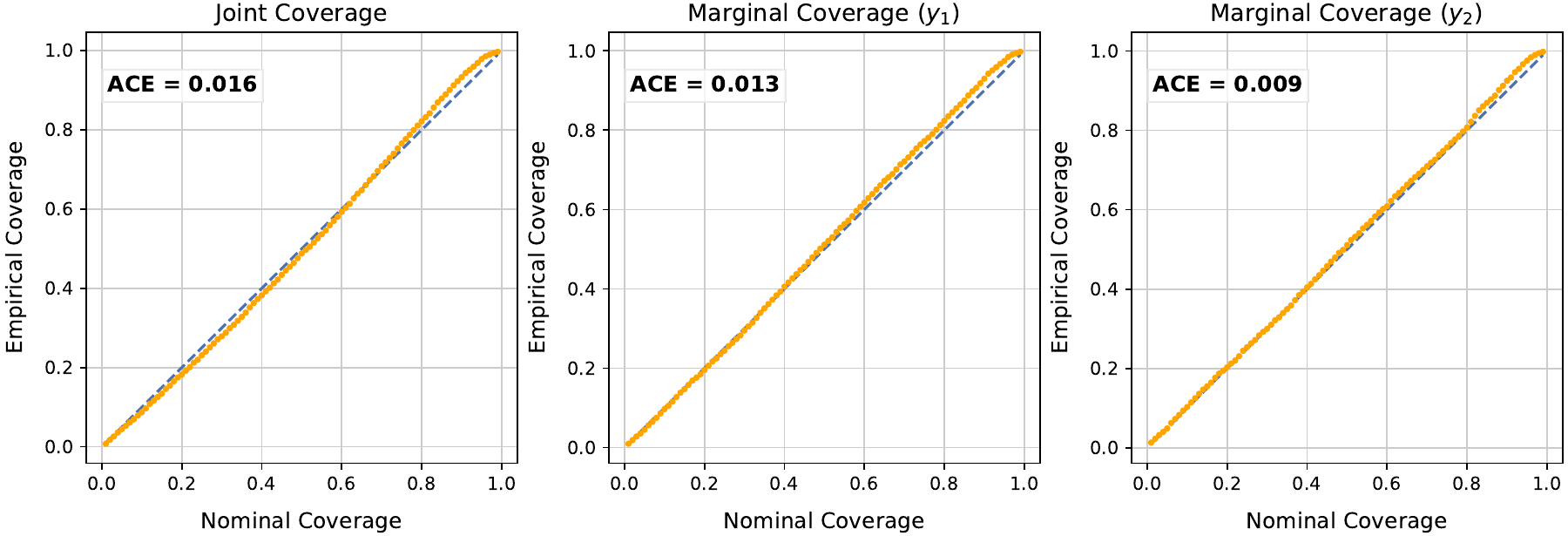}
    \vspace{0.15cm}
    \includegraphics[width=0.85\linewidth]{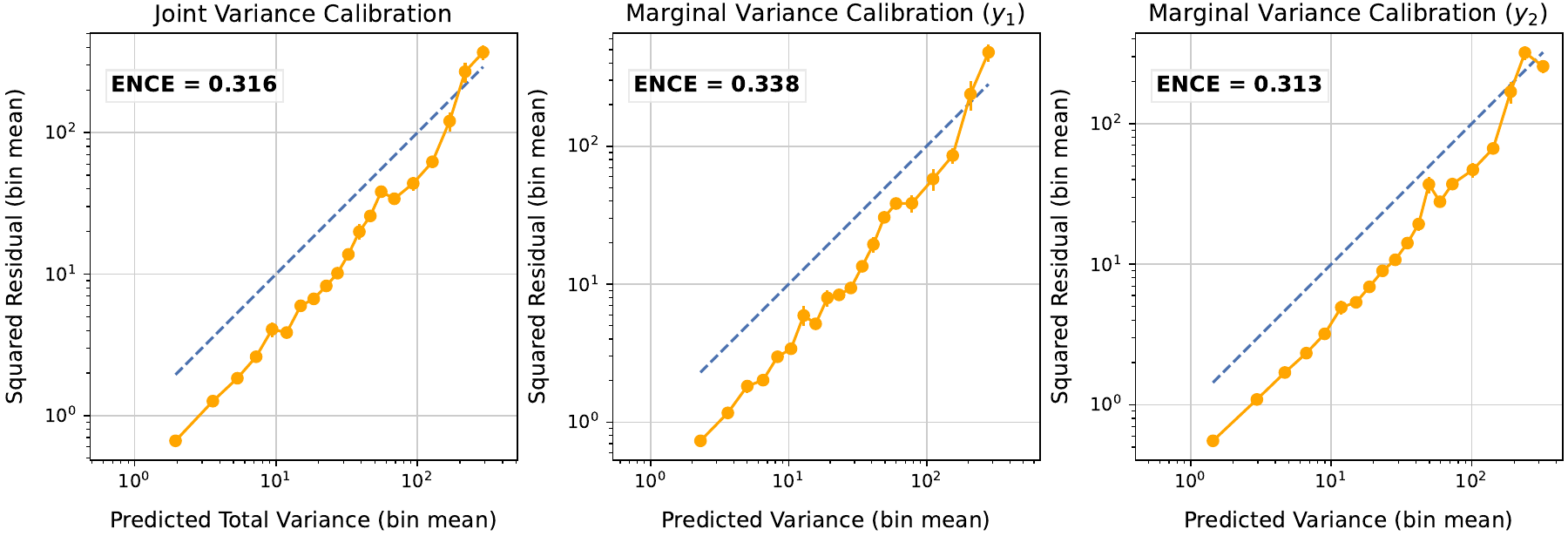}
    \vspace{0.15cm}
    \includegraphics[width=0.85\linewidth]{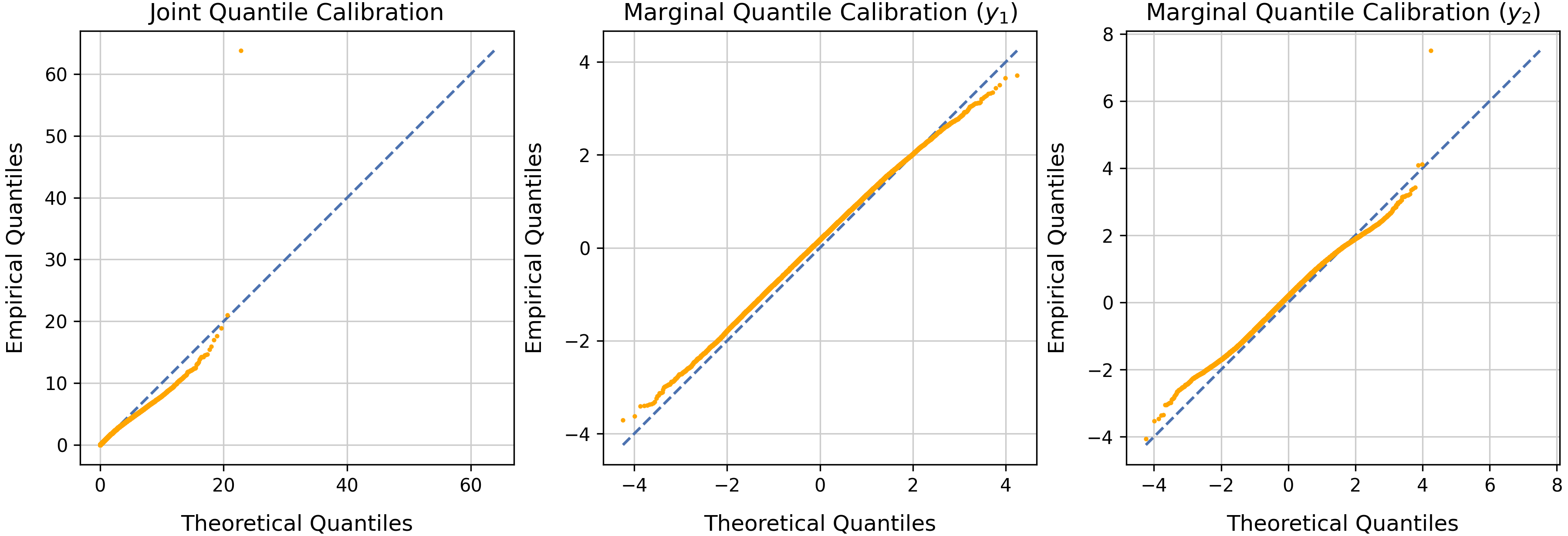}
    \caption{Calibration diagnostics on COCO for YOLO26-Pose Large: calibrated RLE (temperature-scaled). Same layout as \cref{fig:coco_nas_n_base_diag}.}
    \label{fig:coco_y26_l_rle_cal_diag}
\end{figure}

\FloatBarrier


\begin{figure}[p]
    \centering
    \includegraphics[width=0.85\linewidth]{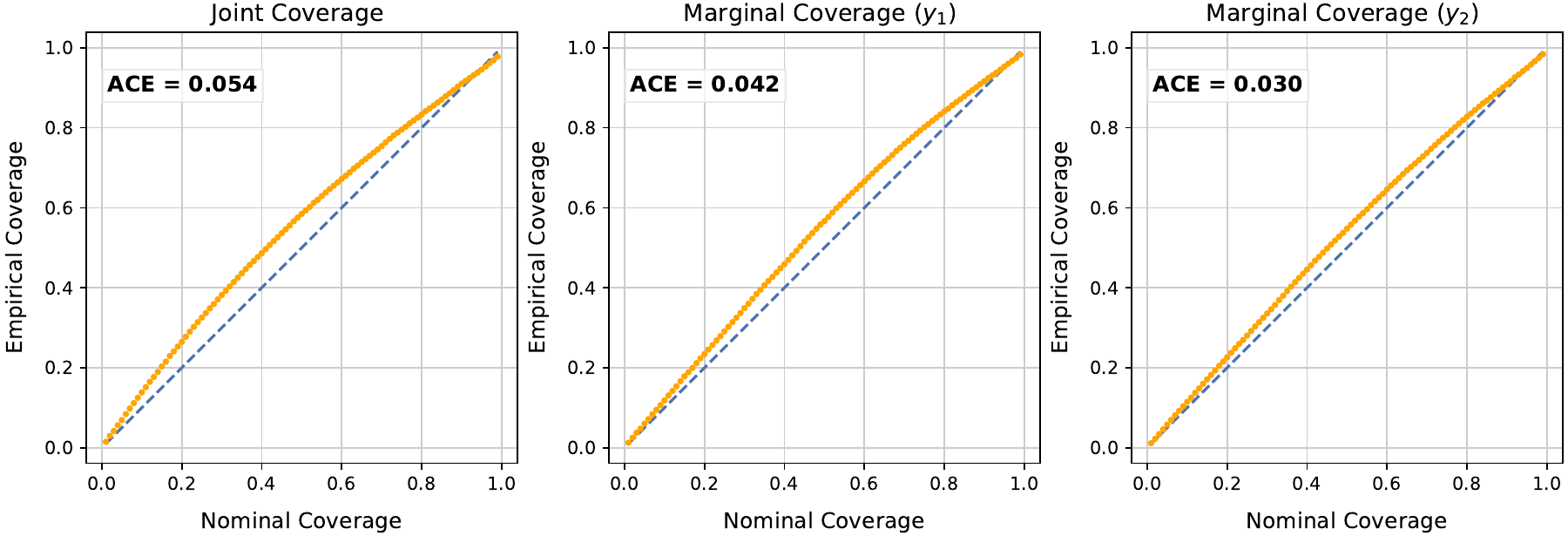}
    \vspace{0.15cm}
    \includegraphics[width=0.85\linewidth]{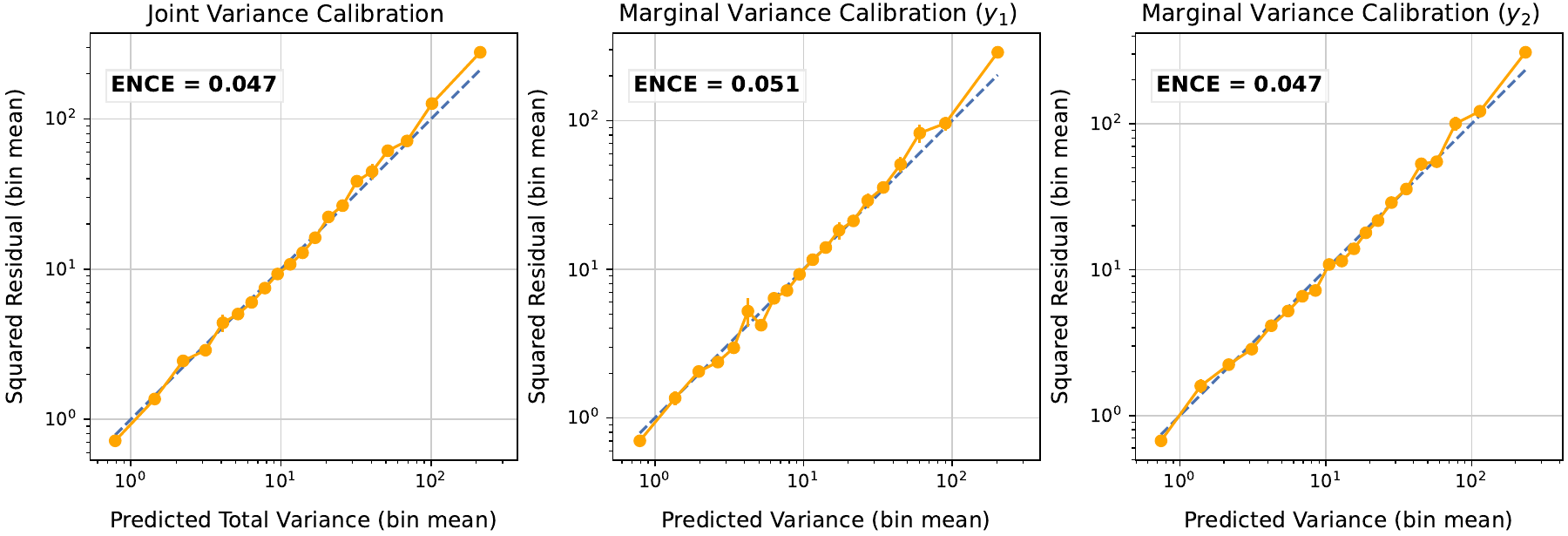}
    \vspace{0.15cm}
    \includegraphics[width=0.85\linewidth]{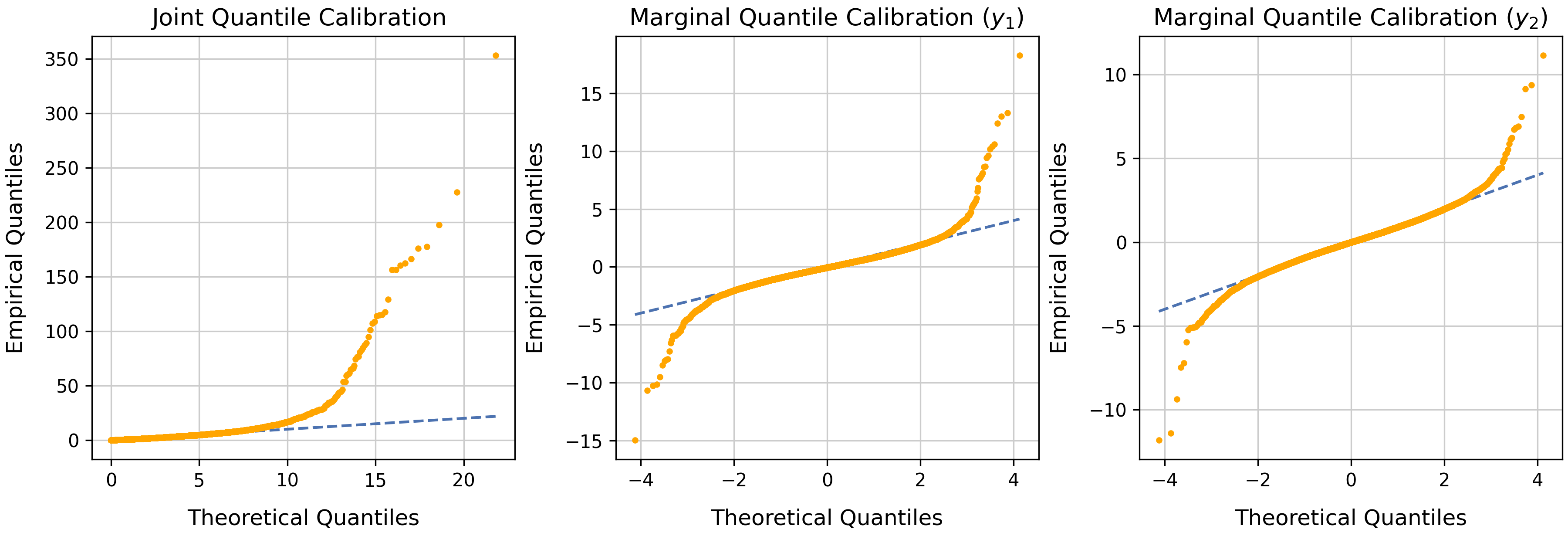}
    \caption{Calibration diagnostics on COCO after keypoint pruning for YOLO-NAS-Pose Nano: Gaussian calibration. Same layout as \cref{fig:coco_nas_n_base_diag}. For the corresponding diagnostics before pruning, see \cref{fig:coco_nas_n_gauss_diag}.}
    \label{fig:coco_kpp_n_gauss_diag}
\end{figure}

\begin{figure}[p]
    \centering
    \includegraphics[width=0.85\linewidth]{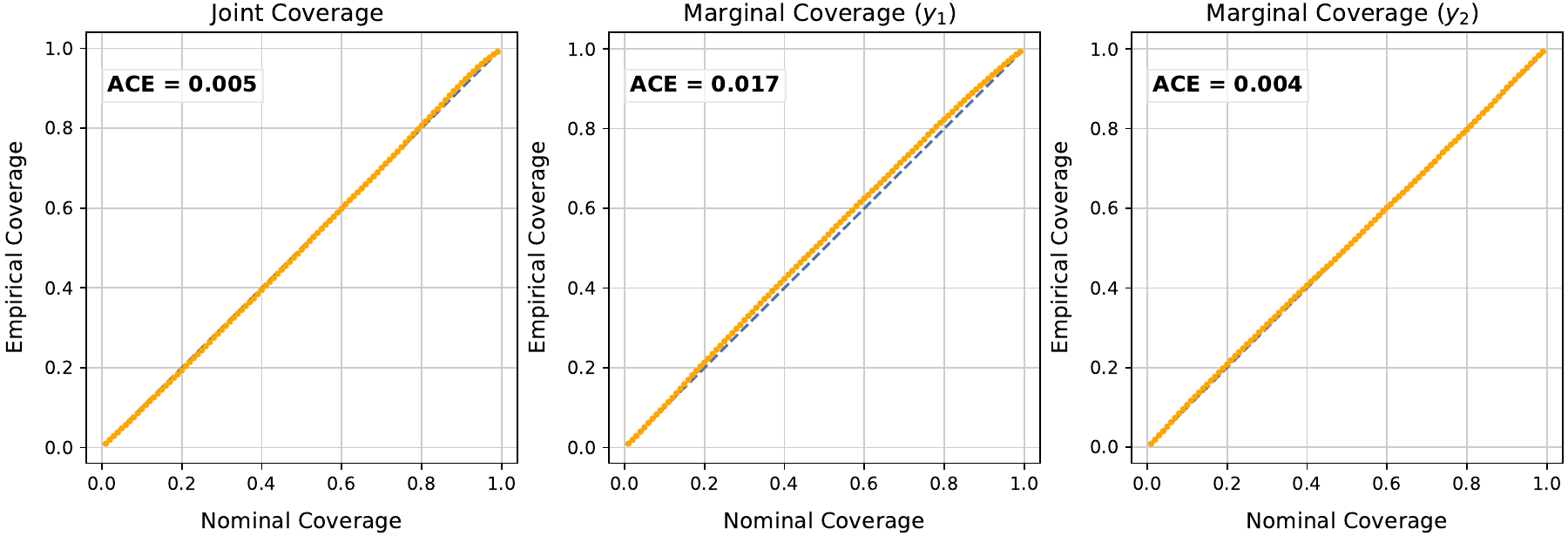}
    \vspace{0.15cm}
    \includegraphics[width=0.85\linewidth]{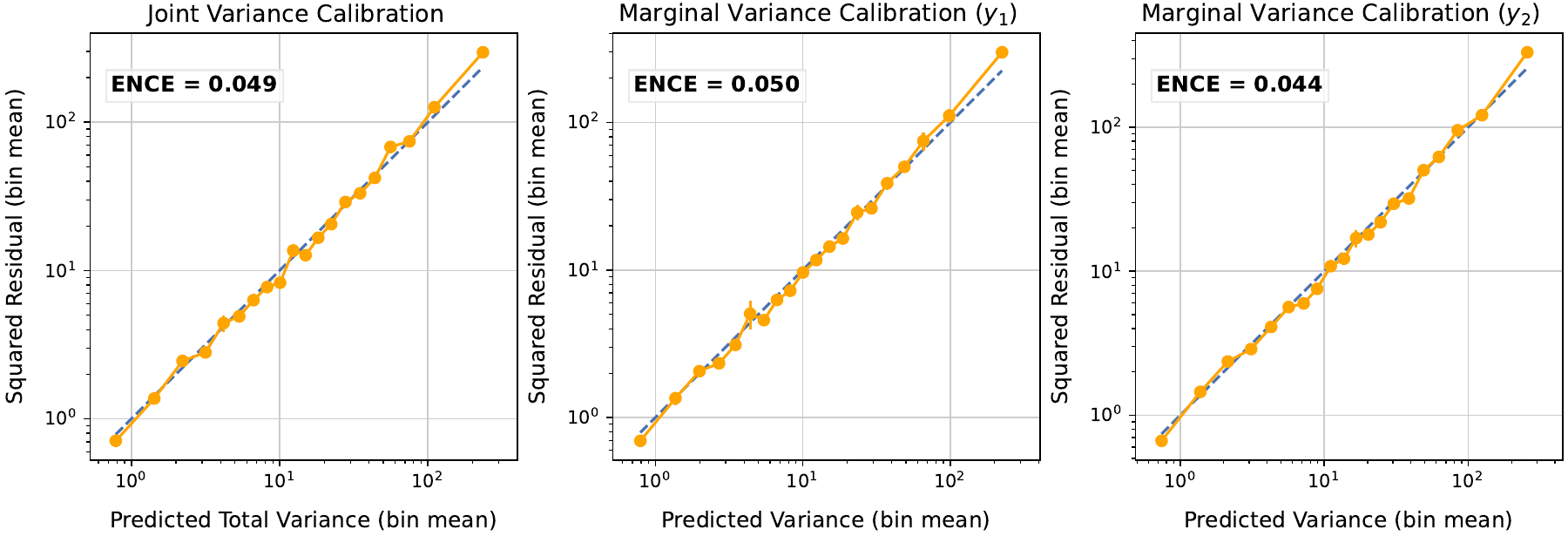}
    \vspace{0.15cm}
    \includegraphics[width=0.85\linewidth]{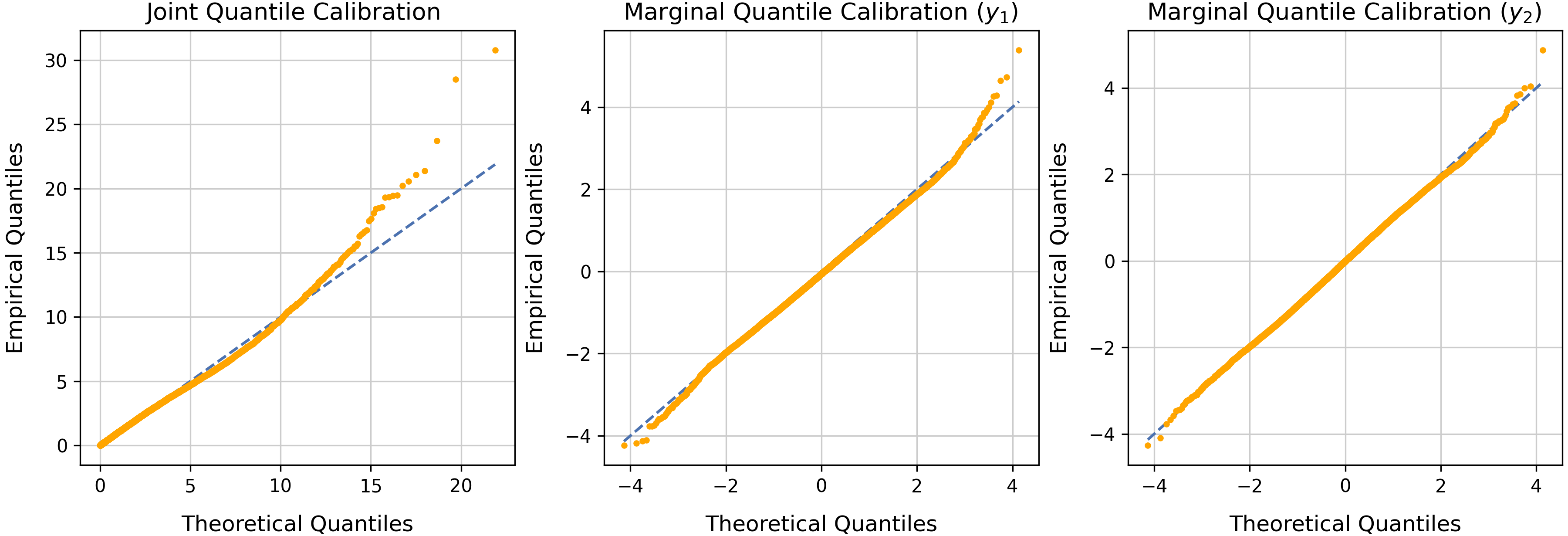}
    \caption{Calibration diagnostics on COCO after keypoint pruning for YOLO-NAS-Pose Nano: Student-$t$ calibration. Same layout as \cref{fig:coco_nas_n_base_diag}. For the corresponding diagnostics before pruning, see \cref{fig:coco_nas_n_t_diag}.}
    \label{fig:coco_kpp_n_t_diag}
\end{figure}

\begin{figure}[p]
    \centering
    \includegraphics[width=0.85\linewidth]{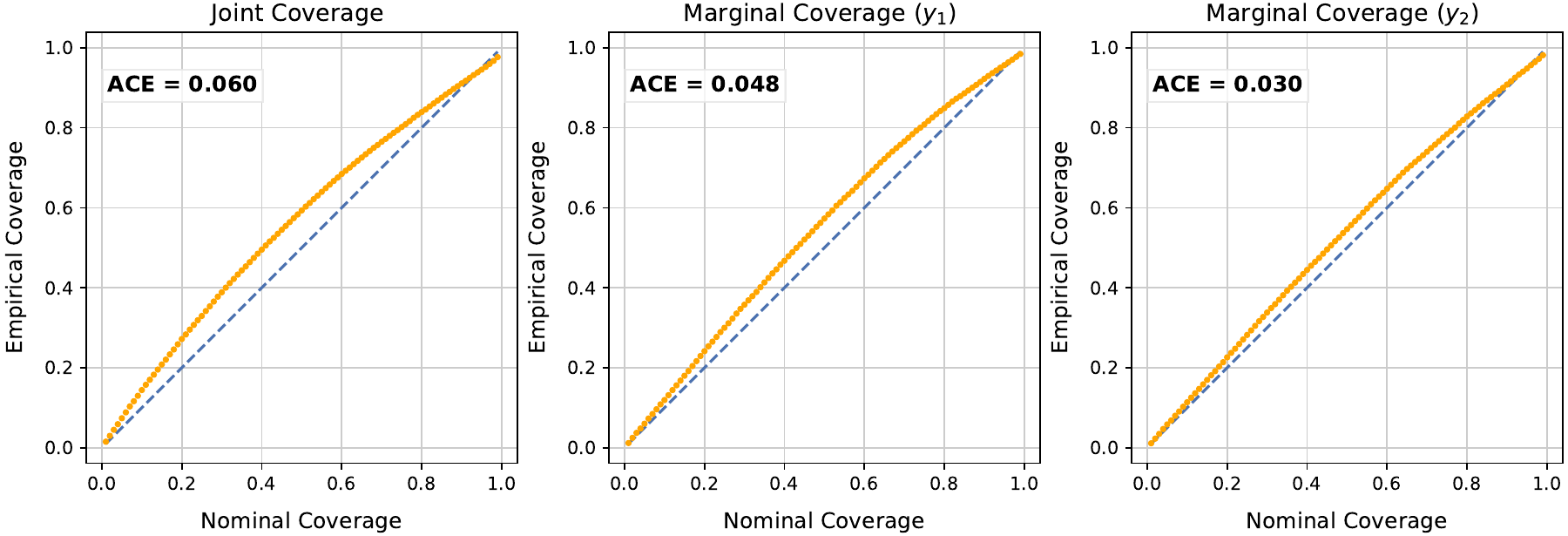}
    \vspace{0.15cm}
    \includegraphics[width=0.85\linewidth]{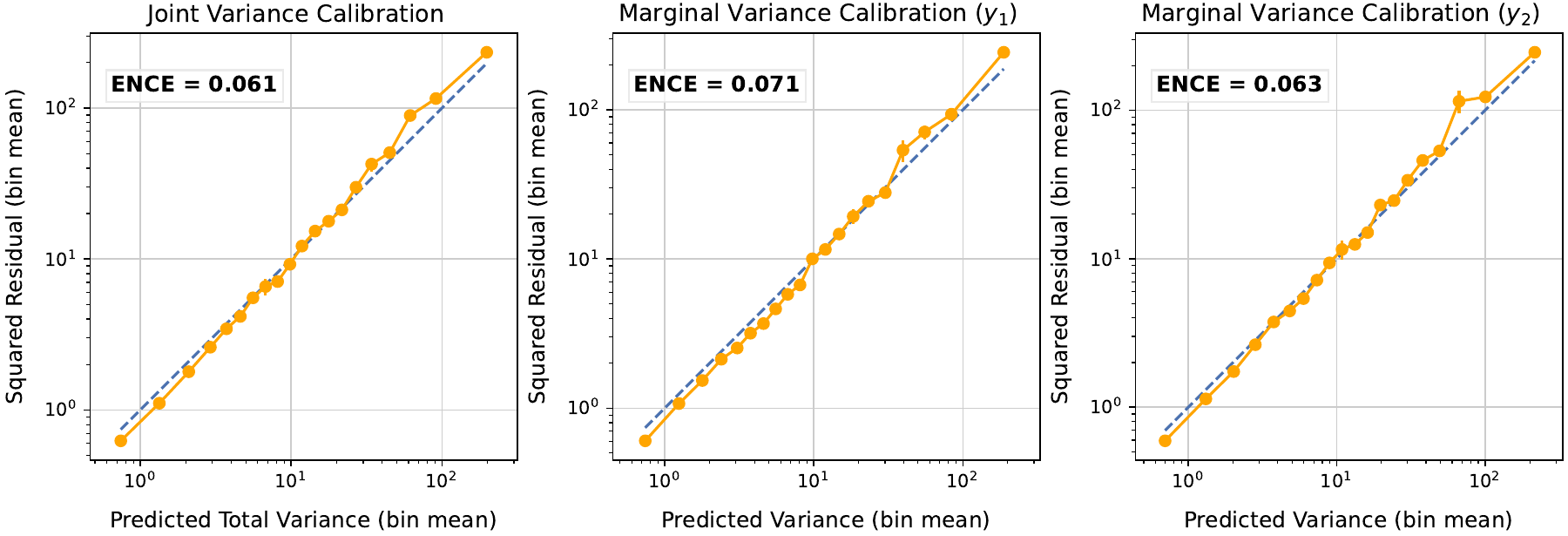}
    \vspace{0.15cm}
    \includegraphics[width=0.85\linewidth]{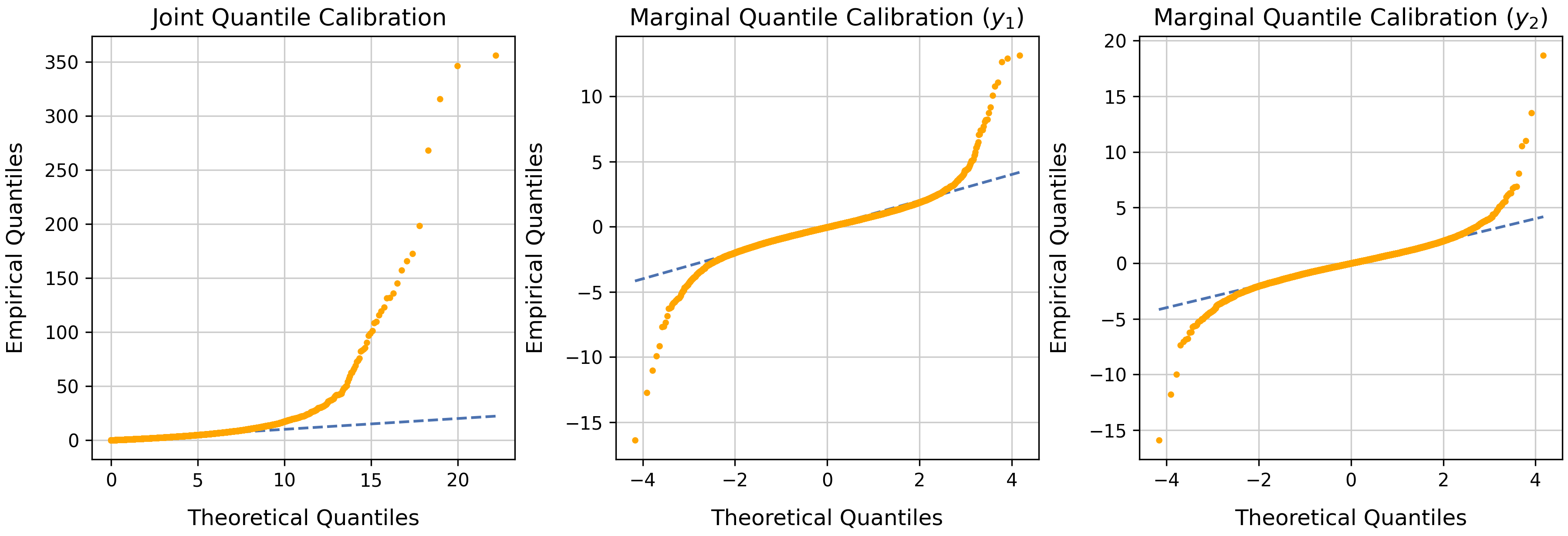}
    \caption{Calibration diagnostics on COCO after keypoint pruning for YOLO-NAS-Pose Large: Gaussian calibration. Same layout as \cref{fig:coco_nas_n_base_diag}. For the corresponding diagnostics before pruning, see \cref{fig:coco_nas_l_gauss_diag}.}
    \label{fig:coco_kpp_l_gauss_diag}
\end{figure}

\begin{figure}[p]
    \centering
    \includegraphics[width=0.85\linewidth]{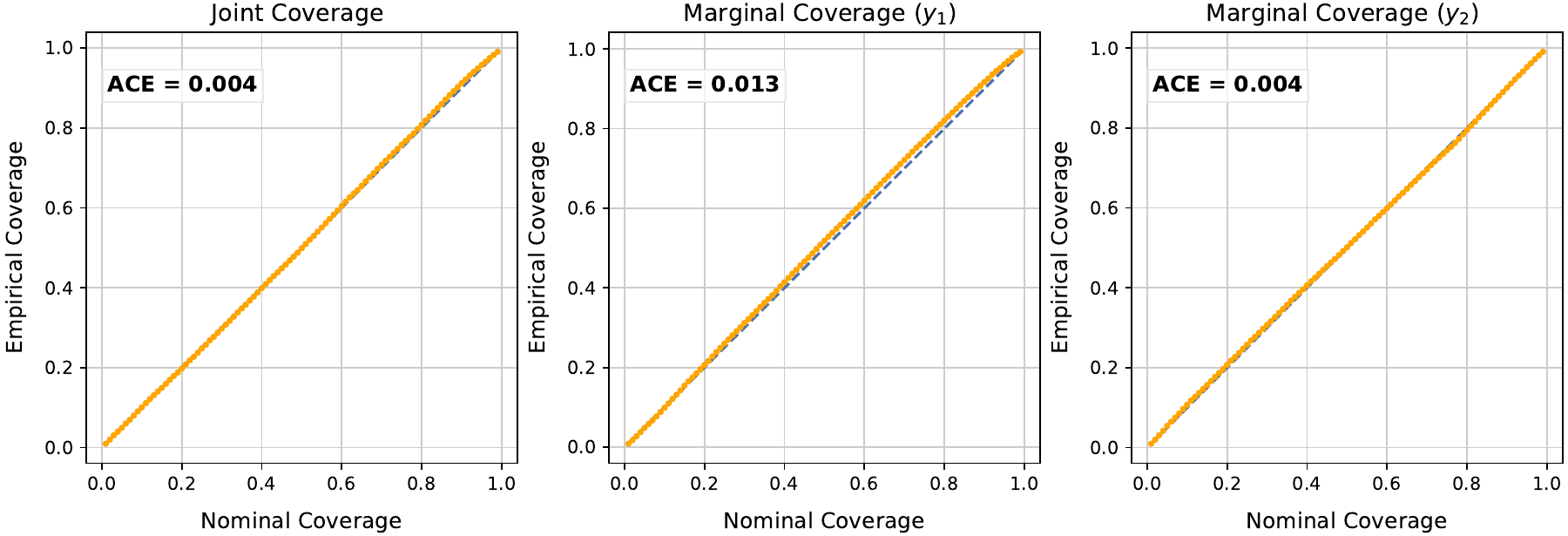}
    \vspace{0.15cm}
    \includegraphics[width=0.85\linewidth]{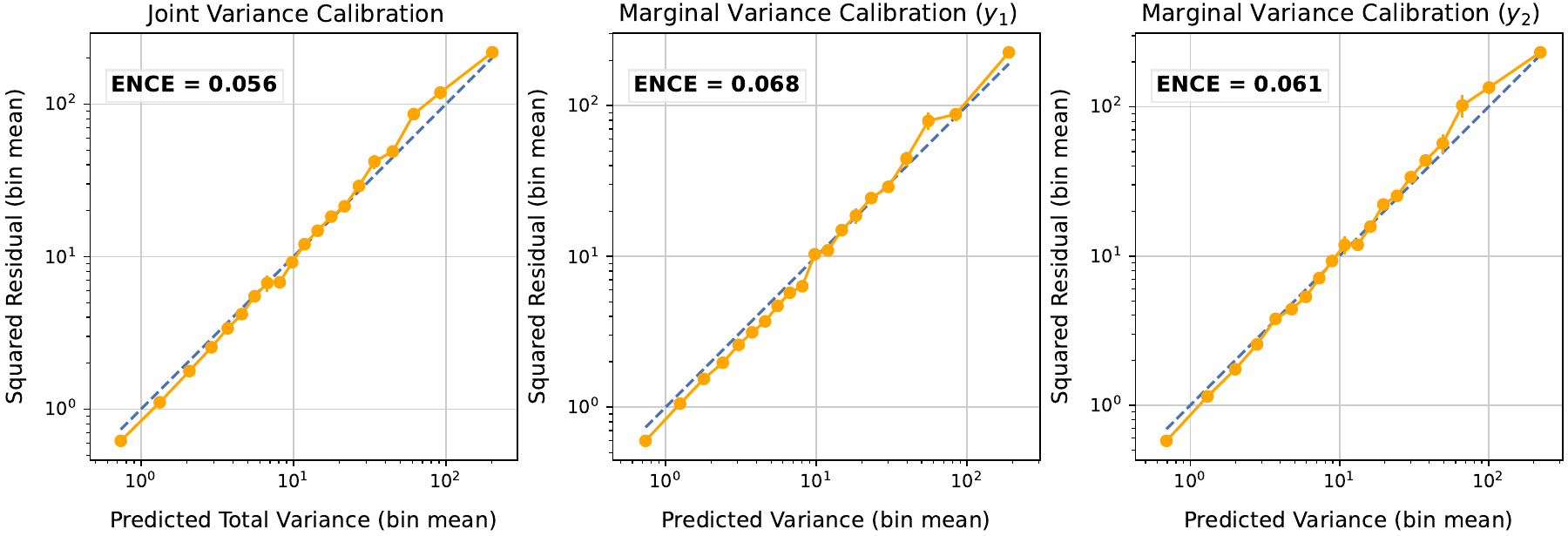}
    \vspace{0.15cm}
    \includegraphics[width=0.85\linewidth]{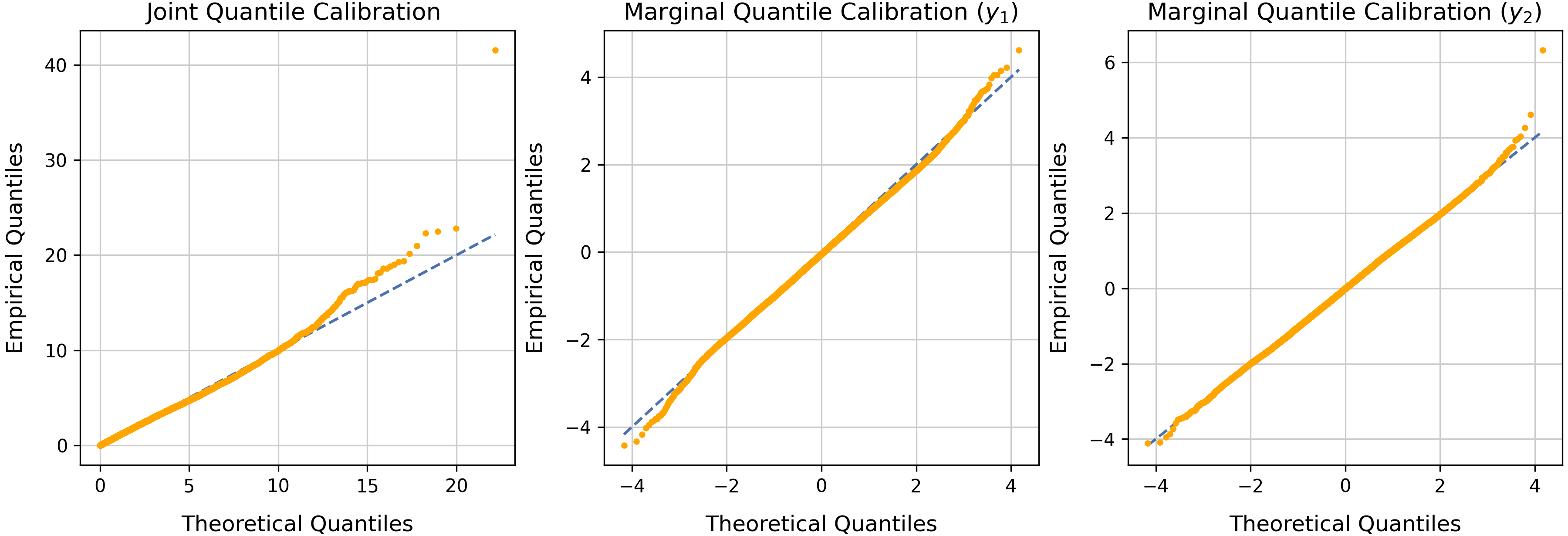}
    \caption{Calibration diagnostics on COCO after keypoint pruning for YOLO-NAS-Pose Large: Student-$t$ calibration. Same layout as \cref{fig:coco_nas_n_base_diag}. For the corresponding diagnostics before pruning, see \cref{fig:coco_nas_l_t_diag}.}
    \label{fig:coco_kpp_l_t_diag}
\end{figure}

\FloatBarrier


\begin{figure}[p]
    \centering
    \includegraphics[width=0.85\linewidth]{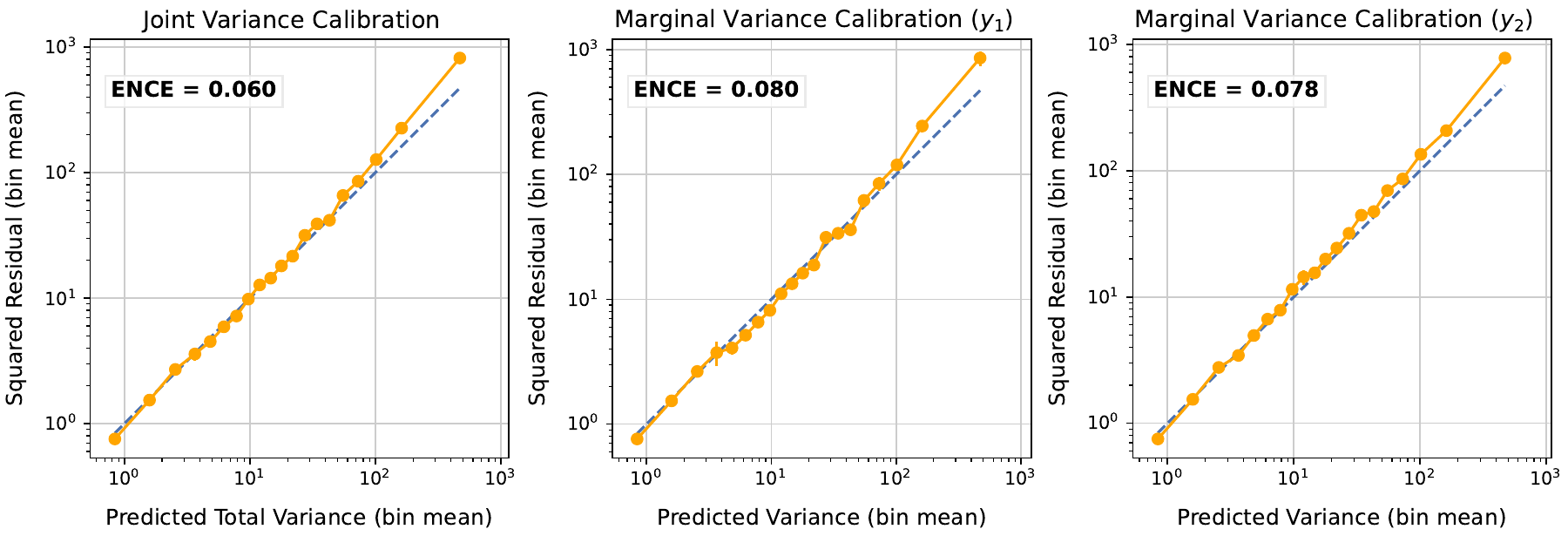}
    \vspace{0.15cm}
    \includegraphics[width=0.85\linewidth]{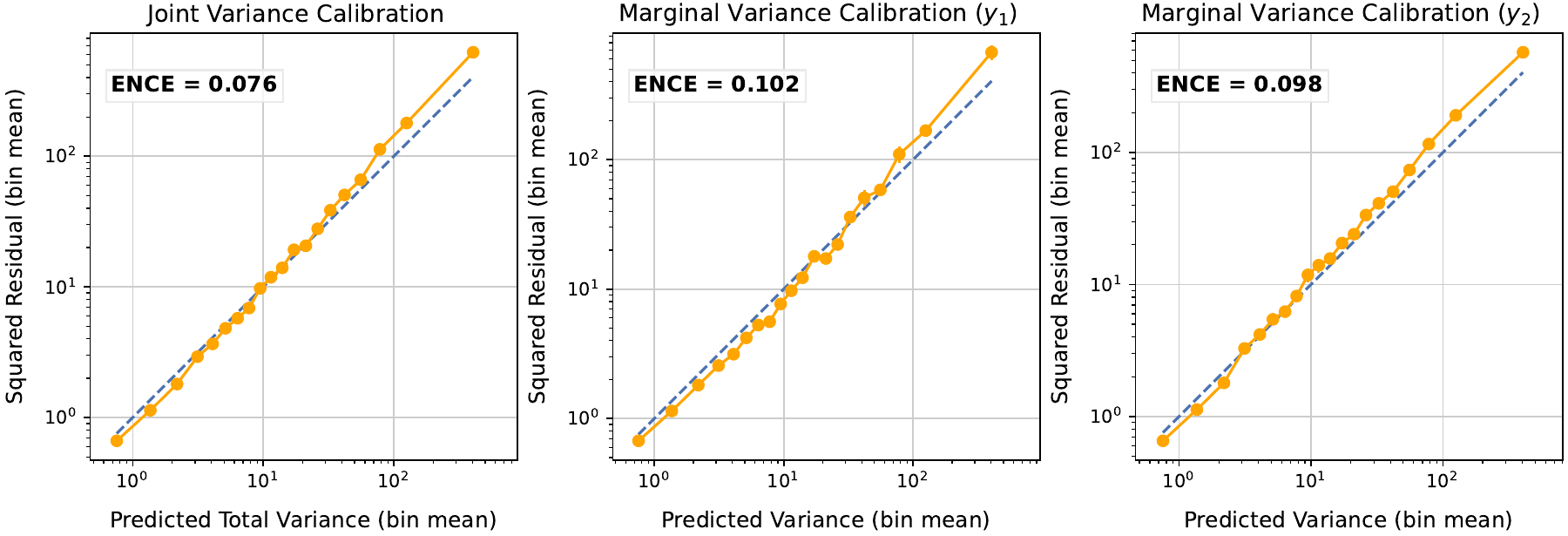}
    \caption{Ablation study of isotropic covariance parameterization on COCO. Variance calibration plots for YOLO-NAS-Pose Nano (top) and Large (bottom) show low joint \ac{ence} but comparatively poor marginal calibration.}
    \label{fig:coco_isotropic_variance_calib_plots}
\end{figure}

\FloatBarrier


\begin{sidewaystable}[p]
\centering
\caption{
Inference latency benchmark on a single NVIDIA A10G GPU at $640 \times 640$ resolution.
Overhead is the relative increase in median latency over the corresponding frozen baseline; FLOP+ is the relative increase in GFLOPs.
Median latency is computed over 200 timed iterations after 50 warmup passes; P95 latency was within 5\% of the median across all configurations, indicating stable measurements.
}
\label{tab:coco_latency}
\setlength{\tabcolsep}{5pt}
\small
\begin{tabular}{clccc cc cc cc}
\toprule
& & & & & \multicolumn{2}{c}{Batch size 1} & \multicolumn{2}{c}{Batch size 4} & \multicolumn{2}{c}{Batch size 16} \\
\cmidrule(lr){6-7} \cmidrule(lr){8-9} \cmidrule(lr){10-11}
Model & Configuration & Params & GFLOPs & FLOP+
& Med.\ (ms) & Overhead
& Med.\ (ms) & Overhead
& Med.\ (ms) & Overhead \\
\midrule
\multirow{7}{*}{\makecell{YOLO-NAS-Pose\\[1pt]Nano}}
& Baseline                         &  9.9M & 10.1 & ---        & 20.67 & ---        & 20.44 & ---        &  70.80 & ---        \\
& Full cov., proposed heads        & 11.3M & 11.6 & $+14.8\%$  & 23.91 & $+15.7\%$  & 25.75 & $+26.0\%$  &  85.18 & $+20.3\%$  \\
& Diagonal cov., proposed heads    & 11.3M & 11.6 & $+14.7\%$  & 22.60 & $+9.3\%$   & 22.72 & $+11.1\%$  &  75.83 & $+7.1\%$   \\
& Isotropic cov., proposed heads   & 11.3M & 11.5 & $+14.6\%$  & 22.56 & $+9.1\%$   & 22.73 & $+11.2\%$  &  75.72 & $+7.0\%$   \\
& Full cov., last-layer heads      & 10.0M & 10.2 & $+1.0\%$   & 23.00 & $+11.3\%$  & 24.76 & $+21.2\%$  &  81.85 & $+15.6\%$  \\
& Diagonal cov., last-layer heads  & 10.0M & 10.2 & $+0.9\%$   & 21.64 & $+4.7\%$   & 21.71 & $+6.2\%$   &  72.48 & $+2.4\%$   \\
& Isotropic cov., last-layer heads & 10.0M & 10.2 & $+0.8\%$   & 21.65 & $+4.7\%$   & 21.79 & $+6.6\%$   &  72.40 & $+2.3\%$   \\
\midrule
\multirow{7}{*}{\makecell{YOLO-NAS-Pose\\[1pt]Large}}
& Baseline                         & 79.4M & 77.5 & ---        & 24.80 & ---        & 61.58 & ---        & 229.68 & ---        \\
& Full cov., proposed heads        & 91.8M & 90.2 & $+16.4\%$  & 28.65 & $+15.5\%$  & 69.22 & $+12.4\%$  & 257.31 & $+12.0\%$  \\
& Diagonal cov., proposed heads    & 91.8M & 90.2 & $+16.3\%$  & 27.67 & $+11.6\%$  & 66.69 & $+8.3\%$   & 247.91 & $+7.9\%$   \\
& Isotropic cov., proposed heads   & 91.8M & 90.1 & $+16.3\%$  & 28.71 & $+15.8\%$  & 66.66 & $+8.2\%$   & 247.79 & $+7.9\%$   \\
& Full cov., last-layer heads      & 80.0M & 78.2 & $+0.9\%$   & 27.52 & $+10.9\%$  & 65.66 & $+6.6\%$   & 242.19 & $+5.4\%$   \\
& Diagonal cov., last-layer heads  & 79.9M & 78.2 & $+0.9\%$   & 26.06 & $+5.1\%$   & 62.72 & $+1.8\%$   & 232.82 & $+1.4\%$   \\
& Isotropic cov., last-layer heads & 79.9M & 78.2 & $+0.9\%$   & 26.07 & $+5.1\%$   & 62.73 & $+1.9\%$   & 232.74 & $+1.3\%$   \\
\midrule
\multirow{2}{*}{\makecell{YOLO26-Pose\\[1pt]Nano}}
& Baseline              &  3.7M &  5.3 & ---        & 17.67 & ---        & 18.07 & ---        &  35.24 & ---        \\
& Probabilistic ext.    &  4.0M &  5.9 & $+10.3\%$  & 19.62 & $+11.0\%$  & 20.01 & $+10.7\%$  &  39.12 & $+11.0\%$  \\
\midrule
\multirow{2}{*}{\makecell{YOLO26-Pose\\[1pt]Large}}
& Baseline              & 28.7M & 52.1 & ---        & 25.69 & ---        & 43.69 & ---        & 164.41 & ---        \\
& Probabilistic ext.    & 29.6M & 54.0 & $+3.6\%$   & 27.64 & $+7.6\%$   & 45.36 & $+3.8\%$   & 169.85 & $+3.3\%$   \\
\bottomrule
\end{tabular}
\end{sidewaystable}

\FloatBarrier


\begin{figure}[p]
    \centering
    \includegraphics[width=0.85\linewidth]{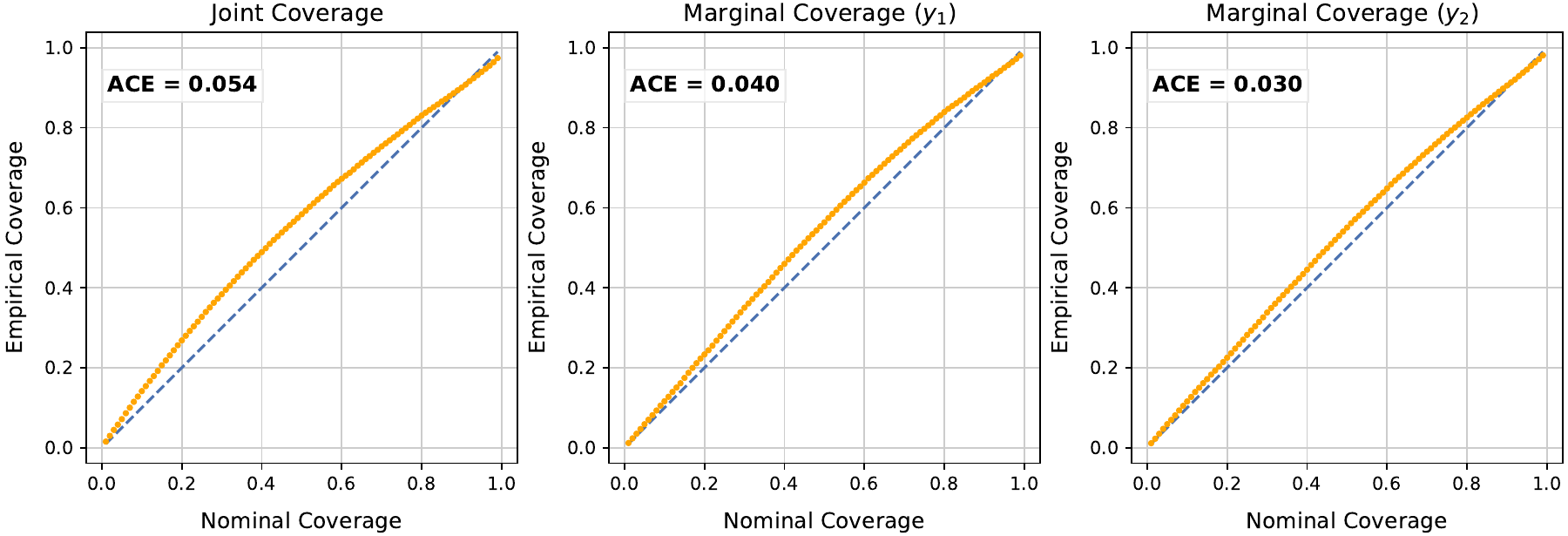}
    \vspace{0.15cm}
    \includegraphics[width=0.85\linewidth]{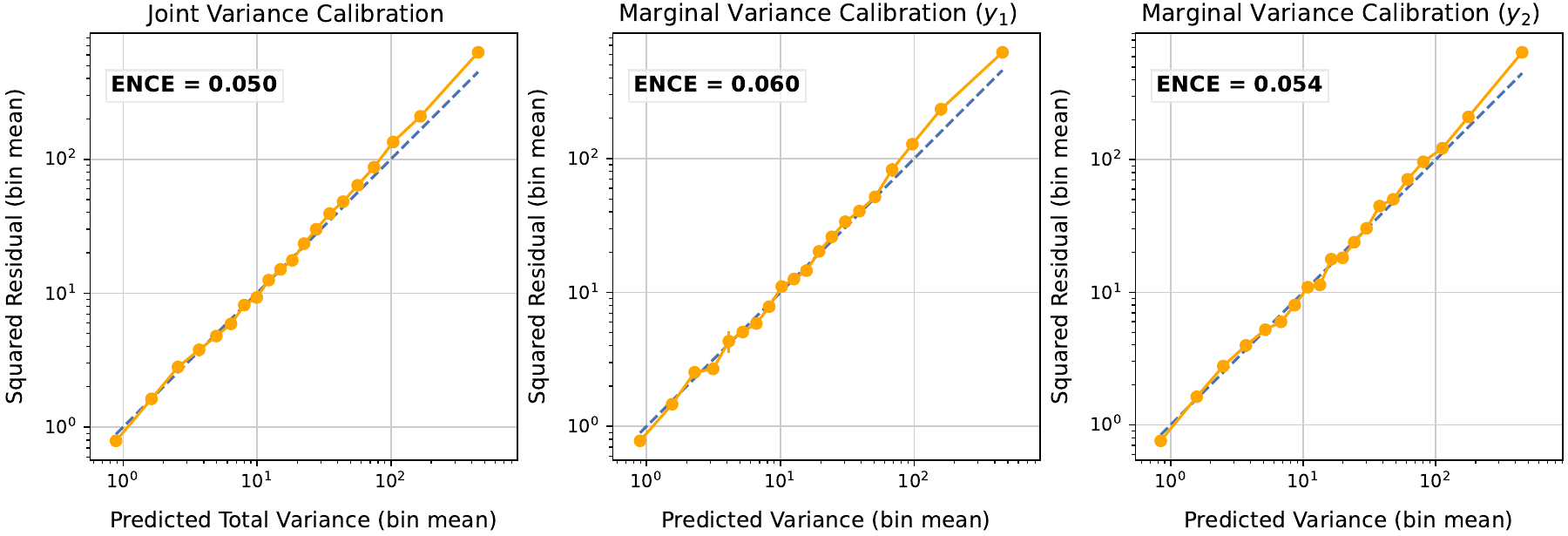}
    \vspace{0.15cm}
    \includegraphics[width=0.85\linewidth]{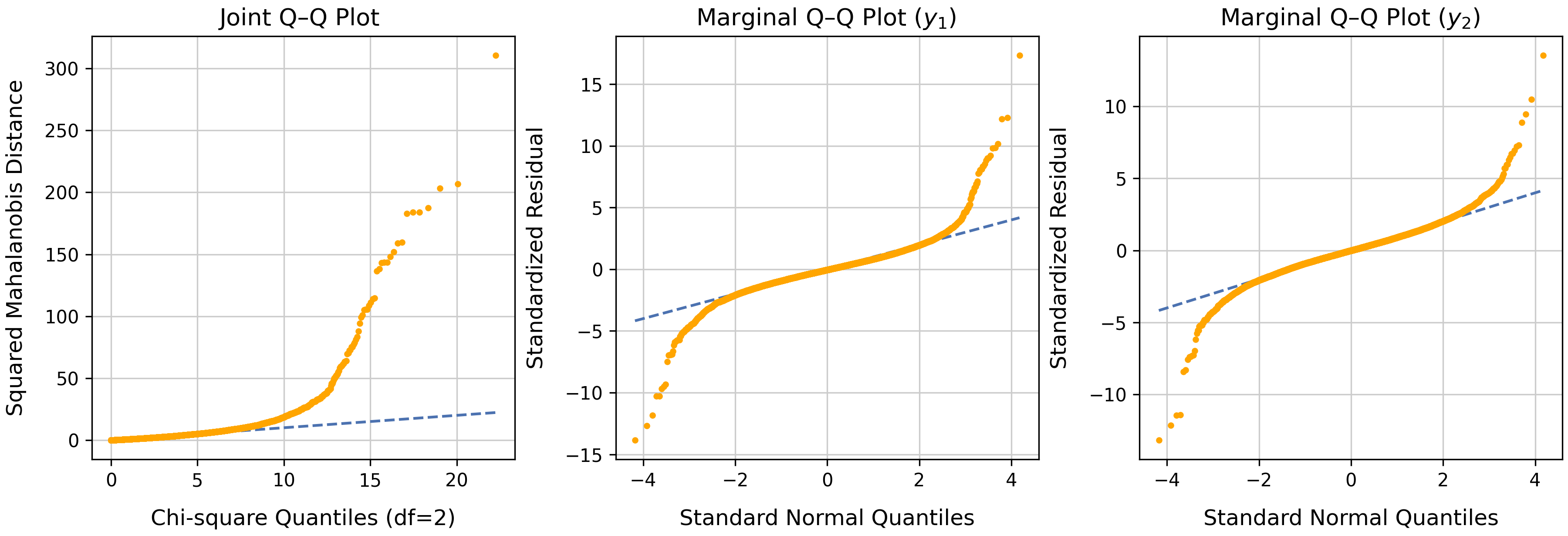}
    \caption{Gaussian Residual Distribution Fine-Tuning (see App.~\ref{app:residual_finetune}). Calibration diagnostics on COCO for YOLO-NAS-Pose Nano: Gaussian calibration. Same layout as \cref{fig:coco_nas_n_base_diag}. Compared to our default approach (\cref{fig:coco_nas_n_gauss_diag}), calibration is particularly improved in the central region of the distribution, while deviations remain in the extreme quantiles.}
    \label{fig:coco_poks_nano_gauss_diag}
\end{figure}

\begin{figure}[p]
    \centering
    \includegraphics[width=0.85\linewidth]{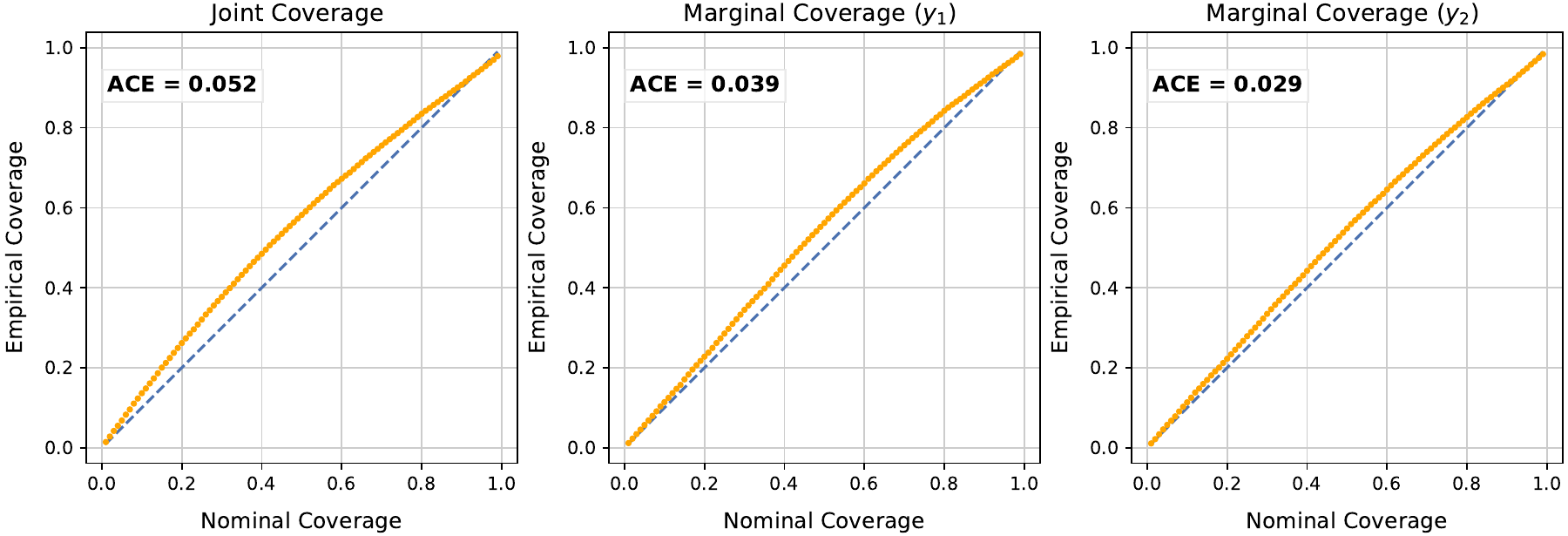}
    \vspace{0.15cm}
    \includegraphics[width=0.85\linewidth]{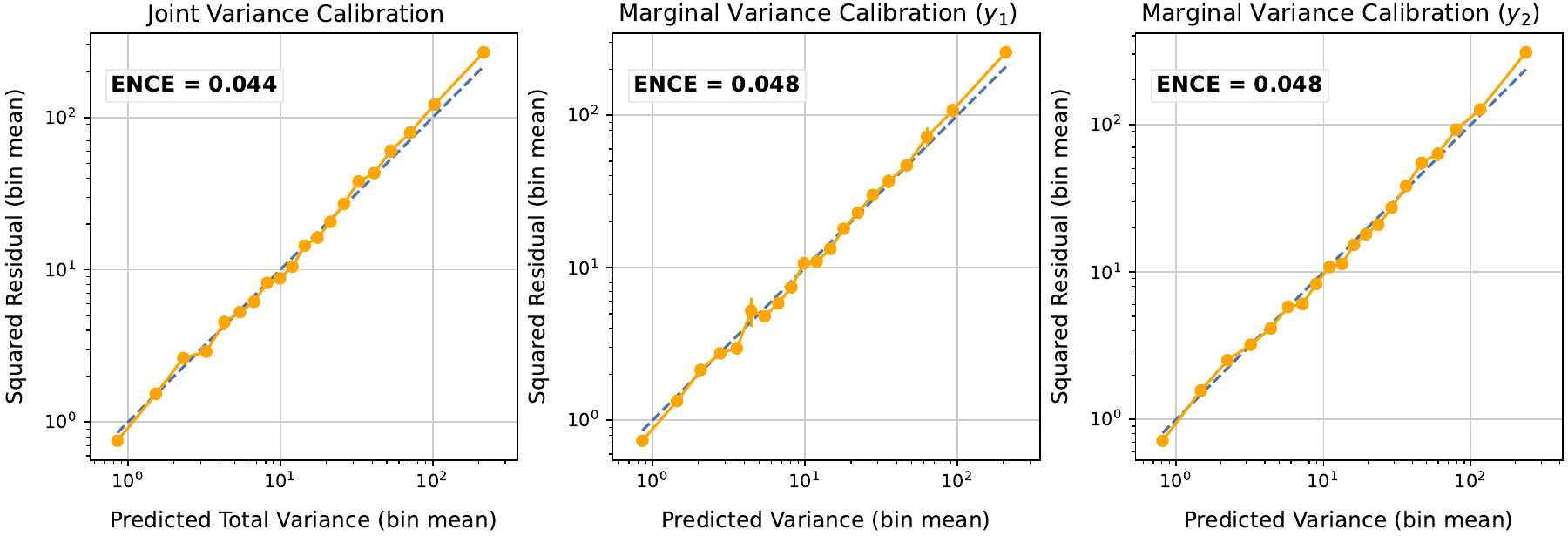}
    \vspace{0.15cm}
    \includegraphics[width=0.85\linewidth]{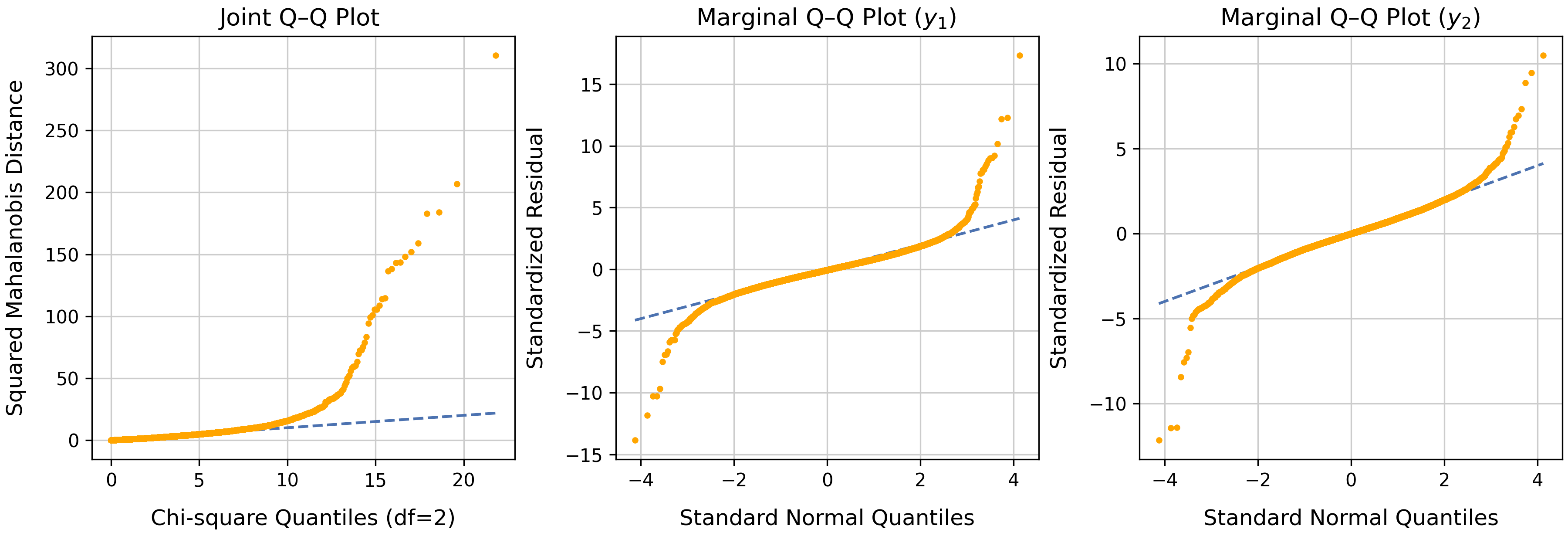}
    \caption{Gaussian Residual Distribution Fine-Tuning (see App.~\ref{app:residual_finetune}). Calibration diagnostics on COCO for YOLO-NAS-Pose Nano: Gaussian calibration with keypoint pruning. Same layout as \cref{fig:coco_nas_n_base_diag}.}
    \label{fig:coco_poks_nano_gauss_kpp_diag}
\end{figure}

\FloatBarrier

\newpage
\section{Vision-Based Aircraft Landing}
\label{app:vbl_details}

This appendix provides additional details on the \acf{vbl} system described in \cref{sec:app_vbl}. A system overview is shown in \cref{fig:vbl_architecture}.

We apply our post-hoc probabilistic extension introduced in \cref{sec:prob_extension} to a YOLO-Pose model that is trained to detect georeferenced runway keypoints, such as runway corners. For each detected keypoint $k\in\mathcal{K}^\mathrm{keep}$, the \ac{vbl} model outputs a mean image-plane location $\hat{\mathbf{y}}_{k}$ together with a calibrated covariance $\tilde{\mathbf{\Sigma}}_{k}$. In the \ac{vbl} pipeline, we use the Gaussian-calibrated covariances from \cref{subsec:calibration}, since downstream \ac{ekf}-based fusion requires an explicit measurement covariance. The underlying YOLO architecture is chosen to satisfy the stringent latency budget of \ac{vbl}, and our lightweight post-hoc uncertainty heads preserve this low-latency deployment regime.

\subsection{Vision-Based Landing as a Complementary Autoland Modality}
\label{app:vbl_motivation}
Although automatic landing systems are technologically mature, their routine use is constrained by operational capacity considerations and infrastructure dependencies.
Traditional \acf{ils} guidance can be affected by radio-frequency multipath and signal disturbances caused by aircraft and vehicles on the airport surface.
To preserve signal integrity during low-visibility and automatic-landing operations, \acf{atc} may need to protect \ac{ils} critical and sensitive areas~\citep{icao2020ilsCriticalSensitiveAreas}.
These protections can reduce runway capacity, which provides an operational incentive to limit routine autoland use when visual or standard instrument conditions allow manual landings.

The modernized \acf{gls}, based on \acf{gbas}, alleviates some limitations of conventional \ac{ils} infrastructure but introduces different dependencies and certification challenges.
First, \ac{gls} relies on \acf{gnss}, making it vulnerable to localized radio-frequency interference, including jamming and spoofing~\citep{moralesFerre2020intentionalInterferenceSatelliteNavigation}.
Second, ionospheric anomalies remain a key integrity concern for high-category \ac{gls} operations, complicating the certification of Category III \ac{gls} autoland capability~\citep{murphy2010gbasIonosphericAnomalyErrors}.
At the same time, there is a strong operational incentive to increase the availability of automatic landing capability, for example, to support robust all-weather operations and reduce flight-crew workload during demanding approach and landing phases~\citep{caldwell2009fatigueCountermeasuresAviation}.

\ac{vbl} provides a complementary landing-navigation modality that is independent of \ac{ils} ground transmitters and does not rely directly on \ac{gnss}.
It is therefore not subject to the same radio-navigation infrastructure and \ac{gnss}-interference limitations, although it introduces its own perception, visibility, and certification challenges.

\subsection{Covariance-Weighted 6-DoF Pose Estimation}
\label{app:pnp}
Let $\mathbf{X}_{k}\in\mathbb{R}^3$ denote the known 3D position of runway keypoint $k$ in a known world reference frame, and let $\pi(\mathbf{X}_{k},\mathbf{R},\mathbf{t})$ denote the perspective projection of $\mathbf{X}_{k}$ into the image under rotation $\mathbf{R}$ and translation $\mathbf{t}$. A covariance-weighted \ac{pnp} formulation~\citep{FerrazBMN2014, vakhitov2021uncertaintyawarecameraposeestimation} estimates the camera pose by
\begin{equation}
\label{eq:vbl_weighted_pnp_full}
(\hat{\mathbf{R}},\hat{\mathbf{t}})
\;=\;
\arg\min_{\mathbf{R},\,\mathbf{t}}
\sum_{k\in\mathcal{K}^\mathrm{keep}}
\left\|
\hat{\mathbf{y}}_{k}-\pi(\mathbf{X}_{k},\mathbf{R},\mathbf{t})
\right\|_{\tilde{\mathbf{\Sigma}}_{k}^{-1}}^2,
\end{equation}
where the notation $\|\mathbf{a}\|_{\mathbf{A}}^2 \triangleq \mathbf{a}^{\top}\mathbf{A}\mathbf{a}$ denotes the squared norm induced by a matrix $\mathbf{A}$.
The rotation $\hat{\mathbf{R}}$, which reflects the aircraft's attitude (i.e., roll, pitch, and yaw), is essential for the uncertainty propagation described in \cref{app:uncertainty_propagation} and is also used for position integrity monitoring.

If the \ac{vbl} system design allows the use of an onboard attitude source, the aircraft attitude can be provided by the \ac{irs}.
Denoting the corresponding rotation by $\mathbf{R}_{\mathrm{IRS}}$, pose estimation reduces to a translation-only problem,
\begin{equation}
\label{eq:vbl_weighted_pnp_translation}
\hat{\mathbf{t}}
\;=\;
\arg\min_{\mathbf{t}}
\sum_{k\in\mathcal{K}^\mathrm{keep}}
\left\|
\hat{\mathbf{y}}_{k}-\pi(\mathbf{X}_{k},\mathbf{R}_{\mathrm{IRS}},\mathbf{t})
\right\|_{\tilde{\mathbf{\Sigma}}_{k}^{-1}}^2.
\end{equation}
By reducing the number of unknowns, this formulation improves robustness and enables stable estimation from as few as two non-degenerate keypoint correspondences.

Since \ac{pnp} estimates the camera pose, with the translation-only formulation estimating only the camera position, the known camera-to-aircraft extrinsic calibration is used to express the resulting position estimate in the aircraft reference frame required by downstream sensor fusion.

\subsection{First-Order Uncertainty Propagation}
\label{app:uncertainty_propagation}
To propagate image-plane keypoint covariances into a 3D position covariance, we stack the predicted keypoints into
\begin{equation}
\hat{\mathbf{y}}
\;=\;
\big[
\hat{\mathbf{y}}_{k_1}^{\top}, \dots, \hat{\mathbf{y}}_{k_{|\mathcal{K}^\mathrm{keep}|}}^{\top}
\big]^{\top}
\end{equation}
and define the corresponding block-diagonal covariance
\begin{equation}
\tilde{\mathbf{\Sigma}}_{\mathbf{y}}
\;=\;
\operatorname{blkdiag}\!\left(
\tilde{\mathbf{\Sigma}}_{k_1}, \dots, \tilde{\mathbf{\Sigma}}_{k_{|\mathcal{K}^\mathrm{keep}|}}
\right).
\end{equation}
Moreover, let $\bar{\mathbf{R}}$ denote the rotation used for uncertainty propagation, with $\bar{\mathbf{R}}\triangleq\hat{\mathbf{R}}$ for the full \ac{pnp} formulation and $\bar{\mathbf{R}}\triangleq\mathbf{R}_{\mathrm{IRS}}$ for the translation-only formulation. 
We define the stacked reprojection function
\begin{equation}
\mathbf{h}(\mathbf{t})
\;=\;
\big[
\pi(\mathbf{X}_{k_1},\mathbf{\bar{R}},\mathbf{t})^\top,\dots,
\pi(\mathbf{X}_{k_{|\mathcal{K}^\mathrm{keep}|}},\mathbf{\bar{R}},\mathbf{t})^\top
\big]^\top
\end{equation}
and let
\begin{equation}
\mathbf{J}
\;=\;
\left.
\frac{\partial \mathbf{h}(\mathbf{t})}{\partial \mathbf{t}}
\right|_{\mathbf{t}\,=\,\hat{\mathbf{t}}}
\end{equation}
be its Jacobian at the estimated position. A first-order linearization yields the approximate translation covariance
\begin{equation}
\label{eq:vbl_uq_prop}
\mathbf{\Sigma}_{\mathbf{t}}
\;\approx\;
\left(
\mathbf{J}^{\top}
\tilde{\mathbf{\Sigma}}_{\mathbf{y}}^{-1}
\mathbf{J}
\right)^{-1}.
\end{equation}
This approximation is computationally efficient and therefore compatible with the real-time constraints of \ac{vbl}. Its accuracy depends on the local linearity of the projection model and can degrade under weak keypoint geometry or large image-plane errors. We refer to \citet{valentin2024probabilisticparameterestimatorscalibration} for further details.

\subsection{Integration into Sensor Fusion}
\label{app:sensor_fusion}
The final \ac{vbl} output consists of the runway-relative position estimate $\hat{\mathbf{t}}$ of the aircraft and its corresponding covariance $\mathbf{\Sigma}_{\mathbf{t}}$.
These quantities are forwarded to the \ac{ekf}-based sensor-fusion module, which combines \ac{vbl} with complementary navigation sensors, such as the \ac{gnss}, the \ac{ils}, the \ac{irs}, and the \ac{ra}.
Since the downstream fusion relies on a time-uncorrelated measurement model, we deliberately avoid temporal smoothing and feature tracking in \ac{vbl}.
Each update is computed from a single image, which keeps the vision measurement model aligned with the covariance interpretation used downstream.
Further details are provided, for example, in \citet{roumeliotis2000ekf_fusion}.

\subsection{Vision-Based Landing Dataset}
\label{app:vbl_dataset}

\begin{figure}[tbp]
    \centering
    \includegraphics[width=0.495\linewidth]{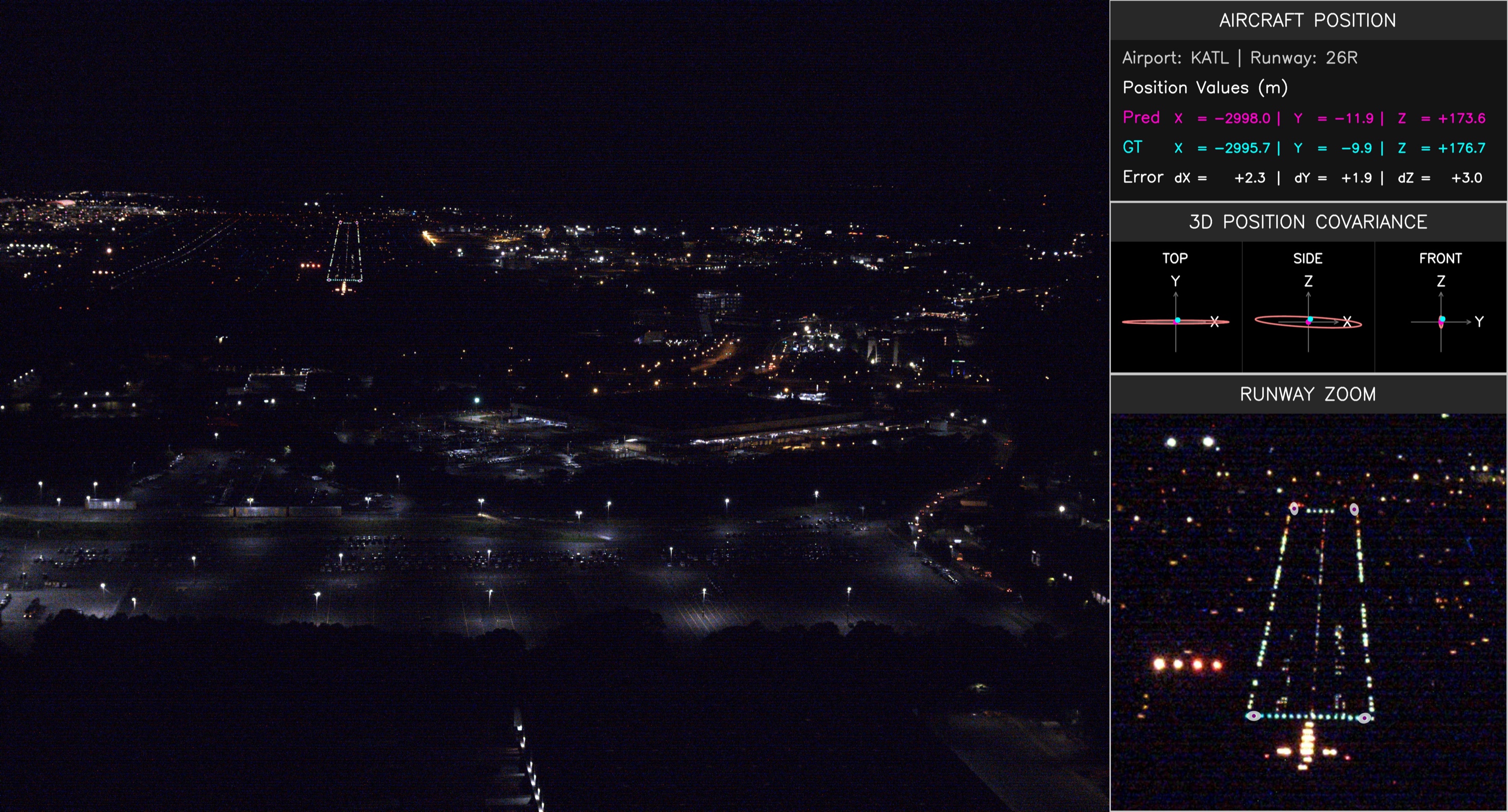}
    \includegraphics[width=0.495\linewidth]{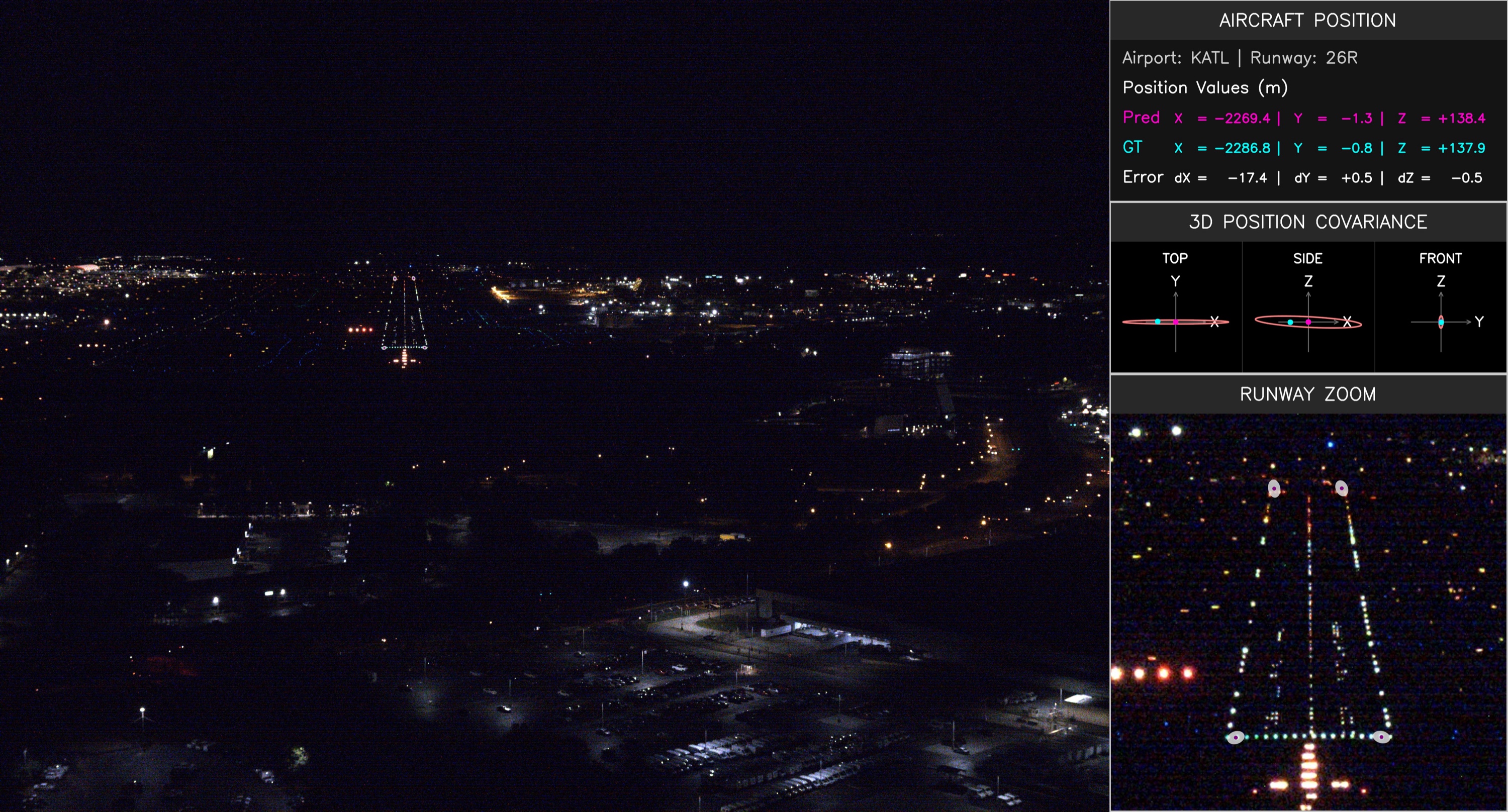}
    \includegraphics[width=0.495\linewidth]{figures/vbl/KATL/KATL_26R_FT237_20_step_00300.jpg}
    \includegraphics[width=0.495\linewidth]{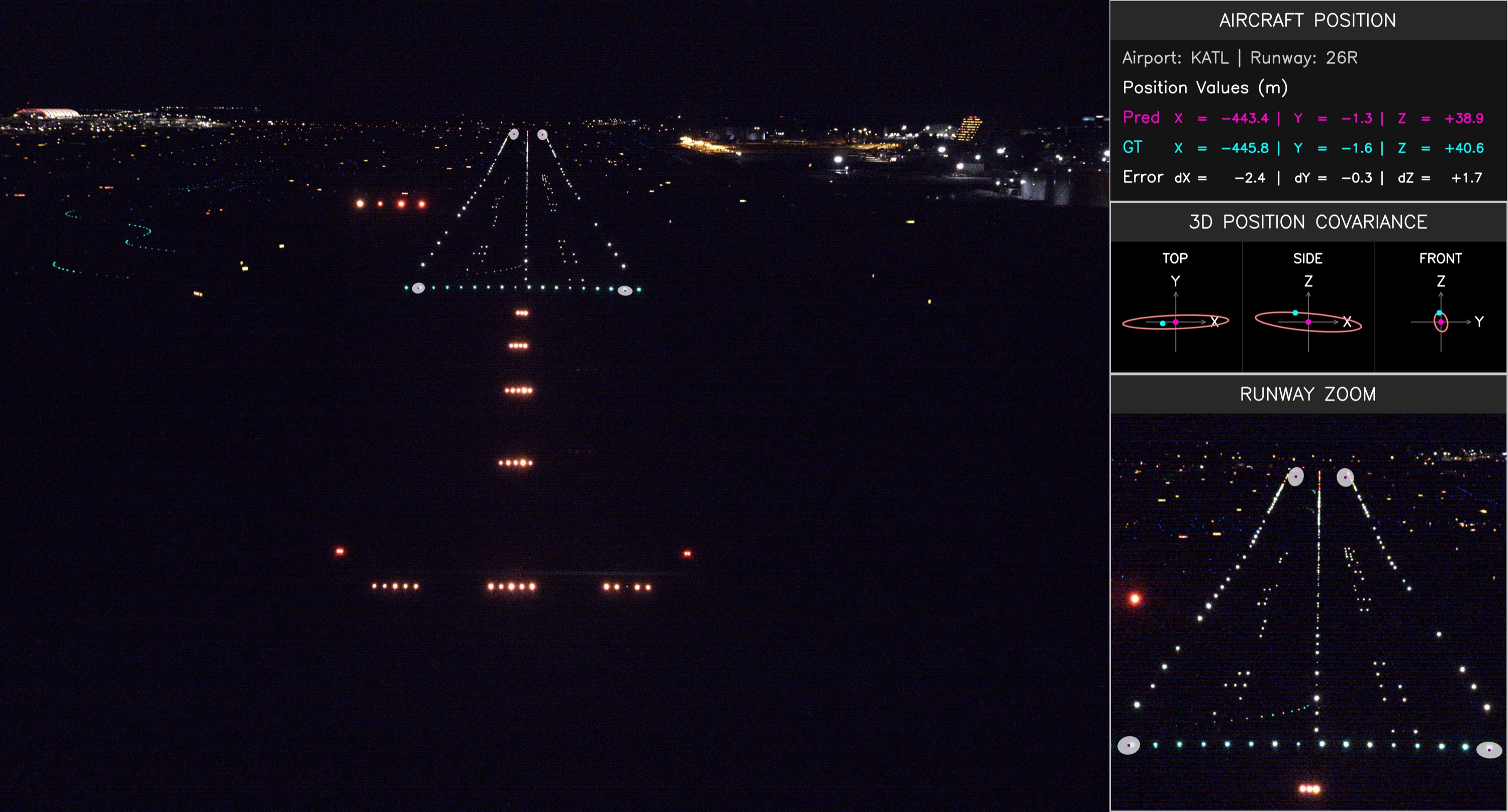}
    \caption{Vision-based landing at night at Hartsfield--Jackson Atlanta International Airport (KATL).}
    \label{fig:vbl_KATL_seq}
\end{figure}

The \ac{vbl} model is trained and evaluated on a large-scale internal aircraft-landing dataset comprising over $10^6$ real images at ${1024\times1024}$ resolution, captured by onboard cameras during approach and landing at multiple airports.
Each sequence begins at approximately $6000\,$m direct distance from the runway threshold and continues through touchdown.
The dataset covers varied operational conditions, including daytime, dusk, dawn, and nighttime (\cref{fig:vbl_KATL_seq}).
Each image is annotated with the 2D locations of four georeferenced runway-corner keypoints per runway, with up to three runways visible per image.
The 3D world coordinates of the runway corners are known from aeronautical databases, and differential-GPS measurements provide reference aircraft positions for evaluating the propagated 3D estimates.

We split the data into training (80\%), validation (5\%), test (10\%), and calibration (5\%) subsets, with the calibration set disjoint from training and test data for post-hoc uncertainty calibration (\cref{subsec:calibration}).
The post-hoc probabilistic extension is trained without data augmentation so that the learned residual distribution matches the distribution encountered at inference time, as described in \cref{sec:app_vbl}.

Although this dataset is proprietary, the methodology is dataset-agnostic.
Applying the pipeline described in this appendix to a different runway dataset requires three components: (i)~georeferenced 3D coordinates for the detected keypoints, (ii)~a calibrated camera model for the perspective projection $\pi$, and (iii)~reference positions for calibration and evaluation.
These components are standard in runway detection and pose estimation settings~\citep{balduzzi2021runway_landing_guidance,valentin2024probabilisticparameterestimatorscalibration}.

\pagebreak

\subsection{Landing Sequences and Detailed Calibration Diagnostics}
\label{app:vbl_seqs_and_detailed_calib_diag}
\FloatBarrier

\begin{figure}[h]
    \centering
    \includegraphics[width=0.495\linewidth]{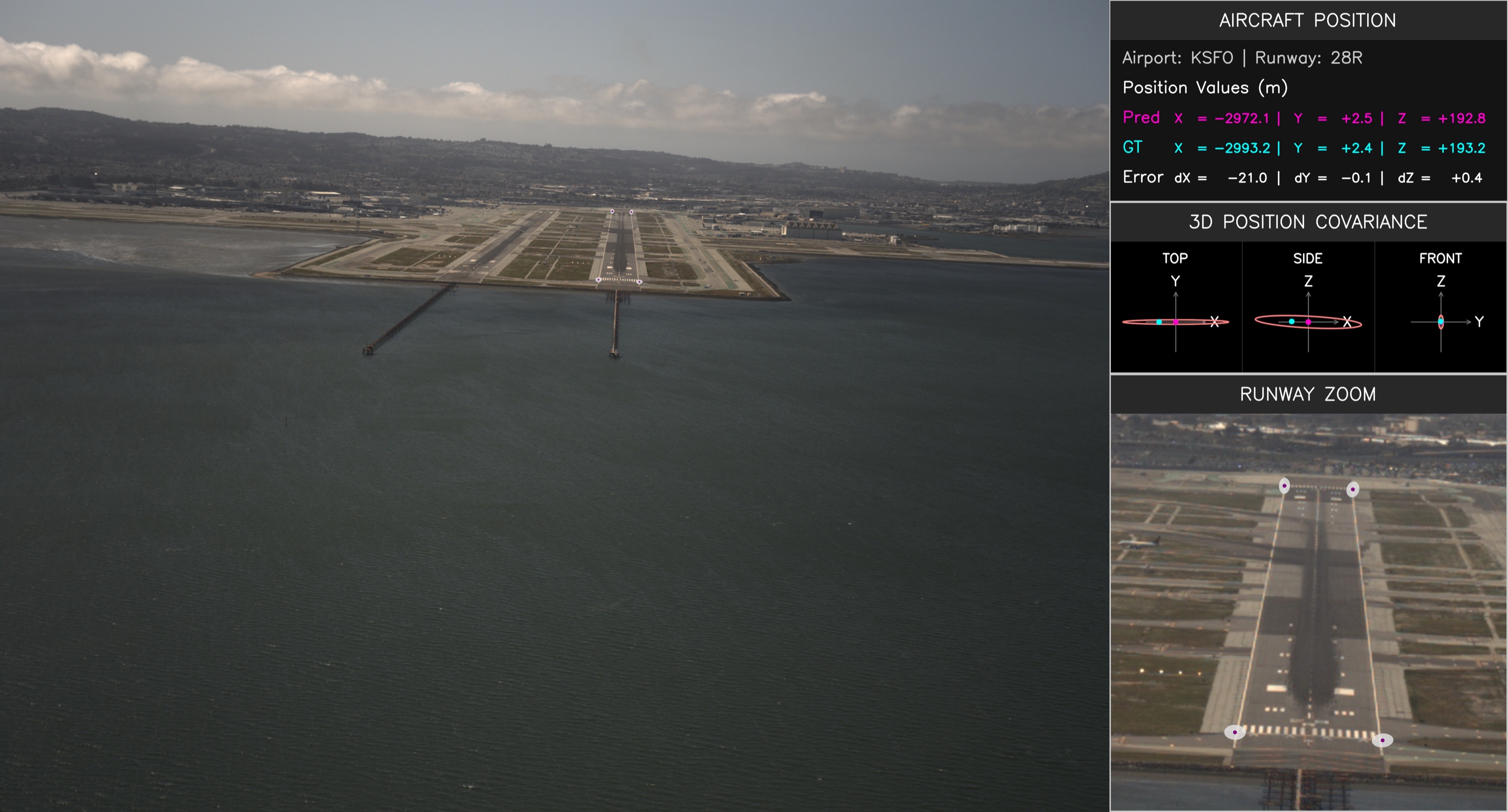}
    \includegraphics[width=0.495\linewidth]{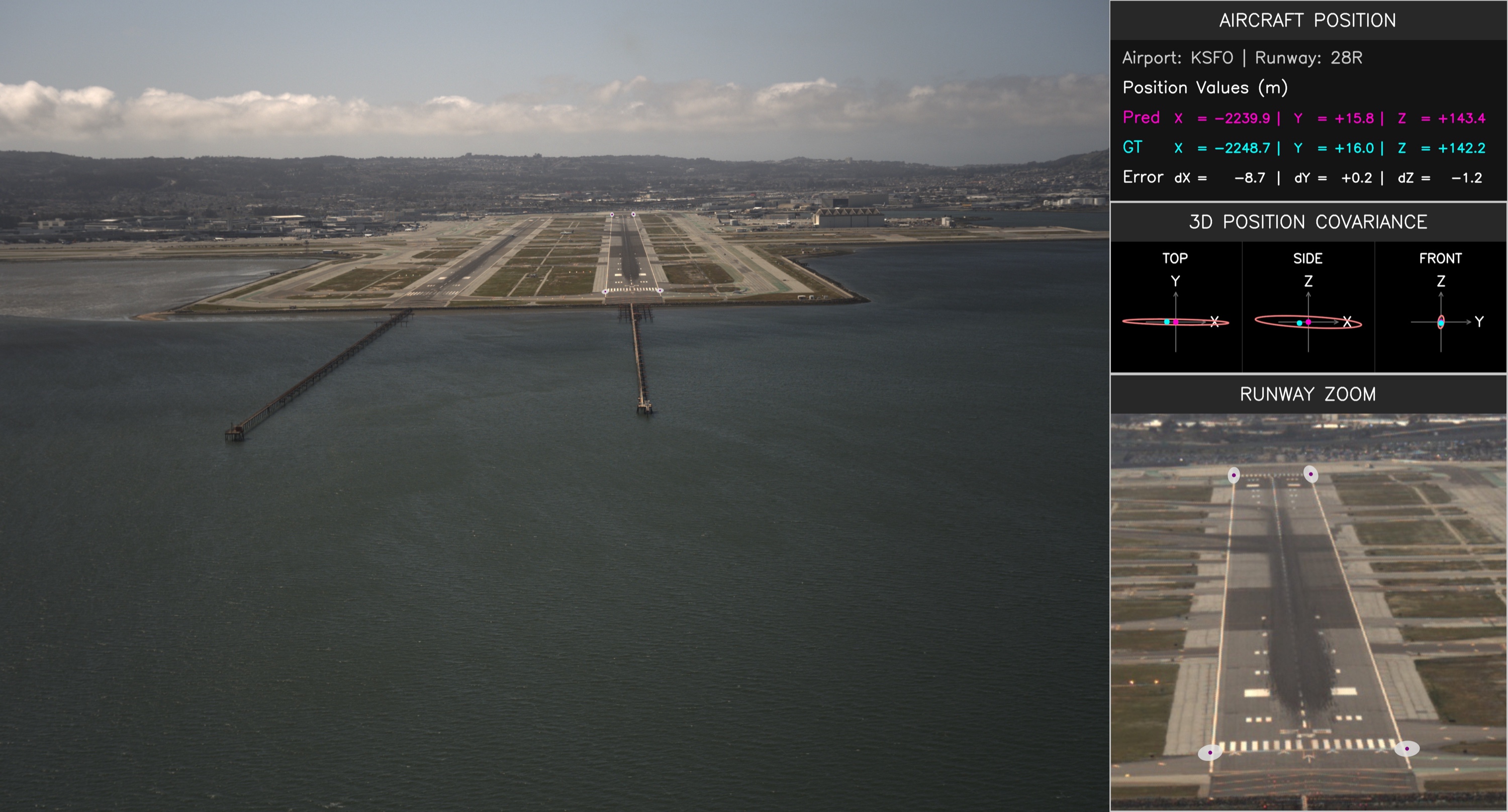}
    \includegraphics[width=0.495\linewidth]{figures/vbl/KSFO/KSFO_28R_FT126_6_step_00200.jpg}
    \includegraphics[width=0.495\linewidth]{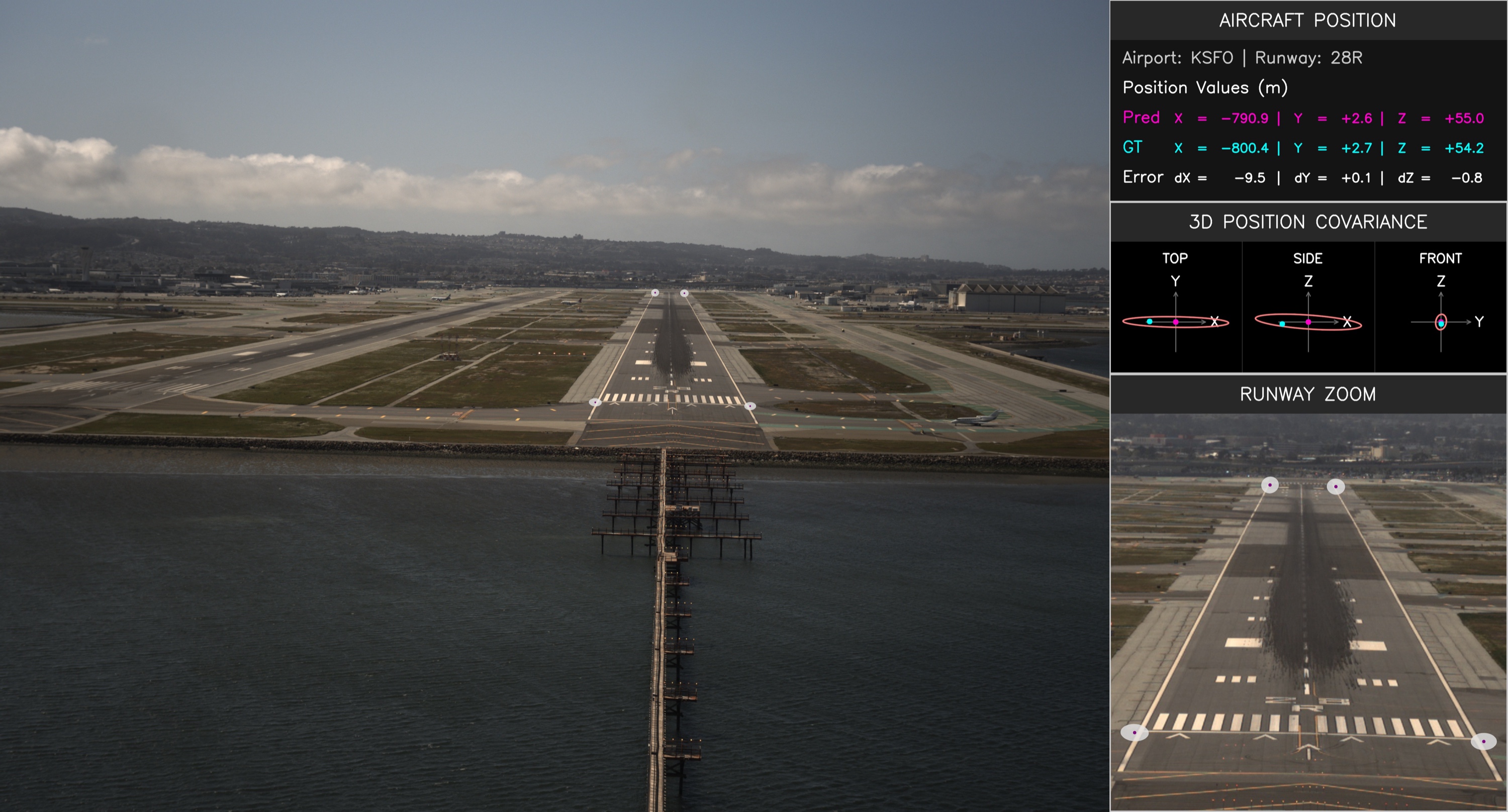}
    \caption{Vision-based landing at San Francisco International Airport (KSFO).}
    \label{fig:vbl_KSFO_seq}
\end{figure}
\begin{figure}[h]
    \centering
    \includegraphics[width=0.495\linewidth]{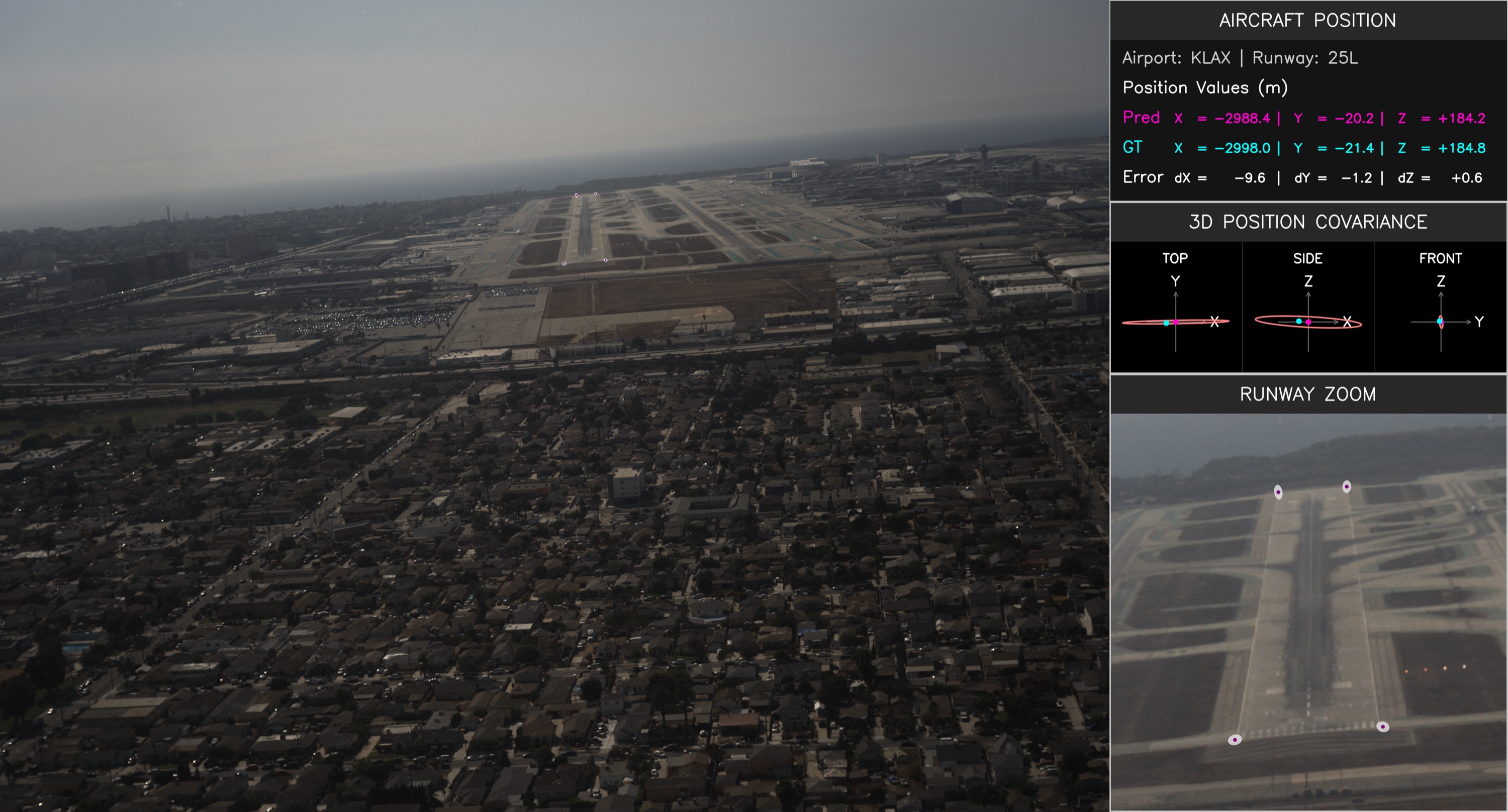}
    \includegraphics[width=0.495\linewidth]{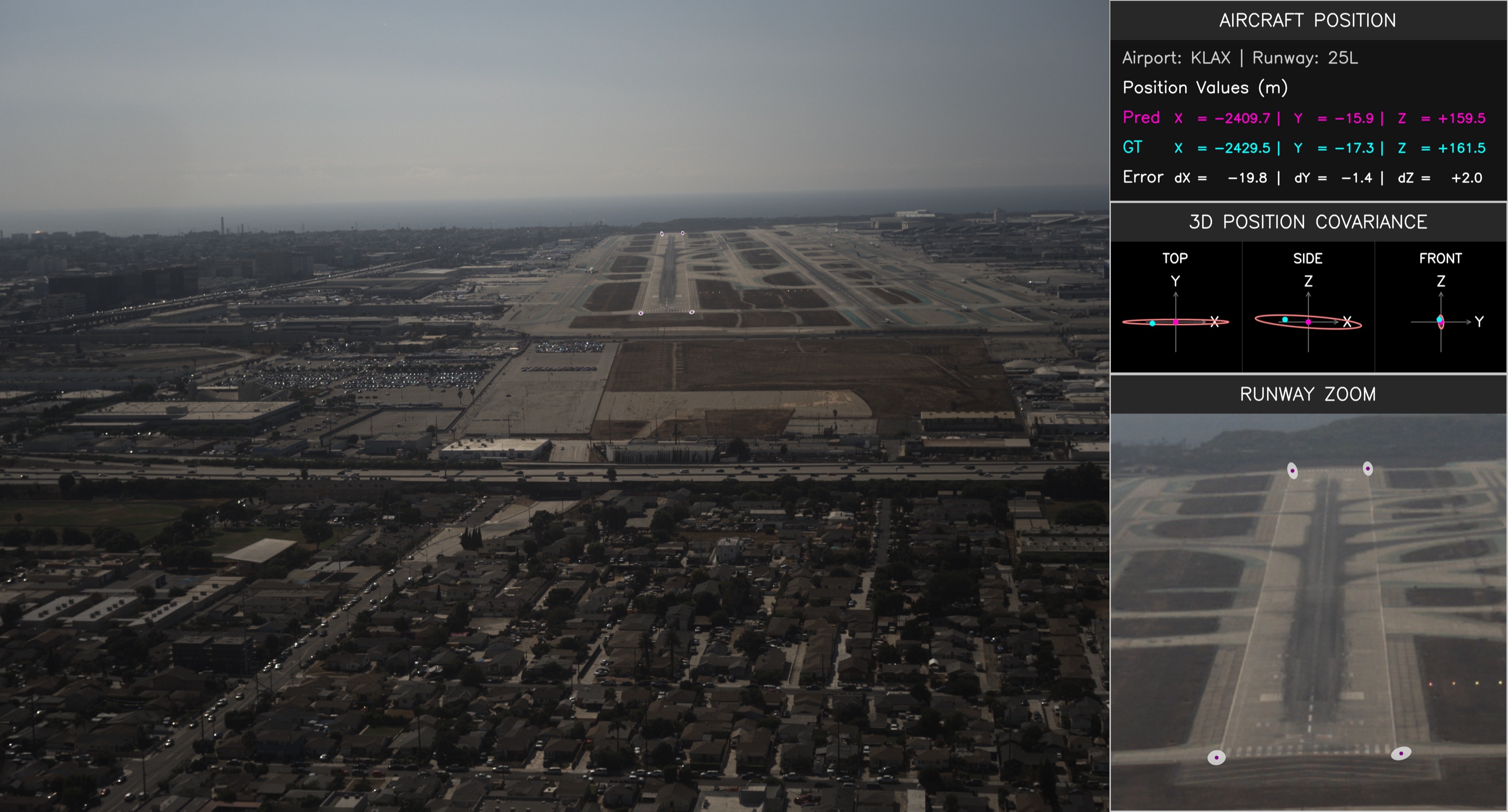}
    \includegraphics[width=0.495\linewidth]{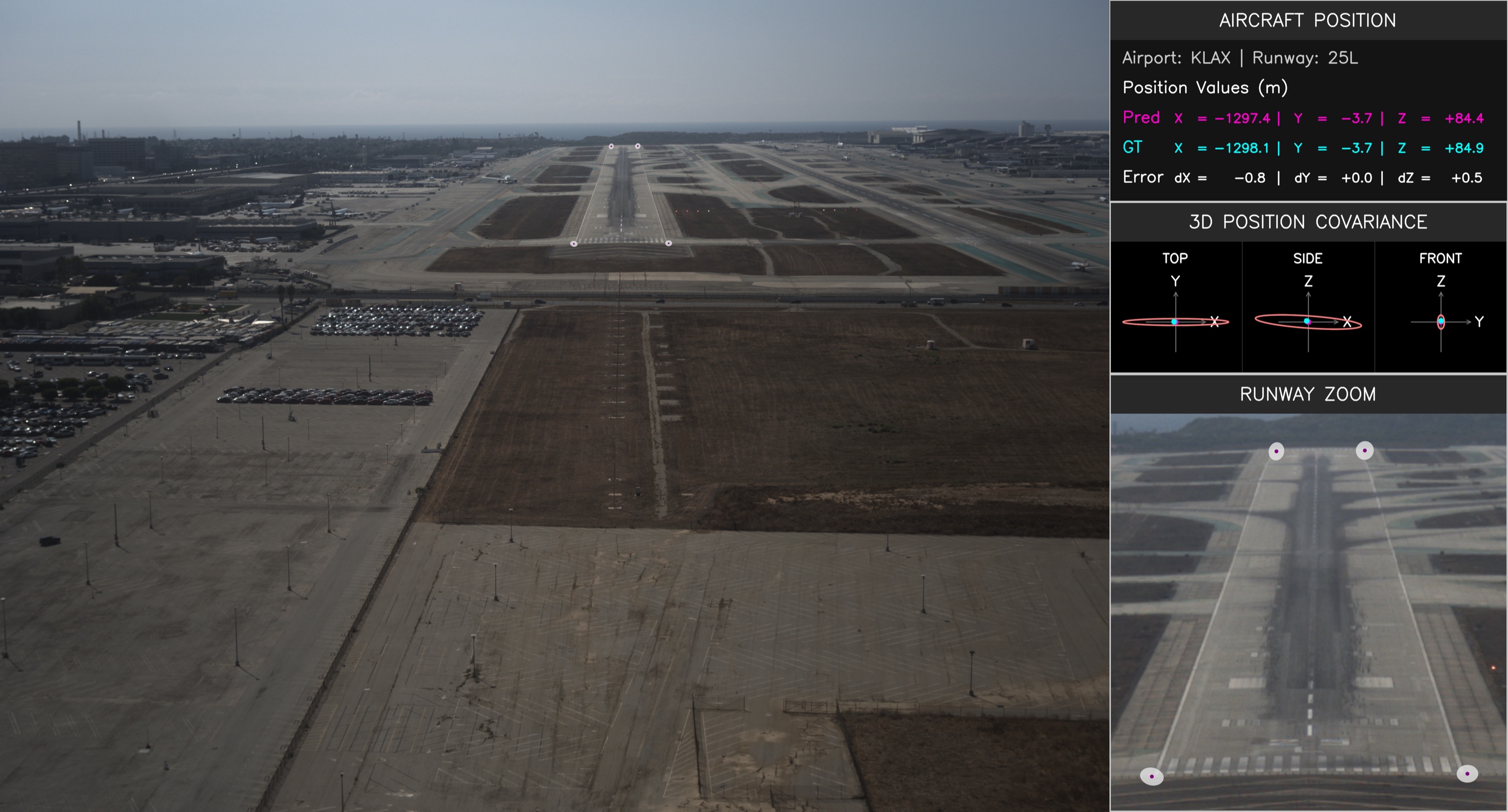}
    \includegraphics[width=0.495\linewidth]{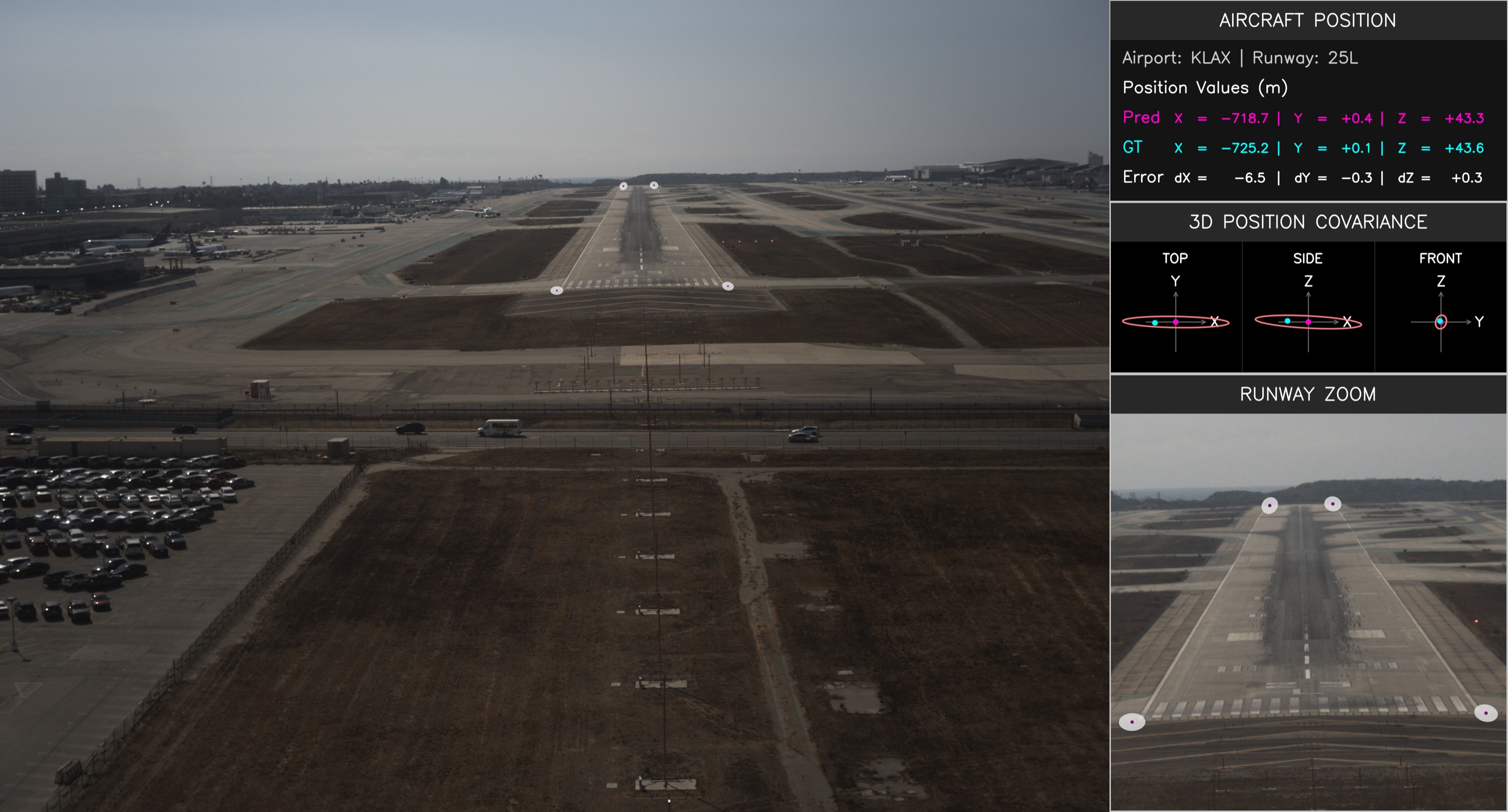}
    \caption{Vision-based landing at Los Angeles International Airport (KLAX).}
    \label{fig:vbl_KLAX_seq}
\end{figure}
\begin{figure}[tbp]
    \centering
    \includegraphics[width=0.495\linewidth]{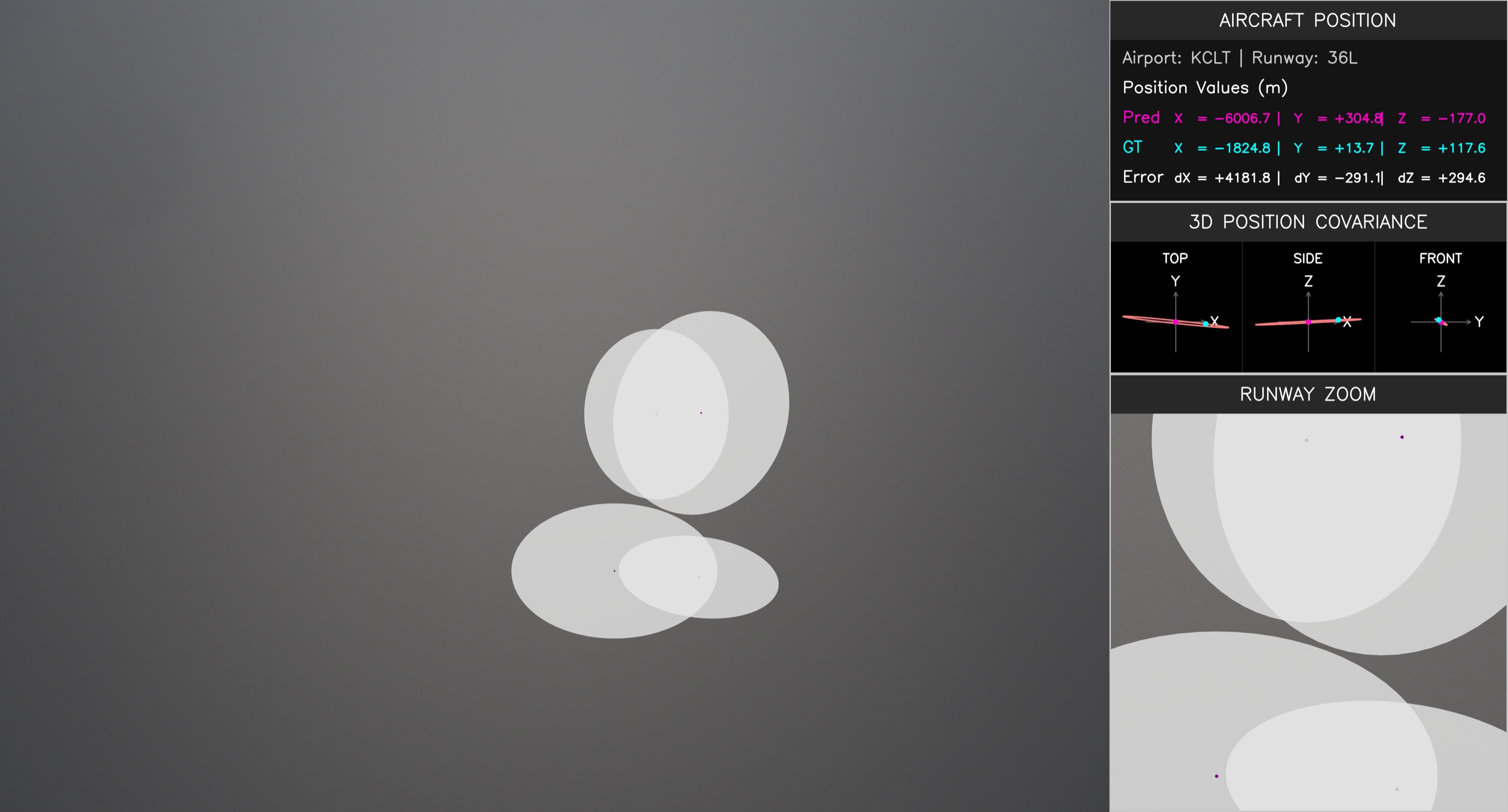}
    \includegraphics[width=0.495\linewidth]{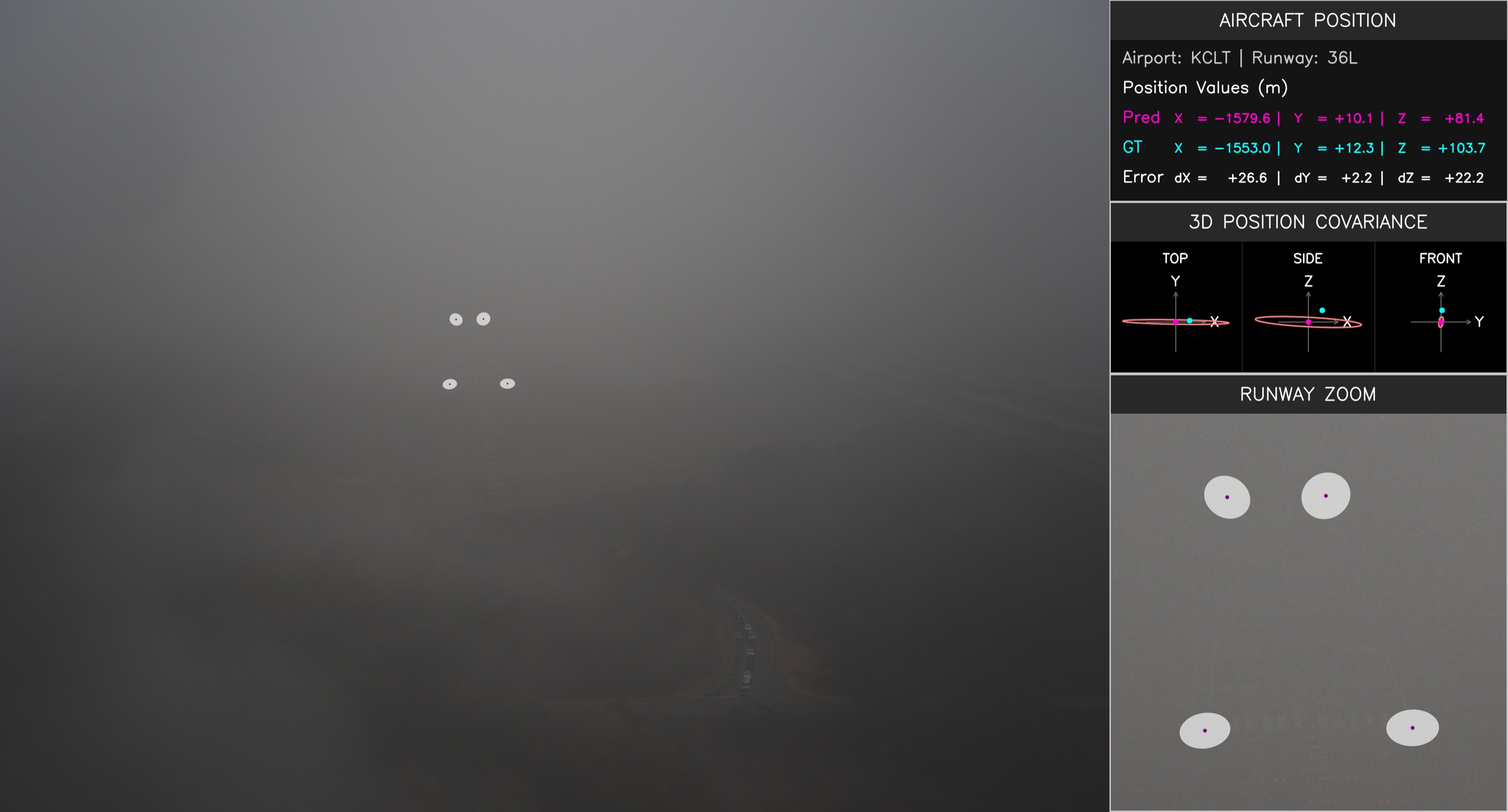}
    \includegraphics[width=0.495\linewidth]{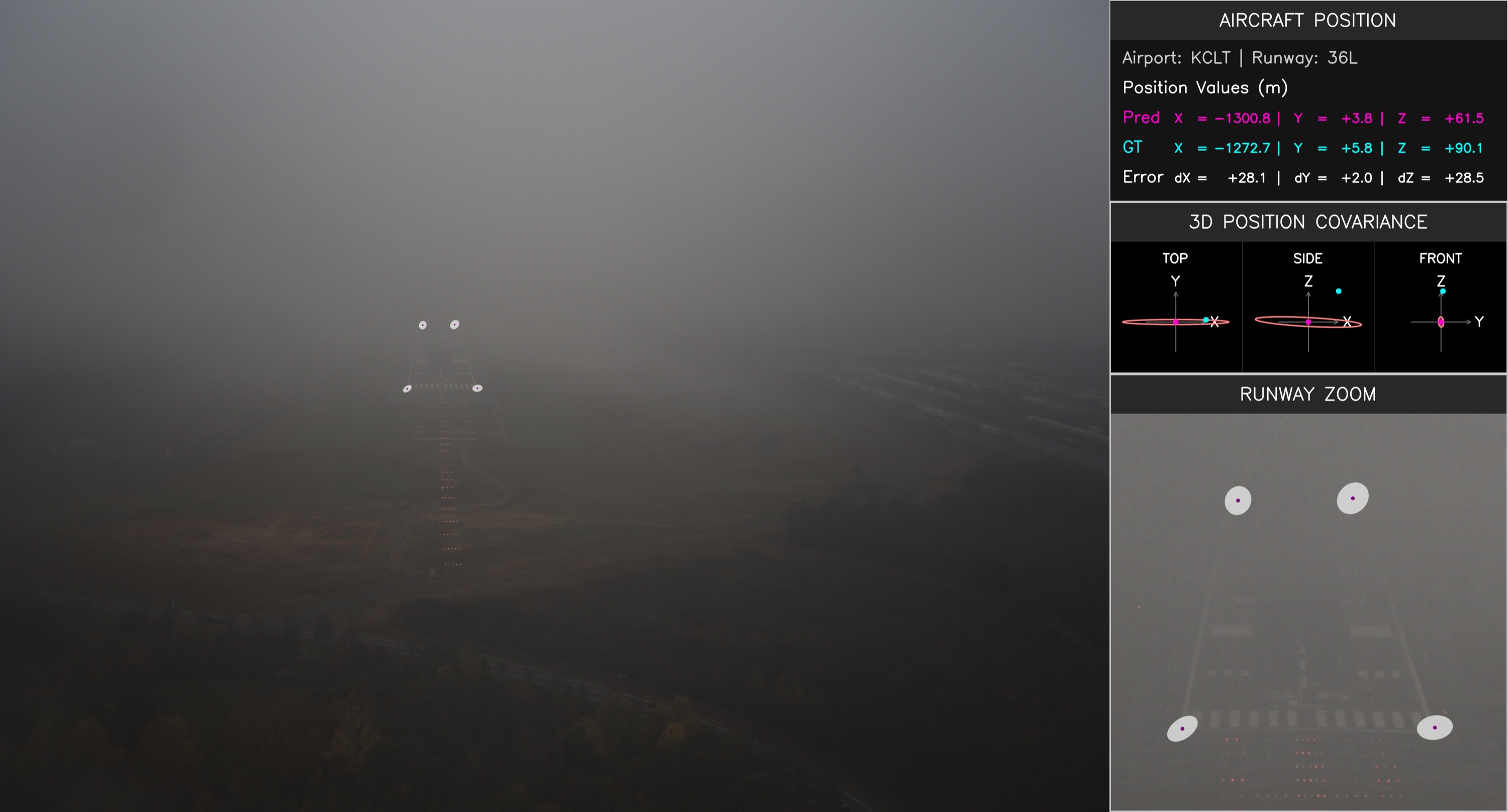}
    \includegraphics[width=0.495\linewidth]{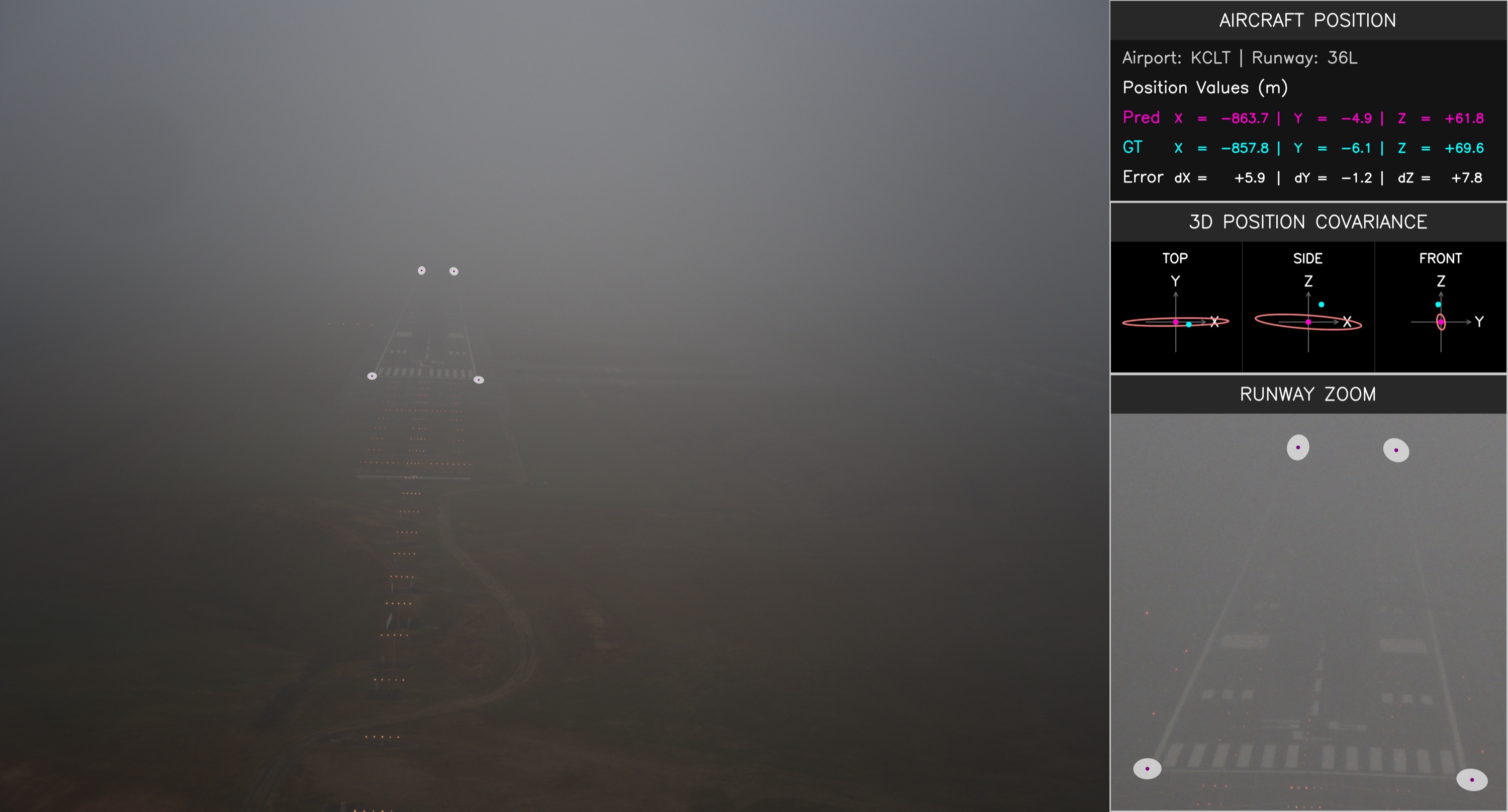}
    \caption{Vision-based landing under fog (distribution shift w.r.t. the training data) at Charlotte Douglas International Airport (KCLT). These examples illustrate that calibrated keypoint covariances learned under in-distribution conditions do not capture the epistemic uncertainty associated with out-of-distribution inputs. Modeling it remains an important direction for future work, particularly in safety-critical applications.}
    \label{fig:vbl_ood_KCLT_seq}
\end{figure}

\begin{figure}[tbp]
    \centering

    \begin{subfigure}[b]{\linewidth}
        \centering
        \includegraphics[
            width=0.49\linewidth,
            trim={0px 137px 0px 137px},
            clip        
        ]{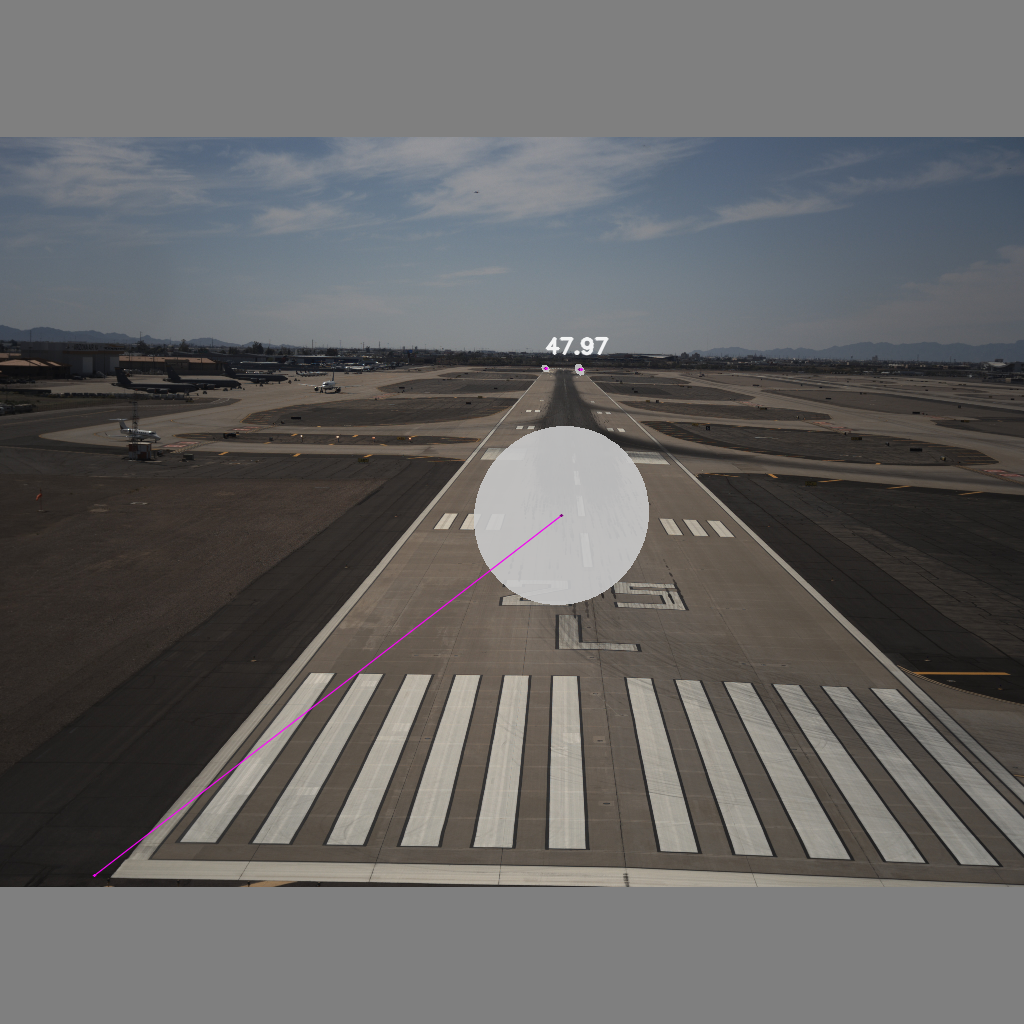}
        \hfill
        \includegraphics[
            width=0.49\linewidth,
            trim={0px 137px 0px 137px},
            clip        
        ]{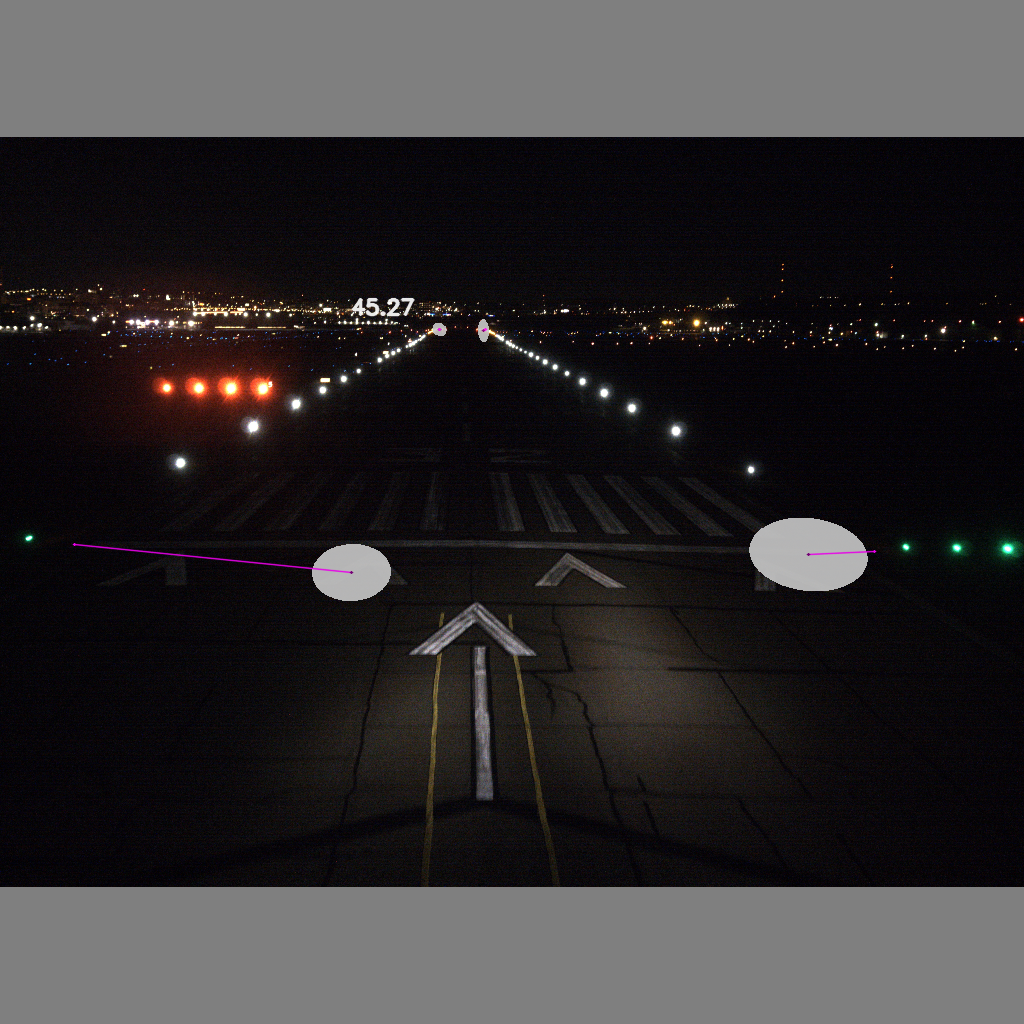}
        \caption{Before keypoint pruning}
    \end{subfigure}
    \hfill

    \vspace{0.4cm}
    
    \begin{subfigure}[b]{\linewidth}
        \centering
        \includegraphics[
            width=0.49\linewidth,
            trim={0px 137px 0px 137px},
            clip        
        ]{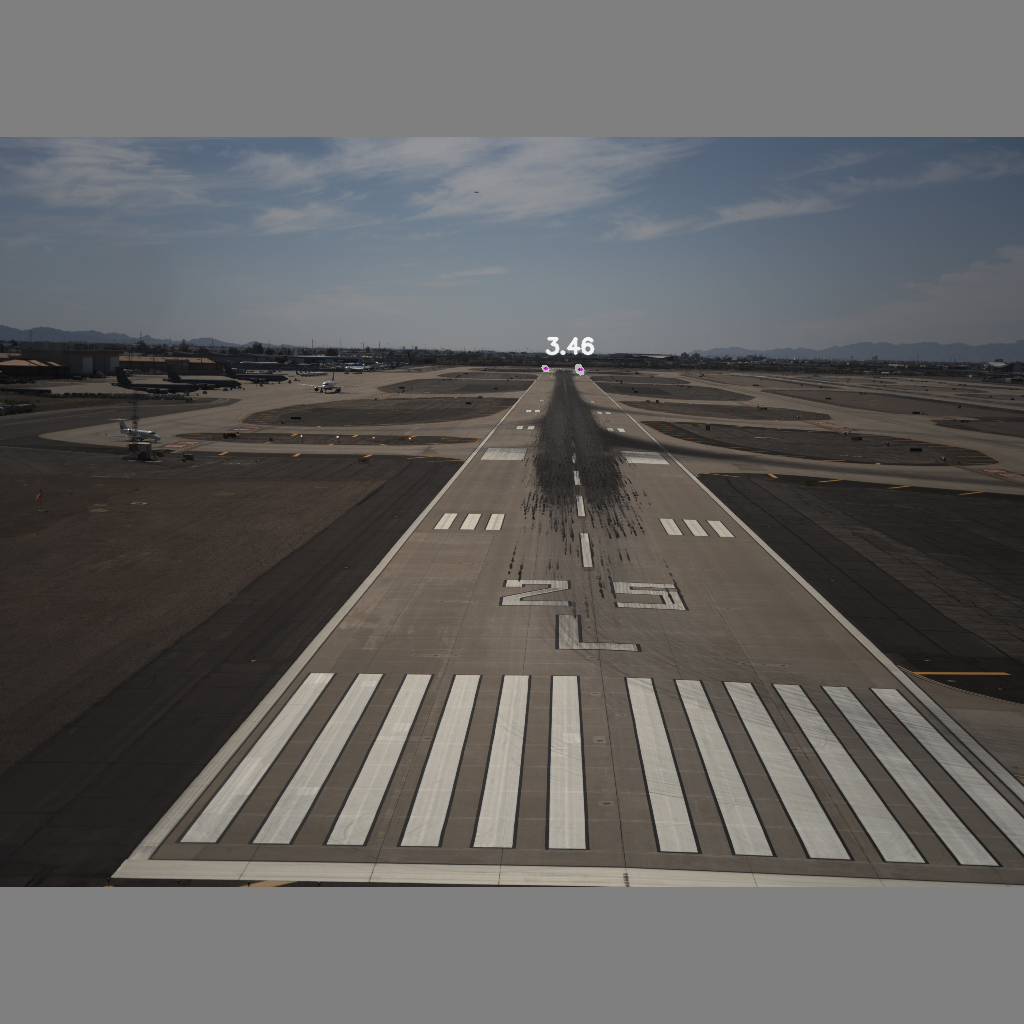}
        \hfill
        \includegraphics[
            width=0.49\linewidth,
            trim={0px 137px 0px 137px},
            clip        
        ]{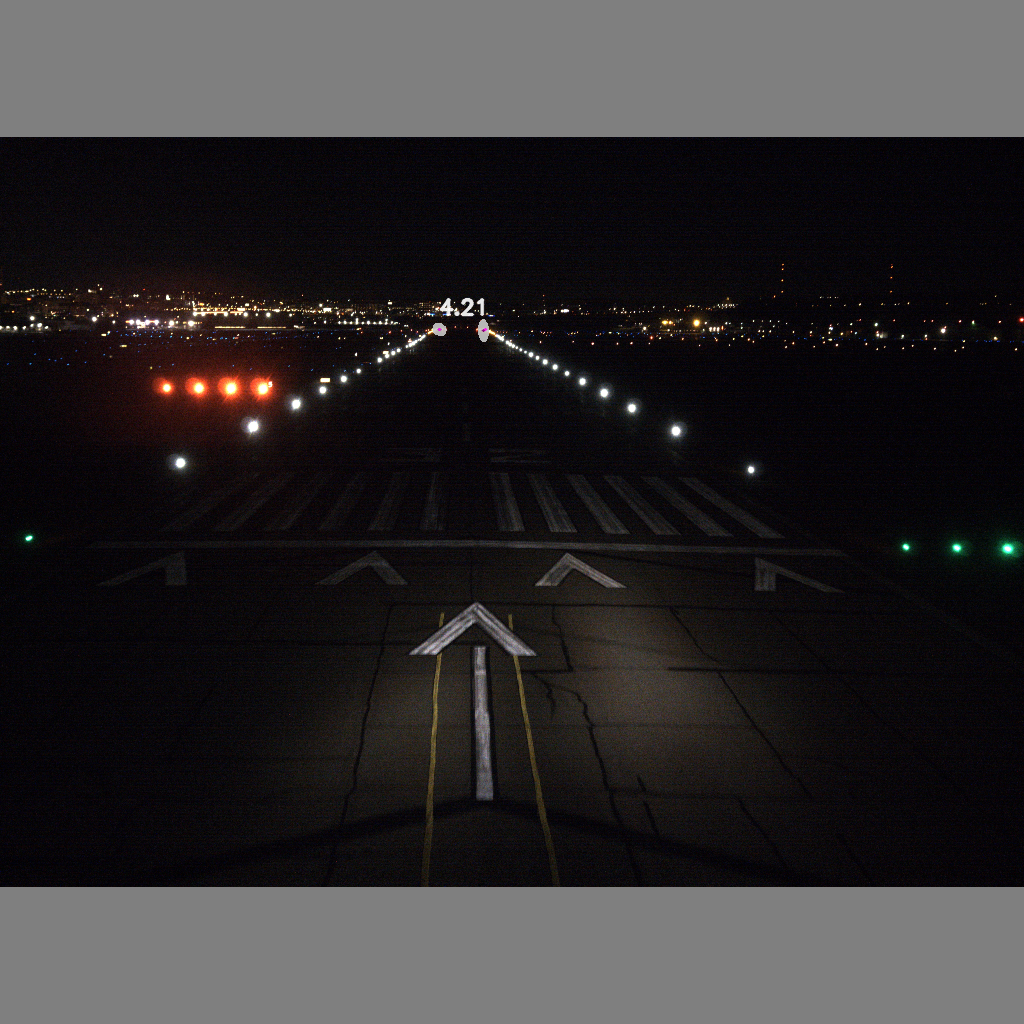}
        \caption{After keypoint pruning}
    \end{subfigure}

    \caption{
    Example keypoint predictions before and after pruning uncertain keypoints. 
    The gray ellipses indicate the 95\% probability mass of the predicted keypoint distributions, while the purple lines visualize the residual between prediction and ground truth. The displayed values correspond to the associated \acp{nll}.
    }
    \label{fig:vbl_keypoint_pruning}
\end{figure}

\begin{figure}[tbp]
    \centering
    \includegraphics[width=0.85\linewidth]{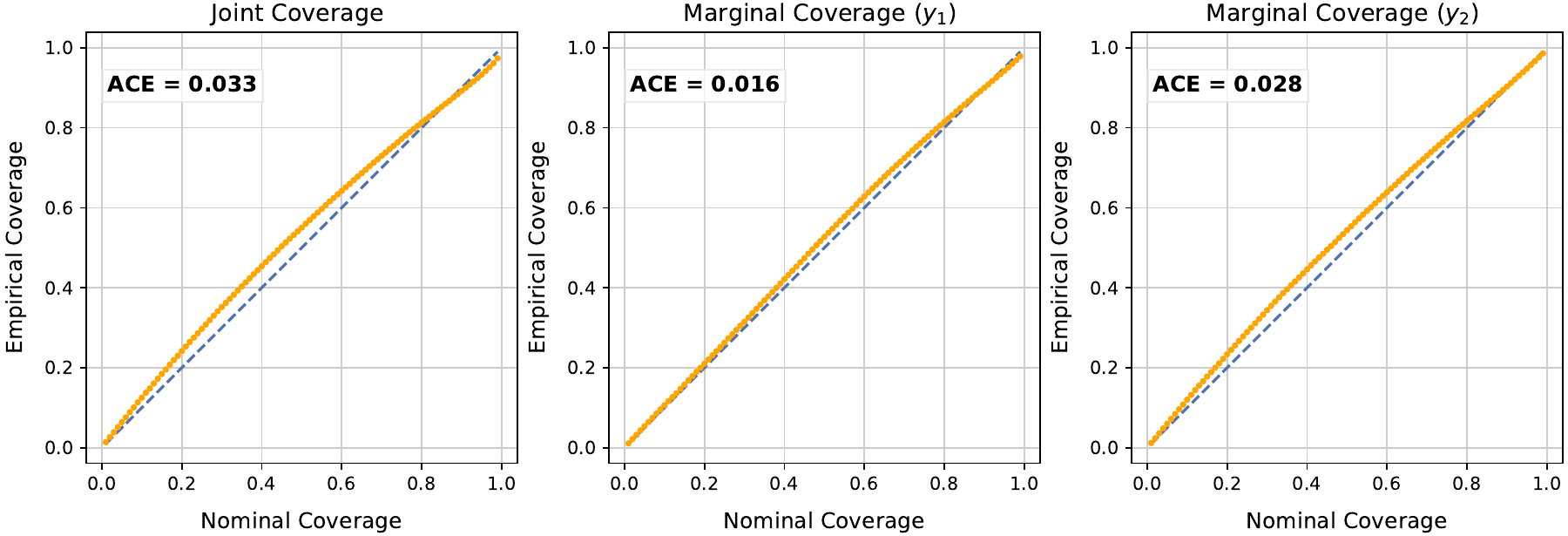}
    \vspace{0.15cm}
    \includegraphics[width=0.85\linewidth]{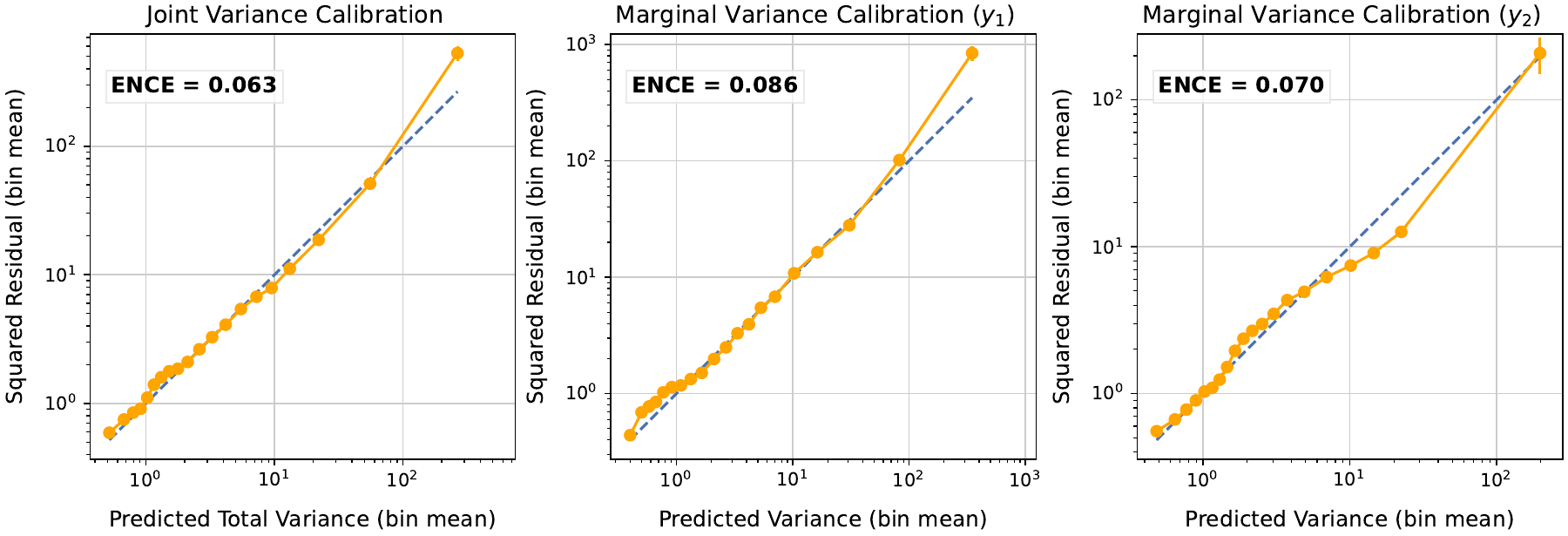}
    \vspace{0.15cm}
    \includegraphics[width=0.85\linewidth]{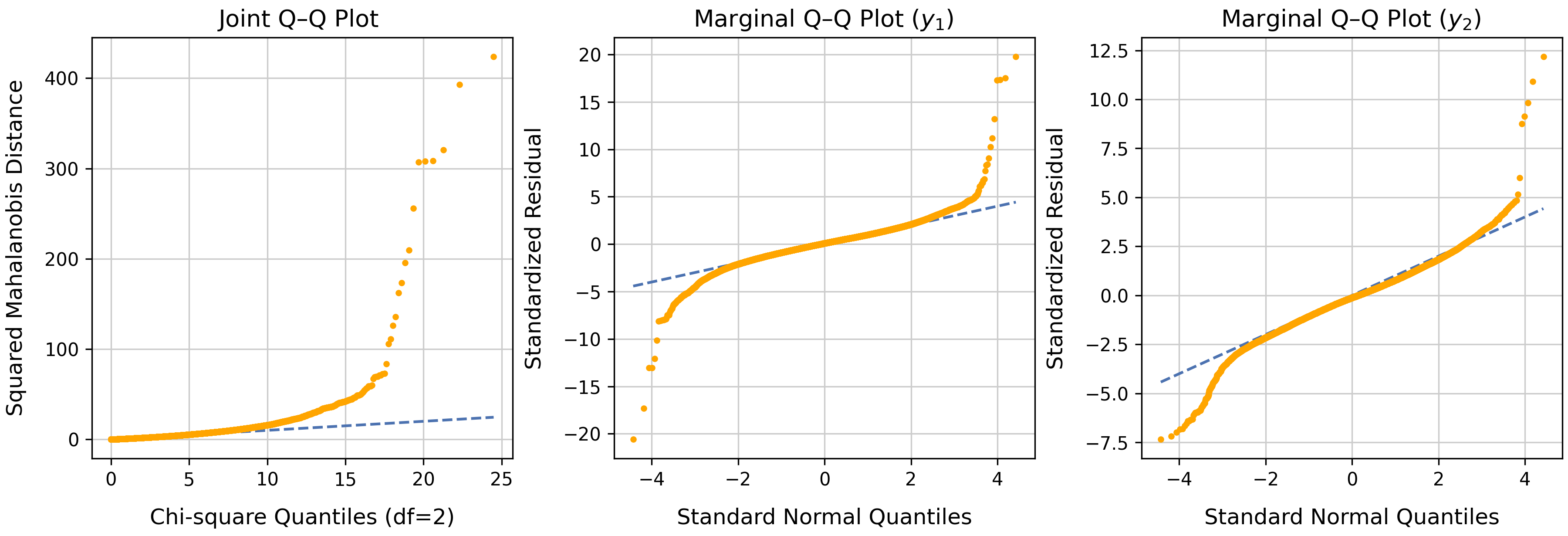}
    \caption{Calibration diagnostics for \ac{vbl}: Gaussian calibration.
    Top to bottom: coverage plots, variance calibration plots, and \ac{qq} plots.
    Each row shows the joint (left) and the marginal diagnostics (center and right).}
    \label{fig:vbl_calib_diag_with_marg}
\end{figure}

\begin{figure}[tbp]
    \centering
    \includegraphics[width=0.85\linewidth]{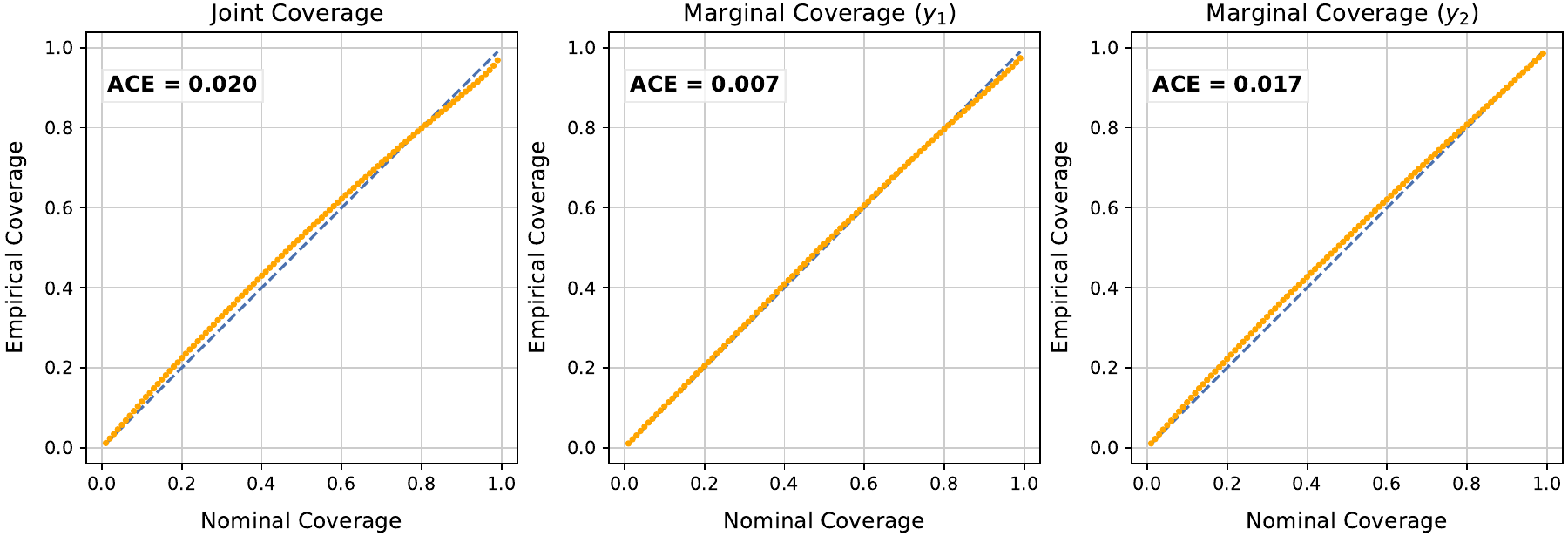}
    \vspace{0.15cm}
    \includegraphics[width=0.85\linewidth]{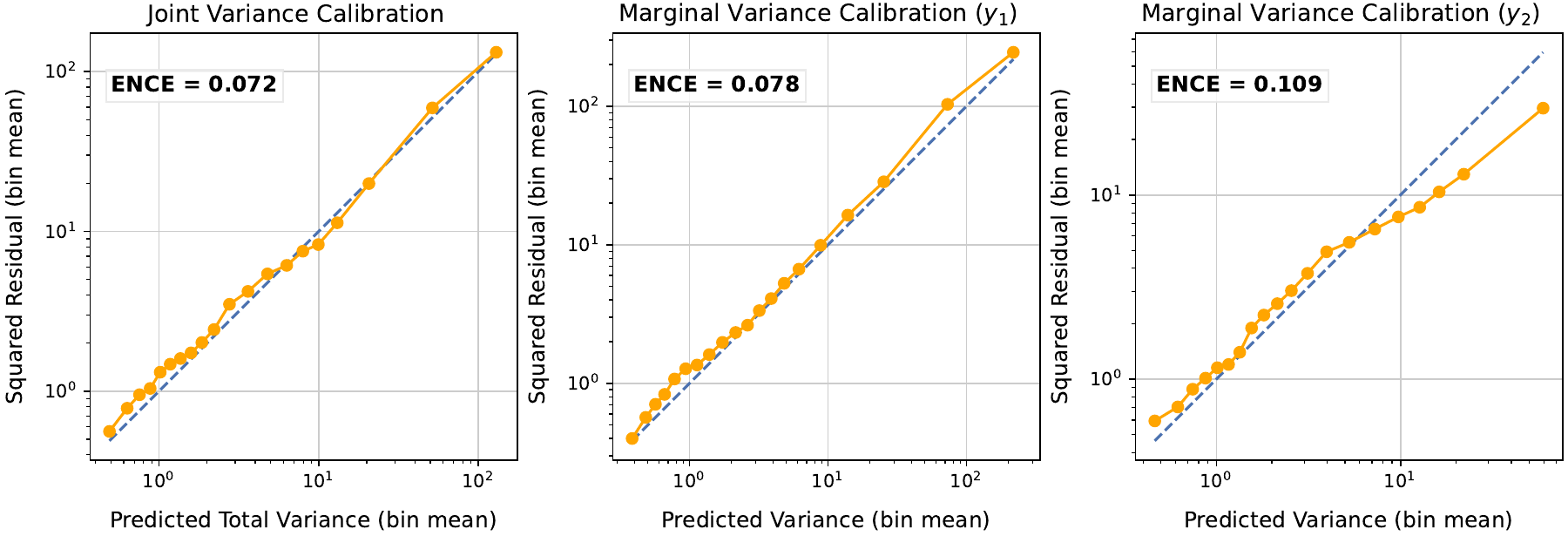}
    \vspace{0.15cm}
    \includegraphics[width=0.85\linewidth]{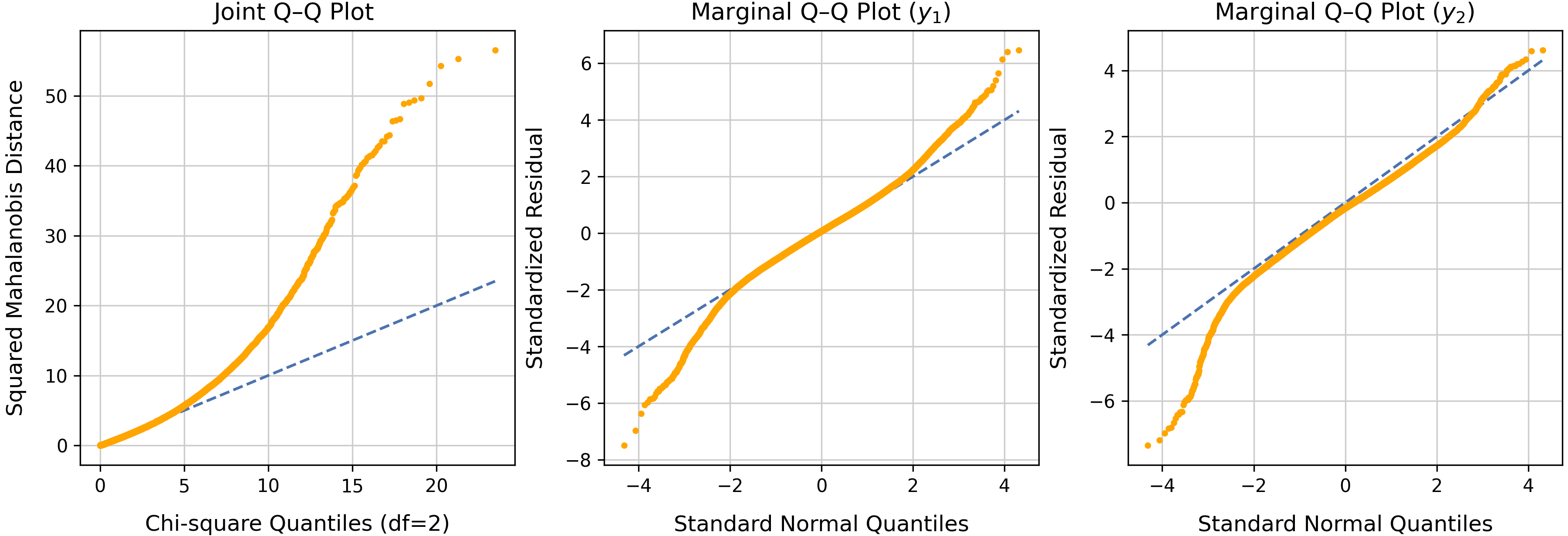}
    \caption{Calibration diagnostics for \ac{vbl}: Gaussian calibration + keypoint pruning. Same layout as \cref{fig:vbl_calib_diag_with_marg}.}
    \label{fig:vbl_qq_plots}
\end{figure}

\end{document}